\documentclass{article}

\usepackage[final,nonatbib]{neurips_2024}

\usepackage[utf8]{inputenc} % allow utf-8 input
\usepackage[T1]{fontenc}    % use 8-bit T1 fonts
\usepackage{hyperref}       % hyperlinks
\usepackage{url}            % simple URL typesetting
\usepackage{booktabs}       % professional-quality tables
\usepackage{amsfonts}       % blackboard math symbols
\usepackage{nicefrac}       % compact symbols for 1/2, etc.
\usepackage{microtype}      % microtypography
\usepackage[usenames,dvipsnames]{xcolor}         % colors
\usepackage{graphicx}
\usepackage{makecell}
\usepackage{algorithm}
\usepackage{algpseudocode}
\usepackage{wrapfig}
\usepackage{amsmath}
\usepackage{utfsym}
\usepackage{diagbox}
\usepackage{multirow}
\usepackage{threeparttable} 
\usepackage{caption}
\usepackage{subcaption}
\usepackage{colortbl}
\usepackage{cleveref}
\usepackage{float}

\usepackage[export]{adjustbox}

\title{Automated Label Unification for Multi-Dataset Semantic Segmentation with GNNs}

\begin{document}

\author{%
  Rong Ma\footnotemark[1], Jie Chen\footnotemark[1], Xiangyang Xue, and Jian Pu\footnotemark[2]\\
  Fudan University\\
  \texttt{rma22@m.fudan.edu.cn, \{chenj19,xyxue,jianpu\}@fudan.edu.cn} \\
}

% \footnote{These authors contributed to the work equally and should be regarded as co-first authors}

\maketitle

\renewcommand{\thefootnote}{\fnsymbol{footnote}}

\footnotetext[1]{These authors contributed to the work equally and should be regarded as co-first authors.}
\footnotetext[2]{Corresponding author.}
% \footnote{These authors contributed to the work equally and should be regarded as co-first authors}
\begin{abstract}
Deep supervised models possess significant capability to assimilate extensive training data, thereby presenting an opportunity to enhance model performance through training on multiple datasets. However, conflicts arising from different label spaces among datasets may adversely affect model performance. In this paper, we propose a novel approach to automatically construct a unified label space across multiple datasets using graph neural networks. This enables semantic segmentation models to be trained simultaneously on multiple datasets, resulting in performance improvements. Unlike existing methods, our approach facilitates seamless training without the need for additional manual reannotation or taxonomy reconciliation. This significantly enhances the efficiency and effectiveness of multi-dataset segmentation model training. The results demonstrate that our method significantly outperforms other multi-dataset training methods when trained on seven datasets simultaneously, and achieves state-of-the-art performance on the WildDash 2 benchmark. Our code can be found in \href{https://github.com/Mrhonor/AutoUniSeg}{https://github.com/Mrhonor/AutoUniSeg}.
\end{abstract}

\section{Introduction}
\label{sec:intro}

Recent advances in computer vision \cite{radford2021learning, kirillov2023segment} have shown the advantages of large datasets in training robust visual models \cite{awais2023foundational}. However, for deep supervised visual models that rely on annotated data, the collection of such extensive annotated datasets can be prohibitively costly \cite{chen2023scaledet}. To address this expense and expand the data available for training, several efforts \cite{bevandic2022multi, lambert2020mseg,meng2023detection} focus on the challenges of multi-dataset training, enabling the use of diverse datasets to train more robust and generalizable models.

% many efforts \cite{bevandic2022multi, lambert2020mseg} have been made to utilize multiple datasets to train more robust and generalizable models, enabling them to address challenges across various domains more effectively and adapt to real-world scenarios \cite{zendel2018wilddash}.
% Semantic segmentation \cite{hao2020brief} has become a fundamental research task in computer vision. Numerous works \cite{cheng2022masked, xie2021segformer} have achieved remarkable results on publicly available datasets. However, due to the high cost of annotation, existing datasets often focus on specific domains. Therefore, many efforts \cite{bevandic2022multi, lambert2020mseg} have been made to utilize multiple datasets to train more robust and generalizable models, enabling them to address challenges across various domains more effectively and adapt to real-world scenarios \cite{zendel2018wilddash}.

%Deep supervised models trained on multiple datasets need to address the issue of conflicting annotation standards and label spaces. 
Models trained on multiple datasets must confront the challenge of reconciling conflicting annotation standards and label spaces. For example, the class \textit{road} in the BDD dataset \cite{yu2018bdd100k} can be further divided into several classes in the Mapillary dataset \cite{neuhold2017mapillary}: \textit{road}, \textit{lane marking} and \textit{crosswalk}. Similarly, the Mapillary dataset labels both \textit{barrier} and \textit{curb} as distinct classes, while in the IDD dataset \cite{varma2019idd}, they are combined under the single label \textit{curb}. These conflicts impact the supervised learning of models, as they may be incorrectly penalized for predicting finer-grained classes from other datasets.

Another challenge is the task of unifying diverse dataset labels to produce outputs in a standardized format \cite{meletis2023training,lin2024universal}. Several methods \cite{chen2023scaledet, meng2023detection} attempt to address this by concatenating the label spaces of all datasets and using language models to encode label names into a text embedding space. However, there approaches introduce redundancy and fail to handle issues where labels share names but differ in annotation granularity. Other methods involve manually constructing universal taxonomies \cite{bevandic2022multi, lin2024universal} or relabeling \cite{lambert2020mseg}, both of which are time-consuming and labor-intensive. Recent approaches aim to automatically construct universal taxonomies \cite{bevandic2022automatic, uijlings2022missing}, but these methods typically identify inter-label relations between only two datasets, involving a time-consuming iterative training process.

% \begin{wraptable}{r}{4.3cm}
% 	\centering
% 	\begin{tabular}{lrrr}
% 	\hline
% Methods & Unified output & W/o manually assitance \\
% \hline
% - & - & - & - & 78.3 & 33.4 & 46.1 & 62.5 & 66.1 & 27.3 & 35.7 \\
% \checkmark & - & - & - & 78.6 & 33.9 & 46.5 & 64.0 & 66.9 & 28.3 & 35.6 \\ 
% \checkmark & \checkmark & - & - &  78.8 & 33.9 & 46.4 & 64.2 & 66.4 & 28.4 & 35.8\\ 

% \hline
% 	\end{tabular}
% \end{wraptable}

% The biggest challenge for simultaneous training is the conflicting label spaces and different labeling standards. Not only the different dataset has diverse label spaces but also the same named class actually has a different semantic definition \cite{uijlings2022missing}. For instance, the class car in Vistas \cite{neuhold2017mapillary} includes car and pickup but in Ade20K \cite{zhou2017scene} only includes car itself. The class pickup and van is included in the class van in Ade20k. Hence, Resolving label space conflicts solely based on class names is far from sufficient. 

% We use GNNs to learn the unified label embedding space and dataset label mapping. Images are encoded into pixel embedding through the segmentation network, projected into the unified label embedding space to provide unified label space predictions, and then mapped to the dataset-specific label space via the dataset label mapping for training.

\begin{figure}
  \centering
   \includegraphics[width=1\linewidth]{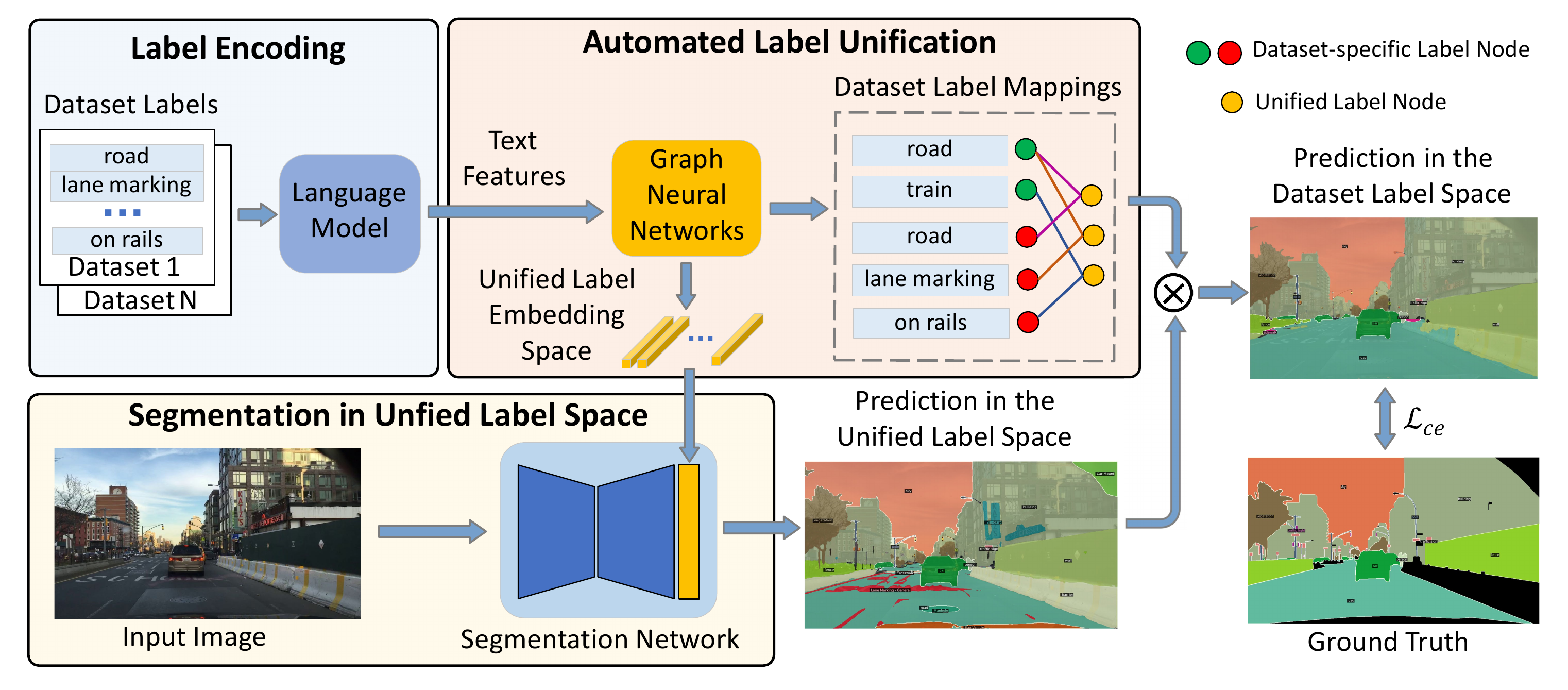}

   \caption{Our method consists of three modules. The label encoding provides the semantic text features of the dataset labels. The GNNs learn the unified label embedding space and dataset label mappings based on the textual features and input images. The segmentation network leverages the unified label embedding space to produce segmentation results in the unified label space.}
   \label{fig:GNNs}
\end{figure}

 % We model the relationship between the dataset-specific label space and the unified label space as a bipartite graph. Each dataset label is encoded by language model into dataset-specific node features, while we define a set of unified label nodes with randomly initialized learnable features. Subsequently, we use GNNs and training images from each dataset to learn label relationships and unified label embedding space. Based on learned label relation, we solve the dataset label mapping via unbalanced optimal transport.

In this paper, we propose a novel approach leveraging Graph Neural Networks (GNNs) \cite{kipf2016semi} to automatically construct unified label space, enabling segmentation model to be trained simultaneously on multiple datasets. In contrast to previous approaches \cite{chen2023scaledet,bevandic2022multi, bevandic2022automatic}, our approach eliminates the need for manual re-annotation or iterative training procedures to construct universal taxonomies, while also addressing the limitation of language-based methods in distinguishing categories with identical semantics. As depicted in \autoref{fig:GNNs}, we utilize a language model to convert the dataset labels into text features. 
Then, we apply GNNs to learn the relationships and associations among these labels. The process creates a unified label embedding space and dataset label mappings. The output head of the segmentation network incorporates the unified label embedding space to generate unified segmentation results within this unified space. The dataset label mappings are subsequently utilized to align the unified segmentation results with the label spaces relevant to each dataset. This enables the training of segmentation models and graph neural networks with dataset-specific annotations.

\section{Related work}
\label{sec:formatting}

%-------------------------------------------------------------------------
\textbf{Multi-datasets Semantic Segmentation.} In recent years, numerous studies \cite{zhou2022lmseg, meletis2023training} focused on training semantic segmentation models on multiple datasets. A simple approach involved incorporating dataset-specific modules in the model, such as dataset-specific output heads \cite{masaki2021multi} or dataset-specific batch normalization layers \cite{wang2022cross}, to produce predictions tailored to each dataset domain. While these methods effectively avoided issues of dataset label space conflicts, they offered limited applicability in real-world scenarios as they could not deliver unified predictions. MSeg \cite{lambert2020mseg} addressed the problem at the data level via manual re-annotation processes to resolve label conflicts. However, this approach was time-consuming, error-prone and not easily scalable. Recent methods employed manual \cite{bevandic2024weakly} or automatic techniques \cite{bevandic2022automatic} to construct universal taxonomies and establish label relationships between datasets label space and the unified label space. These methods, leveraging partial label learning approaches \cite{zhang2017disambiguation, cid2012proper}, enabled training with dataset-specific annotations and producing unified predictions.

\textbf{Construct Universal Taxonomies.} Each dataset has its unique domain, necessitating the construction of universal taxonomies to enable the model to cover all domains. Several approaches \cite{chen2023scaledet, meng2023detection} attempted to concatenate dataset label spaces and differentiated similar classes by aligning text embedding encoded by language model. However, this approach struggled with classes that had the same name but different levels of annotation granularity, and direct concatenation of label spaces could lead to semantic conflicts. Other research \cite{lin2024universal, meletis2023training} attempted to establish a unified label space through the expertise of human annotators. Additionally, other studies \cite{uijlings2022missing, zhou2022simple} aimed to address this issue by developing automated mechanisms to create universal taxonomies. However, these methods either could only construct a unified label space between two datasets at a time \cite{bevandic2022automatic} or could not handle complex class relationships \cite{zhou2022simple}. Our approach stands out by automatically constructing universal taxonomies in a single training session with multiple datasets, resulting in significant time savings compared to methods that require multiple training iterations. 

\textbf{Graph Neural Networks} demonstrated exceptional effectiveness in dealing with complex topological data structures, as highlighted in recent research  \cite{velivckovic2023everything}. The applicability extended across various domains, including recommendation systems  \cite{hao2020p}, knowledge graph construction  \cite{zhong2023comprehensive}, and skeletal action recognition  \cite{zhu2020topology}. In the context of our problem, discovering label relations can be conceptualized as a link prediction task \cite{zhang2018link, kumar2020link}. However, conventional approaches for graph link prediction  \cite{cai2020multi, cai2021line} are not suitable for our model since we do not possess ground truth links. We use the values of the learnable adjacency matrix as predictions of whether nodes are linked. Supervision of the linked predictions relys on the segmentation results of the segmentation network and the corresponding image annotations.

\begin{figure}
    \centering
    \includegraphics[width=1.0\linewidth]{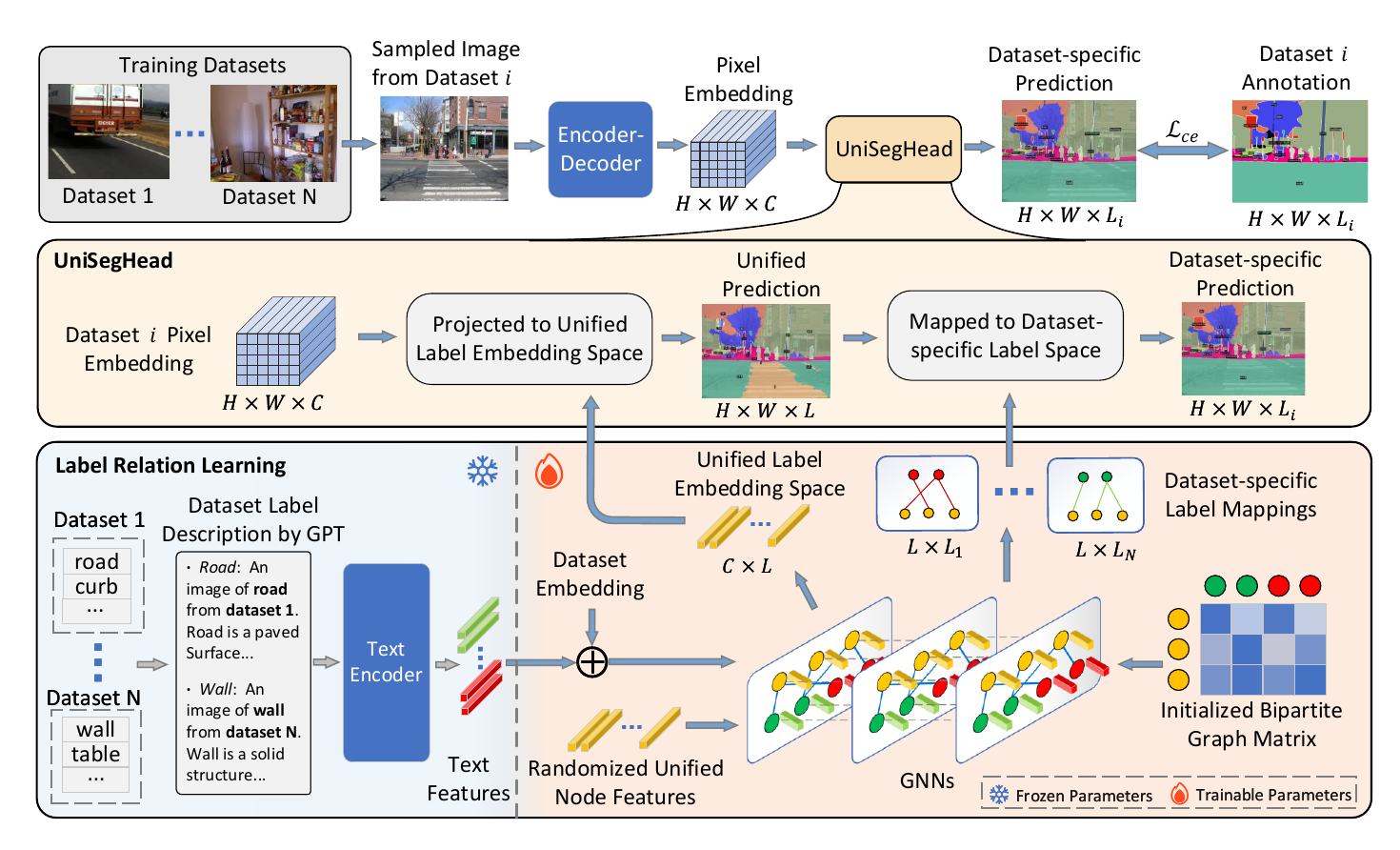}
    \caption{%Our method trains with dataset-specific annotations through label mappings constructed by GNNs. Our semantic segmentation model leverages a UniSegHead to enable simultaneous training on multiple datasets. 
    Illustration of our method that training with dataset-specific annotations through label mappings constructed by GNNs. We leverage a unified segmentation head (UniSegHead) to enable simultaneous training on multiple datasets. In the UniSegHead, we compute the matrix product between pixel embedding and augmented unified node features output by the GNNs, resulting in predictions for the unified label space. We finally utilize the label mappings constructed by GNNs to map the unified predictions to dataset-specific prediction for training.}
    \label{fig:framework}
\end{figure}

\section{Proposed Method}
\label{sec:method}
The comprehensive framework is depicted in \autoref{fig:framework}. We first define the unified representation of the multi-dataset label space. Utilizing this representation, we build a graph neural network to learn the unified label space. Finally, leveraging the unified label space, we train our segmentation network and the graph neural network using the dataset annotations.

% First, we introduce our method for automatically unifying the semantic label space across multiple datasets in \autoref{sec:uni}. Then, in \autoref{sec:gnn}, we explain how we optimize the unified label space using GNNs. Finally, in \autoref{sec:st-train}, we describe our approach for training semantic segmentation models and GNNs with dataset-specific annotations.

\subsection{Unification of Multi-dataset Label Space}
\label{sec:uni}
% During training, an image is randomly sampled from dataset $i$ and fed to an encoder-decoder, which provides embedding for each pixel position $\{\mathbf{p}_1,  \mathbf{p}_2, \ldots , \mathbf{p}_j\}$, where $v_i$ is $D$-dimensional vectors. 
\textbf{Unified Label Space.} Given $K$ datasets with their respective label space $\{L_1, L_2, \ldots, L_K\}$, multi-dataset semantic segmentation requires a model to predict within a consistent label space that encompasses all dataset label spaces $L_{pred} = \bigcup_{i=1}^K L_i$. Each pixel must be assigned to a label in this unified label space. We define $N$ unified label nodes $\mathcal{A}=\{\alpha_1, \alpha_2, \ldots, \alpha_N\}$, serving as the nodes in the graph neural networks. Their corresponding $D$-dimensional learnable embedding $\{\mathbf{x}_1, \mathbf{x}_2, \ldots, \mathbf{x}_N\}$ represent the unified label embedding space. The number of unified label nodes $N$ is often smaller than the total number of dataset classes $|L|$, because we aim to merge multiple identical classes into a single unified label. The image is first encoded into pixel embedding $\mathcal{P}$ by the segmentation network, which is then projected into the unified label embedding space to assign a unified label to each pixel.

\textbf{Label Mappings.} We define a mapping from the unified label space to the dataset-specific label space $\mathbf{M}_i: \mathcal{A} \rightarrow L_i$, which is used to train the model with dataset-specific annotations. Mathematically, $\mathbf{M}_i \in \{0,1\}^{N \times |L_i|}$ is a boolean linear transformation. Each unified class $\alpha$ is at most linked to a class $c$ within a specific dataset $i$ to prevent label conflicts: $\mathbf{M}_\alpha \mathbf{1} \leq \mathbf{1}$. To handle different annotation granularities, we use label mappings to merge multiple unified label nodes representing fine-grained classes into one super-class $c$: $\mathbf{M}_c^\top \mathbf{1} \geq \mathbf{1}$. For example, the \textit{curb} from IDD can be simultaneously mapped by unified label nodes represented \textit{curb} and \textit{barrier}. We use unified label nodes as input nodes for the GNNs, which learn the label mappings and unified label embedding space, thus enabling the automated unification of multi-dataset label spaces.

\subsection{Learning Unified Label Space with Graph Neural Networks}
\label{sec:gnn}

Learning the label mappings between dataset label spaces and the unified label space can be viewed as a bipartite graph matching problem. This makes graph neural networks well-suited to address this issue. Below, we detail our approach of constructing GNNs for learning a unified label space.

% Learning label mapping between the unified labels and the dataset-specific labels can be seen as a bipartite graph matching problem. In this graphical representation, each dataset label $l_m^{(i)}$ and atom $\alpha_m$ is viewed as node and their label mapping is viewed as links. We can efficiently process relationships between label nodes using graph neural networks. With the help of training samples, we can optimize the label mapping during the training process.

\textbf{Input Nodes.} To construct the input feature of dataset-specific label nodes, as illustrated in \autoref{fig:framework}, we used the dataset labels in the template "An image of <label> from the dataset <dataset>" as plain text input and employed ChatGPT to complete the detailed description of each label. Then, we employ llama-2 \cite{touvron2023llama} to encode these label descriptions and generate text features. To further distinguish nodes from different datasets, we introduce learnable dataset embedding for each dataset $\{\mathbf{d}_1, \mathbf{d}_2, \ldots, \mathbf{d}_K\}$. The dataset embedding is combined with the text features to form the input features for dataset-specific label nodes: 
\begin{equation}
\mathbf{x}_{i,m} = f_t(l_{i,m}) + \mathbf{d}_i,
\label{eq:feat}
\end{equation}
where $\mathbf{x}_{i,m}$ is the input feature of the $m$-th label from dataset $i$, and $f_t(l_{i,m})$ is the text feature of the label description $l_{i,m}$ encoded by language model. The input features of unified label nodes are randomly initialized to the same dimension as the dataset-specific label node. To determine the appropriate number of unified label nodes, inspired by the approach in \cite{zhou2022simple}, we used cross-validation results across different datasets to identify the number of mergeable categories, which served as the initial selection for the unified label nodes. The specific algorithm can be found in \autoref{apdx:2}. Together, dataset-specific label nodes and unified label nodes constitute the input nodes of GNNs. During the training process, we maintain a constant number of nodes. After the training is completed, we will remove inactive unified label nodes, meaning those that were not assigned to any dataset-specific label.
% features provide the model with the ability to choose similar dataset-specific nodes to connect and126
% determine a better label mapping
% The continuous adjacency matrix $\mathbf{M}_a$, with values ranging from 0 to 1, could be considered as attention weights that control information propagation.

\textbf{Learnable Adjacency Matrix.} To enable label mappings to be updated via gradient descent, we embed the label mappings as a continuous, learnable graph adjacency matrix $\mathbf{M}_a$. Values in the adjacency matrix represent the weights of corresponding edges. Only the edges between unified label nodes and dataset-specific nodes are learnable, while others are fixed to zero. We apply a softmax operation to the edges connecting each unified label node with nodes of a particular dataset, ensuring that the sum of the edge weights $w$ between each unified label node and nodes of this dataset equals one. The element at the intersection
of the $r$-th row and the $c$-th column in the lower triangular portion of $\mathbf{M}_a$ is formulated as:
\begin{equation}
\mathbf{M}_{r,c}=\left\{\begin{array}{l}
\dfrac{e^{w_{r,c}}}{\sum_{c^\prime \in L_i, r^\prime \in \mathcal{A}}e^{w_{r^\prime,c^\prime}}} \quad \text{if r > }|L|\text{ and c < }|L| \\
0 \qquad \qquad \qquad \qquad  \text {\hspace{1mm} otherwise}. \\
\end{array}\right.
\label{eq:learnable}
\end{equation} 
\textbf{Unified Label Embedding.} The forward propagation of our graph model follows the GraphSAGE framwork \cite{hamilton2017inductive} as formulated in \autoref{eq:at}. $\mathbf{W}^{k}$ and $\mathbf{X}^{k}$ are the weight and feature of the $k$-th GNNs layer. 
%The $\sigma$ indicates nonlinear activation function such as tanh and ReLU.
The $\sigma$ indicates nonlinear activation function, implemented as the tanh in this work.
The output features of the unified label nodes from the final layer serve as the unified label embedding space, $\mathbf{X}_u=[\mathbf{x}_1, \mathbf{x}_2, \ldots, \mathbf{x}_N]^\top$. 
\begin{equation}
\mathbf{X}^{k+1} = \sigma(\mathbf{W}^{k}[\mathbf{X}^{k} \| \mathbf{M}_a \mathbf{X}^{k}]).
\label{eq:at}
\end{equation}
To obtain the dataset label mappings, we partition the adjacency matrix into submatrices, each corresponding to a specific dataset. Each submatrix contains only the unified label nodes and the label nodes specific to that dataset. We compute the label mappings for each dataset based on the values in its corresponding submatrix, divided into the following two cases. During training, we will alternately train the segmentation network and the GNNs. When training the GNNs, we directly use the value of the learnable adjacency matrix to establish label mappings, thereby facilitating weight updates through gradient descent. When training the segmentation network, we utilize the unbalanced optimal transport algorithm to solve for the boolean label mappings that satisfies the many-to-one mapping constraints. This algorithm detailed in \autoref{apdx:1} effectively handles the conversion of the continuous adjacency matrix into a discrete dataset label mappings required for training segmentation network.

\subsection{Training a Universal Model with Dataset-specific Annotations}
\label{sec:st-train}
Training a universal model is divided into two steps. The first step involves training a robust encoder-decoder to provide the pixel embedding for each pixel position in the image, where similar objects should have similar features. The second step focuses on learning label mappings and unified label embedding space by GNNs. During the training phase, we alternate between these two steps, freezing one network while training another network. Both of these steps require supervised training using dataset-specific annotations. Here, we primarily focus on the training of GNNs, while further details of the training strategies are provided in the \autoref{apdx:train}.

\textbf{Training with Dataset-specific Annotations.} During training, an image is randomly sampled from dataset $i$ and fed into an segmentation network, which provides embedding for each pixel position $\mathcal{P}=\{\mathbf{p}_1,  \mathbf{p}_2, \ldots , \mathbf{p}_j\}$, where $\mathbf{p}_k$ is $D$-dimensional vectors. We utilize a unified segmentation head (UniSegHead) to assign a dataset label to each pixel. In the UniSegHead, We first project the pixel embedding into the unified label embedding space by multiplying the pixel embedding by the output feature of unified label node $\mathbf{X}_u$, as shown in \autoref{eq:u}. Then, to train the universal model with dataset-specific annotations, we need to map the predictions in unified label space to dataset-specific label space to obtain the probabilities of dataset-specific classes. This is achieved by computing per-pixel dataset-specific logits 
$\mathbf{s}$ by multiplying the dataset-specific label mappings $\mathbf{M}_i$ at each pixel: 
% We could compute the per-pixel dataset-specific logits $\mathbf{s}$ by multiplying the dataset-specific label mappings $\mathbf{M}_i$ at each pixel:
\begin{align} 
 \label{eq:u} & \mathbf{u}_k = \mathbf{X}_u \mathbf{p}_k, \\ 
 \label{eq:aggr} & \mathbf{s}_k = \mathbf{M}_i \mathbf{u}_k.
\end{align}
% \begin{align} 
%  \label{eq:u} & \mathcal{U} = \mathbf{X}_u\mathcal{V}, \\ 
%  \label{eq:aggr} & s_{j,k}^{(i)} = M_i u_{j,k}.
% \end{align}
Finally, probabilities of dataset-specific classes are computed by per-pixel softmax operation over the logits $s$. This allows us to use dataset-specific annotations to compute the pixel-wise loss function, train the network, and update the label mappings. We formulate the cross-entropy loss for a specific pixel to train the segmentation network:
\begin{equation}
\mathcal{L}_{ce}(\mathbf{y}, \mathbf{s}) = -\sum^{|L_i|}_{c=1} y_{c} \log(\operatorname{softmax}(\mathbf{s})_c),
\label{eq:celoss}
\end{equation}
where $\mathbf{y}$ is the pre-pixel annotation, $|L_i|$ represents the total number of classes in the dataset $i$ from which the image originates. 
% To better understand the process of learning label mapping transformation, we analyze the partial derivative of the loss function with respect to the label mapping matrix $M$, as shown:
% \begin{equation}
% \frac{\partial \mathcal{L}_{ce}}{\partial M_{k, c}}=u_c\left(\sigma(s_k)-y_k\right)
% \end{equation}

% Here, position $k$ corresponds to the $k$-th class in the dataset-specific label space, and position $c$ corresponds to the output of the $c$-th class in the unified label space, we use $u_c$ to present the $c$-th element of unified logit at that pixel. Analyzing the gradient with respect to the partial derivative, the sign and magnitude of the gradient of $a_{k,c}$ are jointly controlled by $u_c$ and $y_k$. When $y_k$ is 1, $(\sigma(s_k) - y_k)$ is always less than 0. For positions where $u_c$ is greater than 0, there will be a negative gradient, which is reasonable because a specific dataset class can be represented by multiple semantic atoms. Additionally, The larger the value of $u_c$, the more likely the current pixel belongs to class label $c$, resulting in a larger absolute gradient magnitude. Therefore, after gradient updating, the value of $M_{k,c}$ corresponding to the largest $u_c$ value is more likely to increase. Indicating that unified label c space is more likely to have similar semantic information to class k in the dataset label space. When $y_k$ is 0, $u_c$ needs to be less than 0 to maintain a negative gradient for $M_{k,c}$. Otherwise, it indicates that unified class $c$ should be mapped to where $y_j=1$ instead of class $k$.

\begin{table}[t]
\centering
\caption{Training and test datasets in our experiments.}
\label{tb:datasets}
\resizebox{\columnwidth}{!}{
\begin{tabular}{ccc}
\hline
Dataset Domain & Training and Validation datasets & Unseen test datasets\\
\hline
Driving scene & CityScapes \cite{cordts2016cityscapes}, Mapillary \cite{neuhold2017mapillary}, BDD \cite{yu2018bdd100k}, IDD \cite{varma2019idd} & WildDash 2 \cite{zendel2022unifying}, KITTI \cite{geiger2013vision}, CamVid \cite{brostow2009semantic}, \\
Indoor scene & SUN RGBD \cite{song2015sun} & ScanNet-20 \cite{dai2017scannet}\\
Everyday objects & ADE20K \cite{zhou2019semantic}, COCO \cite{lin2014microsoft} & PASCAL VOC \cite{everingham2010pascal}, PASCAL Context \cite{mottaghi2014role}\\

\hline
\end{tabular}
}
\end{table}

\textbf{Orthogonality Loss.}
To achieve a conflict-free unified label space and avoid redundant unified label nodes that represent the same class or have overly similar features, we introduce soft constraints to promote orthogonality among the unified label node features, inspired by \cite{toldo2021unsupervised}. This orthogonality loss encourages unified label node embedding to be mutually orthogonal. It not only aligns the unified label nodes with annotation standards for practical use but also enhances the diversity of the model and helps in finding a better label mappings:
\begin{equation}
\label{eq:l_orth}
\mathcal{L}_{orth} = -\sum_{i=1}^{N} \operatorname{softmax}(\mathbf{X}_u\mathbf{x}_{i})_i \log(\operatorname{softmax}(\mathbf{X}_u\mathbf{x}_{i})_i).
\end{equation}
The final loss function used to train the GNNs is represented as follows, with $\lambda$s as hyperparameters to adjust the weights of different loss functions:
%We implemented training for the GNNs based on this loss function, while training the segmentation network only using \autoref{eq:celoss}. 
\begin{equation}
\label{eq:l_GNN}
\mathcal{L} = \lambda_1 \mathcal{L}_{ce} + \lambda_2 \mathcal{L}_{orth} .
\end{equation}
 % Additionally, we introduce the sparsity loss $\mathcal{L}_{spa}$ to encourage sparsity in the adjacency matrix, as shown in \autoref{eq:l_spa}. This could help the model remove some meaningless connections.
    % During the last seg traing phase, we drop the entire GNNs but the $\{M^{(i)}, i=1,...,D \}$. The classifier in this stage is not longer keep frozen and trained with the seg network. In the latter part of this stage, we will use a part of training dataset (randomly selected before the stage one and never been used during previous training stage) to evaluate the model's performance. Those link between the unified label space and dataset-specific label space which is not activated during the evaluation will be removed. After removing the unused connection, we reintroduce the previously selected images back into the dataset and continue the finetune stage. And thus, we obtain the final model. When apply the final model to predict in the wild, we will remove the label mapping transformation and manually assign every semantic atom a class name accord to their connection and actually prediction and semantics.
% \begin{align}
% \label{eq:l_orth} & \mathcal{L}_{orth} = \sum_{i=1}^{A} \sigma_i(\mathcal{X}_{i}) log(\sigma_i(\mathcal{X}_{i})) \\
% % \label{eq:l_spa} & \mathcal{L}_{spa} = \| M_{a} \|_{Fro}^2 \\
% \label{eq:l_GNN} & \mathcal{L} = \lambda_1 \mathcal{L}_{ce} + \lambda_2 \mathcal{L}_{orth} % + \lambda_3 \mathcal{L}_{spa}
% \end{align}

\section{Experiments}
\label{sec:exp}

% In this section, we first introduce our dataset setup and implementation details. 
\textbf{Datasets.} We list the semantic segmentation datasets used for training and testing in \autoref{tb:datasets}. Our training datasets cover a wide range of scenarios, from indoor scenes to driving scenes. We also introduce corresponding test datasets, which are not used in the training process, for the respective scenes to evaluate our generalization capability. %The data in the test datasets is never used for training in our model.

\begin{table}[t]
\caption{Multi-dataset performance compared with other methods.}
\label{tb:main}
\centering

    \begin{threeparttable}
    \resizebox{\columnwidth}{!}{
\begin{tabular}{l|ccc|cccccccc}

\hline
 Methods & Backbone & Venue & Label space\tnote{1} & CS & MPL & SUN & BDD & IDD & ADE & COCO & Mean
 \\
 \hline
MSeg \cite{lambert2020mseg} & HRNet-W48 & CVPR 20  & MR & 76.3 &  51.9\tnote{2} & 46.1 & \underline{63.5} & 61.8 & 42.8\tnote{2} & 48.6\tnote{2} & \underline{55.9} \\
NLL+ \cite{bevandic2022multi} & SNp-RN18 & WACV 22 & MC & 72.6 & 39.1 & 41.7 & 58.5 & 54.4\tnote{3} & 31.0 & 35.4 & 47.5   \\
Uni NLL+ \cite{bevandic2024weakly} & SNp-DN161 & IJCV 24  & MC & 76.1  & \textbf{44.2}  & \underline{46.9} & 60.4 & 56.7\tnote{3} & \underline{35.6} & \underline{39.3} & 51.3 \\
\hline
 Single dataset & HRNet-W48 & - & DS & 77.0 & 30.2 & 43.9 & 62.4 & 66.8 & 34.5 & 38.0 & 50.4 \\
 % Multi-SegHead & HRNet-W48 & - & DS & \underline{77.7} & 33.4 & 45.1 & 63.4 & 65.6 & 28.3 & 32.9 & 49.5 \\
  Multi-SegHead & HRNet-W48 & - & DS & \underline{79.5} & 36.1 & 47.3 & \textbf{65.6} & \underline{67.0} & 30.5 & 36.7 & 51.8 \\
Auto univ. \cite{bevandic2022automatic} & SNp-RN18 & BMVC 22 & Auto  & 72.7 & 35.8 & 42.3 & 59.6 & 55.2\tnote{3} & 30.7 & 35.6 & 47.4 \\
Ours & HRNet-W48 & - & Auto & \textbf{80.7} & \underline{43.7} & \textbf{47.5} & \underline{65.5} & \textbf{68.6} & \textbf{42.0} & \textbf{46.7} & \textbf{56.4}  \\
\hline
\end{tabular}
}
\begin{tablenotes}
\scriptsize
\item[1] Approach to construct label space. MC:Manually Construct, MR:Manually Relabel, DS:Dataset-specific, Auto:Automatically Construct.
\item[2] MSeg train and evaluate on 43 of 65 class in Mapillary dataset, 117 of 150 class in ADE dataset, 122 of 133 class in COCO dataset.
\item[3] These methods were trained and evaluated using 30 classes from the IDD dataset, while we trained and evaluated using the officially \\
recommended 26 classes.
\end{tablenotes}
\end{threeparttable}
\end{table}

\begin{table}[t]
\belowrulesep=0pt
\aboverulesep=0pt
\centering

\caption{Performance comparison with two baselines on training and unseen datasets.}
\label{tb:mean}
\resizebox{0.85\columnwidth}{!}{
\begin{tabular}{c|cccc}
\hline
\multirow{2}{*}{\makecell{Trained dataset \\ or label space}} & \multicolumn{2}{c}{Mean results across training datasets} & \multicolumn{2}{c}{Mean results across unseen datasets}\\
\cmidrule(lr){2-3}\cmidrule(lr){4-5}
& Single dataset & Multi-SegHead & Single dataset & Multi-SegHead \\
\hline
CS &  23.9 & 30.9 & 31.4 & 37.9\\
MPL & \underline{30.6} & 36.1 & 36.9 & 41.0\\
SUN & 12.9 & 22.0 & 16.7 & 28.9\\
BDD &  24.8 & 29.1 & 30.4 & 35.7\\
IDD &  26.0 & 34.0 & 28.9 & 36.4\\
ADE &  28.5 & \underline{40.3} & 37.2 & 49.8\\
COCO & 30.5 & 38.8 & \underline{45.1} & \underline{53.2}\\
\hline
Ours  & \multicolumn{2}{c}{\textbf{56.4}} & \multicolumn{2}{c}{\textbf{56.9}}\\
\hline
\end{tabular}
}
\end{table}

\textbf{Implementaion Details.} Our segmentation model is based on the HRNet-W48 architecture  \cite{sun2019high}, while the GNN model is a three-layer GraphSAGE  \cite{hamilton2017inductive}. We utilize the llama-2-7B model to encode label descriptions into 4096-dimensional text features. These text features, augmented with dataset embedding of the same dimensionality, are then employed as node features input into the GraphSAGE. When forming a minibatch from multiple datasets, we evenly sample 3 images per dataset within a batch for each GPU. For all images, We first apply random resizing with a ratio ranging from 0.5 to 2, followed by a random crop operation to achieve a final image size of 768 × 768 pixels. We use AdamW optimizer \cite{loshchilov2017decoupled} with warmup and polynomial learning rate decay, starting with a learning rate of 0.0001. We train our model for 300k iterations on four 80G A100 GPUs.

\subsection{Comparison on Multiple Datasets} 

% \begin{table}
% \caption{Performance, MC:Manually Construct, MR:Manually relabel, DS:dataset-specific, Auto:Automatically Construct, 1: MSeg only train and evaluate on 43 of 65 class, 2: train and evaluate on 117 of 150 class, 3: MSeg train and evaluate on 122 of 133 class.}
% \label{tb:main}
% \centering
% \resizebox{\columnwidth}{!}{
% \begin{tabular}{l|cc|cccccccc}
% \hline

% \hline
%  Methods & Publication & Label space & CS & VT & SUN & BDD & IDD & ADE & COCO & Mean
%  \\
%  \hline
% MSeg \cite{lambert2020mseg} & CVPR 20  & MR & 76.3 &  51.9^1 & 46.1 & \underline{63.5} & 61.8 & 42.8^2 & 48.6^3 & \underline{55.9} \\
% NLL+ \cite{bevandic2022multi} & WACV 22 & MC & 72.6 & 39.1 & 41.7 & 58.5 & 54.4 & 31.0 & 35.4 & 47.5   \\
% Uni NLL+ \cite{bevandic2024weakly} & IJCV 24  & MC & 76.1  & \textbf{44.2}  & \underline{46.9} & 60.4 & 56.7 & \underline{35.6} & \underline{39.3} & 51.3 \\
% \hline
%  Single dataset & - & DS & 77.0 & 28.8 & 43.9 & 62.4 & \underline{66.8} & 34.5 & 38.0 & 50.2 \\
%  Multi-head & - & DS & \underline{77.7} & 33.4 & 45.1 & 63.4 & 65.6 & 28.3 & 32.9 & 49.5 \\
% Auto univ. \cite{bevandic2022automatic} & BMVC 22 & Auto  & 72.7 & 35.8 & 42.3 & 59.6 & 55.2 & 30.7 & 35.6 & 47.4 \\
% Ours & - & Auto & \textbf{80.7} & \underline{43.7} & \textbf{47.5} & \textbf{65.5} & \textbf{68.6} & \textbf{42.0} & \textbf{46.7} & \textbf{56.4}  \\
% \hline
% \end{tabular}
% }

% \end{table}

In \autoref{tb:main}, we present the accuracy of our methods and compare them to other approaches on the seven training datasets. We use mean Intersection over Union (mIoU) to quantify the performance of models, a common metric used to evaluate the performance of segmentation models. Different methods adopt various approaches to construct their label spaces: \textit{Dataset-Specific} represents a lack of a unified label space, where the model outputs a separate label space for each dataset. \textit{Manually Relabel} means manually re-annotating each image. \textit{Manually Construct} means the label space is constructed through human expertise. \textit{Automatically Construct} includes methods where the unified label space is automatically constructed by the model. We also establish two baseline methods: \textit{Single dataset} and \textit{Multi-SegHead}. \textit{Single dataset} demonstrates the results of training on individual dataset only, while \textit{Multi-SegHead} trains on multiple datasets by using dataset-specific segmentation heads. 

The results demonstrate that our method achieves the best average performance in multi-dataset training, while also achieving significant performance improvements on datasets with a large number of classes such as the ADE and COCO datasets. We attribute this to the construction of a robust unified label space. Leveraging visual connections from the samples, our unified label space can discover label relationships beyond textual similarities. For instance, the visual appearance of the \textit{fireplace} in ADE is similar to the \textit{tunnel} in Mapillary. Despite their different semantic meanings, our model merges these labels for prediction, as detailed in \autoref{sec:vis}. This approach saves model capacity and facilitates knowledge transfer across datasets for improved prediction.

In \autoref{tb:mean}, as a supplement to \autoref{tb:main}, we compare our model with various models trained on single dataset, as well as each segmentation head output of the \textit{Multi-SegHead}. Detailed data for each dataset can be found in the \autoref{apdx:tb}. From the table, it can be observed that training with multiple datasets helps improve the model's generalization performance. However, the performance of different segmentation head outputs in \textit{Multi-SegHead} model shows significant differences due to the lack of a unified form of output that performs well across all datasets. In contrast, our approach provides a unified label space covering all datasets, resulting in a significant advantage in average performance. We also list the performance on the five unseen test datasets mentioned in \autoref{tb:datasets}. To evaluate the performance of our model on unseen datasets, we first evaluate the model results on its training dataset. We search for the optimal label mappings based on the accuracy of label predictions. There are no updates to any model parameters except the label mappings in this process. This process can actually be done manually without any annotation information. The results indicate that our model exhibits better generalization performance. It can handle various scenarios and consistently achieve excellent performance on the test set. Compared to \textit{Multi-SegHead model}, our automatically constructed label space has advantages over dataset label spaces. The unified label space can integrate the semantic information from multiple dataset label spaces.

\begin{figure}[t]
  \centering
  \footnotesize
  \setlength{\tabcolsep}{0.02cm} % 设置列之间的间距
  \begin{tabular}{lccccccc}
      & Image & Ground Truth & BDD Model & MPL Model & ADE Model & Our 7ds Model \\
    \rotatebox[origin=l]{90}{\quad BDD} &
    \includegraphics[width=0.16\linewidth]{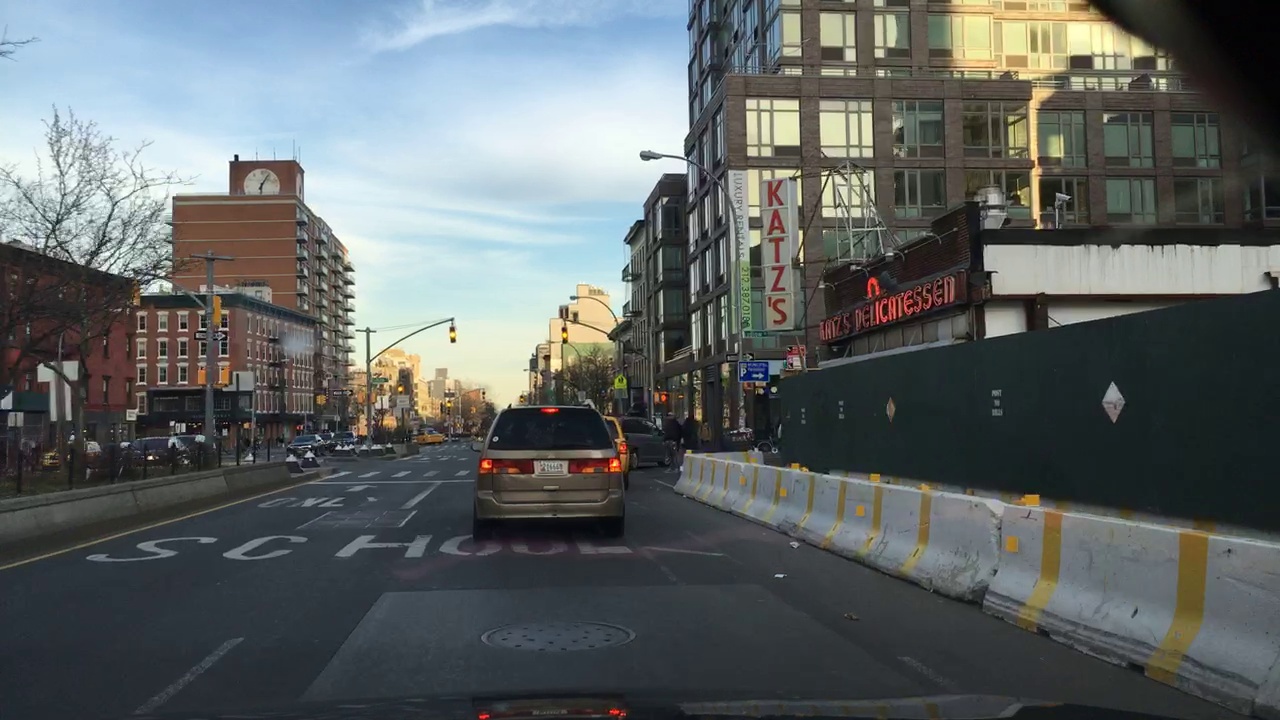} &
    \includegraphics[width=0.16\linewidth]{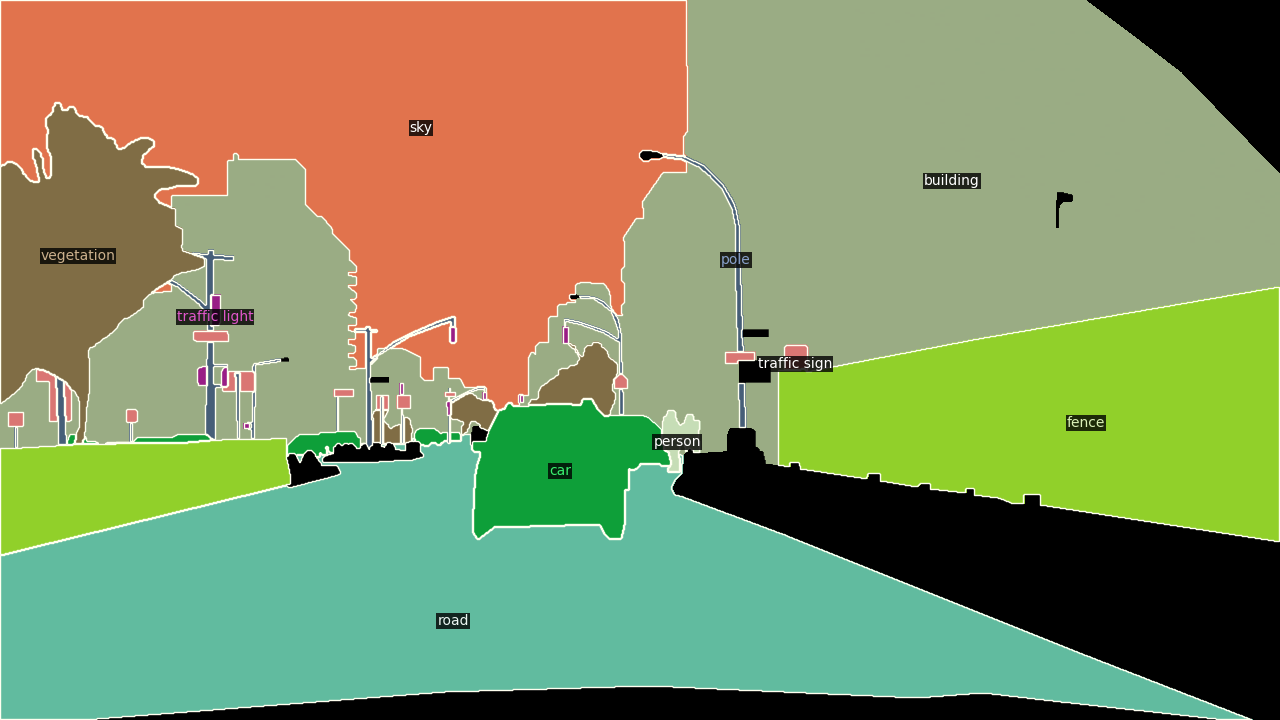} &
    \includegraphics[width=0.16\linewidth]{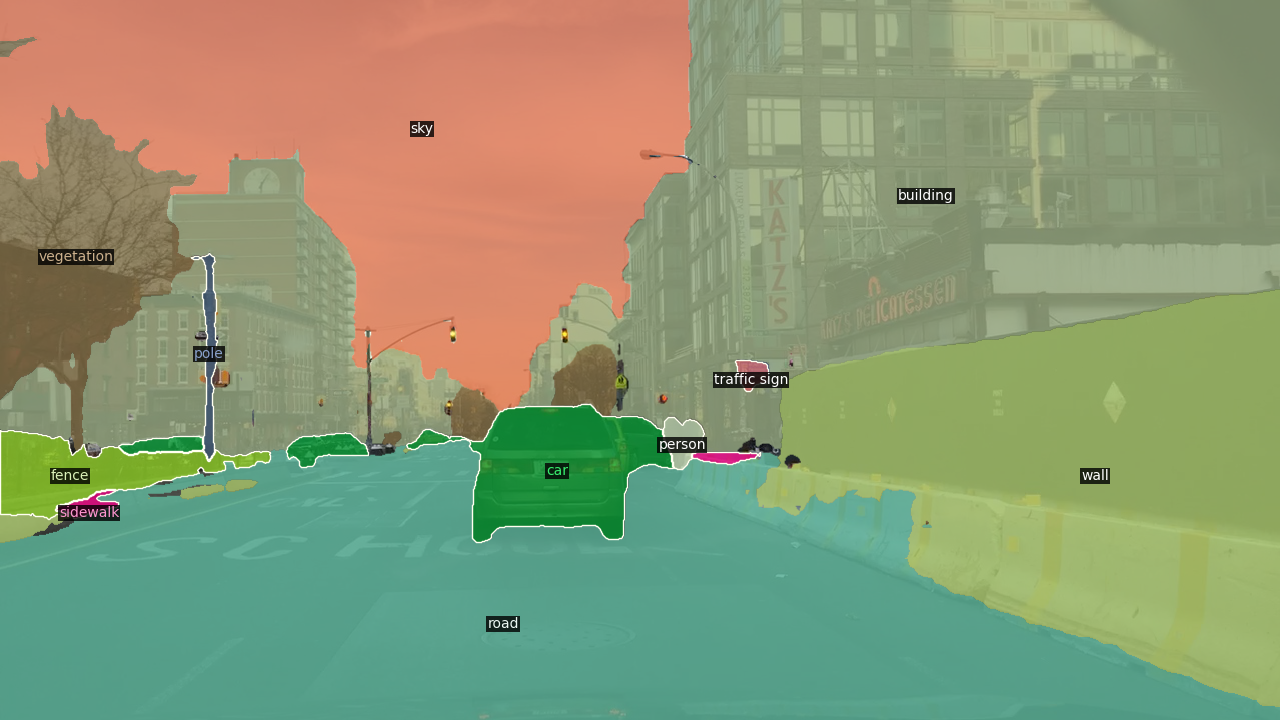} &
    \includegraphics[width=0.16\linewidth]{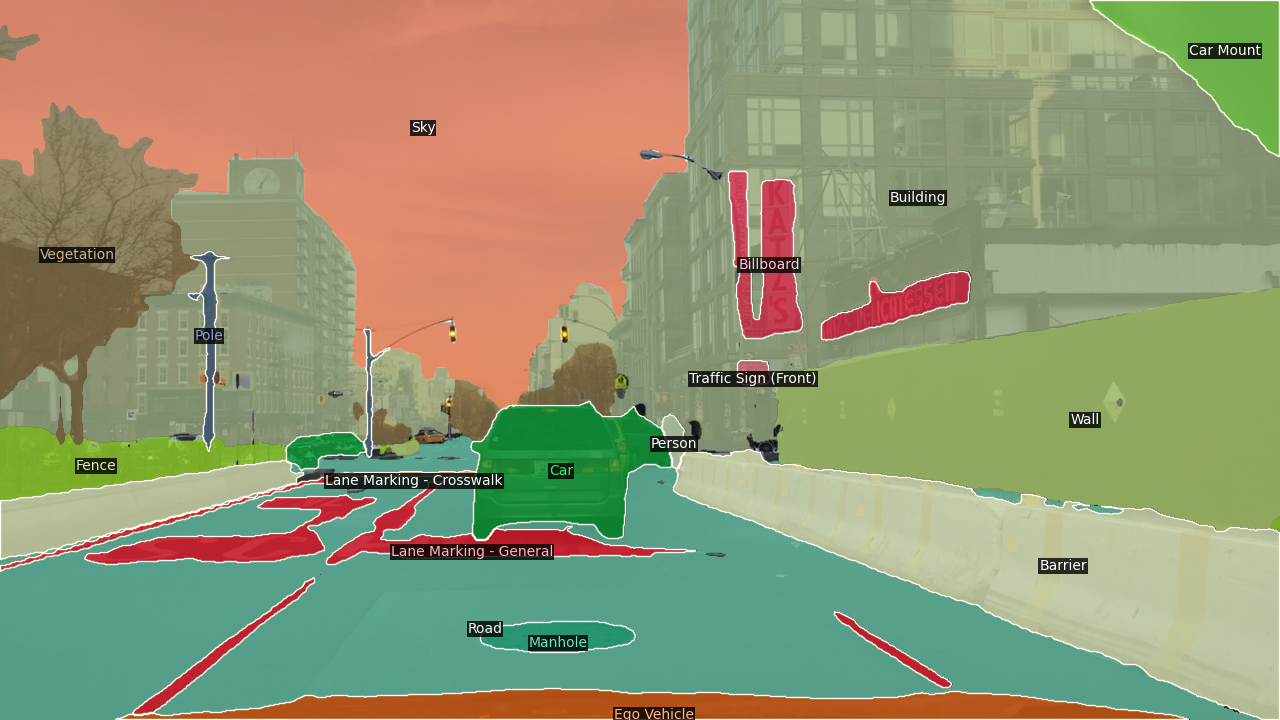} &
    \includegraphics[width=0.16\linewidth]{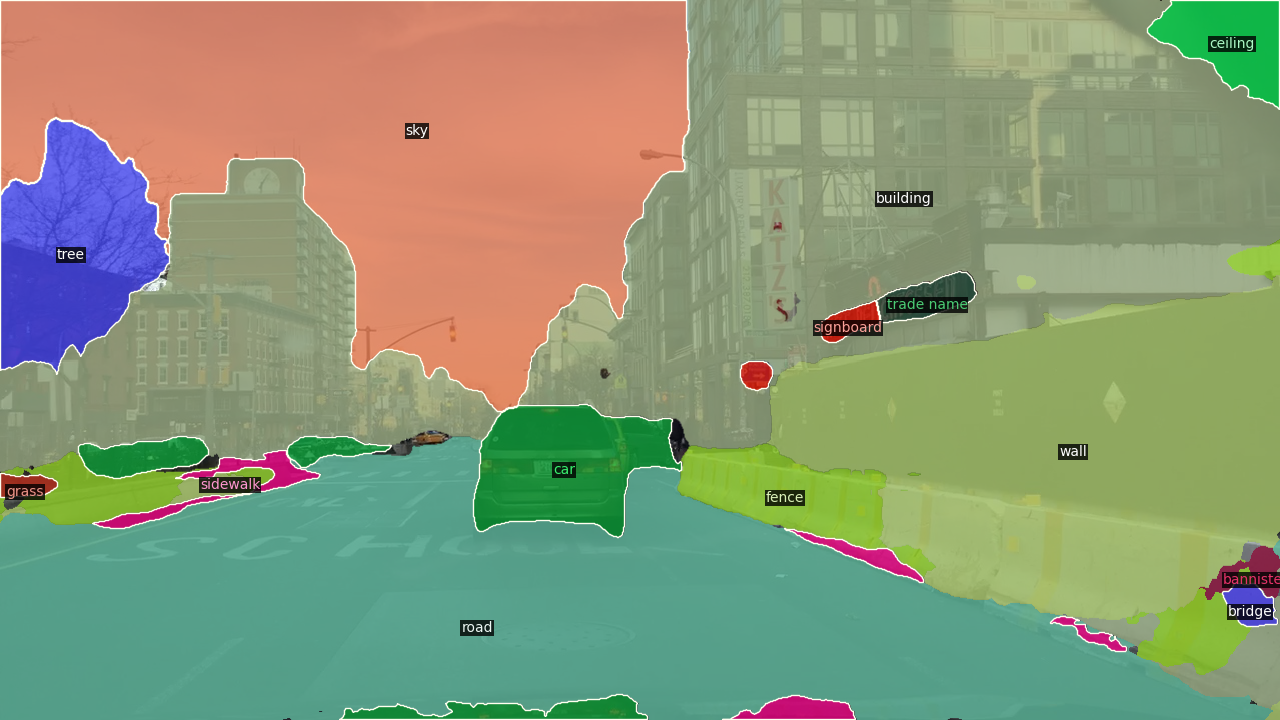} &
    \includegraphics[width=0.16\linewidth]{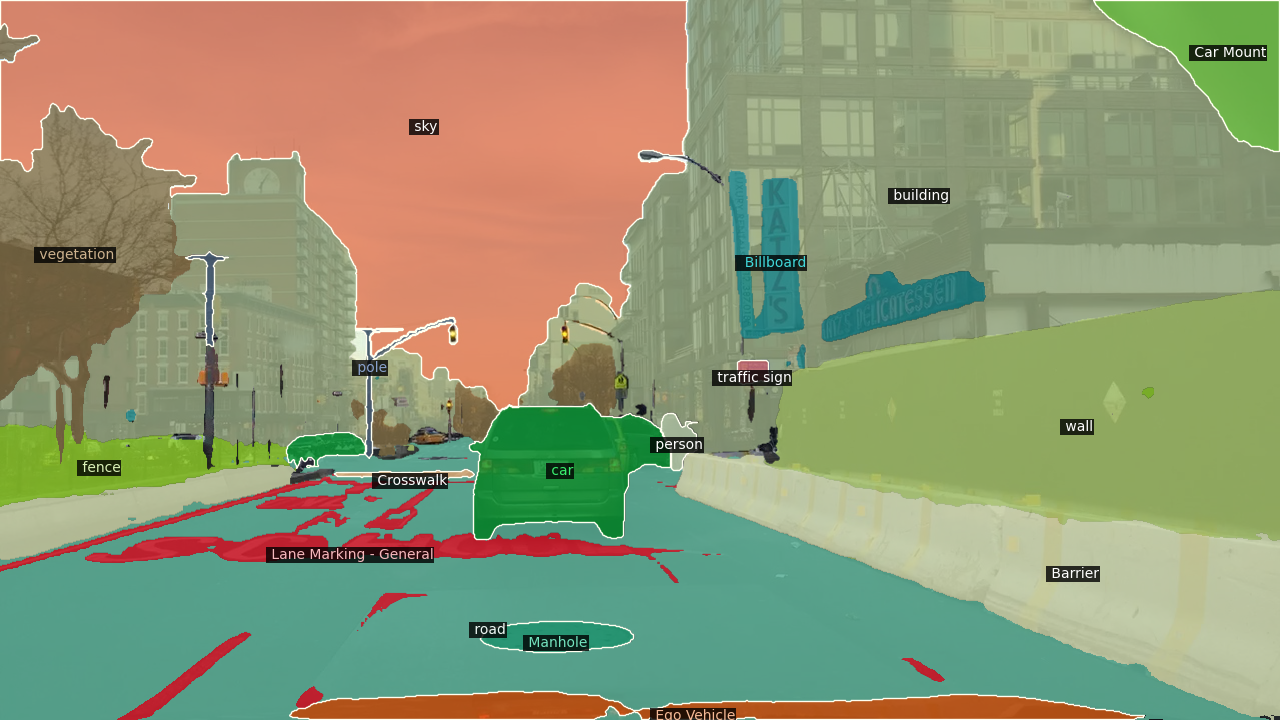} \\

    \rotatebox[origin=l]{90}{\quad MPL} &
    \includegraphics[width=0.16\linewidth]{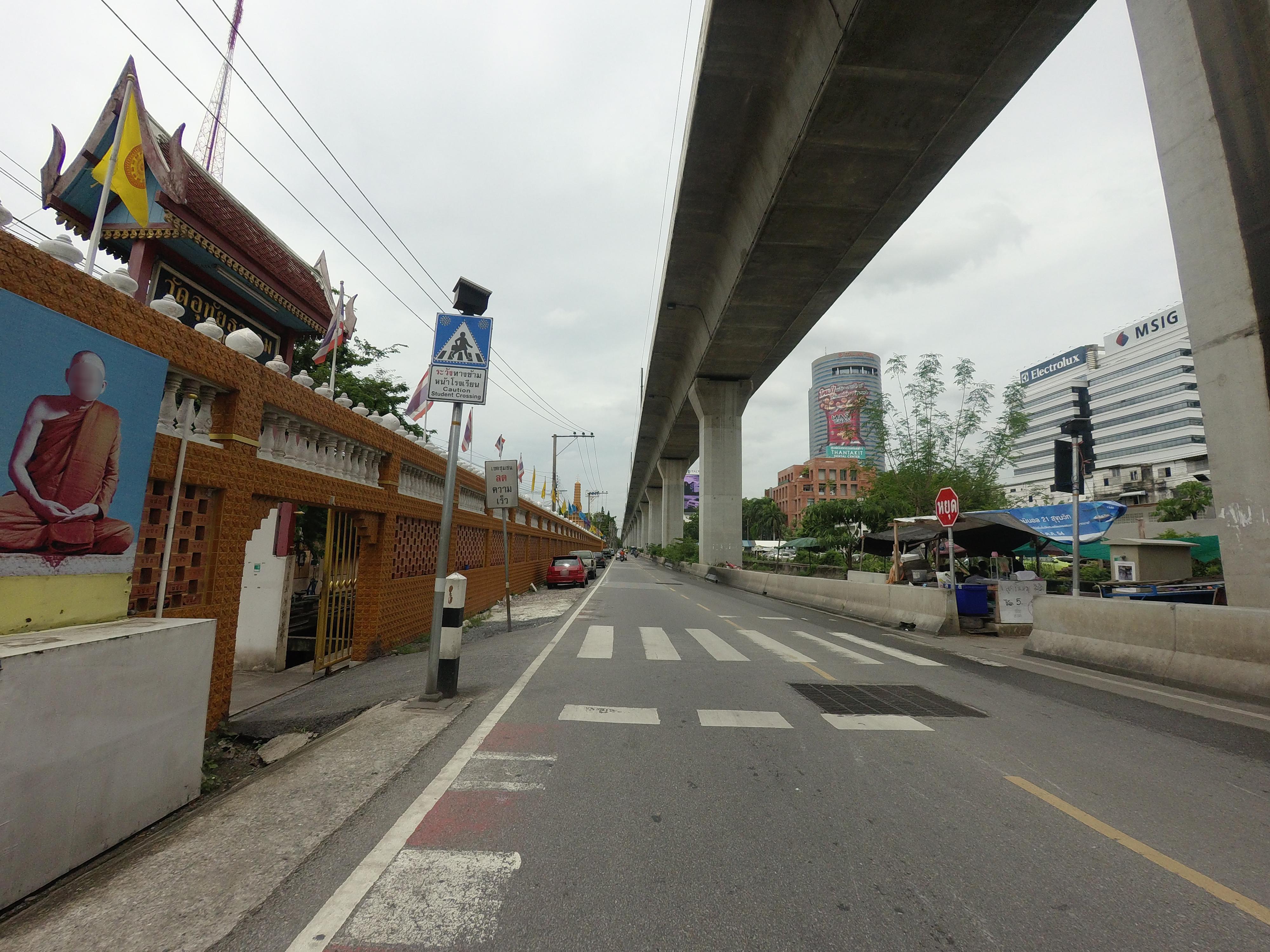} &
    \includegraphics[width=0.16\linewidth]{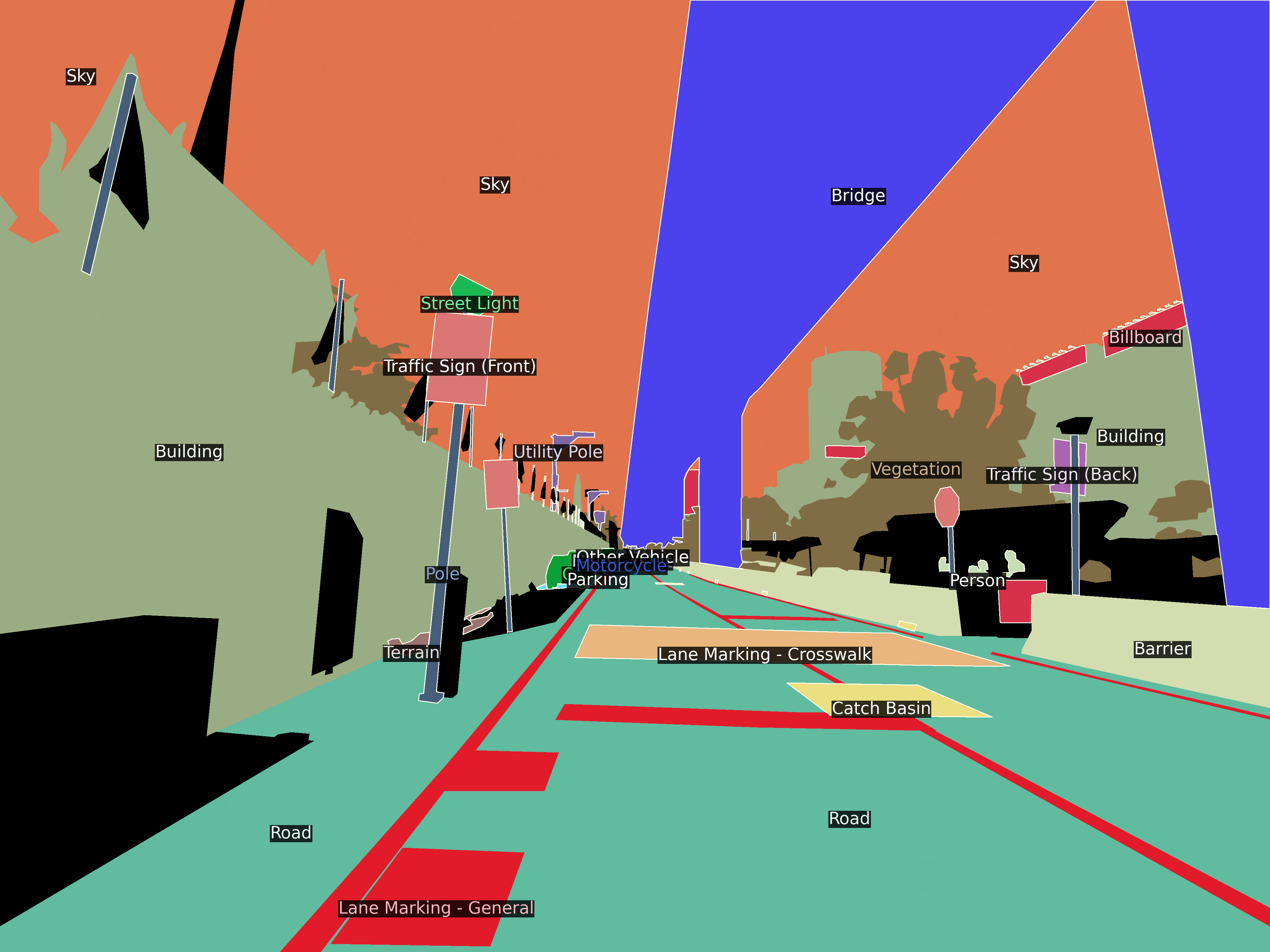} &
    \includegraphics[width=0.16\linewidth]{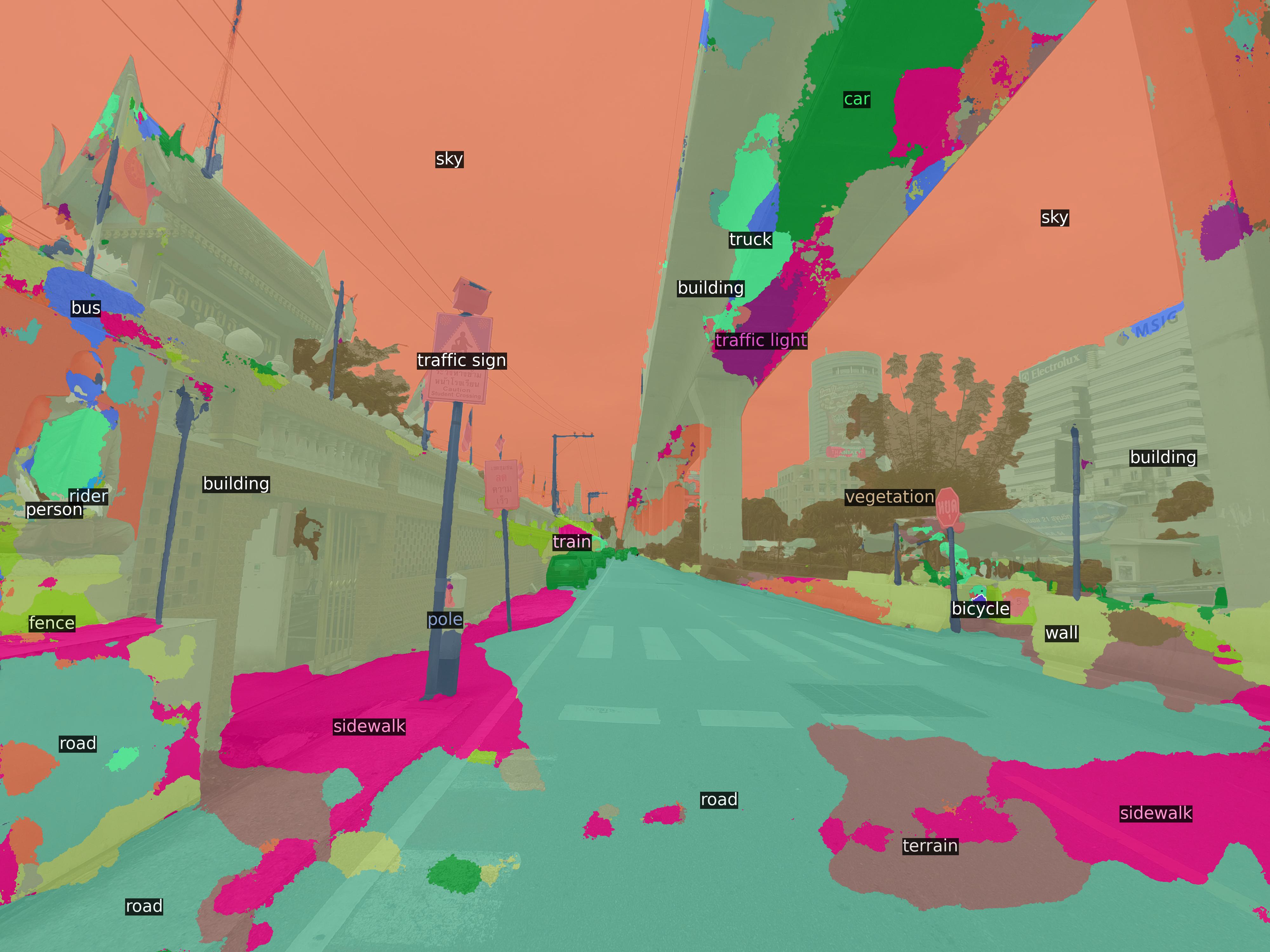} &
    \includegraphics[width=0.16\linewidth]{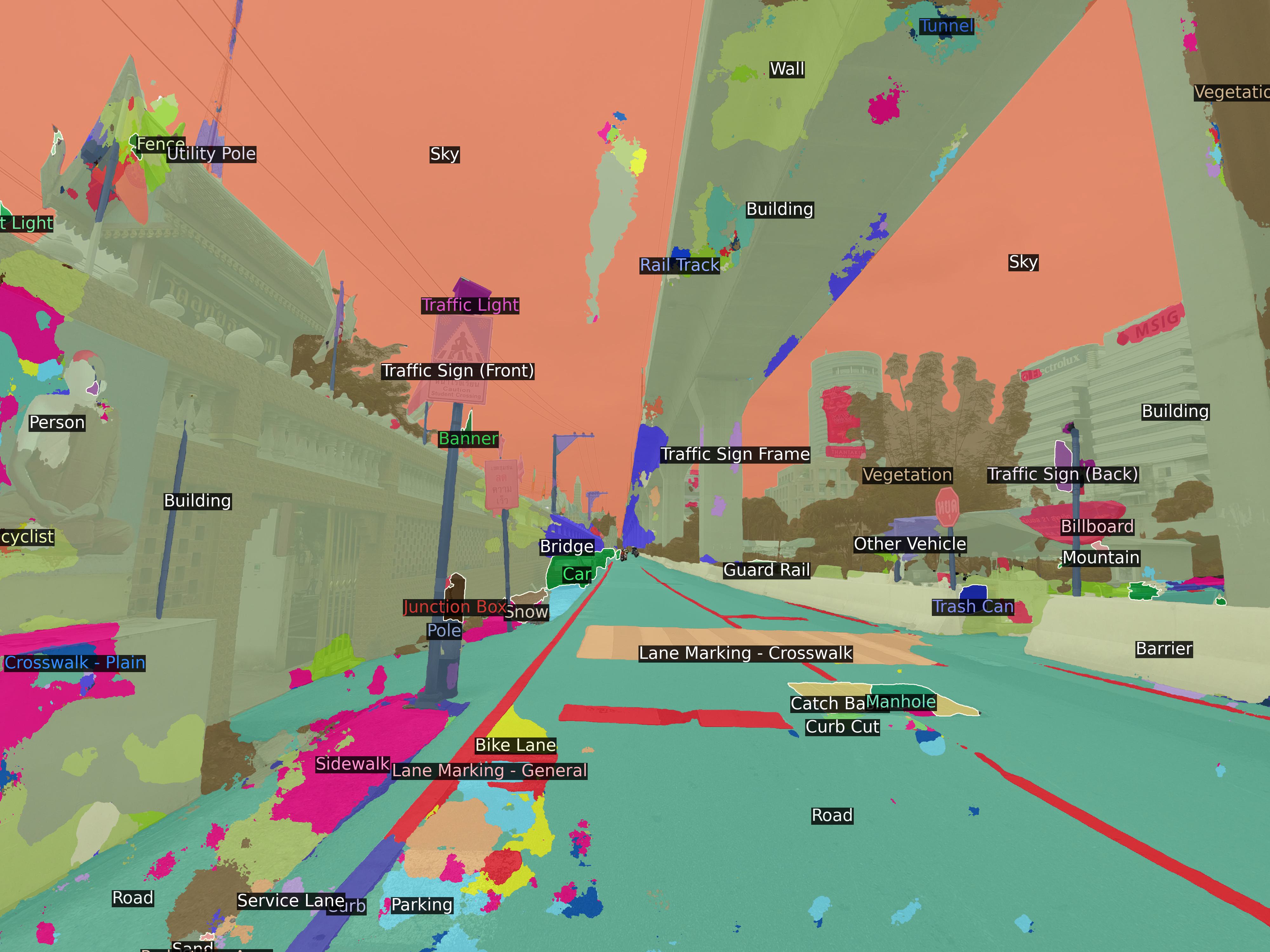} &
    \includegraphics[width=0.16\linewidth]{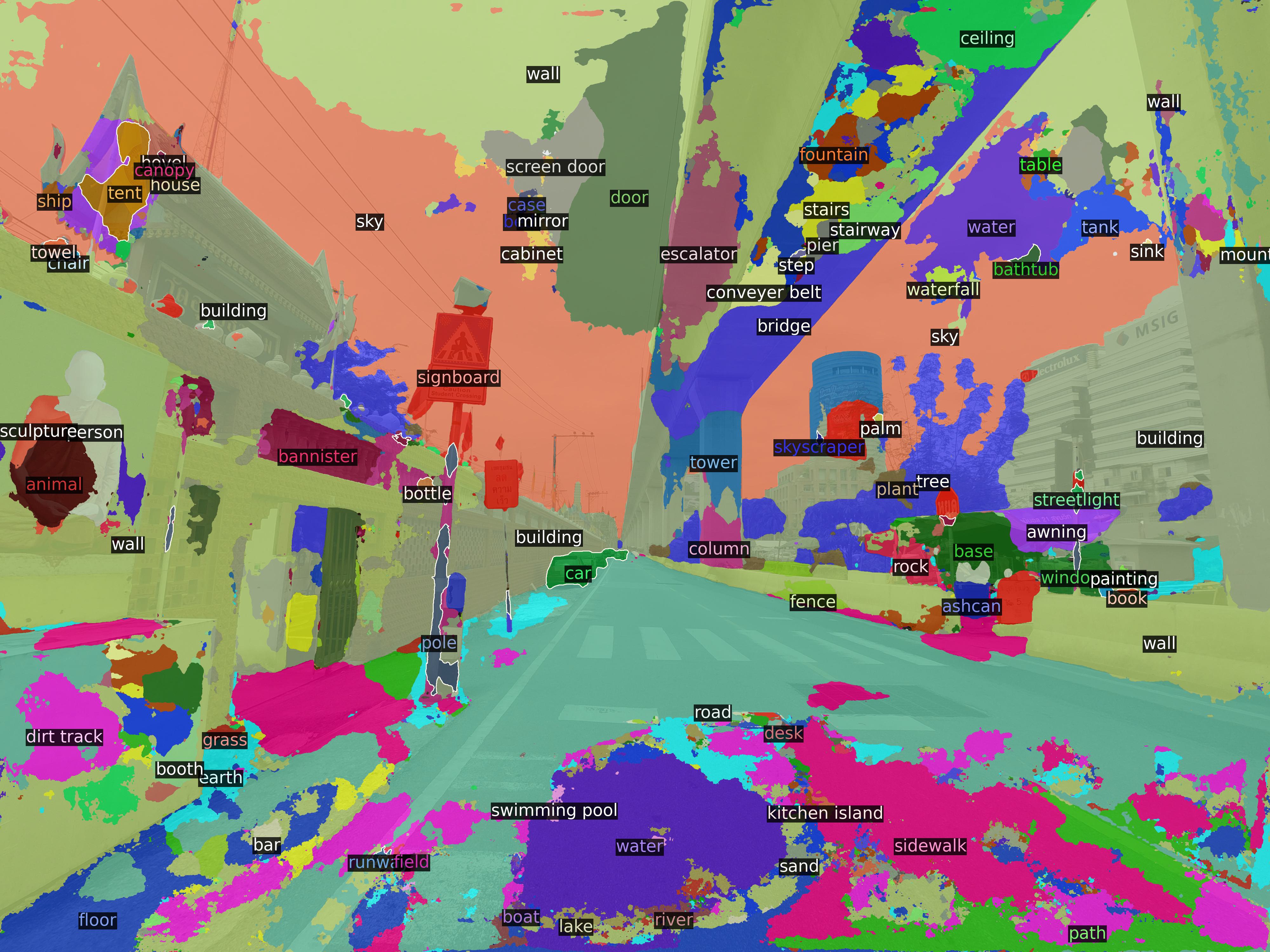} &
    \includegraphics[width=0.16\linewidth]{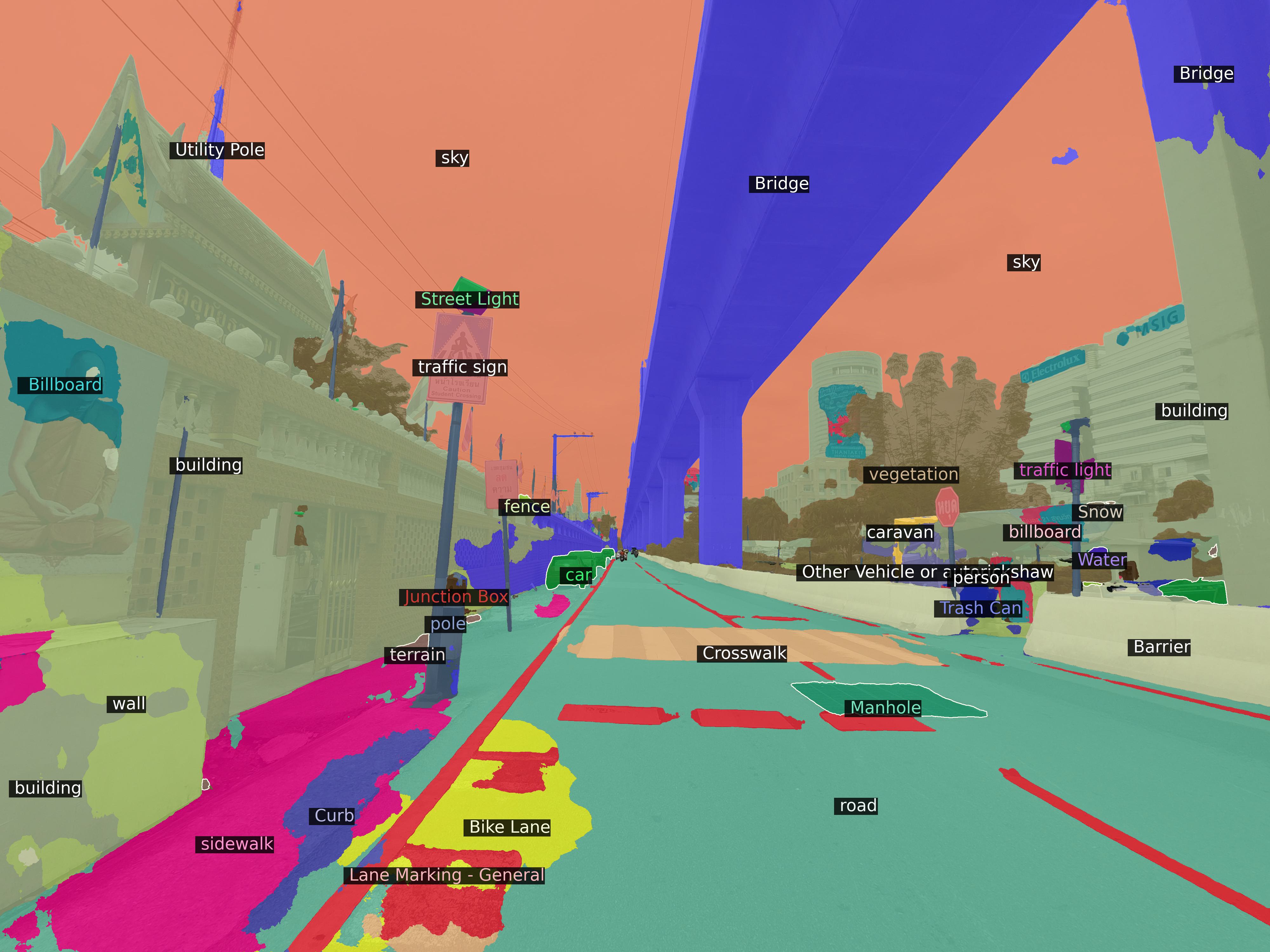} &
     \\

    \rotatebox[origin=l]{90}{\quad ADE} &
    \includegraphics[width=0.16\linewidth]{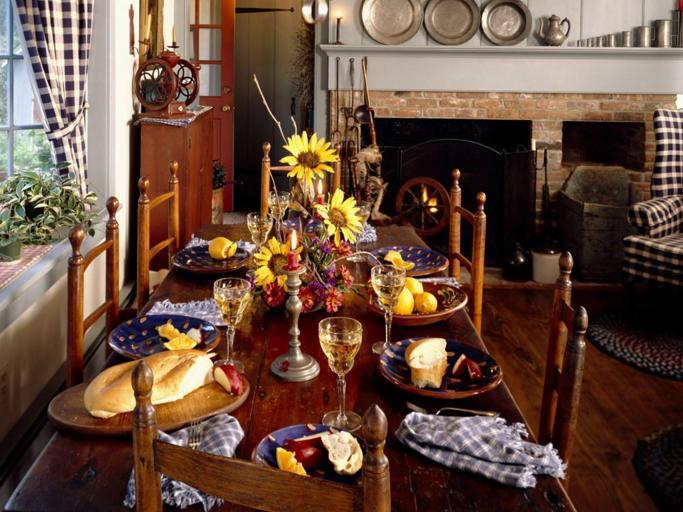} &
    \includegraphics[width=0.16\linewidth]{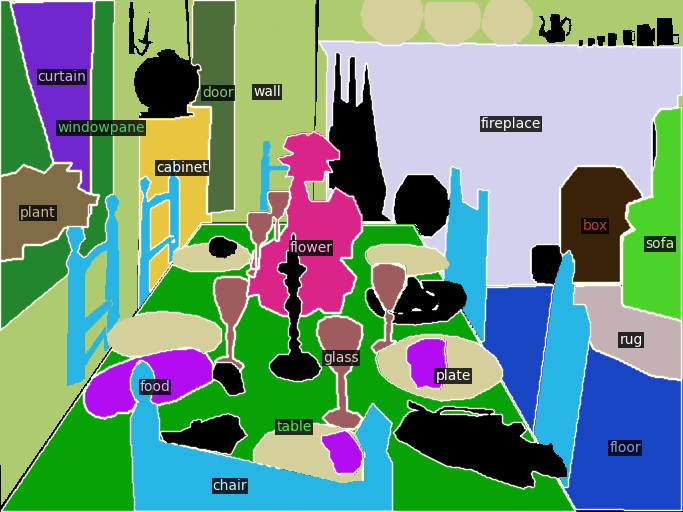} &
    \includegraphics[width=0.16\linewidth]{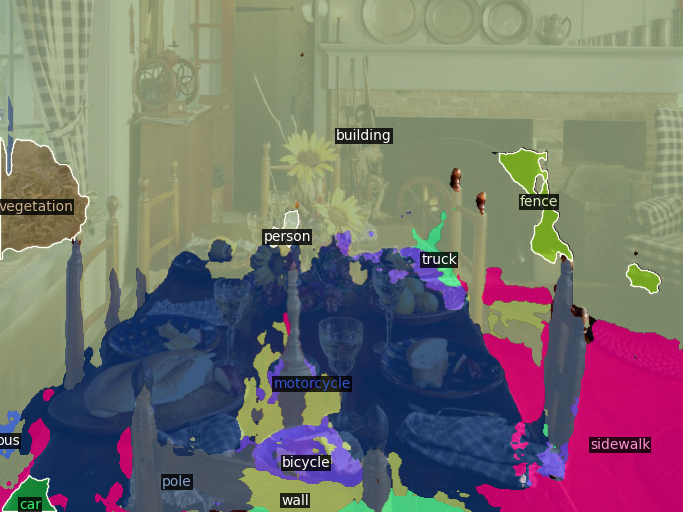} &
    \includegraphics[width=0.16\linewidth]{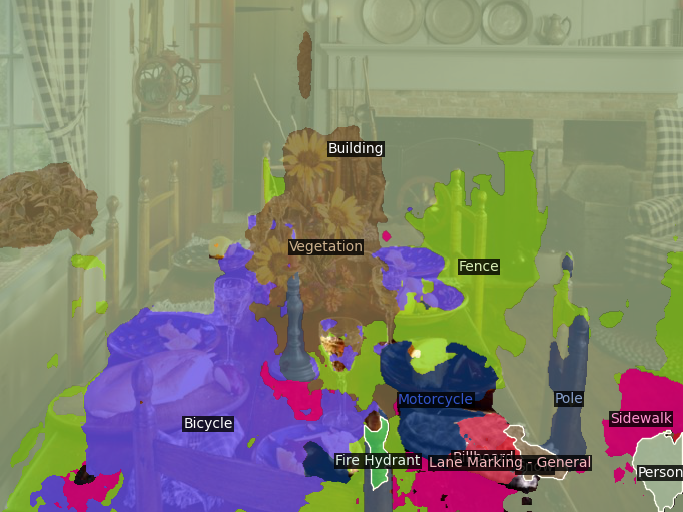} &
    \includegraphics[width=0.16\linewidth]{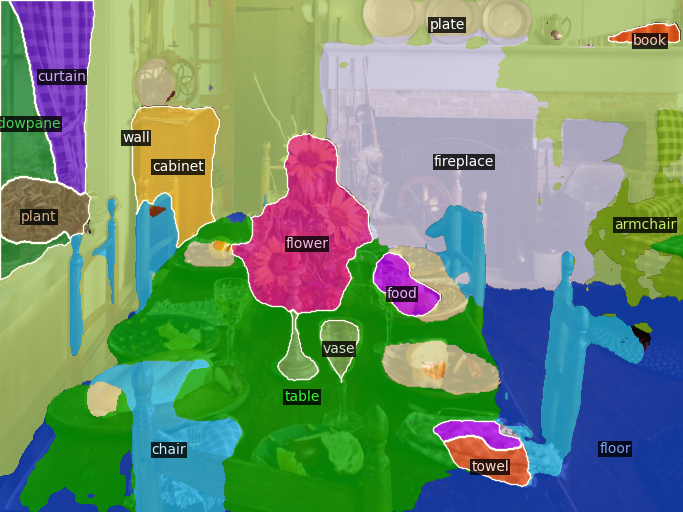} &
    \includegraphics[width=0.16\linewidth]{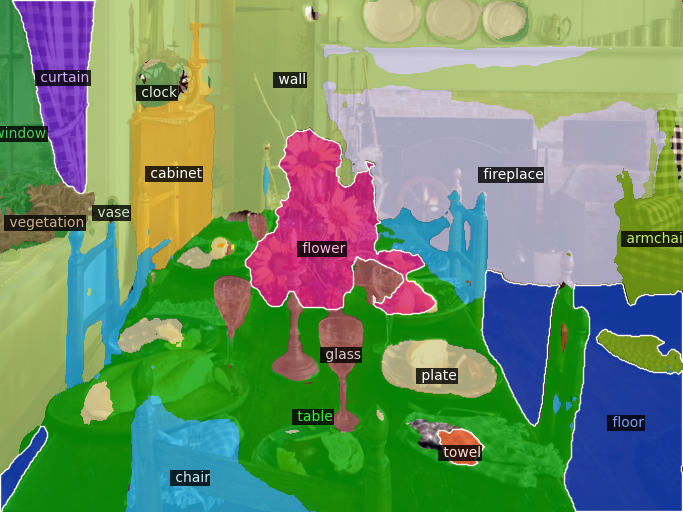} &
     \\
    
  \end{tabular}
  \caption{Visual comparisons with \textit{Single dataset} model on different training datasets.}
  \label{fig4}
\end{figure}

\autoref{fig4} presents the segmentation results on multiple datasets predicted in unified label space. Compared to models trained only on \textit{Single dataset}, our model successfully provides consistent predictions on all datasets. It's worth noting that in the BDD dataset, annotations are not provided for \textit{lane marking}, \textit{crosswalk} and \textit{manhole}, which are only annotated in the Mapillary dataset. Our model successfully integrates the label space of the Mapillary dataset, thereby predicting these classes in the BDD dataset. More results are presented in \autoref{apdx:vis}.

\subsection{Results on WildDash 2 Benchmark.} 
    % The Wilddash \cite{zendel2018wilddash} specifically evaluates the robustness for segmantation model under real-world conditions. Unlike some other benchmarks, Wilddash does not provide a training dataset. Instead, it is intended for testing the robustness of models that have been trained on other datasets, evaluating how well they generalize to real-world scenarios.
    WildDash 2 \cite{zendel2022unifying} provides a benchmark for semantic segmentation, designed to test the robustness of algorithms in real-world driving scenarios. Due to the insufficient number of training samples provided by this dataset, the official recommendation is to use multiple datasets for training. Therefore, this benchmark is well-suited to evaluate the effectiveness of multi-dataset training methods. The WildDash 2 dataset includes negative test cases to challenge the robustness of the model. These negative test cases mainly consist of unconventional driving scenarios, and even non-driving scenarios. Across all pixels within negative test images, a robust model is expected to predict the \textit{void} label for open-set classes and anomalous objects. The WildDash 2 benchmark refers to the metric named Meta Avg mIoU Class, which calculates the mean Intersection over Union for each class by weighting negative and positive test cases according to their occurrence in the benchmark dataset. 
    
    \autoref{tb:wilddash2} presents the current results on the WildDash 2 leaderboard. We present results for zero-shot generalization using our 7ds model. To evaluate in an unseen setting, we map the non-evaluated classes in the unified label space to a \textit{void} label for both positive and negative test frames. To ensure a fair comparison with other works, we also provide evaluation results for models trained using the training datasets from the Robust Vision Challenge 2022\cite{rvc2022}, which include CityScapes, ADE20K, Vistas, VIPER \cite{Richter_2017}, ScanNet, and WildDash 2. The results indicate that our method achieves state-of-the-art performance in both zero-shot and trained settings. Our method exhibits significant performance improvements compared to other methods on negative test cases. We attribute this to the robustness of our model, which has been trained on diverse datasets, enabling it to perform well in unconventional scenarios. Our zero-shot model has been trained on a wider range of datasets compared to the trained model. Therefore, even without training on the WildDash 2 dataset, it achieves similar performance on negative test cases.

\begin{table}[t]
\caption{Performance comparison on WildDash 2 benchmark.}
\label{tb:wilddash2}
\centering
\resizebox{\columnwidth}{!}{
\begin{tabular}{lcccccccc}
\hline
 Model & Venue & Trained & \makecell{Meta Avg \\ mIoU Class} & \makecell{Classic \\ mIoU Class} & \makecell{Classic \\ iIoU Class} & \makecell{Classic \\ mIoU Cat.} & \makecell{Classic \\ iIoU Cat.} & \makecell{Negative \\ mIoU class} \\
\hline
EffPS \cite{mohan2021efficientps} & IJCV 21 & \usym{2717} & 32.2 & 35.7 & 24.4 & 63.8 & 56.0 & 20.4 \\
MSeg \cite{lambert2020mseg} & CVPR 20 & \usym{2717} &  35.2 & 38.7 & 35.4 & 65.1 & 50.7 & 24.7  \\
SeamSeg \cite{porzi2019seamless} & CVPR 19 & \usym{2717} & 37.9 & 41.2 & \textbf{37.2} & 63.1 & \textbf{58.1} & 30.5 \\
UniSeg \cite{kim2022learning} & ECCV 22 & \usym{2717} & 39.4 &  \textbf{41.7} & 35.3 & \textbf{65.8} & 57.4& 34.8 \\
Ours & - & \usym{2717} & \textbf{40.5} & 41.2 & 33.7 & 65.4 & 54.2 & \textbf{43.4}\\
\hline
SNpRN152 \cite{bevandic2020multi} & arXiv 20 & \checkmark & 45.4 &  48.9 & 42.7 & 70.1 & 64.8 & 32.5\\
NLL+ \cite{bevandic2022multi} & WACV 22 & \checkmark & 46.8 &  51.0 & 43.9 & 71.4 & 65.5 & 32.6\\
Uni NLL+ \cite{bevandic2024weakly} & IJCV 24 & \checkmark & 46.9 & 51.6 & 45.9 & 72.8 & 67.5 & 29.0\\
FAN \cite{xiao20221st} & arXiv 22 & \checkmark & 47.5 & 50.8 & 44.0 & \textbf{74.2} & 67.5 & 34.4\\
MIX6D \cite{liu2022empirical} & arXiv 22 & \checkmark & 48.5 & 51.2 & 46.5 & 72.4 & 66.1 & 40.8 \\
Ours & - & \checkmark & \textbf{50.0} & \textbf{52.2} & \textbf{47.5} & 72.4 & \textbf{68.6} & \textbf{44.6} \\
% Ours & - & \checkmark & \textbf{50.2} & \textbf{53.2} & \textbf{49.2} & 73.2 & \textbf{69.5} & \textbf{43.6} \\
\hline
\end{tabular}
}
\end{table}

\begin{table}[t]
\centering
\caption{Comparison of Different Methods of Construct Label Spaces.}
\label{tb:Ablation}
\resizebox{\columnwidth}{!}{
\begin{tabular}{c|c|cccccccc}
\hline

Methods & $|L|$ & CS & MPL & SUN & BDD & IDD & ADE & COCO & Mean
 \\
 \hline
Direct concatenation & 448 & 79.2 & 38.8 & 46.9 & 64.3 & 66.7 & 33.3 & 38.2 & 52.5 \\
 % & \checkmark & \checkmark & \usym{2717} & 79.0 & 37.6 & 46.8 & 64.8 & 67.9 & 30.1 & 34.7 & 51.6 \\
Clustered by text features & 329 & 79.9 & 40.8 & 47.3 & 65.6 & 68.6 & 33.8 & 38.0 & 53.4 \\
Without GNN training & 231 & 79.8 & 39.7 & \textbf{47.7} & \textbf{66.1} & 68.1 & 37.8 & 45.3 & 54.9 \\ 
% Without GNN training & 231 & 80.1 & 42.5 & 47.3 & \textbf{65.7} & 68.6 & 39.2 & 44.5 & 55.4 \\ 
Without GPT's label description  & 226 & 80.2 & 43.4 &  47.2 &  64.7 & \textbf{68.8} & 40.1 & 46.0 & 55.8 \\
Our proposed method & 217 & \textbf{80.7} & \textbf{43.7} & 47.5 & 65.5 & 68.6 & \textbf{42.0} & \textbf{46.7} & \textbf{56.4}   \\
\hline
 % & & & & 76.7 & 34.3 & 43.2 & 61.0 & 64.9 & 15.0 & 11.5 & \\
\end{tabular}
}

\end{table}

\subsection{Ablation Study}
\label{sec:ab study}
To further explore the ability of our GNNs to construct a unified label space, we compared it with four alternative methods. The first method concatenates all dataset label spaces into a single unified space $\bigcup_{i=1}^D L_{i}$. The second approach constructs a unified label space by clustering text features based on cosine similarity using the DBSCAN \cite{bi2012dbscan} algorithm. In the third method, we train a segmentation network using an initial adjacency matrix, as outlined in \autoref{apdx:2}, without incorporating GNN training. The final method involves an ablation experiment, removing the label description module to observe its impact. Experimental results, presented in \autoref{tb:Ablation}, demonstrate that the label space constructed based on GNNs can better assist in the learning of segmentation models. Unlike the first approach, our method optimizes model capacity by focusing on label relationships rather than dataset recognition. Compared to the second approach, our method can differentiate between classes with identical names but differing levels of granularity. By leveraging label descriptions to enrich semantic context, our approach constructs a more refined and functional label space.

\subsection{Exploring the Impact of Training Datasets}
\begin{figure}[t]
    \begin{minipage}{0.65\linewidth}
    \centering
    \includegraphics[width=1.0\linewidth]{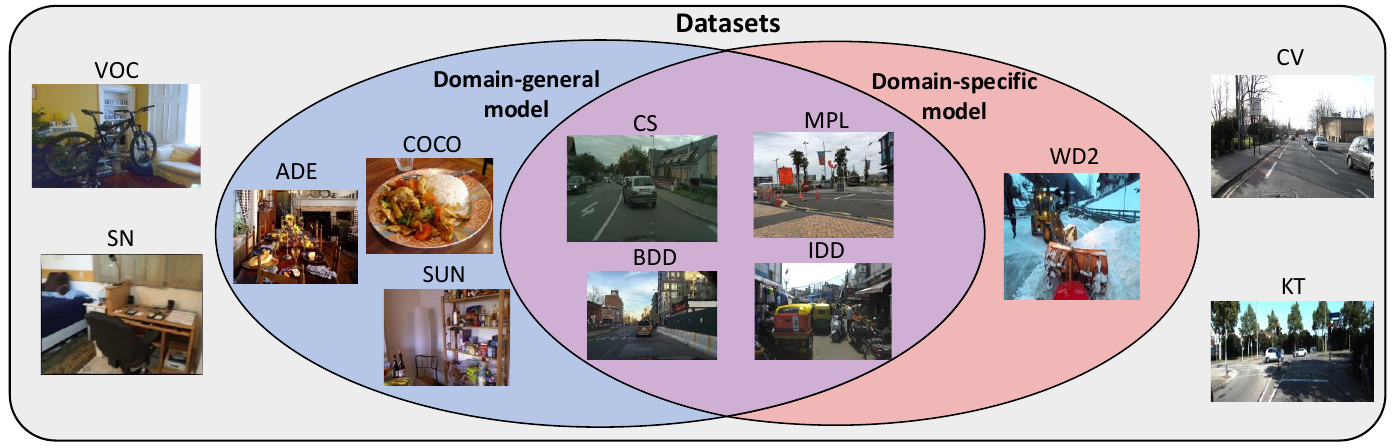}
    \caption{The composition of the training datasets.}
    \label{fig:composition}
    \end{minipage}\hfill
    \begin{minipage}{0.34\linewidth}
        \centering
        \footnotesize
        \captionof{table}{Performance on unseen dataset.}
        \label{tb:both unseen}
        \raisebox{1.7\height}{
        \begin{adjustbox}{height=0.72cm}
        \begin{tabular}{c|cccc}
            \hline
             &  \multicolumn{2}{c}{Driving} & \multicolumn{2}{c}{Non-Driving} \\
            Model & CV & KT & VOC & SN \\
            \hline
            General  & 71.1 & 65.5 & \textbf{69.2} & \textbf{40.6} \\
            Specific & \textbf{74.4} & \textbf{67.4} & 28.7 & 8.1 \\ 
        \end{tabular}
        \end{adjustbox}
        }
    \end{minipage}
\end{figure}

To explore the impact of training dataset selection on model performance, we train a domain-specific model focusing on road driving scenes and a domain-general model on more datasets, as shown in \autoref{fig:composition}. We conduct four sets of comparisons to evaluate the performance across datasets that were trained on both models, trained on one model and not on the other, and not trained on either model. As shown in Table \ref{tb:both unseen} and \autoref{apdx:impact}, the domain-specific model can focus more on learning features specific to the particular scene, resulting in slightly better performance on both trained and unseen driving scene datasets compared to domain-general model. On the other hand, domain-general model trained on more scenes and more data exhibit better generalization performance. Therefore, while it does not lag too far behind in performance on driving scene datasets, it demonstrates overwhelming advantages on other scene datasets.

% The results are shown in the \autoref{tb:impact}, indicating that modes focusing on driving scenes can achieve better performance and generalization on road datasets. However, it falls short in providing predictions for other scenarios, whereas the universal model consistently delivers results.

% \begin{table}
% \centering
% \caption{Impact of Training Datasets}
% \label{tb:impact}
% \footnotesize
% \begin{tabular}{c|cccc|c|ccc|cc}
% \hline
% & \multicolumn{4}{c|}{Both trained} & Universal unseen & \multicolumn{3}{c|}{Road scene unseen} & \multicolumn{2}{c}{Both unseen}\\
% Model & CS & VT &  BDD & IDD & WD2 & SUN & ADE & COCO & CV & VOC\\
% \hline
% Universal & 80.2 & 43.4 & 64.7 & 68.8 & 40.5 & \textbf{47.5} & \textbf{42.0} & \textbf{46.7} & 71.0 & \textbf{69.0}\\
% Road scene & \textbf{82.2} & \textbf{45.7} & \textbf{68.8} & \textbf{71.4} & \textbf{50.2} & 4.4 & 5.4 & 7.3 & \textbf{74.4} & 28.7\\
% \hline
% \end{tabular}
% \end{table}
\definecolor{myblue}{RGB}{181,198,231}
\definecolor{myred}{RGB}{241,182,182}
\definecolor{mypurple}{RGB}{207,175,212}
\definecolor{mygrey}{RGB}{237,237,237}

\subsection{Qualitative Results}
\label{sec:vis}

\begin{figure}[t]
    \centering
    \includegraphics[width=1\linewidth]{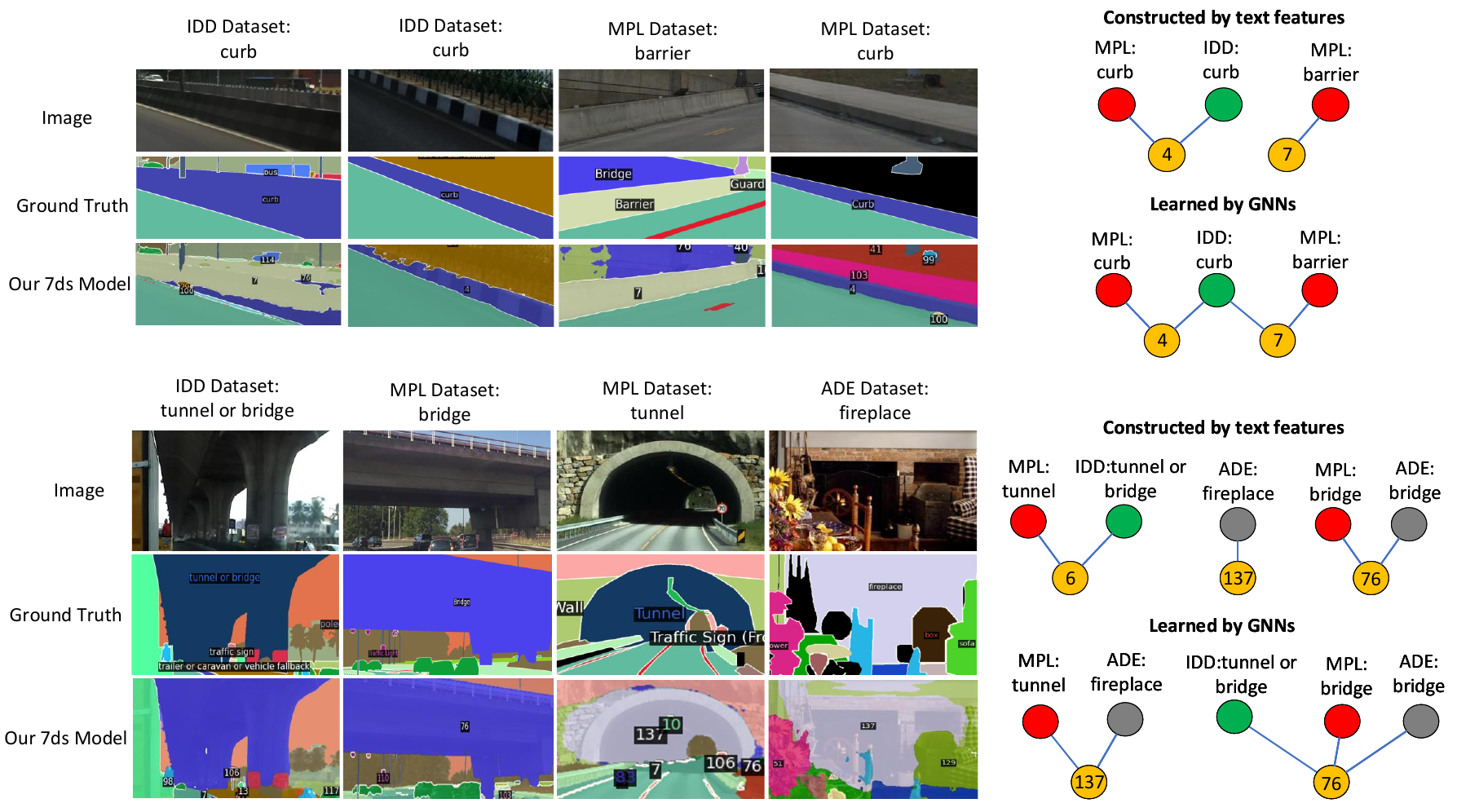}
    \caption{Comparison of unified label space learned by GNNs with constructed by text features.}
    \label{fig:vis}
\end{figure}

\autoref{fig:vis} presents the qualitative analysis results of the label space learned by our model. We compare the label space learned by our model with the label space constructed using text features. The class \textit{curb} in the IDD dataset actually encompasses both the classes \textit{curb} and \textit{barrier} in Mapillary, whereas the constructed label space by text features cannot handle such subclass/superclass relations. In contrast, our GNNs learned label space splits the IDD \textit{curb} into two classes for prediction, effectively handling such label relationships. Similar situations also include the class \textit{tunnel or bridge} in the IDD dataset. However, since the proportion of bridge pixels is orders of magnitude greater than that of tunnels ($10^8$ pixels for \textit{bridge} and $10^4$ pixels for \textit{tunnel}), a more reasonable approach is to merge it with the class \textit{bridge}, as our GNNs have done. Additionally, it is worth noting that due to visual similarities, the Mapillary \textit{tunnel} has been merged with the ADE \textit{fireplace}. This does not actually introduce any conflict because no dataset simultaneously annotates both the \textit{tunnel} and \textit{fireplace}. The model will construct the label space in a way that facilitates its learning process.

\section{Conclusion}
\label{sec:dis}
We propose a novel approach that leverages graph neural networks to construct a unified label space for training semantic segmentation models across multiple datasets. Our method addresses the challenge of label conflicts in multi-dataset semantic segmentation and demonstrates performance improvements across various datasets. The unified label space generated during training, generalizes well to unseen datasets, showcasing the effectiveness of our approach. 

\textbf{Broader Impact.} Our work explores the use of graph neural networks to unify label spaces between datasets, providing a new direction for achieving robust and efficient multi-dataset training. By enabling semantic segmentation models to be trained on multiple datasets with a unified label space, our method can potentially reduce human effort required for re-labeling images and facilitate the expansion of training datasets. This can lead to the development of models that are more universally applicable across various datasets, benefiting a wide range of applications.

\textbf{Limitations.} 
Although our approach does not require manual relabeling efforts, it still relies on fully annotated datasets for training, in contrast to weakly-supervised and unsupervised methods. We aim to explore ways to integrate these alternative methods in future research. Errors in the fully automated construction of a unified label space do present some safety risks for autonomous driving tasks. Therefore, we recommend introducing a manual review mechanism to address these issues. Additionally, using ChatGPT may generate inaccurate label descriptions, which could affect the prediction of label relationships. Therefore, we aim to improve the accuracy of label descriptions by incorporating label descriptions provided by official datasets as prompts.

\section{Acknowledgments}
This work was supported in part by  NSFC Project (62176061),  Shanghai Municipal Science and Technology Major Project (2021SHZDZX0103). The computations in this research were performed using the CFFF platform of Fudan University. 

\bibliographystyle{plain}
\bibliography{neurips_2024}

\begin{thebibliography}{10}

\bibitem{rvc2022}
Robust vision challenge.
\newblock \url{http://www.robustvision.net/index.php}.
\newblock Accessed: 2022-12-02.

\bibitem{awais2023foundational}
Muhammad Awais, Muzammal Naseer, Salman Khan, Rao~Muhammad Anwer, Hisham Cholakkal, Mubarak Shah, Ming-Hsuan Yang, and Fahad~Shahbaz Khan.
\newblock Foundational models defining a new era in vision: A survey and outlook.
\newblock {\em arXiv preprint arXiv:2307.13721}, 2023.

\bibitem{bevandic2020multi}
Petra Bevandi{\'c}, Marin Or{\v{s}}i{\'c}, Ivan Grubi{\v{s}}i{\'c}, Josip {\v{S}}ari{\'c}, and Sini{\v{s}}a {\v{S}}egvi{\'c}.
\newblock Multi-domain semantic segmentation with pyramidal fusion.
\newblock {\em arXiv preprint arXiv:2009.01636}, 2020.

\bibitem{bevandic2022multi}
Petra Bevandi{\'c}, Marin Or{\v{s}}i{\'c}, Ivan Grubi{\v{s}}i{\'c}, Josip {\v{S}}ari{\'c}, and Sini{\v{s}}a {\v{S}}egvi{\'c}.
\newblock Multi-domain semantic segmentation with overlapping labels.
\newblock In {\em Proceedings of the IEEE/CVF Winter Conference on Applications of Computer Vision}, pages 2615--2624, 2022.

\bibitem{bevandic2024weakly}
Petra Bevandi{\'c}, Marin Or{\v{s}}i{\'c}, Josip {\v{S}}ari{\'c}, Ivan Grubi{\v{s}}i{\'c}, and Sini{\v{s}}a {\v{S}}egvi{\'c}.
\newblock Weakly supervised training of universal visual concepts for multi-domain semantic segmentation.
\newblock {\em International Journal of Computer Vision}, pages 1--23, 2024.

\bibitem{bevandic2022automatic}
Petra Bevandi{\'c} and Sini{\v{s}}a {\v{S}}egvi{\'c}.
\newblock Automatic universal taxonomies for multi-domain semantic segmentation.
\newblock {\em British Machine Vision Conference}, 2022.

\bibitem{bi2012dbscan}
FM~Bi, WK~Wang, and L~Chen.
\newblock Dbscan: density-based spatial clustering of applications with noise.
\newblock {\em J. Nanjing Univ}, 48(4):491--498, 2012.

\bibitem{brostow2009semantic}
Gabriel~J Brostow, Julien Fauqueur, and Roberto Cipolla.
\newblock Semantic object classes in video: A high-definition ground truth database.
\newblock {\em Pattern Recognition Letters}, 30(2):88--97, 2009.

\bibitem{cai2020multi}
Lei Cai and Shuiwang Ji.
\newblock A multi-scale approach for graph link prediction.
\newblock In {\em Proceedings of the AAAI conference on artificial intelligence}, pages 3308--3315, 2020.

\bibitem{cai2021line}
Lei Cai, Jundong Li, Jie Wang, and Shuiwang Ji.
\newblock Line graph neural networks for link prediction.
\newblock {\em IEEE Transactions on Pattern Analysis and Machine Intelligence}, 44(9):5103--5113, 2021.

\bibitem{chen2023scaledet}
Yanbei Chen, Manchen Wang, Abhay Mittal, Zhenlin Xu, Paolo Favaro, Joseph Tighe, and Davide Modolo.
\newblock Scaledet: A scalable multi-dataset object detector.
\newblock In {\em Proceedings of the IEEE/CVF Conference on Computer Vision and Pattern Recognition}, pages 7288--7297, 2023.

\bibitem{cid2012proper}
Jes{\'u}s Cid-Sueiro.
\newblock Proper losses for learning from partial labels.
\newblock {\em Advances in neural information processing systems}, 25, 2012.

\bibitem{cordts2016cityscapes}
Marius Cordts, Mohamed Omran, Sebastian Ramos, Timo Rehfeld, Markus Enzweiler, Rodrigo Benenson, Uwe Franke, Stefan Roth, and Bernt Schiele.
\newblock The cityscapes dataset for semantic urban scene understanding.
\newblock In {\em Proceedings of the IEEE conference on computer vision and pattern recognition}, pages 3213--3223, 2016.

\bibitem{dai2017scannet}
Angela Dai, Angel~X Chang, Manolis Savva, Maciej Halber, Thomas Funkhouser, and Matthias Nie{\ss}ner.
\newblock Scannet: Richly-annotated 3d reconstructions of indoor scenes.
\newblock In {\em Proceedings of the IEEE conference on computer vision and pattern recognition}, pages 5828--5839, 2017.

\bibitem{everingham2010pascal}
Mark Everingham, Luc Van~Gool, Christopher~KI Williams, John Winn, and Andrew Zisserman.
\newblock The pascal visual object classes (voc) challenge.
\newblock {\em International journal of computer vision}, 88:303--338, 2010.

\bibitem{geiger2013vision}
Andreas Geiger, Philip Lenz, Christoph Stiller, and Raquel Urtasun.
\newblock Vision meets robotics: The kitti dataset.
\newblock {\em The International Journal of Robotics Research}, 32(11):1231--1237, 2013.

\bibitem{hamilton2017inductive}
Will Hamilton, Zhitao Ying, and Jure Leskovec.
\newblock Inductive representation learning on large graphs.
\newblock {\em Advances in neural information processing systems}, 30, 2017.

\bibitem{hao2020p}
Junheng Hao, Tong Zhao, Jin Li, Xin~Luna Dong, Christos Faloutsos, Yizhou Sun, and Wei Wang.
\newblock P-companion: A principled framework for diversified complementary product recommendation.
\newblock In {\em Proceedings of the 29th ACM International Conference on Information \& Knowledge Management}, pages 2517--2524, 2020.

\bibitem{kim2022learning}
Dongwan Kim, Yi-Hsuan Tsai, Yumin Suh, Masoud Faraki, Sparsh Garg, Manmohan Chandraker, and Bohyung Han.
\newblock Learning semantic segmentation from multiple datasets with label shifts.
\newblock In {\em European Conference on Computer Vision}, pages 20--36. Springer, 2022.

\bibitem{kipf2016semi}
Thomas~N Kipf and Max Welling.
\newblock Semi-supervised classification with graph convolutional networks.
\newblock In {\em International Conference on Learning Representations}, 2016.

\bibitem{kirillov2023segment}
Alexander Kirillov, Eric Mintun, Nikhila Ravi, Hanzi Mao, Chloe Rolland, Laura Gustafson, Tete Xiao, Spencer Whitehead, Alexander~C Berg, Wan-Yen Lo, et~al.
\newblock Segment anything.
\newblock {\em arXiv preprint arXiv:2304.02643}, 2023.

\bibitem{kumar2020link}
Ajay Kumar, Shashank~Sheshar Singh, Kuldeep Singh, and Bhaskar Biswas.
\newblock Link prediction techniques, applications, and performance: A survey.
\newblock {\em Physica A: Statistical Mechanics and its Applications}, 553:124289, 2020.

\bibitem{lambert2020mseg}
John Lambert, Zhuang Liu, Ozan Sener, James Hays, and Vladlen Koltun.
\newblock Mseg: A composite dataset for multi-domain semantic segmentation.
\newblock In {\em Proceedings of the IEEE/CVF conference on computer vision and pattern recognition}, pages 2879--2888, 2020.

\bibitem{lin2024universal}
Feng Lin, Wenze Hu, Yaowei Wang, Yonghong Tian, Guangming Lu, Fanglin Chen, Yong Xu, and Xiaoyu Wang.
\newblock Universal object detection with large vision model.
\newblock {\em International Journal of Computer Vision}, 132(4):1258--1276, 2024.

\bibitem{lin2014microsoft}
Tsung-Yi Lin, Michael Maire, Serge Belongie, James Hays, Pietro Perona, Deva Ramanan, Piotr Doll{\'a}r, and C~Lawrence Zitnick.
\newblock Microsoft coco: Common objects in context.
\newblock In {\em Computer Vision--ECCV 2014: 13th European Conference, Zurich, Switzerland, September 6-12, 2014, Proceedings, Part V 13}, pages 740--755. Springer, 2014.

\bibitem{liu2022empirical}
Yajie Liu, Pu~Ge, Qingjie Liu, Shichao Fan, and Yunhong Wang.
\newblock An empirical study on multi-domain robust semantic segmentation.
\newblock {\em arXiv preprint arXiv:2212.04221}, 2022.

\bibitem{loshchilov2017decoupled}
Ilya Loshchilov and Frank Hutter.
\newblock Decoupled weight decay regularization.
\newblock {\em arXiv preprint arXiv:1711.05101}, 2017.

\bibitem{masaki2021multi}
Shota Masaki, Tsubasa Hirakawa, Takayoshi Yamashita, and Hironobu Fujiyoshi.
\newblock Multi-domain semantic-segmentation using multi-head model.
\newblock In {\em 2021 IEEE International Intelligent Transportation Systems Conference (ITSC)}, pages 2802--2807. IEEE, 2021.

\bibitem{meletis2023training}
Panagiotis Meletis and Gijs Dubbelman.
\newblock Training semantic segmentation on heterogeneous datasets.
\newblock {\em arXiv preprint arXiv:2301.07634}, 2023.

\bibitem{meng2023detection}
Lingchen Meng, Xiyang Dai, Yinpeng Chen, Pengchuan Zhang, Dongdong Chen, Mengchen Liu, Jianfeng Wang, Zuxuan Wu, Lu~Yuan, and Yu-Gang Jiang.
\newblock Detection hub: Unifying object detection datasets via query adaptation on language embedding.
\newblock In {\em Proceedings of the IEEE/CVF Conference on Computer Vision and Pattern Recognition}, pages 11402--11411, 2023.

\bibitem{mohan2021efficientps}
Rohit Mohan and Abhinav Valada.
\newblock Efficientps: Efficient panoptic segmentation.
\newblock {\em International Journal of Computer Vision}, 129(5):1551--1579, 2021.

\bibitem{mottaghi2014role}
Roozbeh Mottaghi, Xianjie Chen, Xiaobai Liu, Nam-Gyu Cho, Seong-Whan Lee, Sanja Fidler, Raquel Urtasun, and Alan Yuille.
\newblock The role of context for object detection and semantic segmentation in the wild.
\newblock In {\em Proceedings of the IEEE conference on computer vision and pattern recognition}, pages 891--898, 2014.

\bibitem{neuhold2017mapillary}
Gerhard Neuhold, Tobias Ollmann, Samuel Rota~Bulo, and Peter Kontschieder.
\newblock The mapillary vistas dataset for semantic understanding of street scenes.
\newblock In {\em Proceedings of the IEEE international conference on computer vision}, pages 4990--4999, 2017.

\bibitem{porzi2019seamless}
Lorenzo Porzi, Samuel~Rota Bulo, Aleksander Colovic, and Peter Kontschieder.
\newblock Seamless scene segmentation.
\newblock In {\em Proceedings of the IEEE/CVF conference on computer vision and pattern recognition}, pages 8277--8286, 2019.

\bibitem{radford2021learning}
Alec Radford, Jong~Wook Kim, Chris Hallacy, Aditya Ramesh, Gabriel Goh, Sandhini Agarwal, Girish Sastry, Amanda Askell, Pamela Mishkin, Jack Clark, et~al.
\newblock Learning transferable visual models from natural language supervision.
\newblock In {\em International conference on machine learning}, pages 8748--8763. PMLR, 2021.

\bibitem{Richter_2017}
Stephan~R. Richter, Zeeshan Hayder, and Vladlen Koltun.
\newblock Playing for benchmarks.
\newblock In {\em {IEEE} International Conference on Computer Vision, {ICCV} 2017, Venice, Italy, October 22-29, 2017}, pages 2232--2241, 2017.

\bibitem{song2015sun}
Shuran Song, Samuel~P Lichtenberg, and Jianxiong Xiao.
\newblock Sun rgb-d: A rgb-d scene understanding benchmark suite.
\newblock In {\em Proceedings of the IEEE conference on computer vision and pattern recognition}, pages 567--576, 2015.

\bibitem{sun2019high}
Ke~Sun, Yang Zhao, Borui Jiang, Tianheng Cheng, Bin Xiao, Dong Liu, Yadong Mu, Xinggang Wang, Wenyu Liu, and Jingdong Wang.
\newblock High-resolution representations for labeling pixels and regions.
\newblock {\em arXiv preprint arXiv:1904.04514}, 2019.

\bibitem{toldo2021unsupervised}
Marco Toldo, Umberto Michieli, and Pietro Zanuttigh.
\newblock Unsupervised domain adaptation in semantic segmentation via orthogonal and clustered embeddings.
\newblock In {\em Proceedings of the IEEE/CVF Winter conference on Applications of Computer Vision}, pages 1358--1368, 2021.

\bibitem{touvron2023llama}
Hugo Touvron, Louis Martin, Kevin Stone, Peter Albert, Amjad Almahairi, Yasmine Babaei, Nikolay Bashlykov, Soumya Batra, Prajjwal Bhargava, Shruti Bhosale, et~al.
\newblock Llama 2: Open foundation and fine-tuned chat models.
\newblock {\em arXiv preprint arXiv:2307.09288}, 2023.

\bibitem{uijlings2022missing}
Jasper Uijlings, Thomas Mensink, and Vittorio Ferrari.
\newblock The missing link: Finding label relations across datasets.
\newblock In {\em European Conference on Computer Vision}, pages 540--556. Springer, 2022.

\bibitem{varma2019idd}
Girish Varma, Anbumani Subramanian, Anoop Namboodiri, Manmohan Chandraker, and CV~Jawahar.
\newblock Idd: A dataset for exploring problems of autonomous navigation in unconstrained environments.
\newblock In {\em 2019 IEEE Winter Conference on Applications of Computer Vision (WACV)}, pages 1743--1751. IEEE, 2019.

\bibitem{velivckovic2023everything}
Petar Veli{\v{c}}kovi{\'c}.
\newblock Everything is connected: Graph neural networks.
\newblock {\em Current Opinion in Structural Biology}, 79:102538, 2023.

\bibitem{wang2022cross}
Li~Wang, Dong Li, Han Liu, Jinzhang Peng, Lu~Tian, and Yi~Shan.
\newblock Cross-dataset collaborative learning for semantic segmentation in autonomous driving.
\newblock In {\em Proceedings of the AAAI Conference on Artificial Intelligence}, pages 2487--2494, 2022.

\bibitem{xiao20221st}
Junfei Xiao, Zhichao Xu, Shiyi Lan, Zhiding Yu, Alan Yuille, and Anima Anandkumar.
\newblock 1st place solution of the robust vision challenge 2022 semantic segmentation track.
\newblock {\em arXiv preprint arXiv:2210.12852}, 2022.

\bibitem{yu2018bdd100k}
Fisher Yu, Wenqi Xian, Yingying Chen, Fangchen Liu, Mike Liao, Vashisht Madhavan, Trevor Darrell, et~al.
\newblock Bdd100k: A diverse driving video database with scalable annotation tooling.
\newblock {\em arXiv preprint arXiv:1805.04687}, 2(5):6, 2018.

\bibitem{zendel2022unifying}
Oliver Zendel, Matthias Sch{\"o}rghuber, Bernhard Rainer, Markus Murschitz, and Csaba Beleznai.
\newblock Unifying panoptic segmentation for autonomous driving.
\newblock In {\em Proceedings of the IEEE/CVF Conference on Computer Vision and Pattern Recognition}, pages 21351--21360, 2022.

\bibitem{zhang2017disambiguation}
Min-Ling Zhang, Fei Yu, and Cai-Zhi Tang.
\newblock Disambiguation-free partial label learning.
\newblock {\em IEEE Transactions on Knowledge and Data Engineering}, 29(10):2155--2167, 2017.

\bibitem{zhang2018link}
Muhan Zhang and Yixin Chen.
\newblock Link prediction based on graph neural networks.
\newblock {\em Advances in neural information processing systems}, 31, 2018.

\bibitem{zhong2023comprehensive}
Lingfeng Zhong, Jia Wu, Qian Li, Hao Peng, and Xindong Wu.
\newblock A comprehensive survey on automatic knowledge graph construction.
\newblock {\em arXiv preprint arXiv:2302.05019}, 2023.

\bibitem{zhou2019semantic}
Bolei Zhou, Hang Zhao, Xavier Puig, Tete Xiao, Sanja Fidler, Adela Barriuso, and Antonio Torralba.
\newblock Semantic understanding of scenes through the ade20k dataset.
\newblock {\em International Journal of Computer Vision}, 127:302--321, 2019.

\bibitem{zhou2022lmseg}
Qiang Zhou, Yuang Liu, Chaohui Yu, Jingliang Li, Zhibin Wang, and Fan Wang.
\newblock Lmseg: Language-guided multi-dataset segmentation.
\newblock In {\em The Eleventh International Conference on Learning Representations}, 2022.

\bibitem{zhou2022simple}
Xingyi Zhou, Vladlen Koltun, and Philipp Kr{\"a}henb{\"u}hl.
\newblock Simple multi-dataset detection.
\newblock In {\em Proceedings of the IEEE/CVF Conference on Computer Vision and Pattern Recognition}, pages 7571--7580, 2022.

\bibitem{zhu2020topology}
Guangming Zhu, Liang Zhang, Hongsheng Li, Peiyi Shen, Syed Afaq~Ali Shah, and Mohammed Bennamoun.
\newblock Topology-learnable graph convolution for skeleton-based action recognition.
\newblock {\em Pattern Recognition Letters}, 135:286--292, 2020.

\end{thebibliography}
%%%%%%%%%%%%%%%%%%%%%%%%%%%%%%%%%%%%%%%%%%%%%%%%%%%%%%%%%%%%
\newpage
\appendix

\section{Training Strategy}
\label{apdx:train}
The training process is divided into three distinct stages: Multi-SegHead Training Stage, Alternating GNNs Training Stage and Alternating SegNet Training Stage, as shown in \autoref{algo:1}. Initially, we commence with the Multi-SegHead Training Stage for 100k iterations. After completing Multi-SegHead Training Stage, we discard the multi-heads and initialize the UniSegHead for predicting class probabilities during subsequent training phases. Subsequently, we alternate between the Alternating GNNs Training Stage and Alternating SegNet Training Stage for a total of three cycles, each lasting 20k iterations. Notably, we prioritize the Alternating GNNs Training stage at the beginning of each cycle.

\textbf{Multi-SegHead Training Stage} aims to equip the segmentation network with fundamental segmentation capabilities. By commencing with a Multi-SegHead model, we provide a robust starting point for the subsequent stages, enhancing both performance and training efficiency. During this stage, the network is trained on multiple datasets using specific segmentation heads for each dataset. Each segmentation head comprises feature weights $\mathbf{W}_{i}$ of dimension $C \times |L_i|$, where $|L_i|$ represents the number of classes of dataset $i$. Pixel embedding $\mathcal{P}$ is multiplied by the $i$-th feature weight to make predictions within the $i$-th dataset's label space. This facilitates the utilization of dataset-specific annotations and cross-entropy loss for network training.
\begin{equation}
\mathcal{L}_{m}(\mathbf{y}, \mathbf{s}) = -\sum^{|L_i|}_{c=1} y_{c} \log(\operatorname{softmax}(\mathbf{W}_{i}\textbf{p})_c).
\label{eq:MHloss}
\end{equation}
\textbf{Alternating GNNs Training Stage.} Training SegNet and GNNs simultaneously is challenging and impractical because training the segmentation network requires a discrete label mapping, while training the GNNs necessitates a continuous and differentiable adjacency matrix. Hence, we alternately training the GNNs and the segmentation network. The Alternating GNNs Training Stage focuses on optimizing a unified label space. During this stage, the GNNs are trained while keeping the segmentation network frozen. When mapping the prediction from the unified label space to the dataset label space, we directly multiply the adjacency matrix $\mathbf{M}_a$. We calculate the loss according to \autoref{eq:l_GNN}. 

% In spired by \cite{zhou2022simple}, we determine the number of unified label nodes and initialize the adjacency matrix $M_a$ based on cross-dataset evaluation outcomes, as detailed in \autoref{apdx:2}. While this approach primarily focuses on merging identical classes, it lays the groundwork for our Graph Neural Networks to refine a cohesive unified label space further.

% Specifically, we leverage all segment heads of Multi-Seghead model to evaluate a particular dataset, discerning co-occurrence patterns across different label spaces. For instance, frequent classification of \textit{potted plant} from the COCO dataset as \textit{vegetation} in the Cityscapes dataset suggests merging these labels into a unified node. To address the multitude of potential candidates, we employ a minimization strategy for a predefined loss function to derive universal taxonomies, detailed in \autoref{apdx:2}. 

\textbf{Alternating SegNet Training Stage.} During this stage, the GNNs remain frozen while the segmentation network is trained. We will use the technique detailed in \autoref{apdx:1} to convert the continuous adjacency matrix into a discrete label mappings that satisfies the constraints, and then use it to compute the cross-entropy loss function and train according to \autoref{eq:celoss}. During the last Alternating SegNet training stage, we extend the training procedure to 100k iterations and drop the entire GNNs but retain the dataset label mappings $\{\mathbf{M}_{i}, i=1,...,K \}$. The unified label embedding $\mathbf{X}_{u}$ in this stage is no longer keep frozen and trained with the SegNet. In the latter part of this stage, we will use training dataset to evaluate the model's performance. Those links between the unified label space and dataset-specific label space which is not activated during the evaluation will be removed. After removing the unused connections, we continue training with the new label mappings. And thus, we obtain the final model. When applying the final model to predict in the wild, we could manually assign each unified label node a class name accord to their prediction and semantics.

\newcommand{\SWITCH}[1]{\State \textbf{switch} #1 \textbf{do }}
\newcommand{\ENDSWITCH}{\State \textbf{end switch}}
\newcommand{\CASE}[1]{\State \textbf{case} #1\textbf{: } }
\newcommand{\ENDCASE}{\State \textbf{end case}}
\newcommand{\DEFAULT}{\textbf{default:} }

\begin{algorithm}[t]
\renewcommand{\algorithmicrequire}{\textbf{Input:}}
\renewcommand{\algorithmicensure}{\textbf{Output:}}
\caption{The training pipeline of our model}
\label{algo:1}

\begin{algorithmic}

\State \textbf{Input:} the number of datasets $K$, the Segmentation Network $SegNet$, the Graph Neural Networks $GNNs$, Multi-head iters $I_p$, GNNs training iters $I_g$ and SegNet training iters $I_s$
\State Stage = Multi-head stage; iter = 0
\For{a multi-datasets sampled mini-batch $\{x_i, y_i\}^K_{i=1}$}
    % \State $\{\mathcal{V}^{(i)}\}^D_{i=1} = \{SegNet(x^{(i)})\}^D_{i=1}$
    % \State $\mathcal{X}^{n} = GNNs(\mathcal{X}^{0}, M_a)$
    % \State $\mathcal{X}^{n}_u, \{\mathcal{X}^{n}_{(i)}\}^D_{i=1} = \mathcal{X}^{n}$ \textcolor{gray}{\# Separate $\mathcal{X}^{n}$ into unify and dataset-specific classifier}

    % \SWITCH {Stage}
    \State \textbf{if} Stage == \textcolor{purple}{Multi-SegHead Training}:
        % \State $\mathcal{L}_p = \frac{1}{K} \sum^D_{i=1} \lambda^{(i)} \sum_{K} y_k^{(i)} log(\sigma(W^{(i)}\mathcal{V}^{(i)}_{k}))$
        \State \qquad Calculate $\mathcal{L}_m$ by \autoref{eq:MHloss}
        \State \qquad Update $SegNet$ to minimize $\mathcal{L}_{m}$
        \State \qquad \textbf{if} iter++ > $I_p$: Replace Multi-SegHead with UniSegHead; Stage = \textcolor{blue}{GNNs training}; iter = 0
    \State \textbf{if} Stage == \textcolor{blue}{GNNs training}:
        % \State $\mathcal{L}_{ce} = \frac{1}{K} \sum^D_{i=1} \lambda^{(i)} \sum_{K} y_k^{(i)} log(\sigma(M^{(i)}\mathcal{X}^n_u\mathcal{V}^{(i)}_{k}))$
        % \State $\mathcal{L}_{aux} = \frac{1}{K} \sum^D_{i=1} \lambda^{(i)} \sum_{K} y_k ^{(i)}log(\sigma(M^{(i)}\mathcal{X}^n_{(i)}\mathcal{V}^{(i)}_{k}))$
        % \State $\mathcal{L}_{orth} = \frac{1}{\mathbf{C}} \sum_{i=1}^{\mathbf{C}} \sigma_i(\mathcal{X}^{n}_{u,i}) log(\sigma_i(\mathcal{X}^{n}_{u,i}))$
        % % \State 
        % \State $\mathcal{L}_{adj}=\frac{1}{n}\|M_a -M_y\|_{Fro}^2$ 
        % \textcolor{gray}{\# $M_y$ is the supervise of learnable adjacency, obtain by training dataset evaluation.}
        % \State $\mathcal{L}_{spa}=\|M_a\|_{Fro}^2$
        % \State $\mathcal{L}_g = \mathcal{L}_{ce}+\mathcal{L}_{aux}+\mathcal{L}_{orth}+\mathcal{L}_{adj}+\mathcal{L}_{spa}$
        \State \qquad Calculate $\mathcal{L}_g$ by \autoref{eq:l_GNN}
        \State \qquad Update $GNNs$ to minimize $\mathcal{L}_g$
        \State \qquad \textbf{if} iter++ > $I_g$: Solve dataset label mappings by \autoref{algo:2}; Stage = \textcolor{red}{Seg training}; iter = 0
        
    \State \textbf{if} Stage == \textcolor{red}{SegNet training}:
        % \State $\mathcal{L}_{ce} = \frac{1}{K} \sum^D_{i=1} \lambda^{(i)} \sum_{K} y_k^{(i)} log(\sigma(M^{(i)}\mathcal{X}^n_u\mathcal{V}^{(i)}_{k}))$
        % \State $\mathcal{L}_{aux} = \frac{1}{K} \sum^D_{i=1} \lambda^{(i)} \sum_{K} y_k^{(i)} log(\sigma(M^{(i)}\mathcal{X}^n_{(i)}\mathcal{V}^{(i)}_{k}))$
        \State \qquad Calculate $\mathcal{L}_{ce}$ by \autoref{eq:celoss}
        \State \qquad Update $SegNet$ to minimize $\mathcal{L}_{ce}$
        \State \qquad \textbf{if} iter++ > $I_s$: Stage = \textcolor{blue}{GNNs training}; iter = 0

\ENDSWITCH

\EndFor
\end{algorithmic}
\end{algorithm}

\section{Selection of Unified Label Node Quantity and Initialization of Adjacency Matrix}
\label{apdx:2}

The appropriate quantity of unified label nodes is crucial for balancing model performance and computational expenses. An excessive number of unified label nodes not only increases computational costs but may also result in redundancy in both the model and semantic space, leading to situations where multiple unified label nodes represent similar semantic concepts. Conversely, an insufficient number of unified label nodes can prevent the model from fully expressing the entire semantic space $\bigcup_{i=1}^D L_{i}$, making it difficult to learn certain classes in the datasets, thus impacting model performance. Inspired by  \cite{zhou2022simple}, we determine the number of unified label nodes using the following steps. We first enumerate all feasible schemes for merging dataset labels, which form the Cartesian product of all dataset label sets $\mathbb{L}=L_1 \times L_2 \times \ldots \times L_K $. Label merging does not require the participation of all datasets simultaneously, so $L_{i}$ can be an empty element, indicating no participation in label merging.

We evaluate the quality of label merging using the predefined loss function $\mathcal{L}_c$, where merging classes with the same semantic meaning results in smaller losses. Additionally, we require that the merging satisfies the constraint: each dataset node has only one corresponding merged node. It is worth noting that this constraint is stronger than our label mapping constraint described in \autoref{sec:uni}, thus ensuring that our adjacency matrix is always valid. We formulate optimization objective function as:
\begin{equation}
\begin{array}{ll}
\text{minimize}_x & \sum\limits_{\boldsymbol{t} \in \mathbb{T}} x_{\boldsymbol{t}} E_{\mathcal{S}_k}\left[\sum\limits_{c \in L_k \mid \boldsymbol{t}(c)=1} \mathcal{L}_c\left(S_c^k, \tilde{S}_c^k\right)\right]+\lambda | L |  \\
\text { subject to } & \sum\limits_{\boldsymbol{t} \in \mathbb{T} \mid \boldsymbol{t}_c=1} x_{\boldsymbol{t}}=1 \quad \forall_c,
\end{array}
\label{eq:simple}
\end{equation}
where $\mathbb{T}$ represents the set of all feasible edges connecting dataset label nodes to unified label nodes, $ x_{\boldsymbol{t}} \in \{0,1\}$ indicates whether edge $\boldsymbol{t} \in \mathbb{T}$ is selected, with $x_{\boldsymbol{t}} = 1$ indicating the presence of edge $\boldsymbol{t}$ and 0 indicating its absence. $\mathcal{S}_k$ denotes the dataset to which edge $\boldsymbol{t}$ belongs, $L_k$ represents the label space of $\mathcal{S}_k$, and $\boldsymbol{t}(c)$ is the label connected by edge $\boldsymbol{t}$. $\lambda | L |$ is the penalty term for the number of nodes, encouraging the optimizer to merge similar nodes to obtain a more compact unified label space. Based on experimental insights and referencing paper [52], we selected the hyperparameter $\lambda=0.5$. $\mathcal{L}_c\left(S_c^k, \tilde{S}_c^k\right)$ is primarily used to evaluate the quality of label merging, where $S_c^k$ represents the output results on class $c$ using the original dataset segmentation head, and $\tilde{S}_c^k$ represents the output results on class $c$ using segmentation heads from other datasets participating in the merging. 
\begin{equation}
\mathcal{L}_c\left(S_c^k, \tilde{S}_c^k\right) = \sum_{s_c \in \tilde{S}_c^k} IoU(S_c^k)-IoU(s_c),
\label{eq:iou}
\end{equation}
where, $IoU$ is a metric commonly used to measure the performance of semantic segmentation models. Therefore, we use this metric here to evaluate the assistance of merged nodes to the model. By using a linear optimization solver to solve the optimization objective \autoref{eq:simple}, we can obtain a unified label space. We use the number of labels in this label space as the quantity of our unified label nodes and initialize our adjacency matrix based on their label mapping relationships.

\section{Solving the dataset Label Mappings}
\label{apdx:1}

Given the definition of the label mappings and the constraints outlined in \autoref{sec:uni}, solving the dataset label mappings could be conceptualized as a weighted bipartite graph matching problem. The weight of each edge is determined by the values of the GNNs learnable adjacency matrix. The objective is to maximize the sum of edge weights while ensuring that each node of the specific dataset is connected. The unbalanced optimal transport algorithm offers an approximate solution to such problems, but it does not guarantee adherence to the constraint of connecting each node in the specific dataset. Therefore, based on the optimal matching result obtained from the algorithm, we employ a greedy strategy to adjust the matching scheme for all unconnected nodes within the specific dataset.

Initially, for every unconnected dataset-specific node, we sort all unified label nodes based on their matching preferences, from highest to lowest. Subsequently, we examine these unified label nodes and their associated dataset-specific nodes. In cases where the associated dataset-specific node is also connected by another unified label node, we adjust the edge so that this unified label node links to the unconnected specific dataset node, following this process until all nodes are connected. The specific procedure is outlined in \autoref{algo:2}.

\begin{algorithm}[t]
\renewcommand{\algorithmicrequire}{\textbf{Input:}}
\renewcommand{\algorithmicensure}{\textbf{Output:}}
\caption{The procedure of solving the dataset label mappings}
\label{algo:2}
\begin{algorithmic}
\Require  Submatrix of adjacency matrix for each dataset $\{S^{(i)},i=1,...,K\}$, the classes of each dataset $ \{{c^{(i)}}, i=1,...,K\}$, the number of unified label node $N$
\Ensure Mapping matrices $M^{(1)},...,M^{(K)}$
\State Initialize  $\alpha = \dfrac{1}{\mathbf{N}} \mathbf{1}_{\mathbf{N}\times 1}$, $\beta^{(i)} = \{\dfrac{1}{c^{(i)}} \mathbf{1}_{c^{(i)} \times 1}, i = 1,...,K\}$, $d=0$, $\mu$

\For{$i = 1$ to $K$}
    \State $Q^{(i)}_{c^{(i)} \times \mathbf{N}} = \text{UOT}(\alpha, \beta^{(i)}, S^{(i)})$
    \State \textcolor{gray}{\# Assign the label to the node with the highest score}
    \State $M^{(i)}_{j,k} = \mathop{\arg\max}_j \{Q^{(i)}_{j,1},...,Q^{(i)}_{j,\mathbf{N}}\}$ 
    \For{$j = 1$ to $c^{(i)}$}
        \State \textcolor{gray}{\# Find an unlinked dataset-specific node}
        \If{$\max M^{(i)}_{j} == 0$}
            \State $q_j = \text{sort}(Q^{(i)}_{j})$
            \State \textcolor{gray}{\# Find first multi-mapped node, where $sum (M^{(i)}_{j^{\prime}}) > 1$}
            \State $j^{\prime},k^{\prime} = \text{findFirstMulMapped}(q_j, M^{(i)}) $
            \State \textcolor{gray}{\# Replace the corresponding label mappings}
            \State $M^{(i)}_{j^{\prime},k^{\prime}} = 0, M^{(i)}_{j,k^{\prime}} = 1$ 
            
        \EndIf
    \EndFor
    \State $\Bar{\beta}^{(i)}_j = \sum_{k=1}^{c^{(i)}} Q^{(i)}_{j,k} $ \;
    \State $\beta^{(i)} = \mu \beta^{(i)} + (1-\mu) \Bar{\beta}^{(i)}$
\EndFor
\end{algorithmic}
\end{algorithm}

\section{Comparison with Baselines on Multiple Datasets}
\label{apdx:tb}
\autoref{tb:ts} lists the results of mutual evaluation among training datasets. The results indicate that individually-trained models generally demonstrate good accuracy when tested on the same dataset, but perform poorly on other datasets. Our method can leverage knowledge from multiple datasets to improve performance on individual datasets. \autoref{tb:tm} lists the results of the Multi-SegHead model's predictions on different datasets using different segmentation head outputs. It can be observed that none of the segmentation head outputs could consistently provide accurate predictions across all datasets, indicating that their label spaces do not effectively cover all datasets. \autoref{tb:unseen_s} and \autoref{tb:unseen_m} list the performance of different models on unseen datasets. The results indicate that our method has better generalization performance on unseen datasets compared to other methods. Additionally, the unified label space we construct contains richer semantic information, enabling flexible adaptation to datasets from different scenarios.
% Additionally, it can be observed that datasets with large sample sizes and numerous categories, such as VISTAS, COCO, and ADE, exhibit better generalization performance.

\begin{table}[h!]
\centering
\caption{Semantic segmentation accuracy (mIoU) on training datasets compared with Single dataset model.}
\label{tb:ts}
\footnotesize
\begin{tabular}{c|cccccccc}
\hline
Train$\backslash$Test & CS & MPL & SUN & BDD & IDD & ADE & COCO & Mean\\
\hline
CS & \underline{77.0} & 7.2 & 3.7 & 43.4 & 30.4 & 2.6 & 2.8 & 
 23.9\\
MPL & 65.6 & \underline{30.2} & 4.8 & 54.5 & 48.4 & 4.7 & 5.9 & \underline{30.6}\\
SUN & 13.0 & 3.4 & \underline{43.9} & 12.4 & 9.5 & 5.1 & 3.0 & 12.9\\
BDD & 58.9 & 12.1 & 3.6 & \underline{62.4} & 30.9 & 2.7 & 2.7 & 24.8\\
IDD & 50.9 & 10.4 &  4.3 & 43.5 & \underline{66.8} & 3.1 &  2.8 & 26.0 \\
ADE & 42.7 & 11.5 & 35.5 & 34.7 & 27.5 & \underline{34.5} & 12.8 & 28.5\\
COCO & 45.5 & 14.0 & 29.1 & 42.0 & 30.1 & 14.8 & \underline{38.0} & 30.5\\
\hline
Ours & \textbf{80.7} & \textbf{43.7} & \textbf{47.5} & \textbf{65.5} & \textbf{68.6} & \textbf{42.0} & \textbf{46.7} & \textbf{56.4}\\
\hline
\end{tabular}
\end{table}
\vspace{-0.5cm}
\begin{table}[h!]
\centering
\caption{Semantic segmentation accuracy (mIoU) on training datasets compared with Multi-SegHead.}
\label{tb:tm}
\footnotesize
\begin{tabular}{c|cccccccc}
\hline
Label space$\backslash$Test & CS & MPL & SUN & BDD & IDD & ADE & COCO & Mean\\
\hline
CS & \underline{79.5} & 16.1 & 10.4 & 61.6 & 42.8 & 2.5 & 3.6 & 30.9
 \\
MPL & 72.9 & \underline{36.1} & 12.9 & 60.4 & 54.7 & 7.0 & 8.9 & 36.1 \\
SUN & 33.5 & 9.3 & \underline{47.3} & 29.5 & 22.7 & 6.4 & 5.3 & 22.0 \\
BDD & 67.1 & 16.6 & 8.9 & \textbf{65.6} & 39.0 & 2.6 & 3.6 & 29.1\\
IDD & 72.3 & 20.3 & 10.8 & 59.8 & \underline{67.0} & 3.3 & 4.3  &  34.0 \\
ADE & 62.7 & 20.8 & 43.5 & 57.6 & 46.6 & \underline{30.5} & 20.5 & \underline{40.3}\\
COCO & 59.5 & 21.4 & 34.4 & 57.5 & 44.4 & 18.0 & \underline{36.7} & 38.8\\
\hline
Ours & \textbf{80.7} & \textbf{43.7} & \textbf{47.5} & \underline{65.5} & \textbf{68.6} & \textbf{42.0} & \textbf{46.7} & \textbf{56.4}\\
\hline
\end{tabular}
\end{table}
\vspace{-0.5cm}
\begin{table}[h!]
\centering
\footnotesize
\caption{Semantic segmentation accuracy (mIoU) on unseen datasets compared with Single dataset.}
\label{tb:unseen_s}
\begin{tabular}{c|cccccc}
\hline
Train$\backslash$Test & KT & SN & CV & VOC & CT & Mean\\
\hline
CS & 62.3 & 4.3  & 67.5 & 16.1 & 7.0 & 31.4\\
MPL & \textbf{67.3}  & 6.1 & \textbf{73.1} & 24.6 & 13.3 & 36.9\\
SUN & 12.4  & \underline{32.2} & 20.2 & 13.2 & 5.4 & 16.7\\
BDD & 56.0  & 5.3 & 66.7 & 17.3 & 6.7 & 30.4\\
IDD & 50.6  &  6.1 & 62.6 & 17.7 & 7.4 &  28.9\\
ADE & 45.6  & 25.8 & 51.7 & 39.0 & 23.7 & 37.2\\
COCO & 45.4  & 27.4 & 53.9 & \underline{63.0} & \underline{35.8} & \underline{45.1}\\
\hline
Ours & \underline{65.5}  & \textbf{40.6} & \underline{71.1} & \textbf{69.2} & \textbf{38.1} & \textbf{56.9}\\
\hline
\end{tabular}
\end{table}
\vspace{-0.5cm}
\begin{table}[h!]
\centering
\footnotesize
\caption{Semantic segmentation accuracy (mIoU) on unseen datasets compared with Multi-SegHead.}
\label{tb:unseen_m}
\begin{tabular}{c|cccccc}
\hline
Label space$\backslash$Test & KT  & SN & CV & VOC & CT & Mean\\
\hline
CS & 59.7  & 14.6 & \underline{72.3} & 29.0 & 9.3 & 37.9\\
MPL & \underline{63.7}  & 18.3 & \textbf{72.8} & 33.6 & 16.4 & 41.0\\
SUN & 27.0  & \textbf{42.9} & 34.9 & 29.9 & 10.0 & 28.9\\
BDD & 61.0  & 15.7 & 71.2 & 21.9 & 8.7 & 35.7\\
IDD & 56.8  & 16.3 & 69.0 & 29.8 & 10.3 & 36.4\\
ADE & 56.4  & 40.8 & 65.1 & 56.2 & 30.7 & 49.8\\
COCO & 55.9  & 37.6 & 64.6 & \textbf{69.6} & \textbf{38.2} & \underline{53.2}\\
\hline
Ours & \textbf{65.5}  & \underline{40.6} & 71.1 & \underline{69.2} & \underline{38.1} & \textbf{56.9}\\
\hline
\end{tabular}
\end{table}

\clearpage
\section{Exploring the Impact of Training Datasets}
\label{apdx:impact}

\begin{figure}[t]
  \centering
  \footnotesize
  \setlength{\tabcolsep}{0.02cm} % 设置列之间的间距
  \begin{tabular}{lcccccc}
      & CS & IDD & WD2 & SUN & COCO & CV  \\
    \rotatebox[origin=l]{90}{Image} &
    \includegraphics[width=0.16\linewidth]{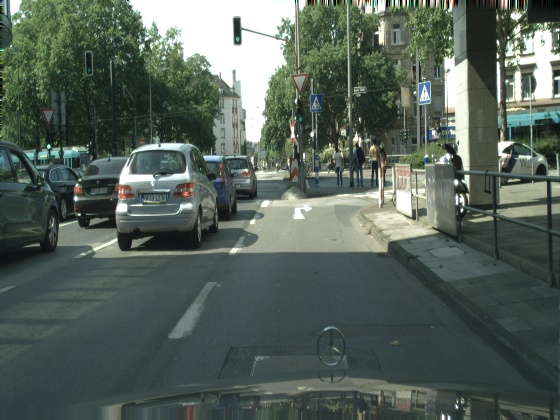} &
    \includegraphics[width=0.16\linewidth]{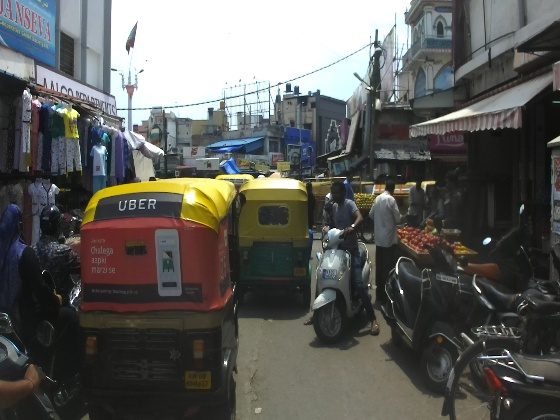} &
    \includegraphics[width=0.16\linewidth]{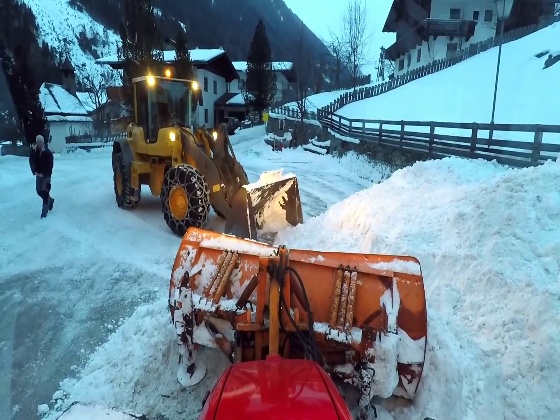} &
    \includegraphics[width=0.16\linewidth]{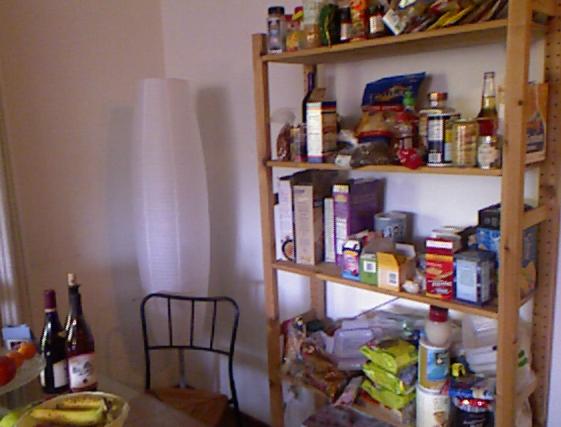} &
    \includegraphics[width=0.16\linewidth]{image/coco_000000000632.jpg} &
    \includegraphics[width=0.16\linewidth]{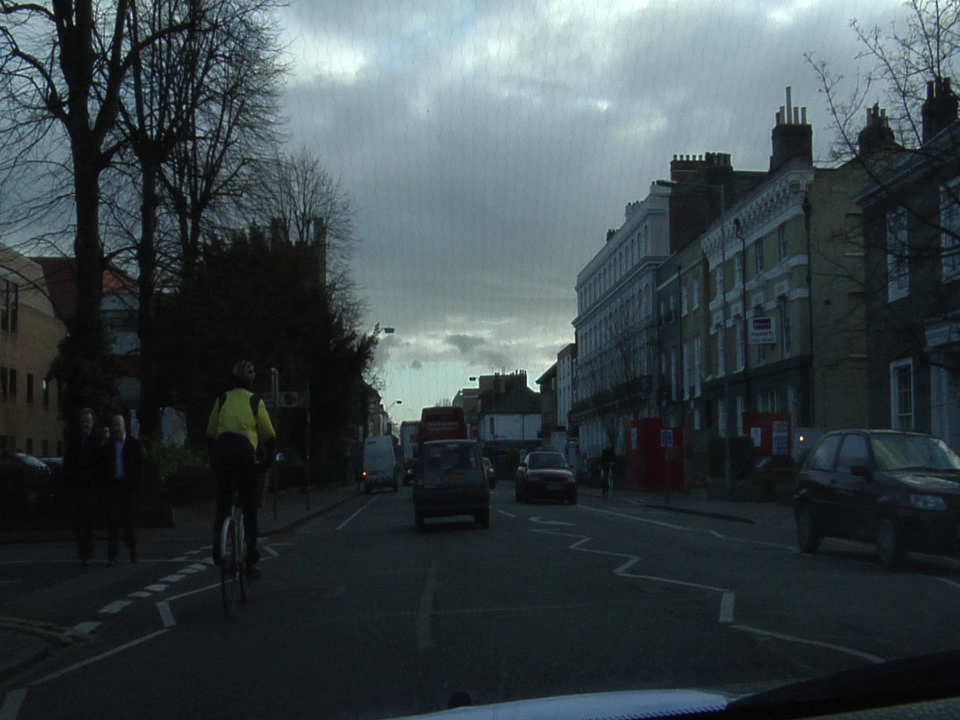} \\

    \rotatebox[origin=l]{90}{\quad GT} &
    \includegraphics[width=0.16\linewidth]{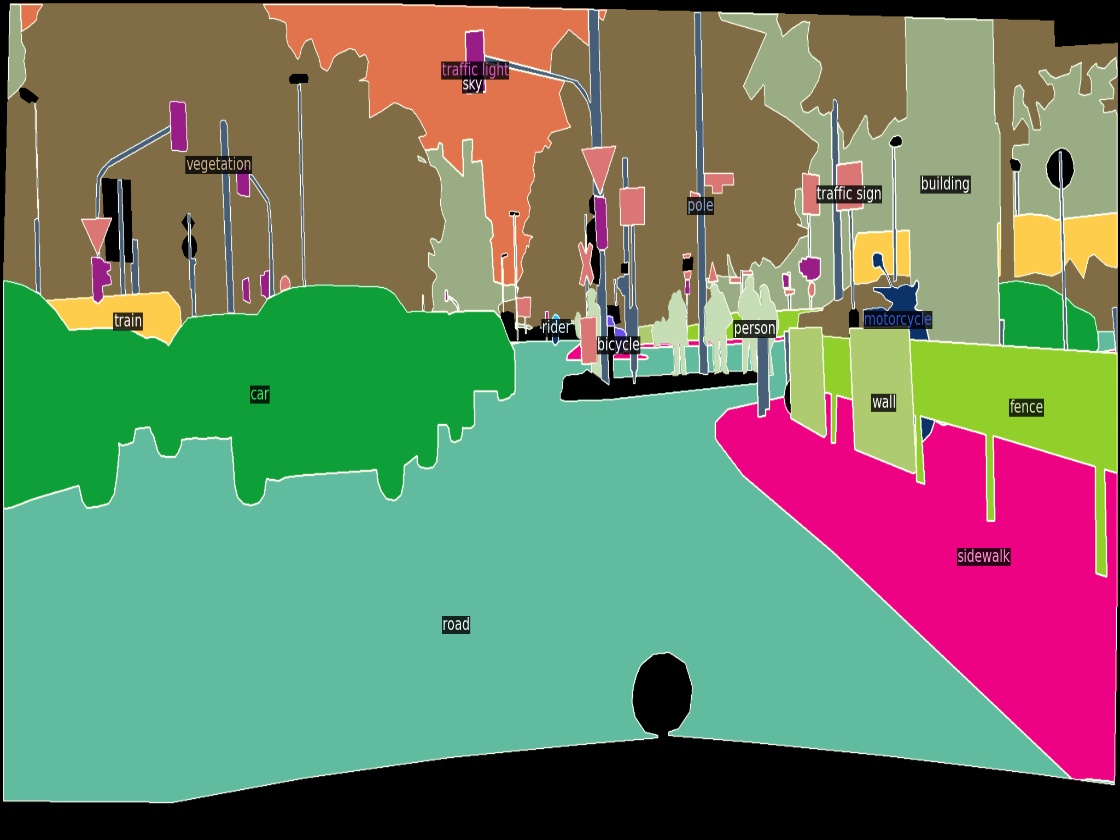} &
    \includegraphics[width=0.16\linewidth]{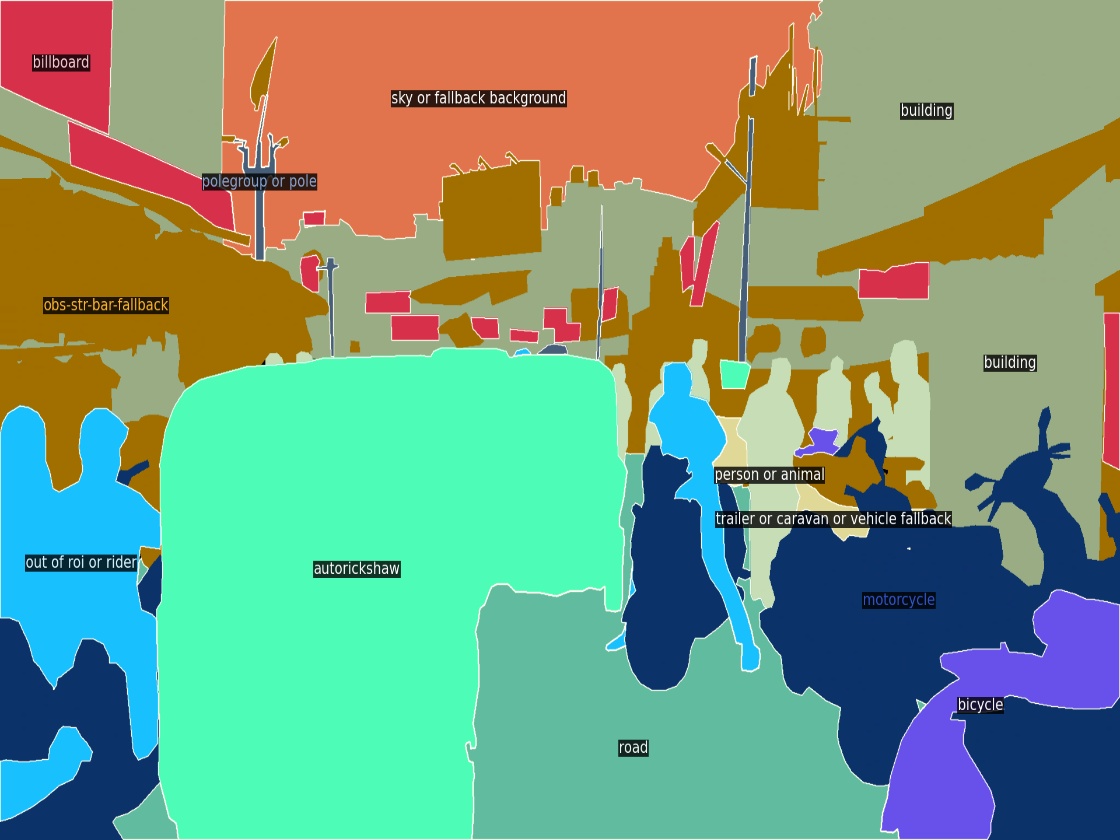} &
    \includegraphics[width=0.16\linewidth]{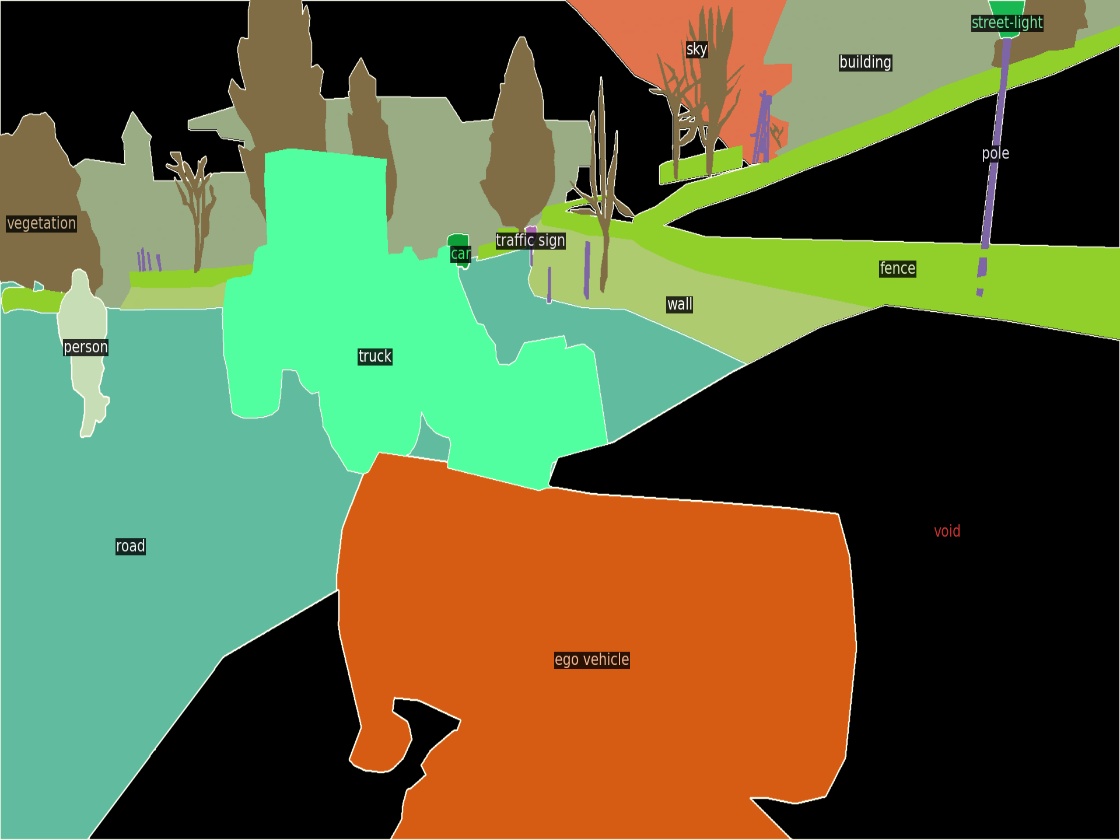} &
    \includegraphics[width=0.16\linewidth]{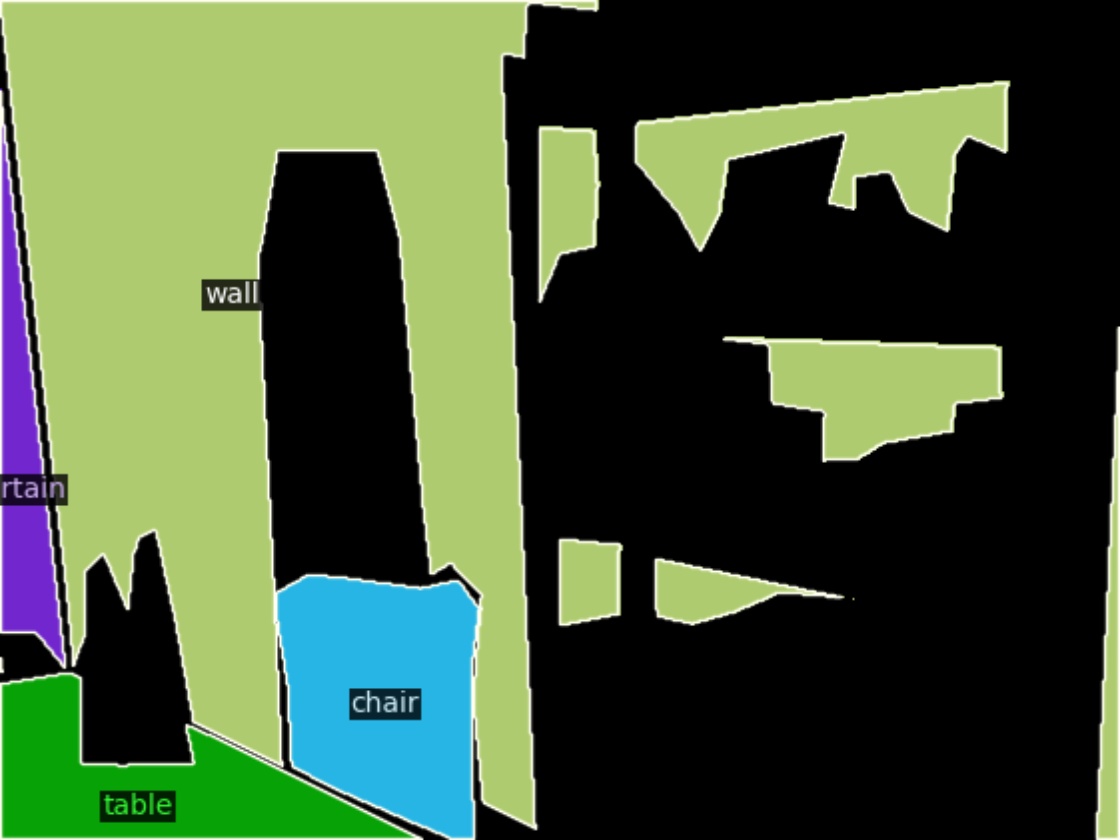} &
    \includegraphics[width=0.16\linewidth]{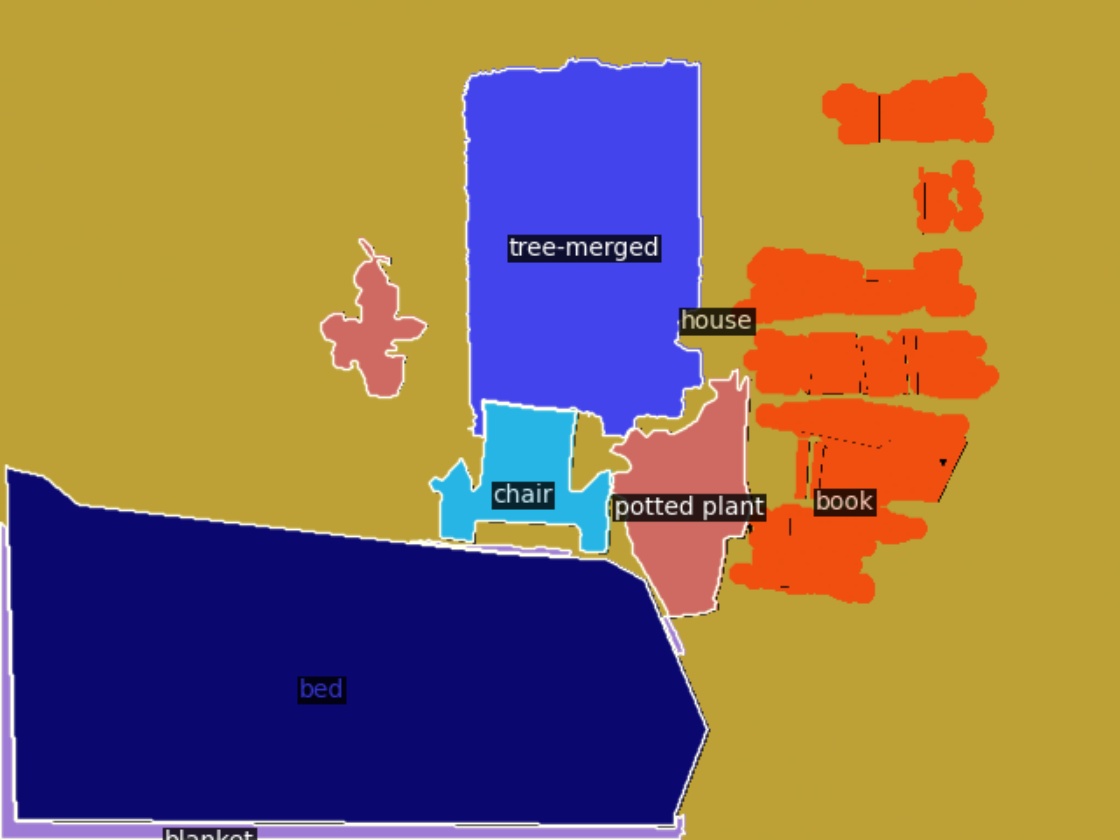}     &
    \includegraphics[width=0.16\linewidth]{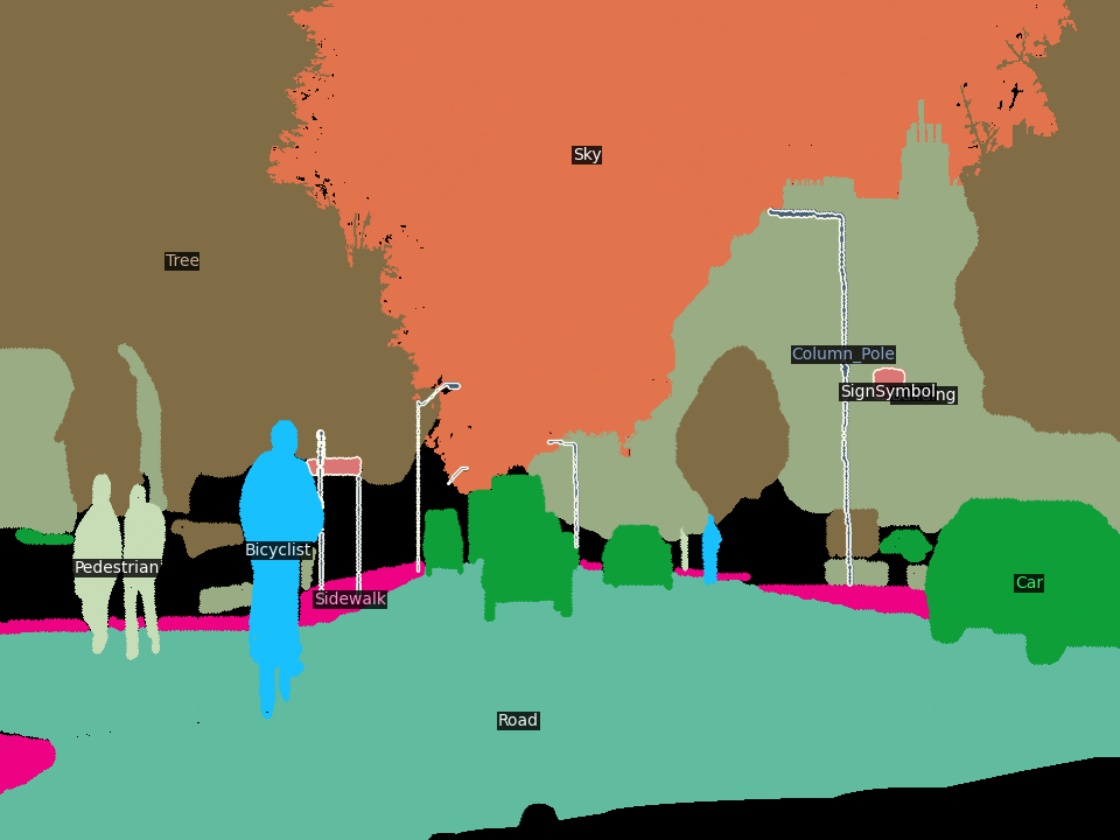}  \\

    \rotatebox[origin=l]{90}{General} &
    \includegraphics[width=0.16\linewidth]{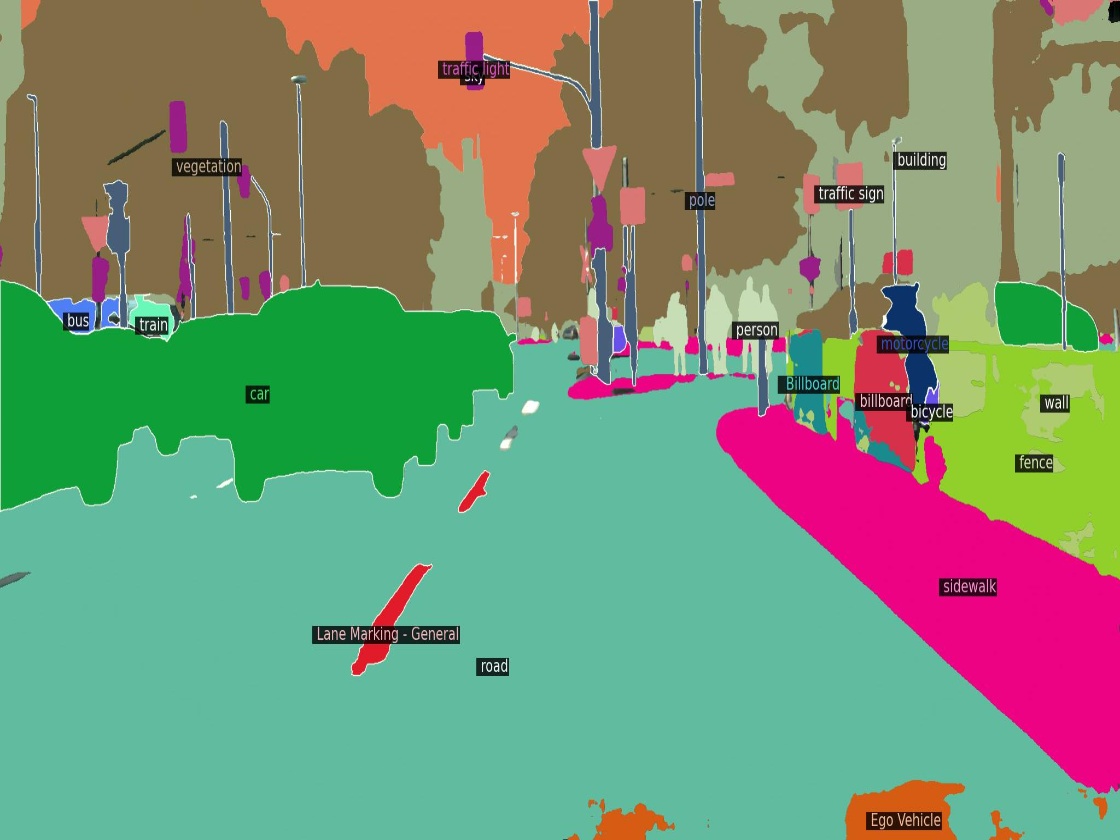} &
    \includegraphics[width=0.16\linewidth]{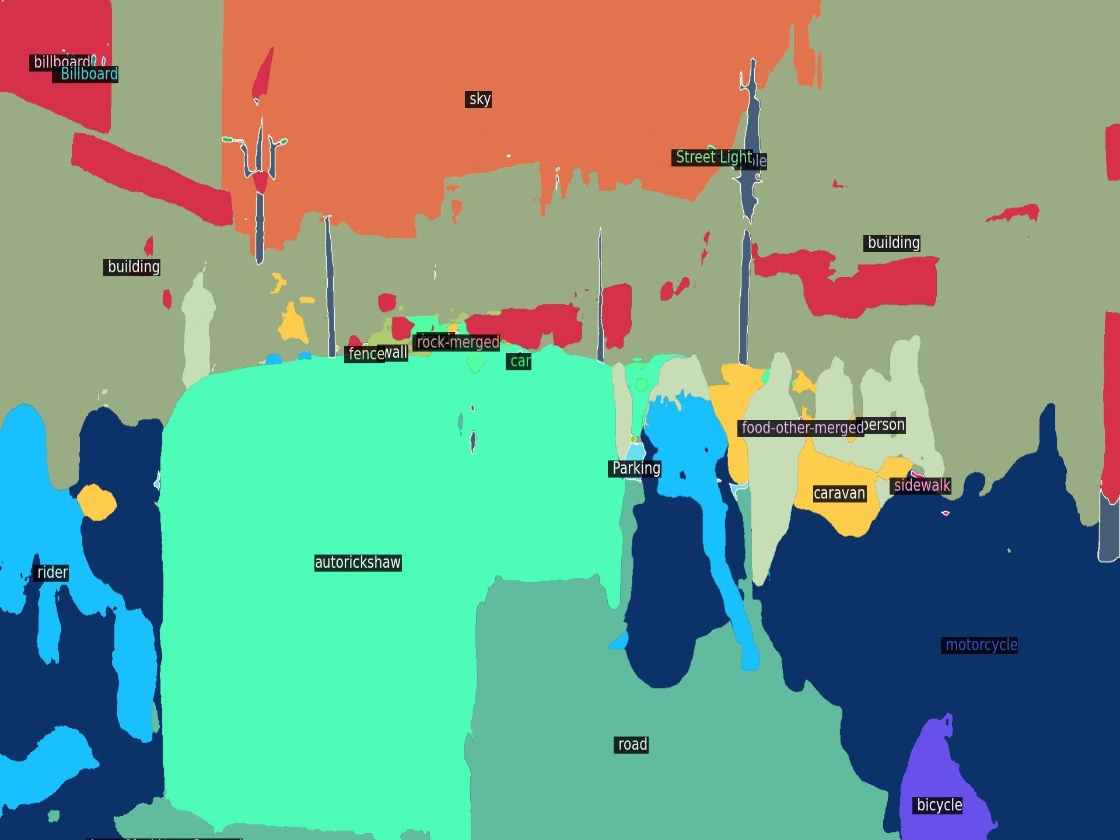} &
    \includegraphics[width=0.16\linewidth]{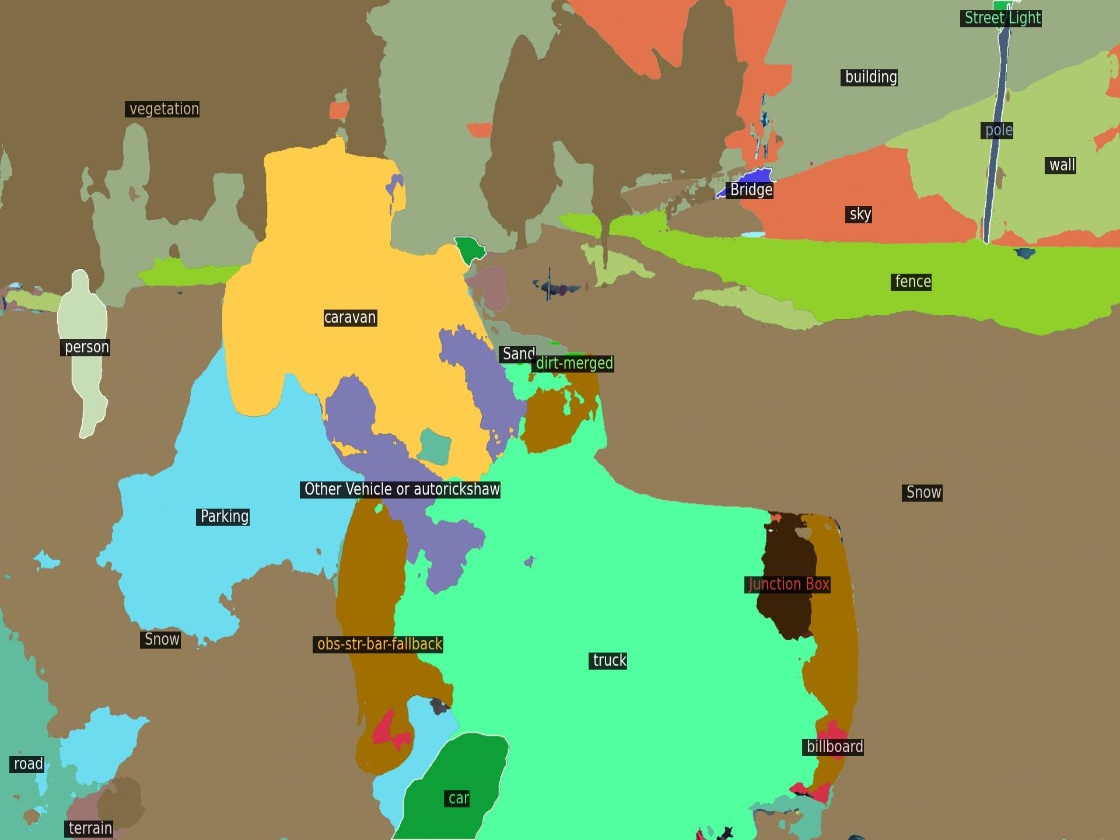} &
    \includegraphics[width=0.16\linewidth]{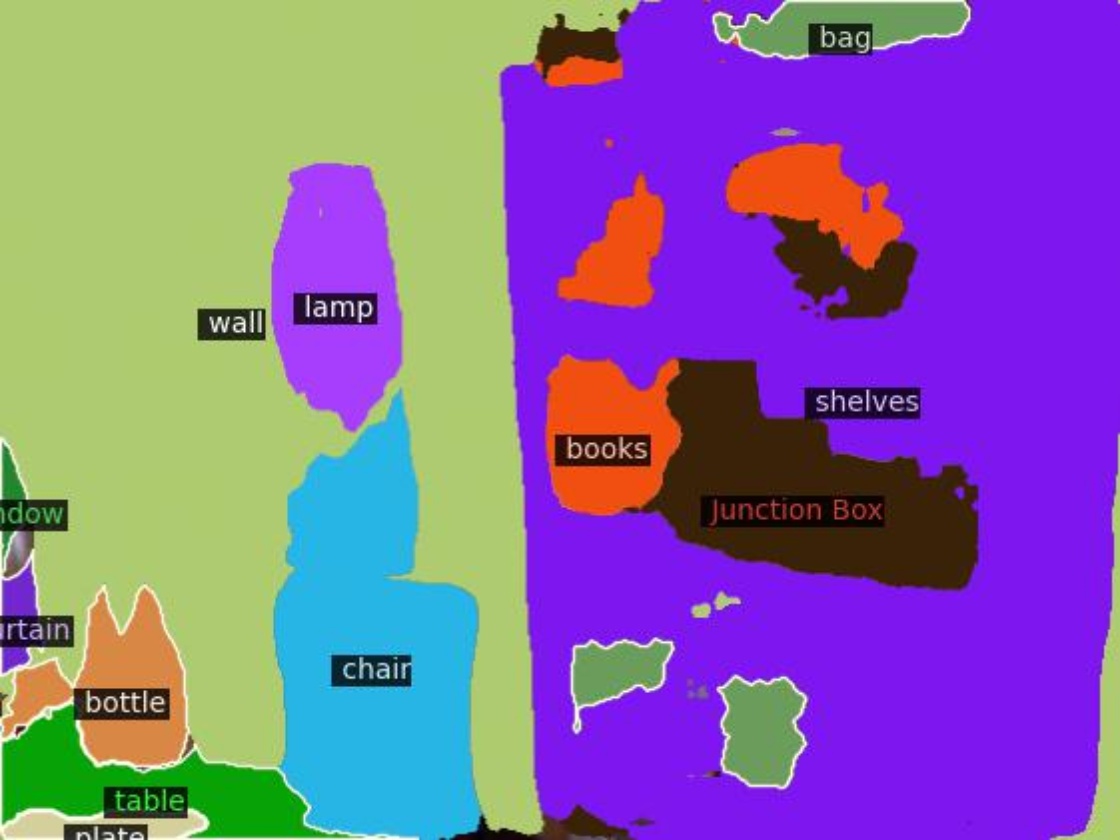} &
    \includegraphics[width=0.16\linewidth]{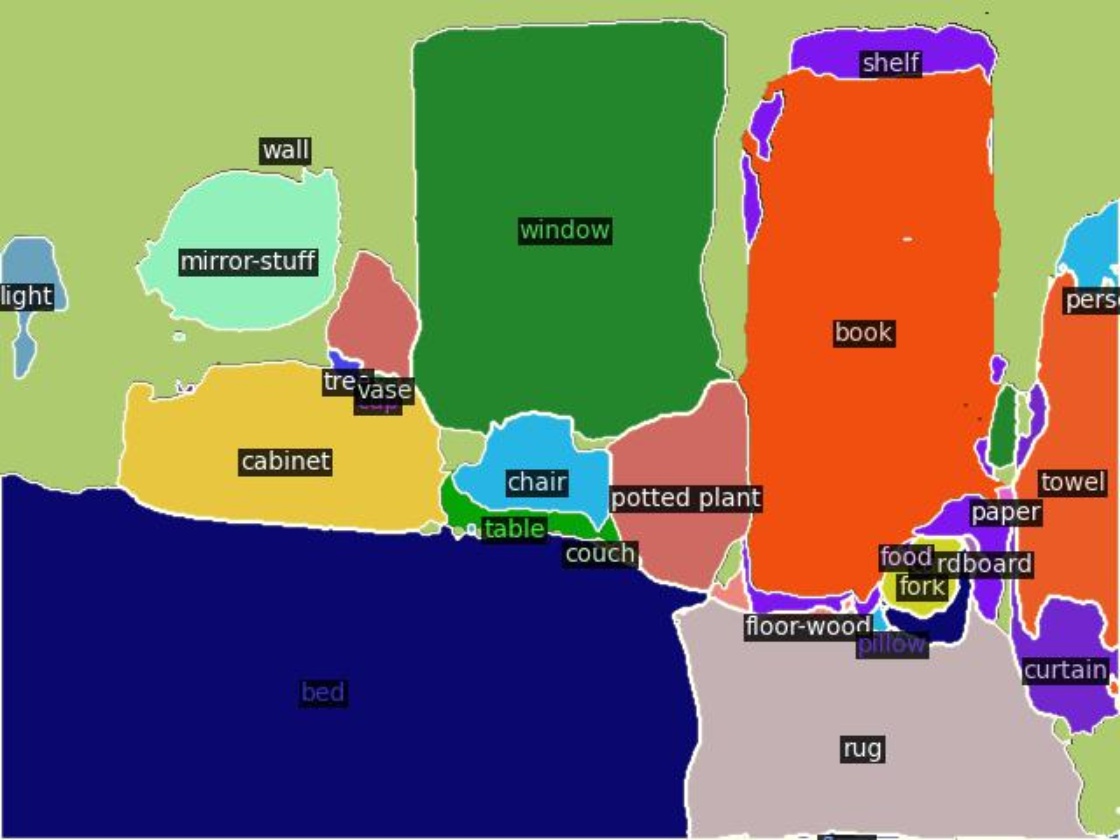}     &
    \includegraphics[width=0.16\linewidth]{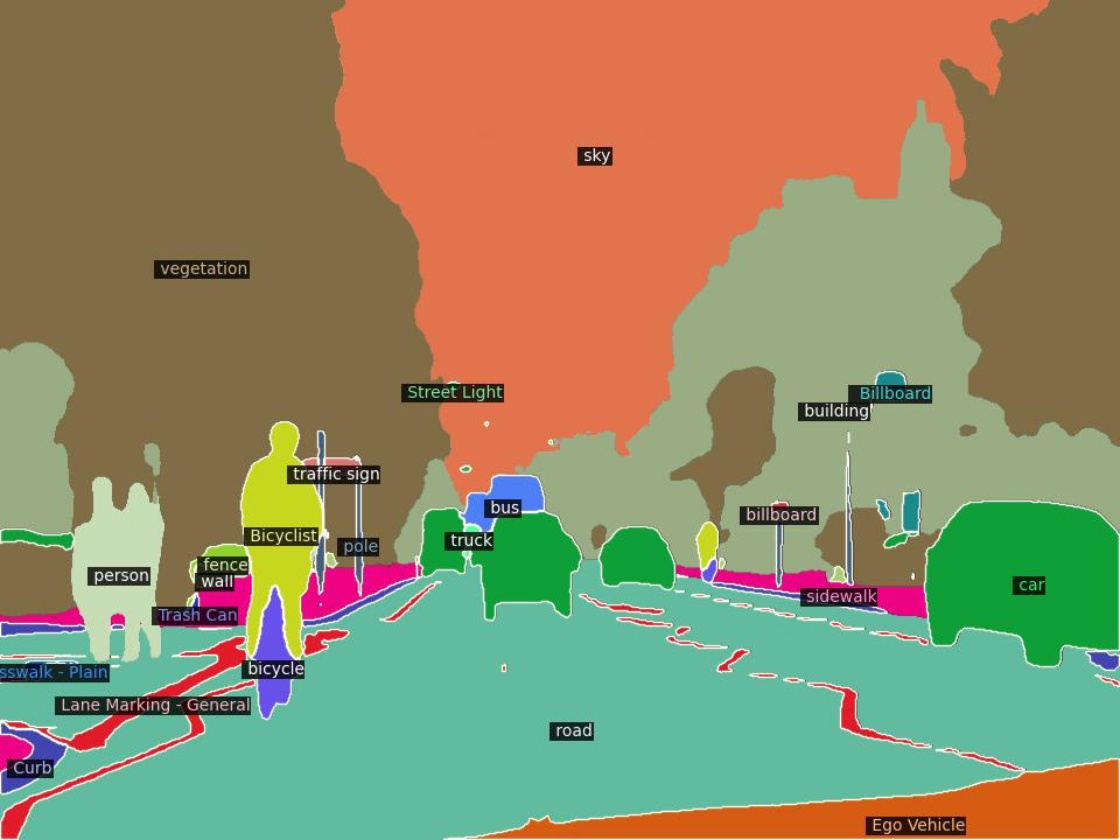}  \\
    
    \rotatebox[origin=l]{90}{Specific} &
    \includegraphics[width=0.16\linewidth]{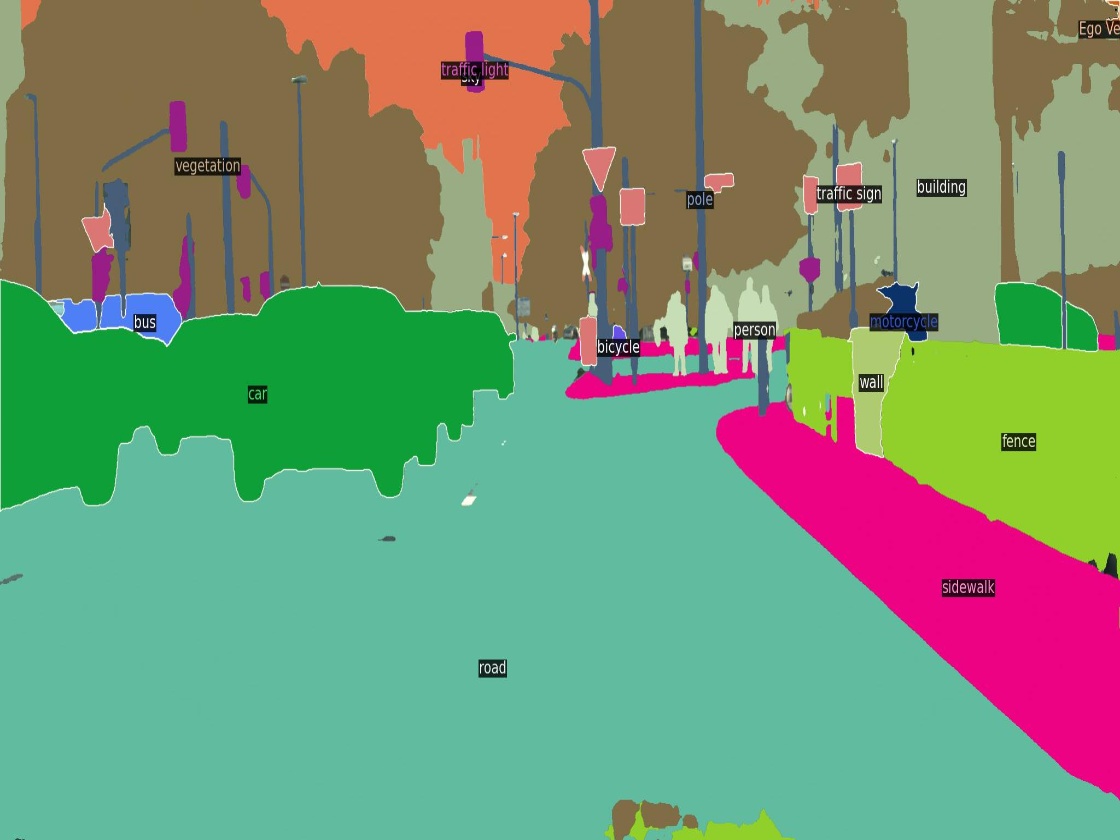} &
    \includegraphics[width=0.16\linewidth]{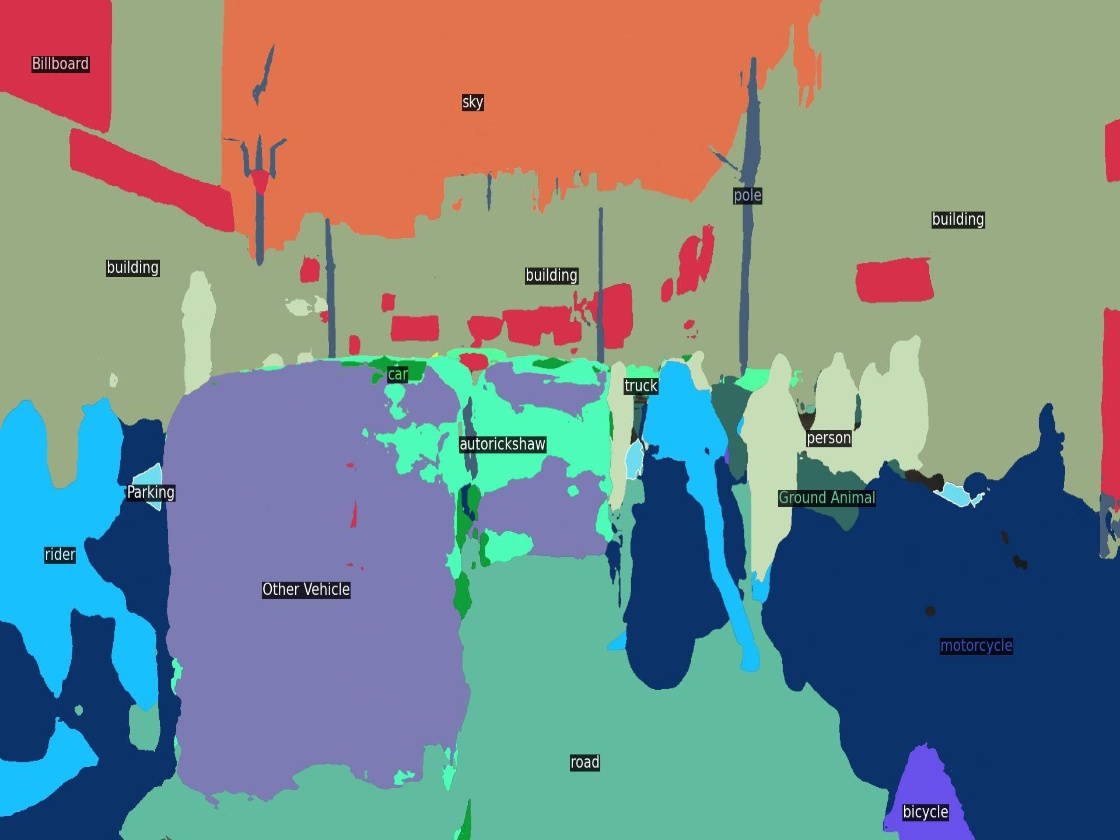} &
    \includegraphics[width=0.16\linewidth]{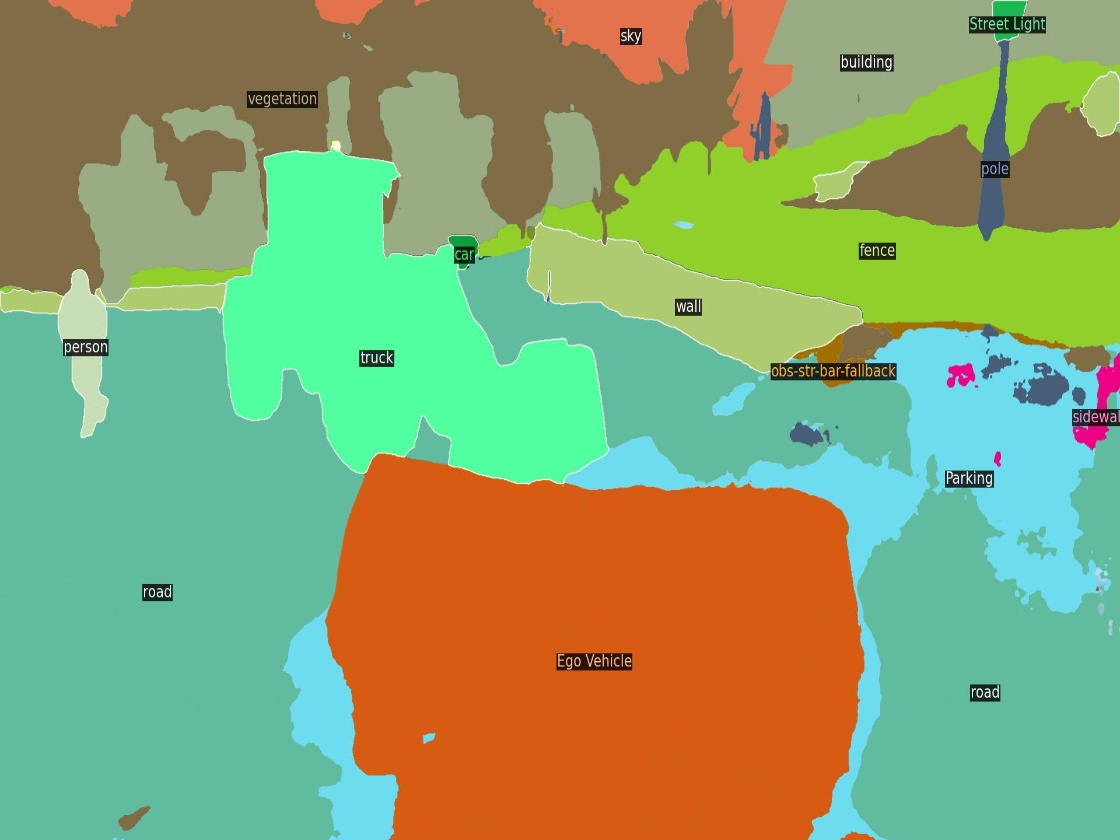} &
    \includegraphics[width=0.16\linewidth]{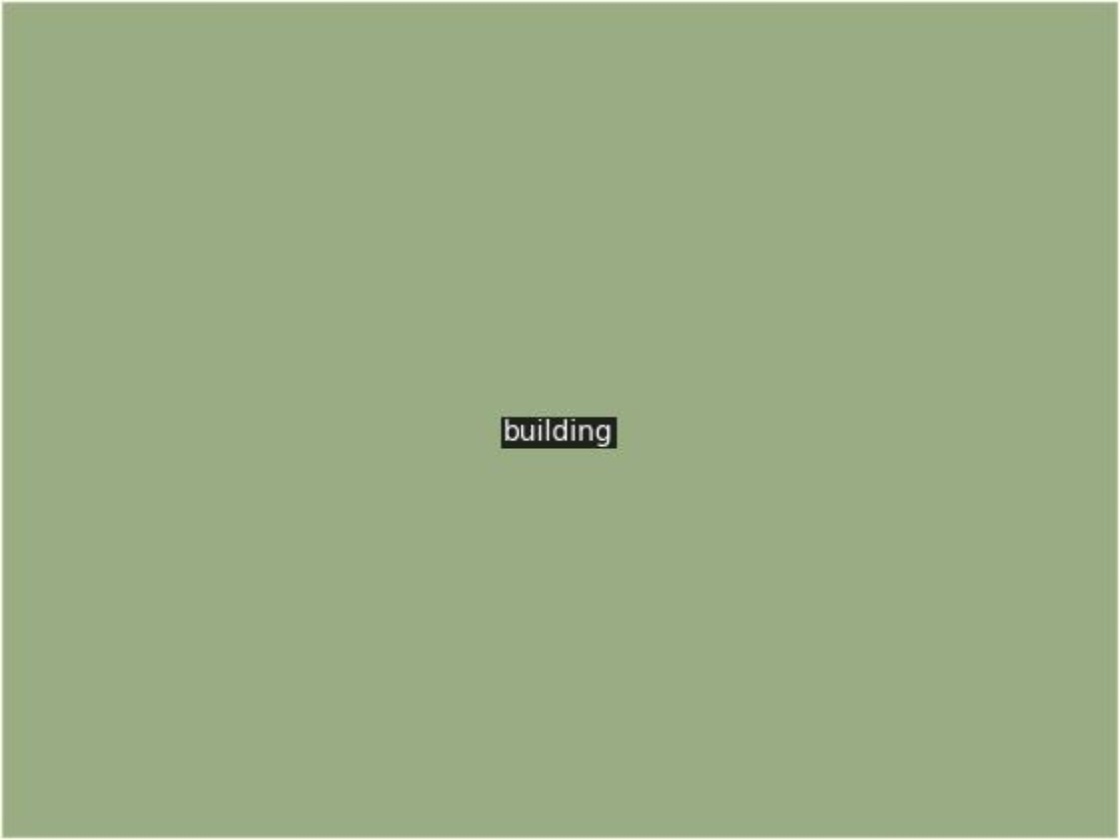} &
    \includegraphics[width=0.16\linewidth]{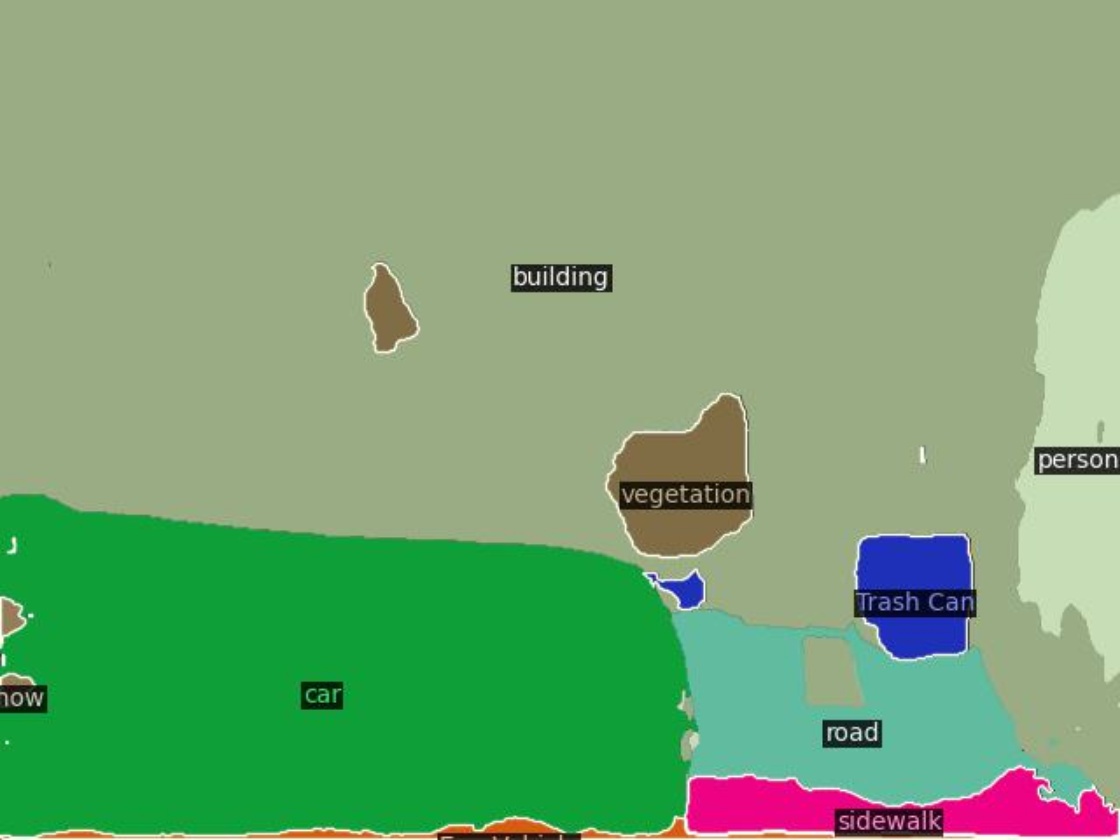}    &
    \includegraphics[width=0.16\linewidth]{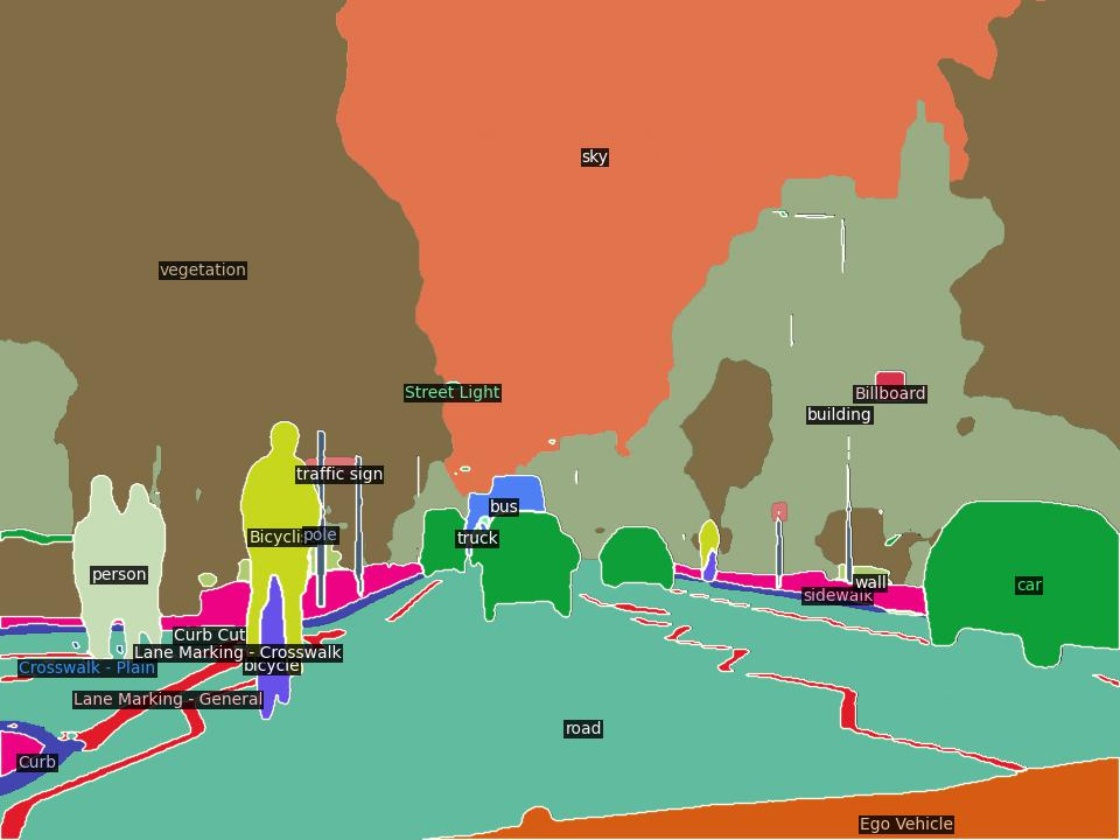}  \\

      & BDD & MPL & ADE & KT & SN & VOC  \\
    \rotatebox[origin=l]{90}{Image} &
    \includegraphics[width=0.16\linewidth]{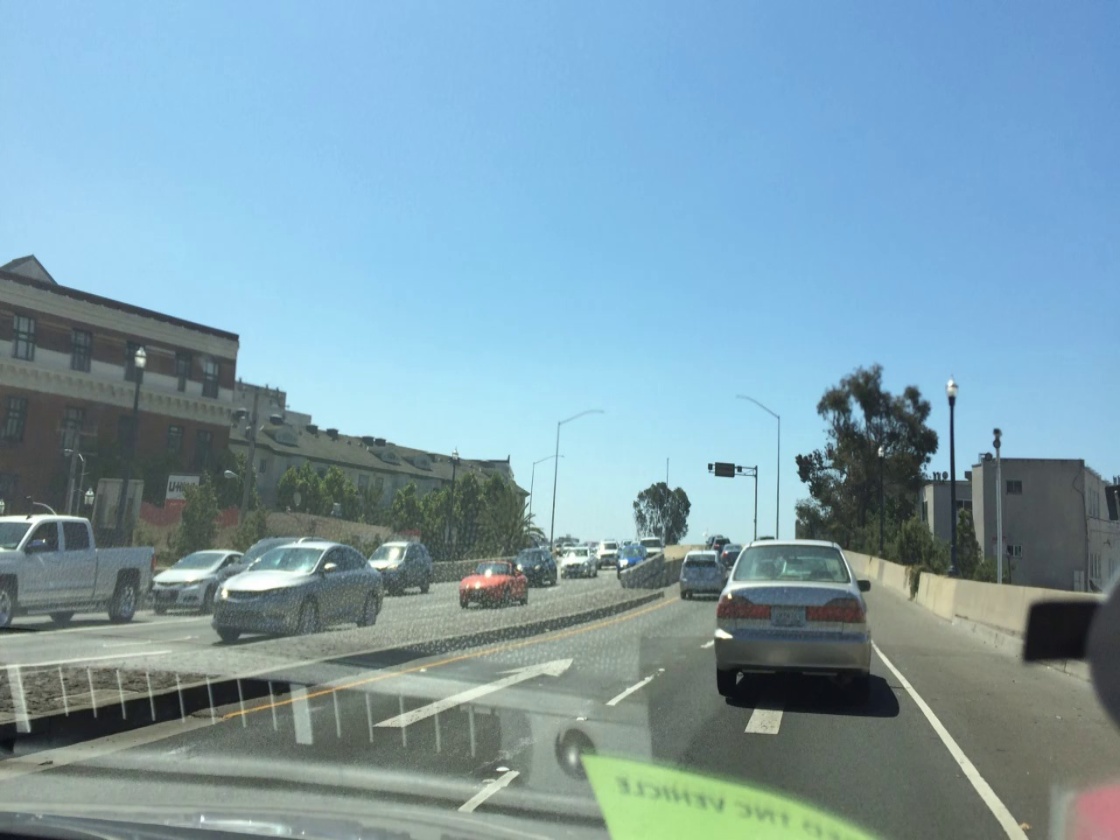} &
    \includegraphics[width=0.16\linewidth]{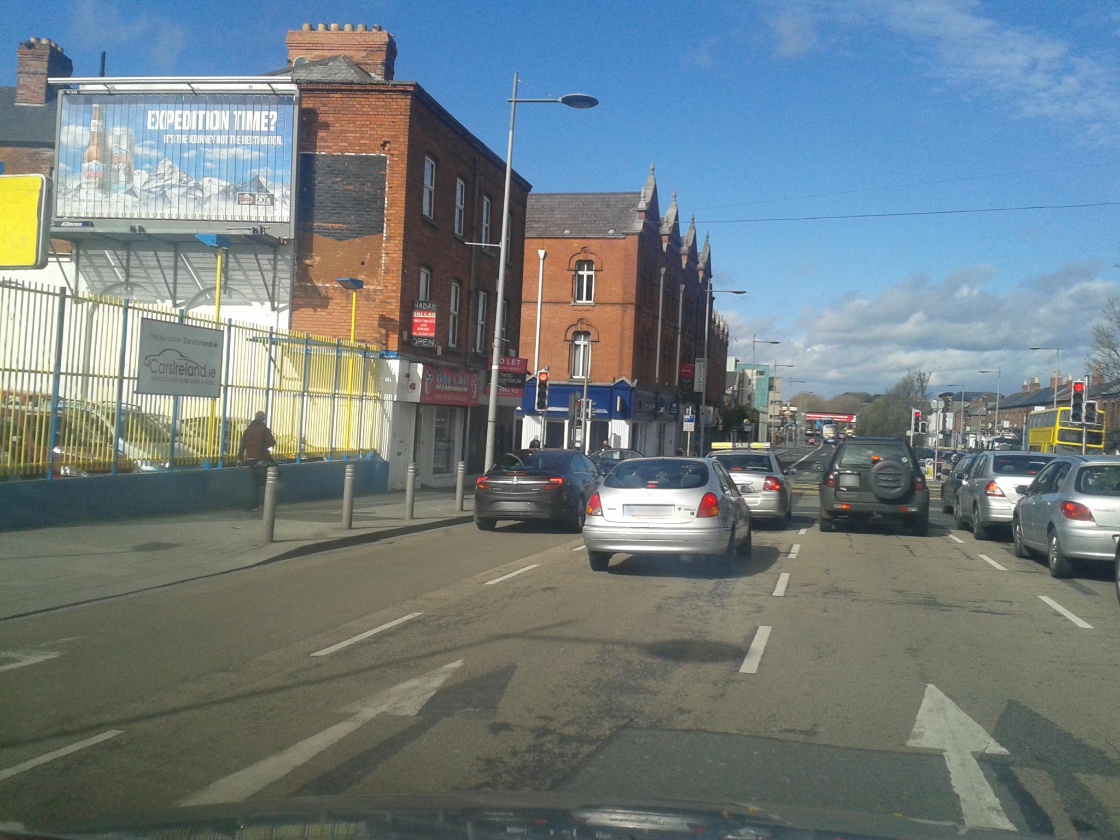} &
    \includegraphics[width=0.16\linewidth]{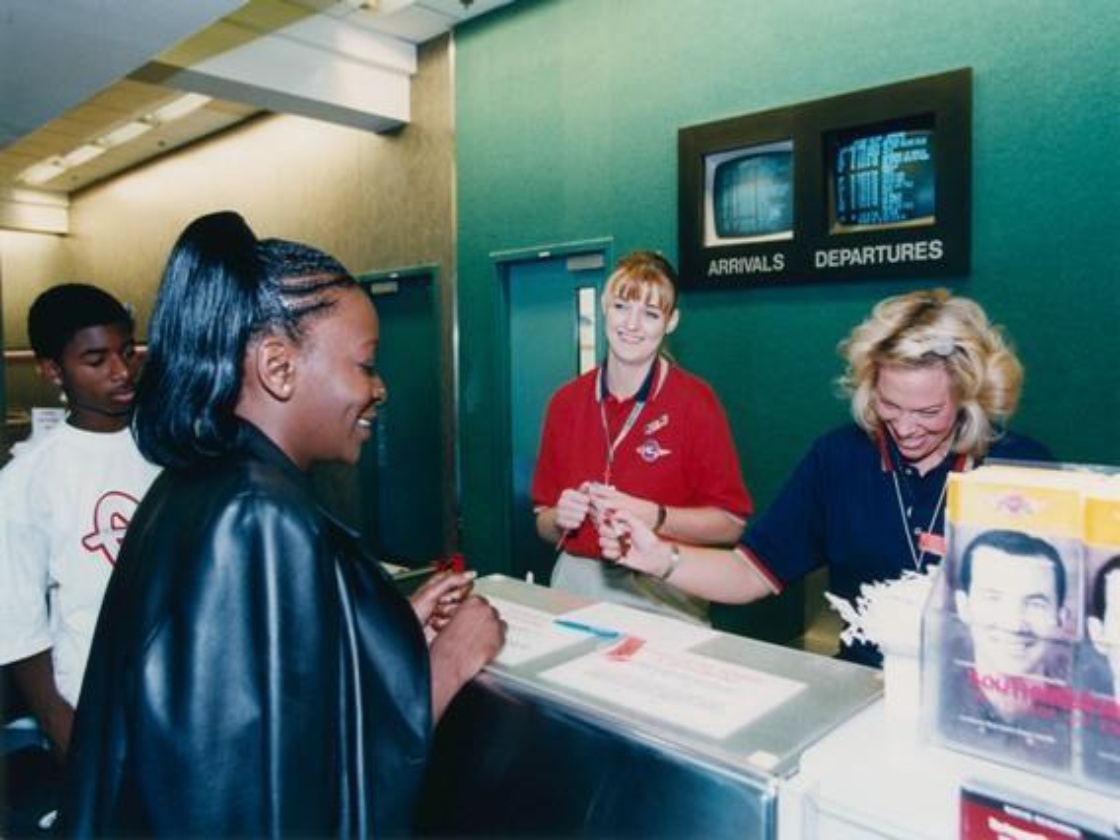} &
    \includegraphics[width=0.16\linewidth]{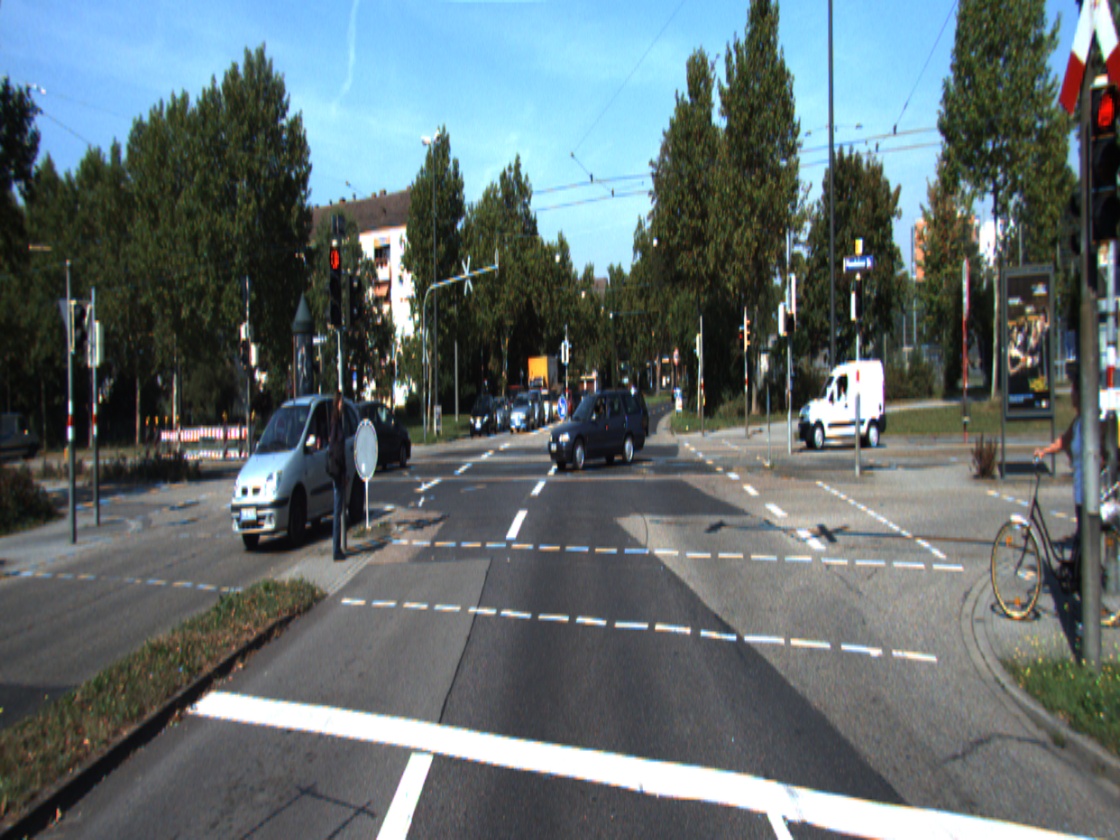} &
    \includegraphics[width=0.16\linewidth]{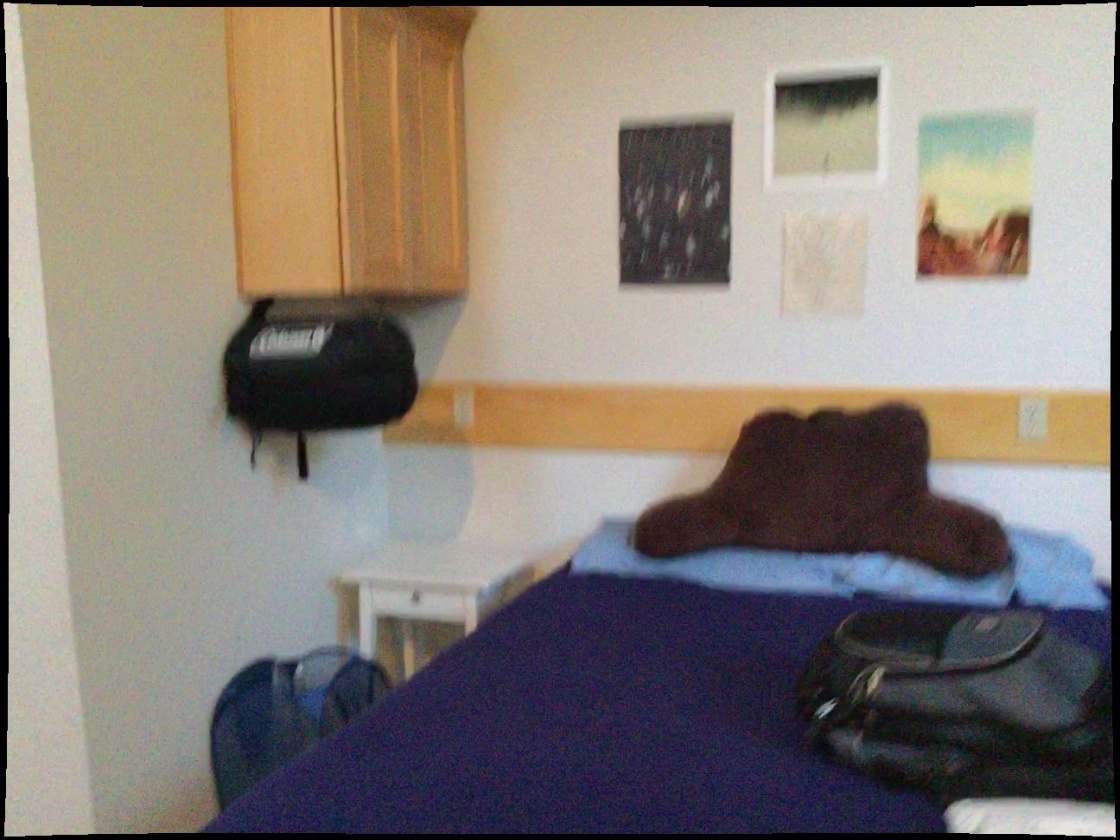} &

    \includegraphics[width=0.16\linewidth]{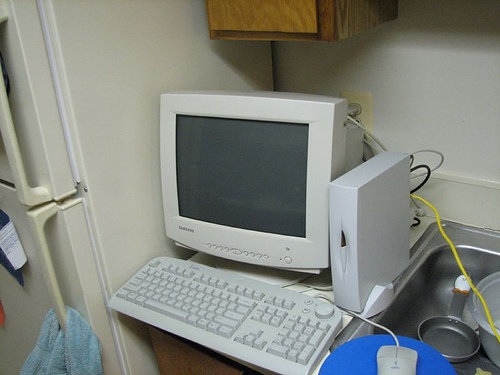} \\

    \rotatebox[origin=l]{90}{\quad GT} &
    \includegraphics[width=0.16\linewidth]{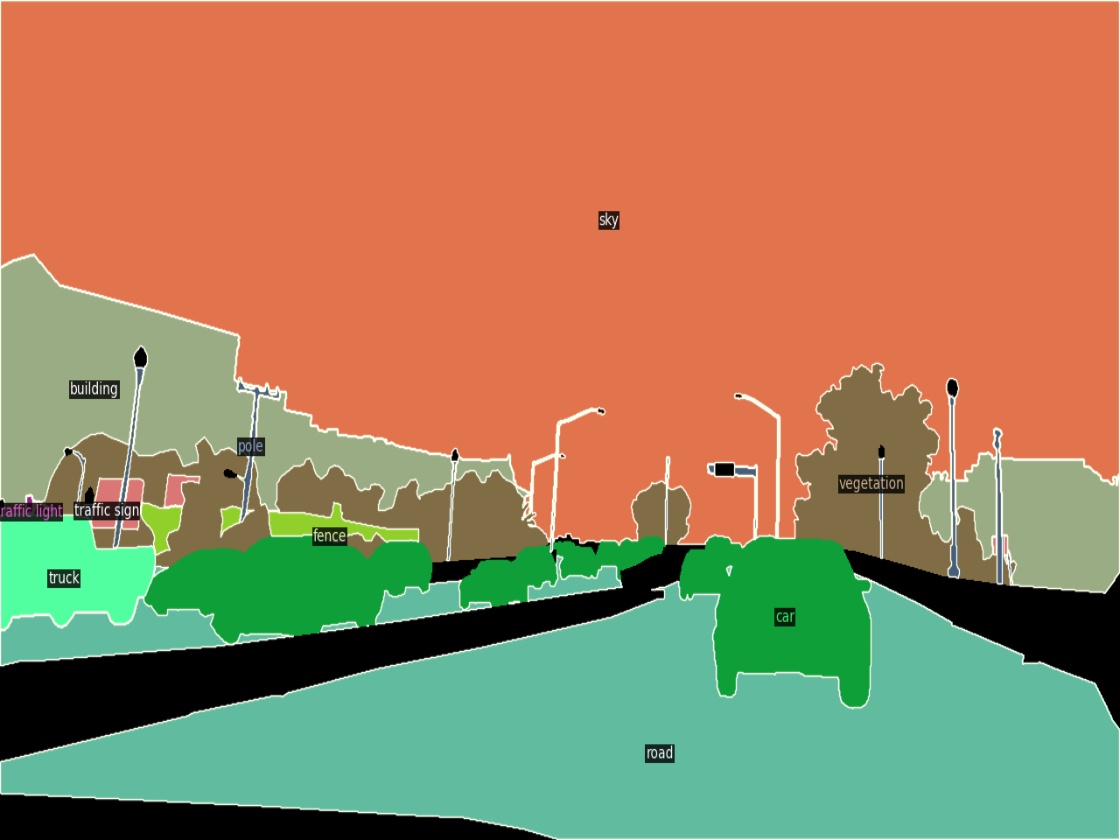} &
    \includegraphics[width=0.16\linewidth]{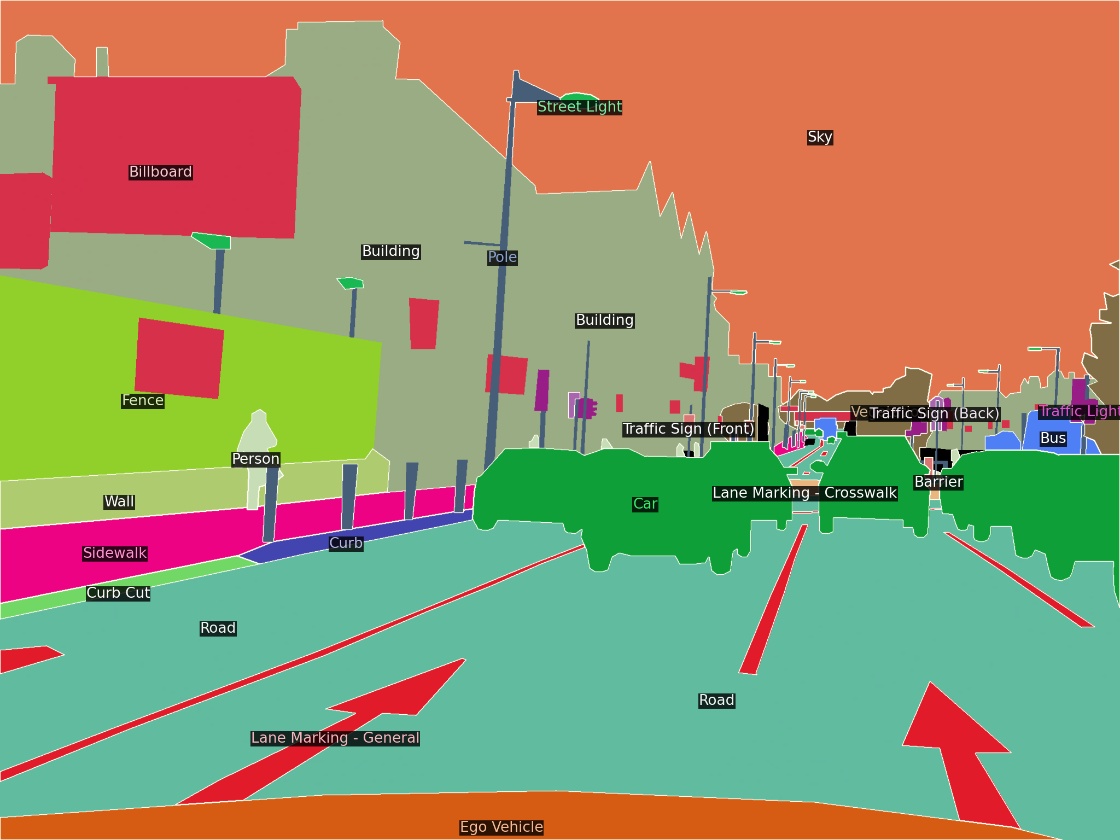} &
    \includegraphics[width=0.16\linewidth]{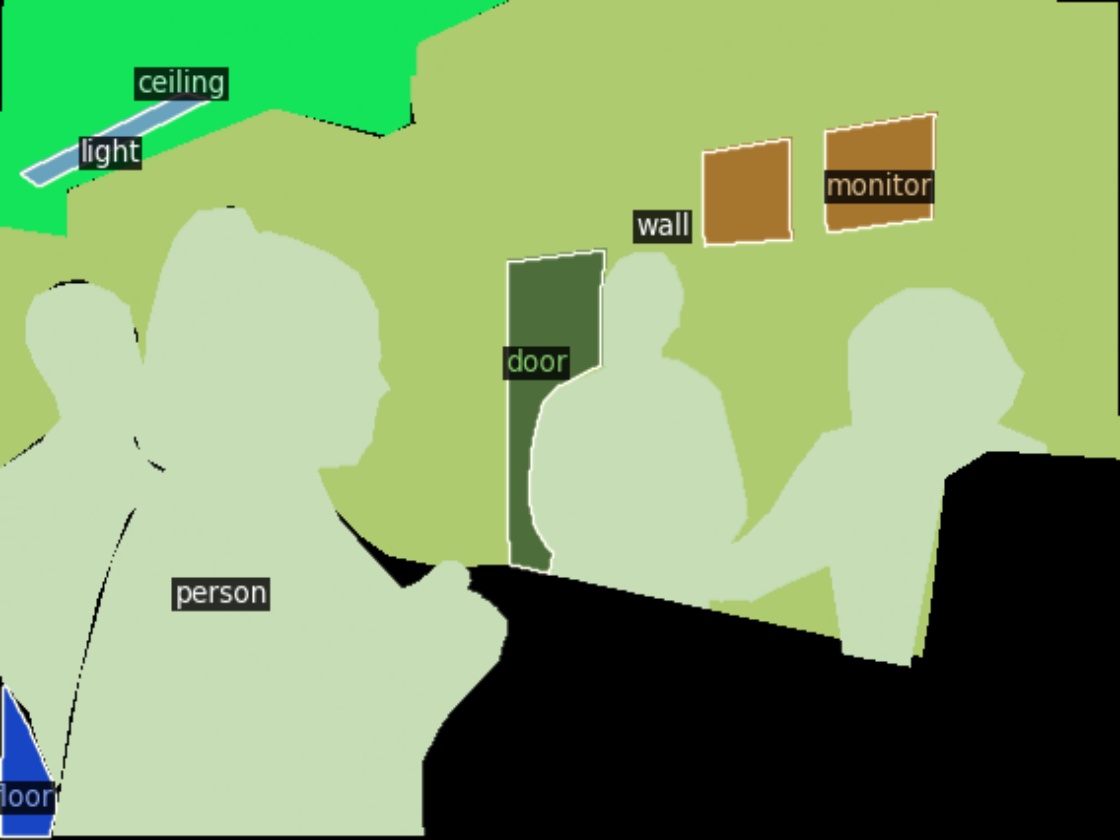} &
    \includegraphics[width=0.16\linewidth]{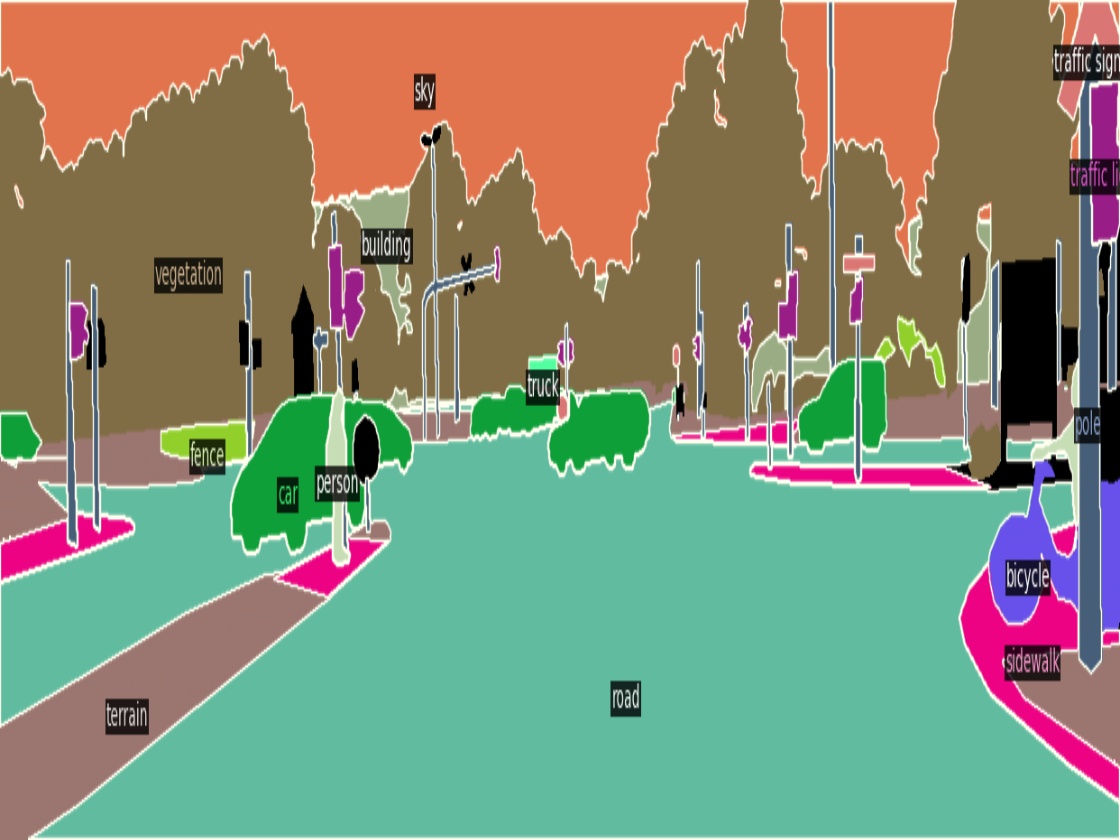} &
    \includegraphics[width=0.16\linewidth]{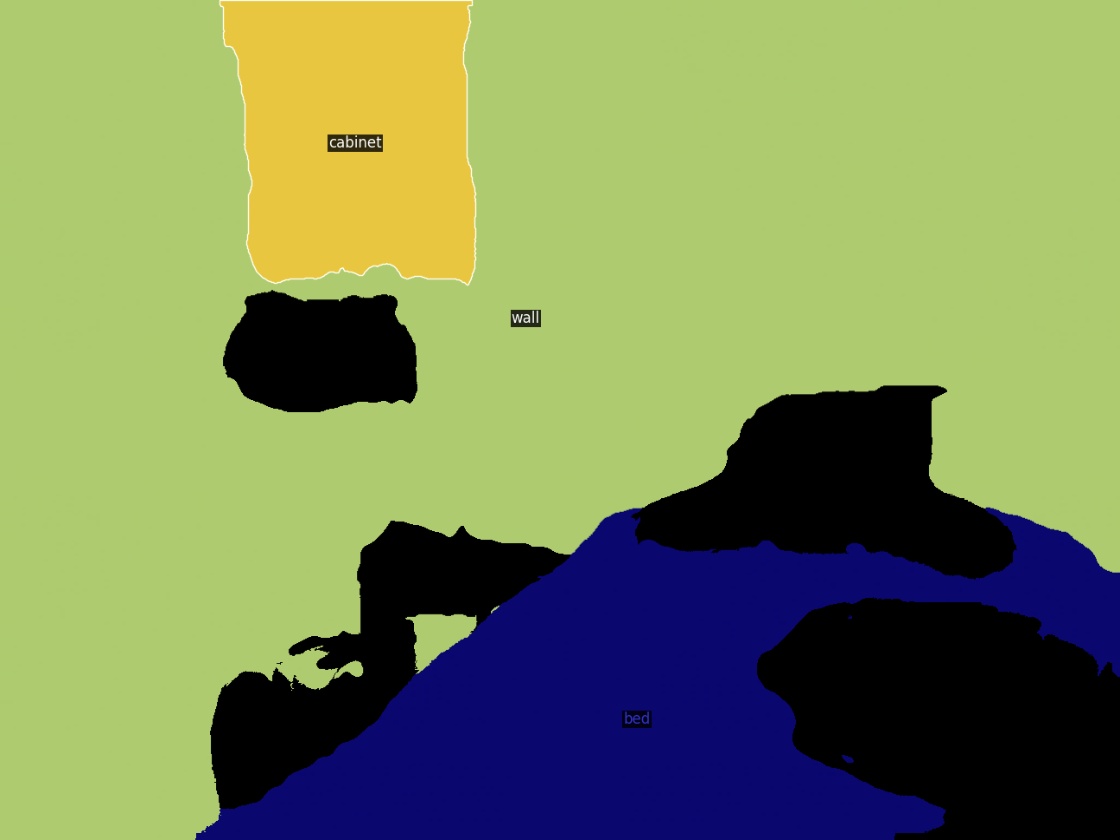}     &

        \includegraphics[width=0.16\linewidth]{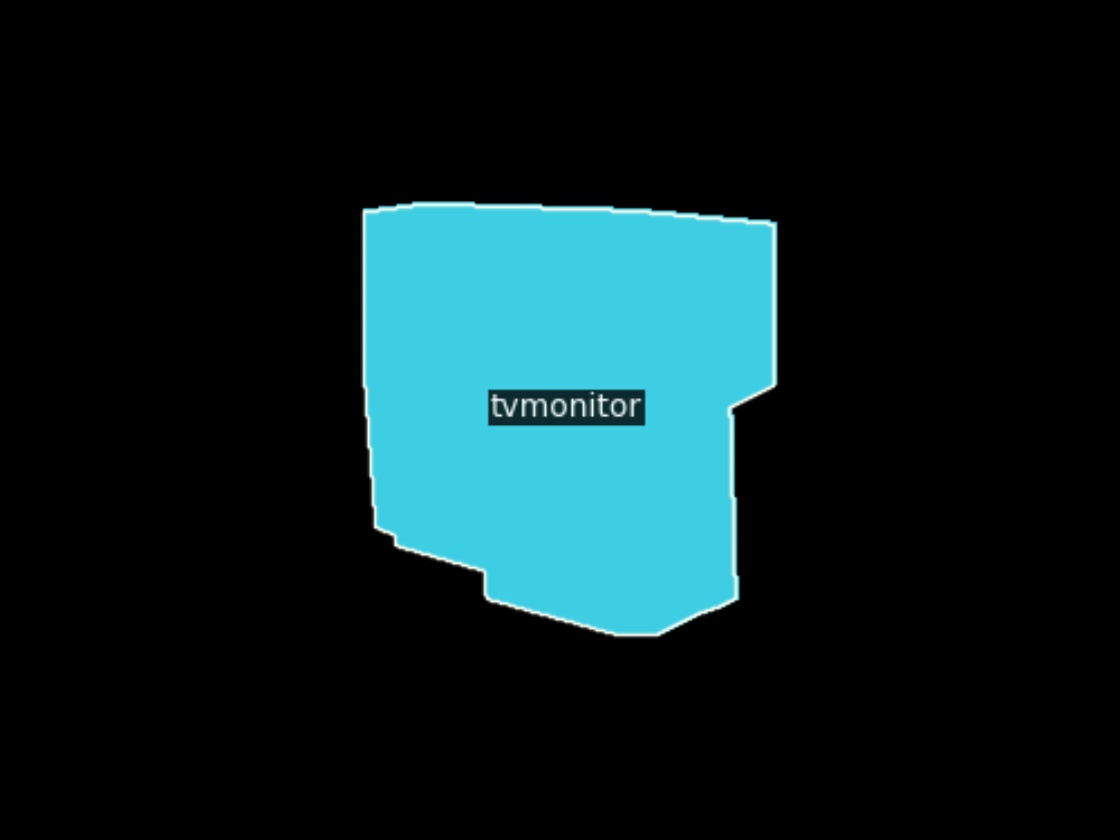} \\

    \rotatebox[origin=l]{90}{General} &
    \includegraphics[width=0.16\linewidth]{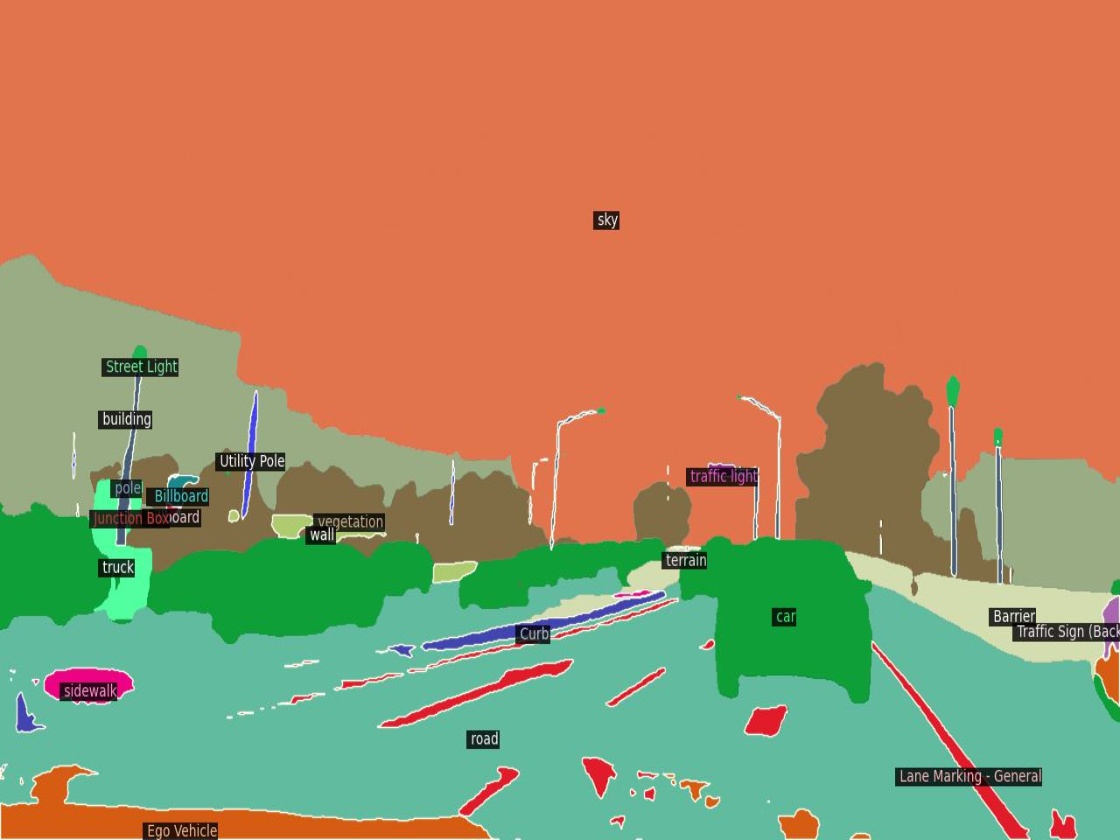} &
    \includegraphics[width=0.16\linewidth]{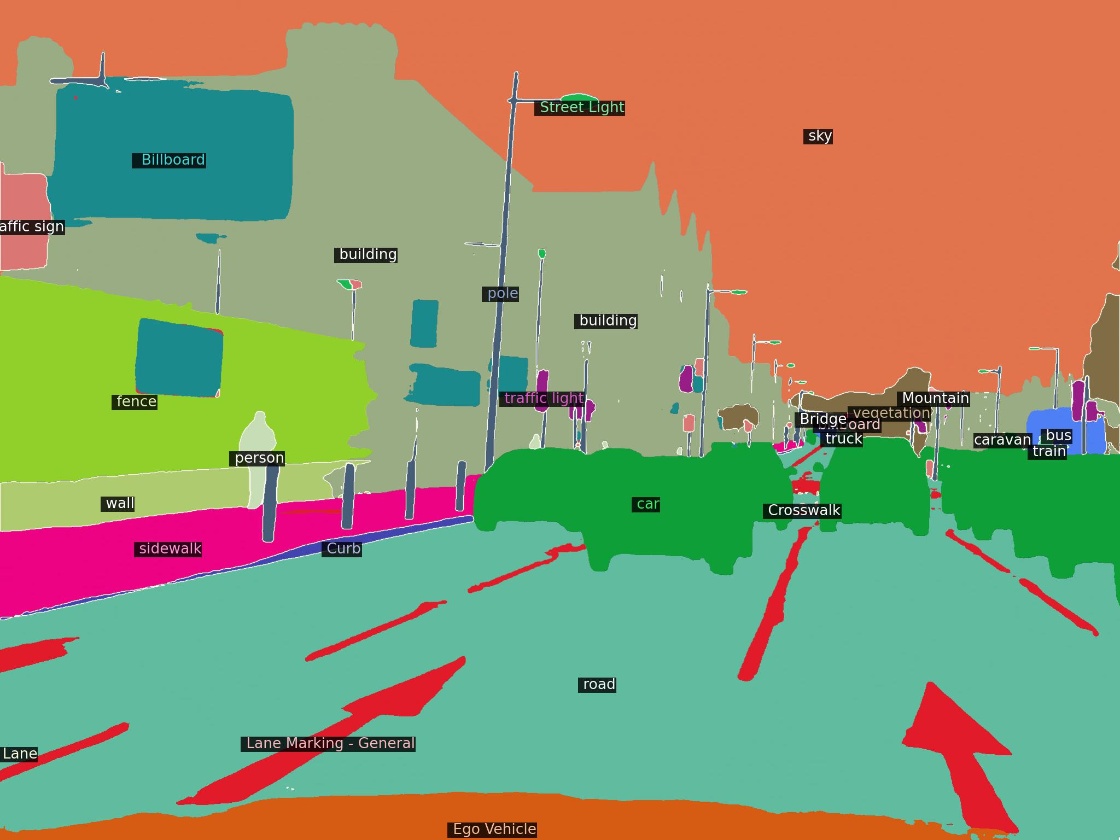} &
    \includegraphics[width=0.16\linewidth]{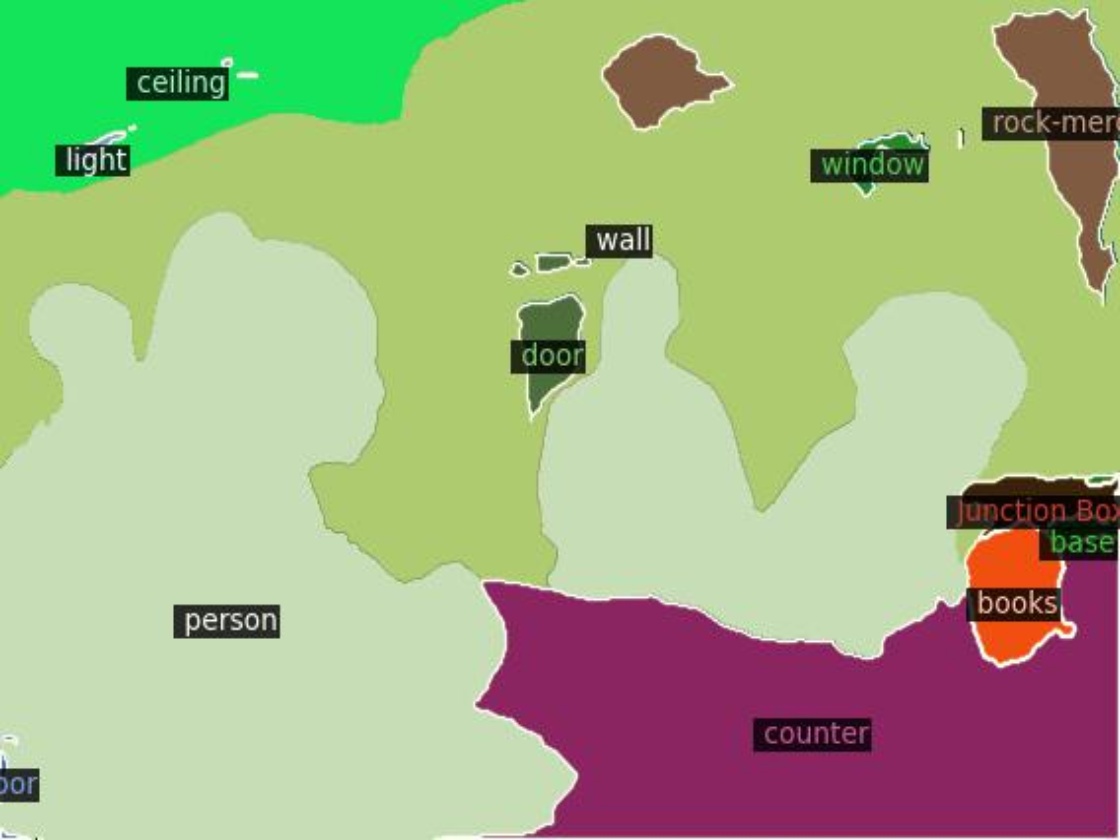} &
    \includegraphics[width=0.16\linewidth]{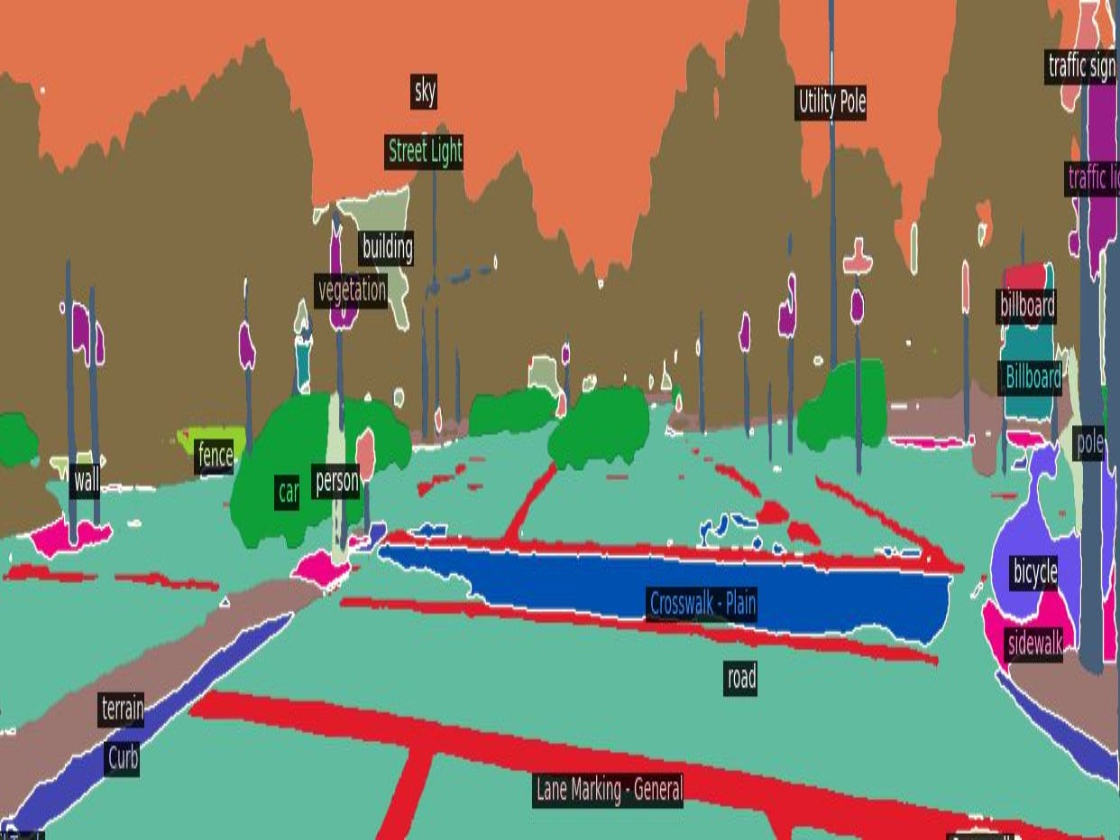} &
    \includegraphics[width=0.16\linewidth]{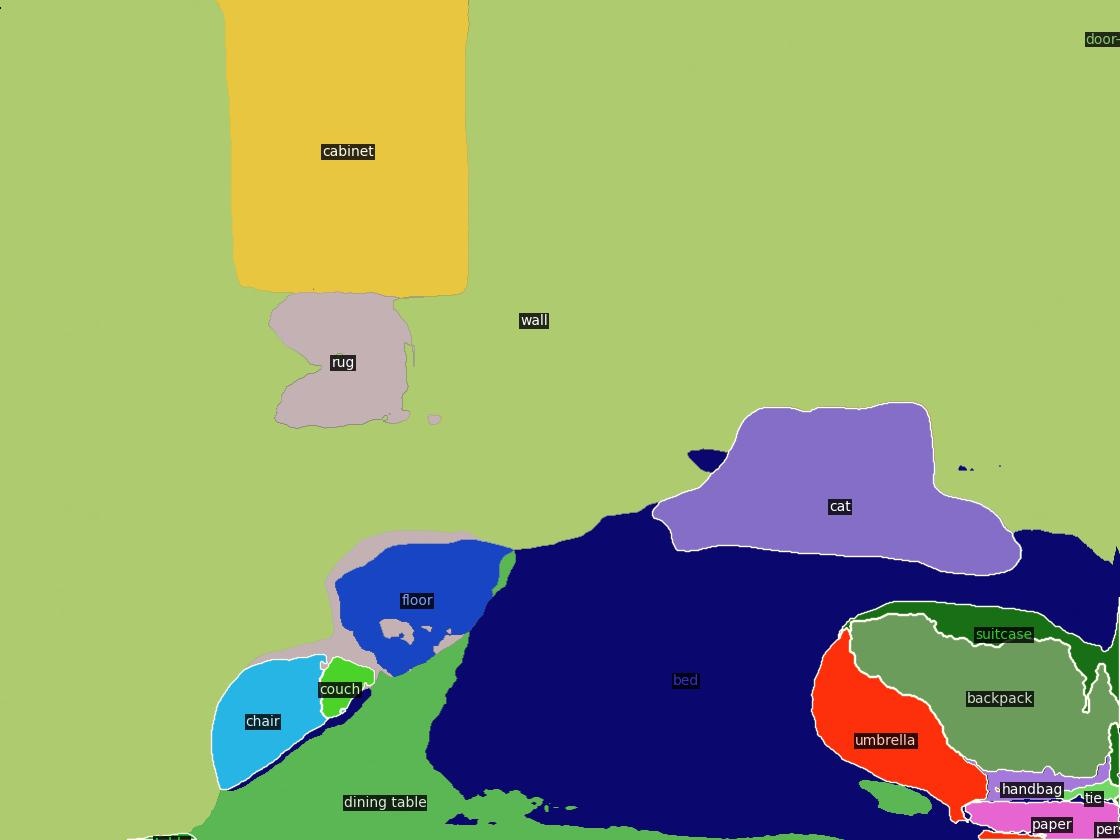}     &

        \includegraphics[width=0.16\linewidth]{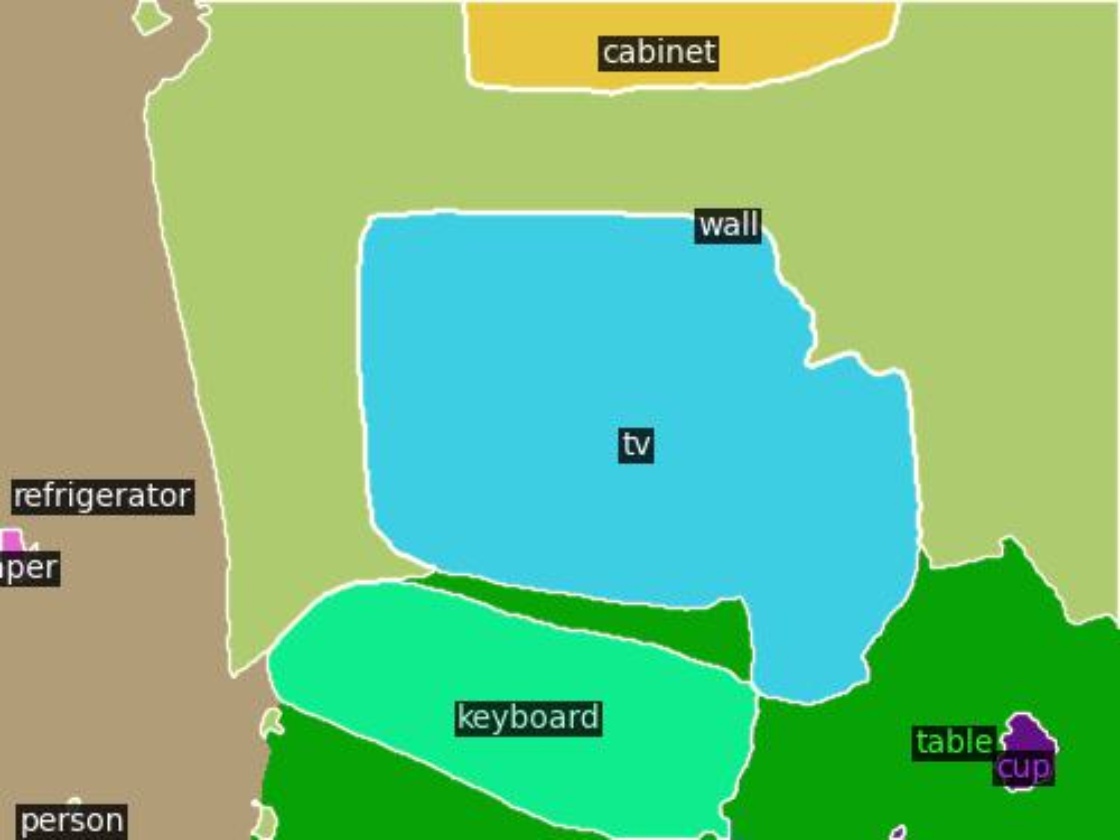} \\
    
    \rotatebox[origin=l]{90}{Specific} &
    \includegraphics[width=0.16\linewidth]{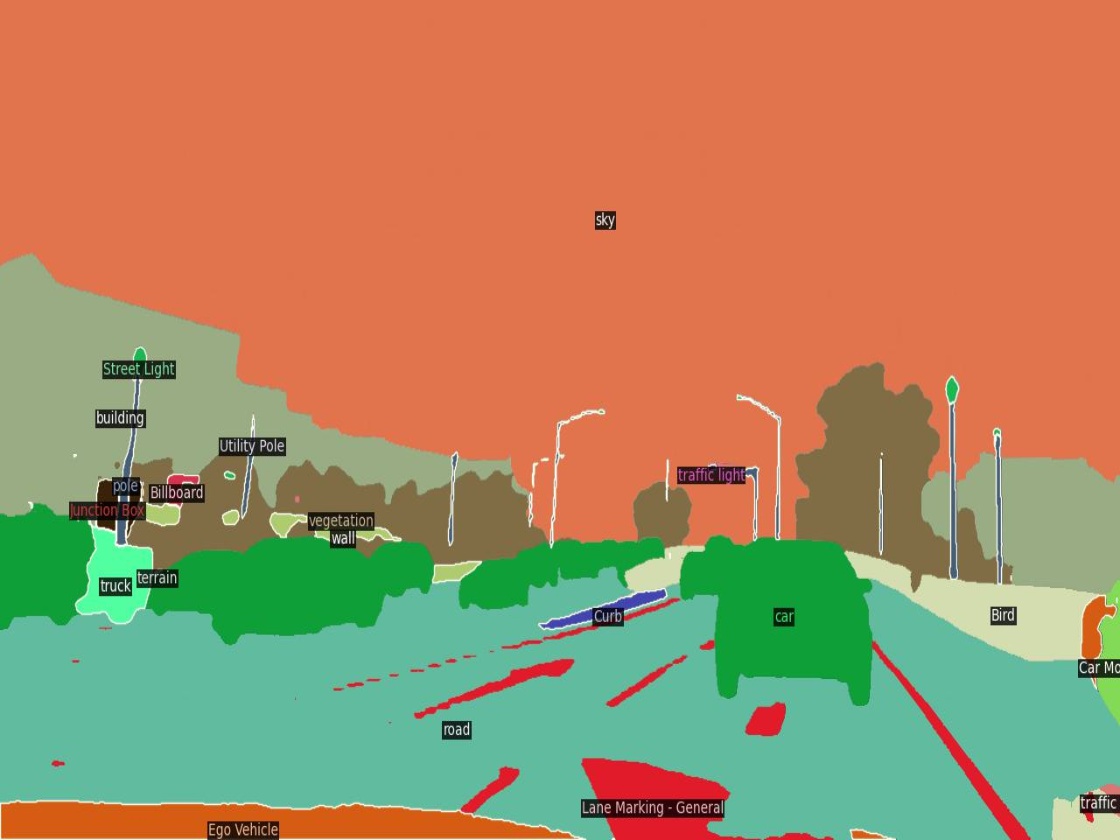} &
    \includegraphics[width=0.16\linewidth]{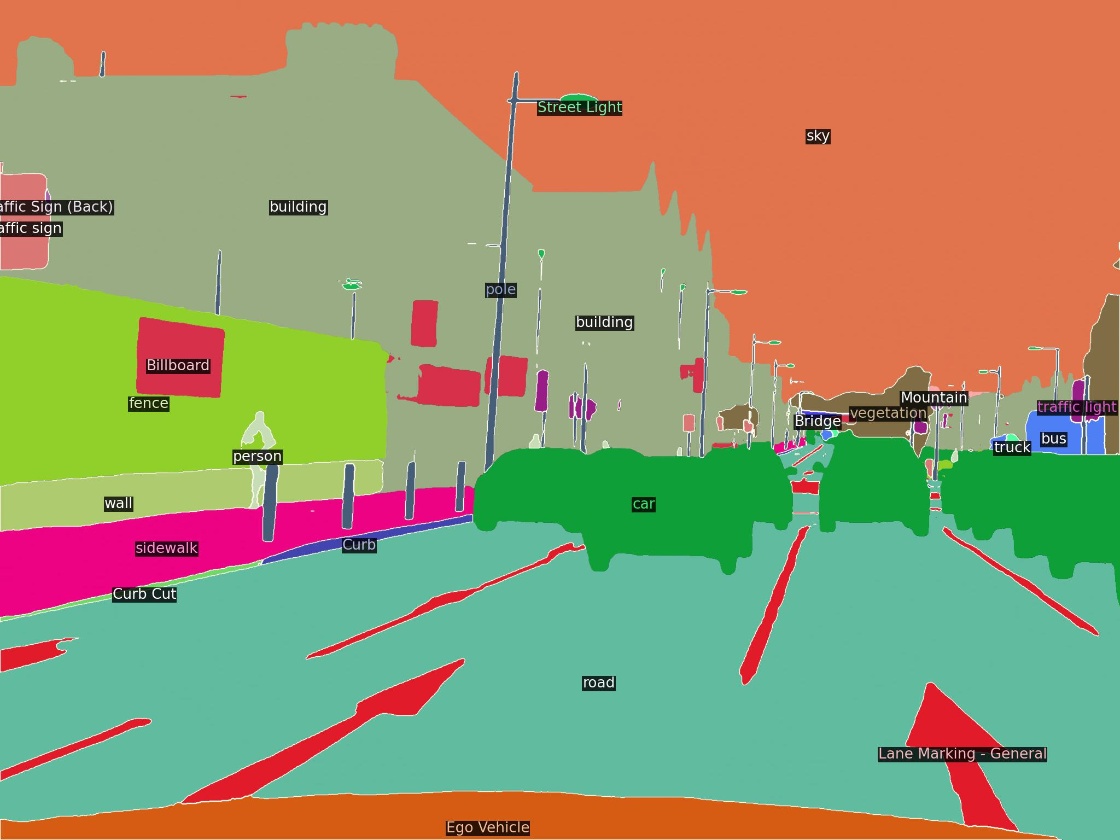} &
    \includegraphics[width=0.16\linewidth]{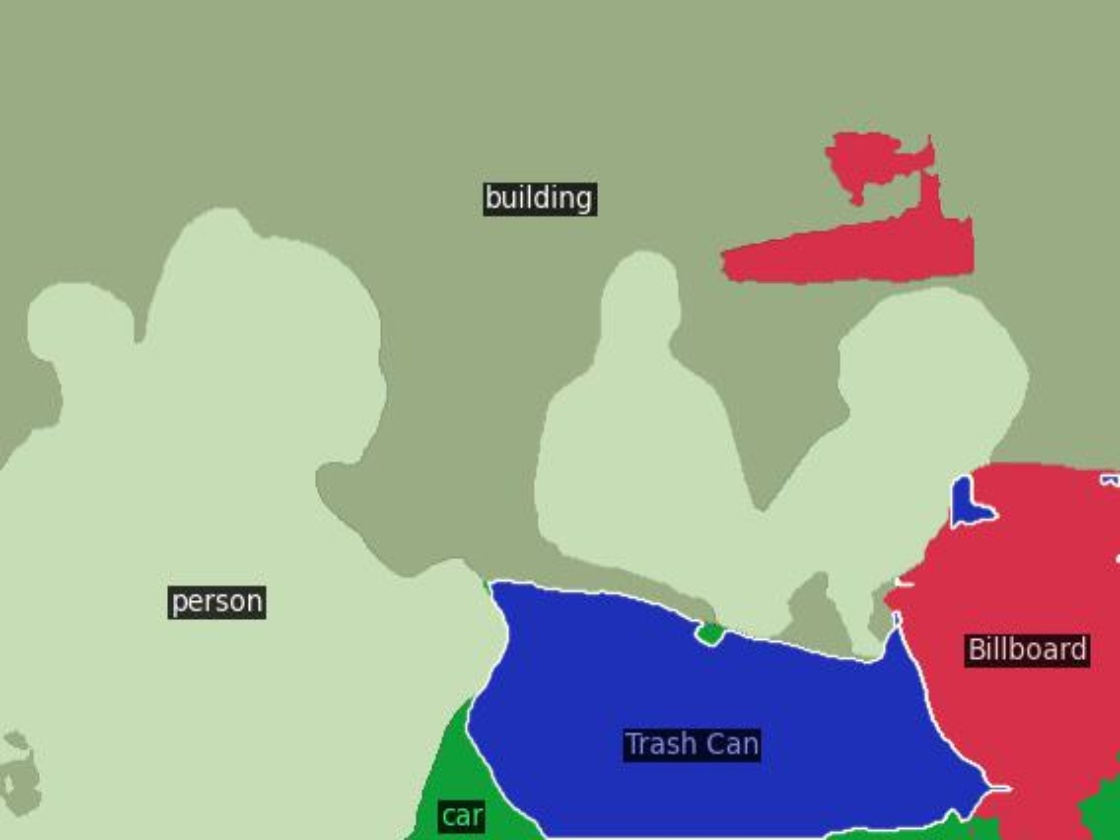} &
    \includegraphics[width=0.16\linewidth]{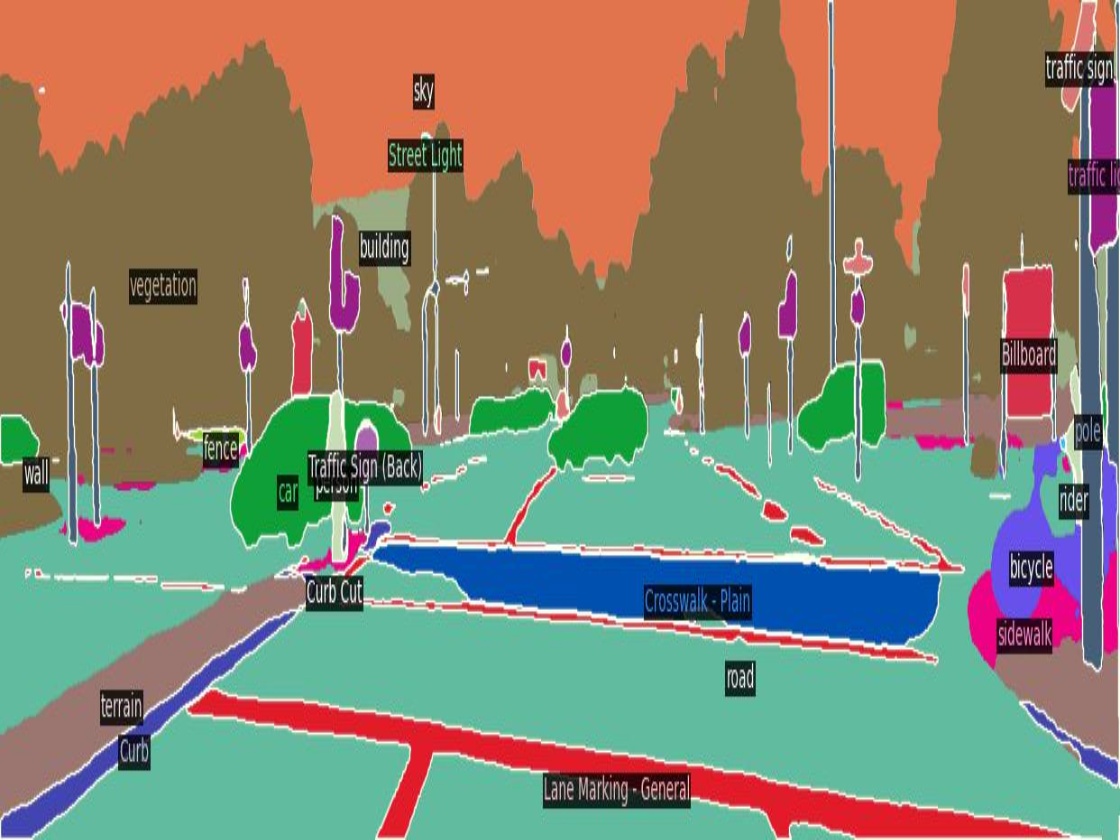} &
    \includegraphics[width=0.16\linewidth]{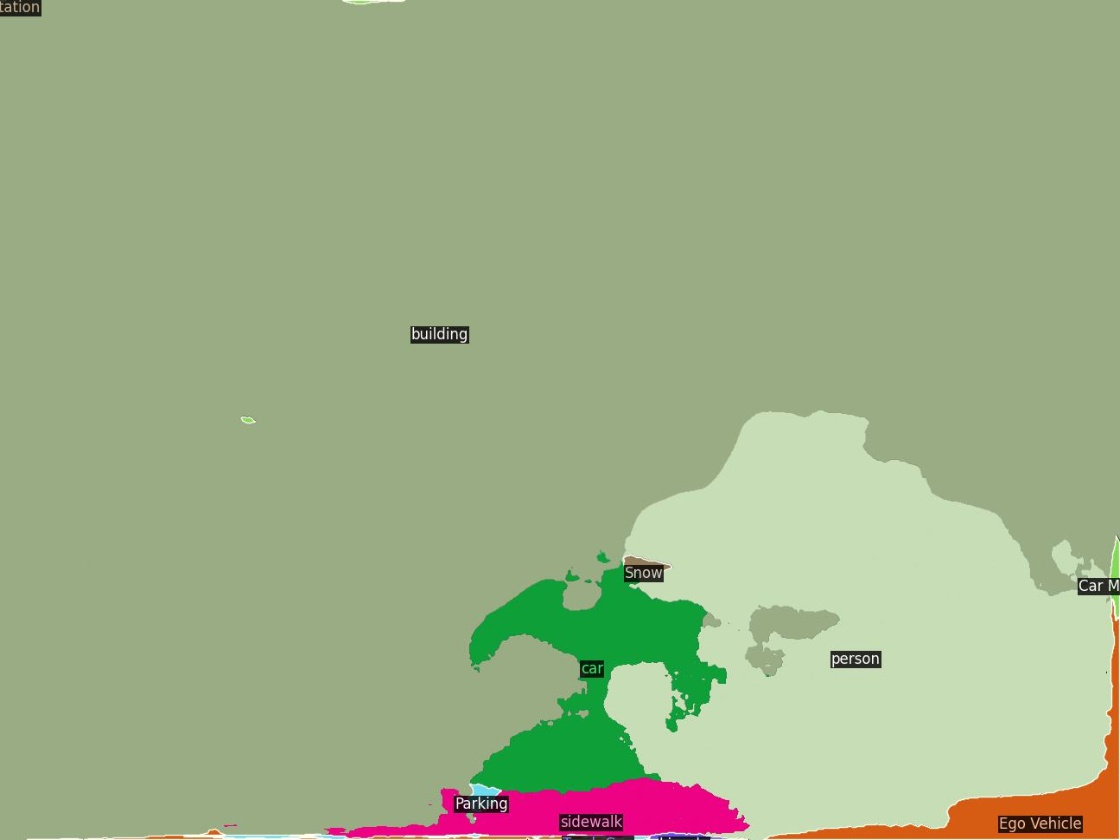}    &

        \includegraphics[width=0.16\linewidth]{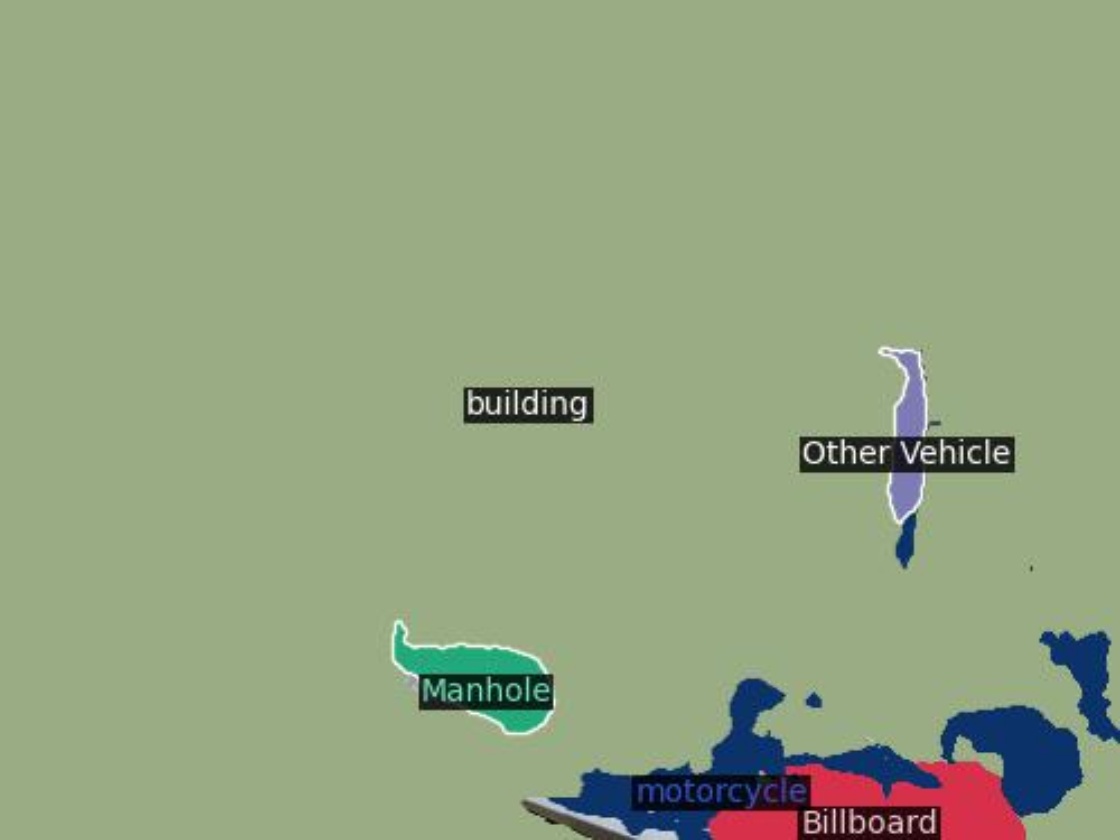} \\
  \end{tabular}
  \caption{Visual comparisons of different training dataset models.}
  \label{fig:5ds}
\end{figure}

Tables \ref{tab:sub1} to \ref{tab:sub4} present a performance comparison between the domain-general and domain-specific models across various training settings and datasets. \autoref{fig:5ds} illustrates the segmentation results of these models on the different datasets.

\textbf{Performance Comparison on Both Trained Datasets.} From \autoref{tab:sub1},  it can be concluded that the performance of the domain-specific model focusing on driving scenes is higher on all trained datasets compared to the domain-general model, despite the domain-general model being trained on five times more samples than the domain-specific model. This is likely attributed to the fact that the domain-specific model needs to predict fewer classes and is less affected by label space conflicts. The model can therefore focus on learning features specific to the particular scene. 

\textbf{Performance Comparison: Unseen Domain-general Model vs. Trained Domain-specific Model.} In \autoref{tab:sub2}, we selected a driving scene dataset that the domain-specific model was trained on, while the domain-specific model was not trained on it, for comparison. As expected, the domain-specific model demonstrates superior performance, and the domain-general model closely follows suit. 

\textbf{Performance Comparison: Trained Domain-general Model vs. Unseen Domain-specific Model.} In \autoref{tab:sub3}, we selected several non-driving scene datasets that the domain-general model was trained on, while the domain-specific model was not trained on it. The domain-specific model fails to predict in non-driving scenes. Compared to \autoref{tab:sub2}, the generalization performance of the domain-specific model in unseen scenes markedly trails behind the domain-general model. Although the domain-specific model may have encountered similar objects in driving scene datasets like \textit{wall}, it still struggles to predict these seen objects well in non-driving scene datasets, as shown in \autoref{fig:5ds}.

\textbf{Performance Comparison on Both Unseen Datasets.} In \autoref{tab:sub4}, we selected datasets that were unseen to both models, including driving scene datasets and non-driving scene datasets. We can observe that the generalization performance of the domain-specific model is slightly better in driving scenes, but its performance in other scenes lags far behind that of the domain-general model. Overall, models trained on more scenes and more data tend to achieve better generalization performance.

\begin{table}[t]
    \footnotesize
    \centering
    \begin{minipage}{0.45\linewidth}
        \centering
        \footnotesize
        \caption{Performance on both trained datasets.}
        \label{tab:sub1}
        \begin{tabular}{c|c|cccc}
            \hline
             & & \multicolumn{4}{c}{Driving scene dataset} \\
            Model & Trained & CS & MPL & BDD & IDD \\
            \hline
            General & \checkmark & 80.7 & 43.7 & 65.5 & 68.6 \\
            Specific & \checkmark & \textbf{82.2} & \textbf{45.7} & \textbf{68.8} & \textbf{71.4} \\ 
        \end{tabular}
    \end{minipage}
    \hfill
    \begin{minipage}{0.45\linewidth}
        \centering
        \footnotesize
        \caption{Unseen domain-general model vs. Trained domain-specific model.}
        \label{tab:sub2}
        \begin{tabular}{c|c|c}
            \hline
             &  & Driving scene dataset \\
            Model & Trained & WD2 \\
            \hline
            General & \usym{2717} & 40.5 \\
            Specific & \checkmark & \textbf{50.2} \\ 
        \end{tabular}
    \end{minipage}
    \vskip\baselineskip
    \begin{minipage}{0.45\linewidth}
        \centering
        \footnotesize
        \caption{Trained domain-general model vs. Unseen domain-specific model.}
        \label{tab:sub3}
        \begin{tabular}{c|c|ccc}
            \hline
             & & \multicolumn{3}{c}{Non-Driving scene dataset} \\
            Model & Trained & SUN & ADE & COCO \\
            \hline
            General & \checkmark & \textbf{47.5} & \textbf{42.0} & \textbf{46.7} \\
            Specific & \usym{2717} & 4.4 & 5.4 & 7.3 \\ 
        \end{tabular}
    \end{minipage}
    \hfill
    \begin{minipage}{0.47\linewidth}
        \centering
        \footnotesize
        \caption{Performance on both unseen datasets.}
        \label{tab:sub4}
        \begin{tabular}{c|c|cccc}
            \hline
             & & \multicolumn{2}{c}{Driving} & \multicolumn{2}{c}{Non-Driving} \\
            Model & Trained & CV & KT & VOC & SN \\
            \hline
            General & \usym{2717} & 71.1 & 65.5 & \textbf{69.2} & \textbf{40.6} \\
            Specific & \usym{2717} & \textbf{74.4} & \textbf{67.4} & 28.7 & 8.1 \\ 
        \end{tabular}
    \end{minipage}

\end{table}

% \clearpage
% \section{Sensitivity Analysis of the Number of Semantic Atoms}
% \label{apdx:sense}

% \autoref{tb:sense} lists the performance of our methods with different number of semantic atoms. 

% \begin{table}[h!]
% \centering
% \caption{Sensitivity analysis of the number of semantic atoms}
% \label{tb:sense}
% \footnotesize
% \begin{tabular}{c|cccccccc}
% \hline
% Number of semantic atoms & CS & MPL & SUN & BDD & IDD & ADE & COCO & Mean\\
% \hline
% 231 & \textbf{80.7} & 43.7 & \textbf{47.5} & \textbf{65.5} & \textbf{68.6} & \textbf{42.0} & \textbf{46.7} & \textbf{56.4}\\
% 294 & 80.4 & \textbf{43.8} & \textbf{47.5} & 65.2 & 68.3 & 37.4 & 40.9 & 54.8\\
% \hline
% \end{tabular}
% \end{table}

\clearpage
\section{Visualization}
\label{apdx:vis}

\begin{figure}[h!]
  \centering
  \footnotesize
  \setlength{\tabcolsep}{0.02cm} % 设置列之间的间距
  \begin{tabular}{lccccccc}
      & CS & MPL & SUN & BDD & IDD & ADE & COCO \\
    \rotatebox[origin=l]{90}{Image} &
    \includegraphics[width=0.13\linewidth]{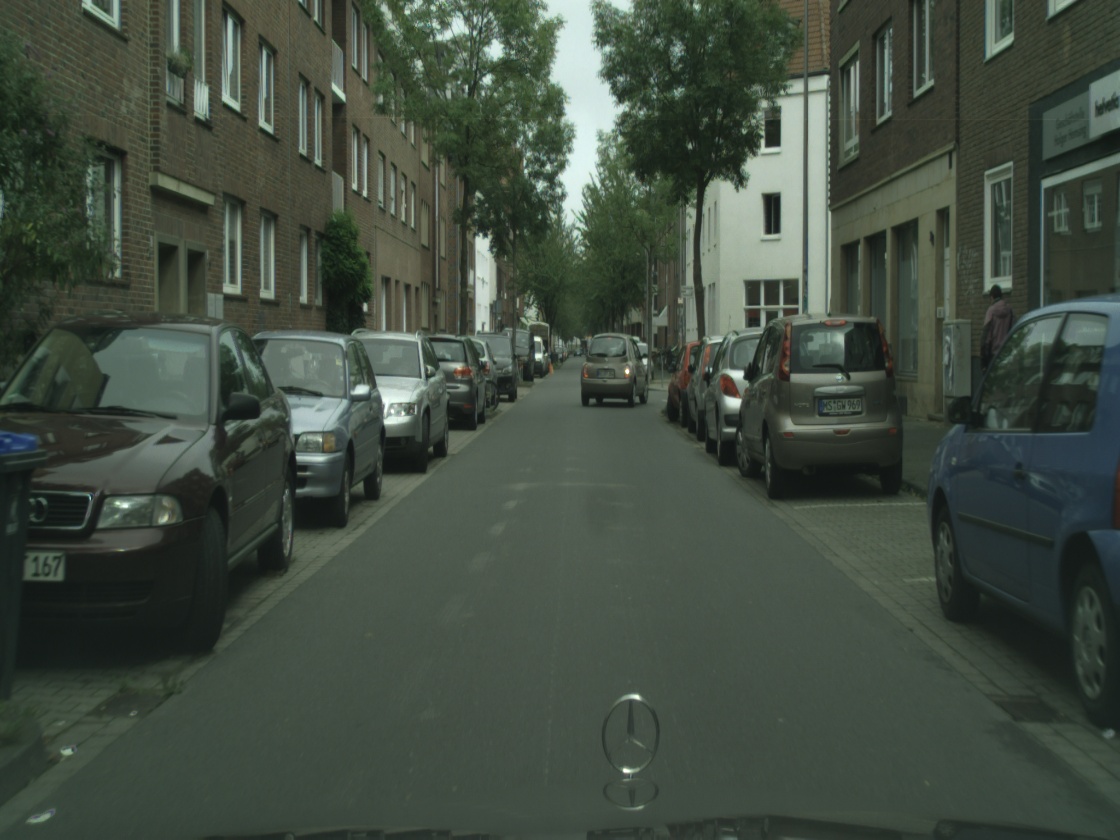} &
    \includegraphics[width=0.13\linewidth]{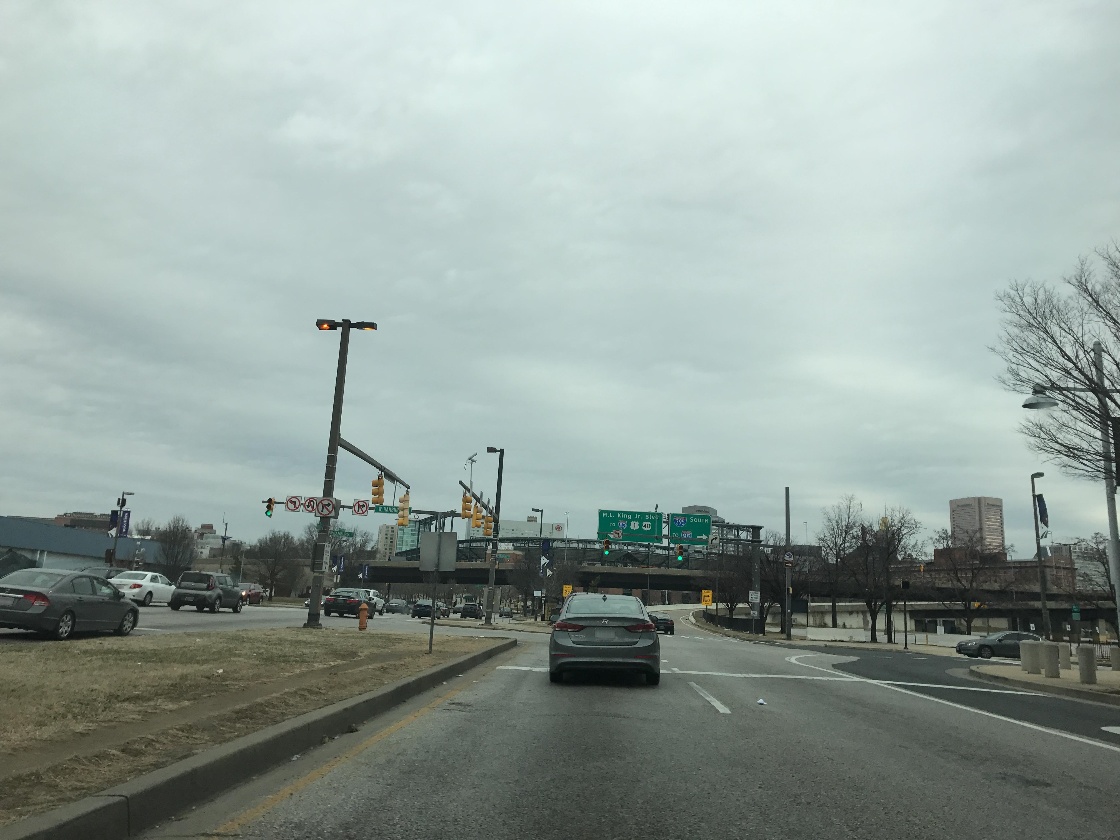} &
    \includegraphics[width=0.13\linewidth]{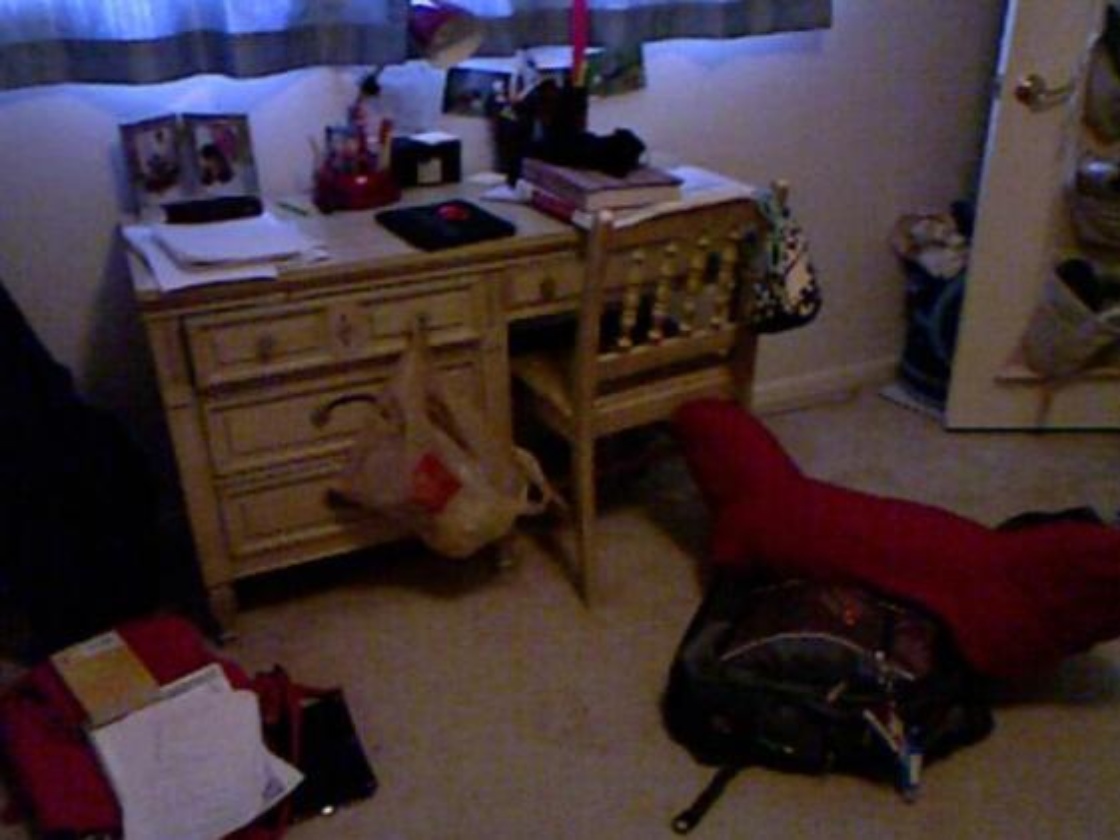} &
    \includegraphics[width=0.13\linewidth]{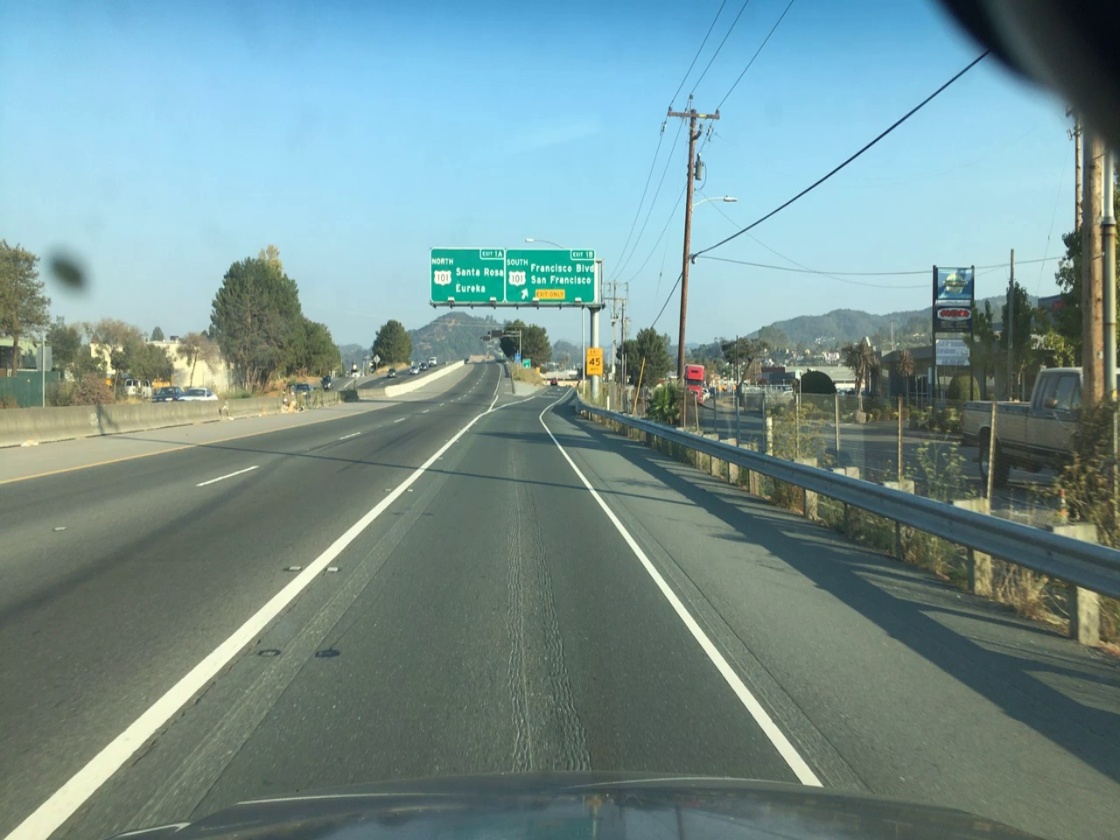} &
    \includegraphics[width=0.13\linewidth]{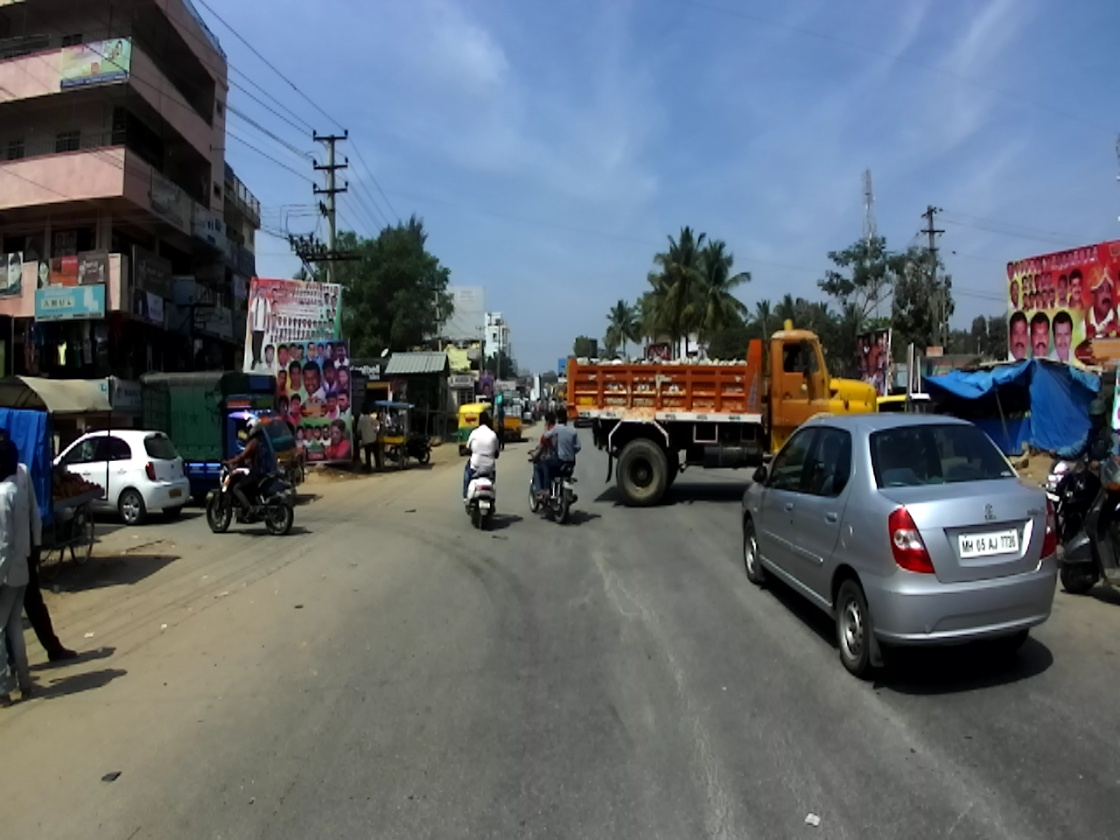} &
    \includegraphics[width=0.13\linewidth]{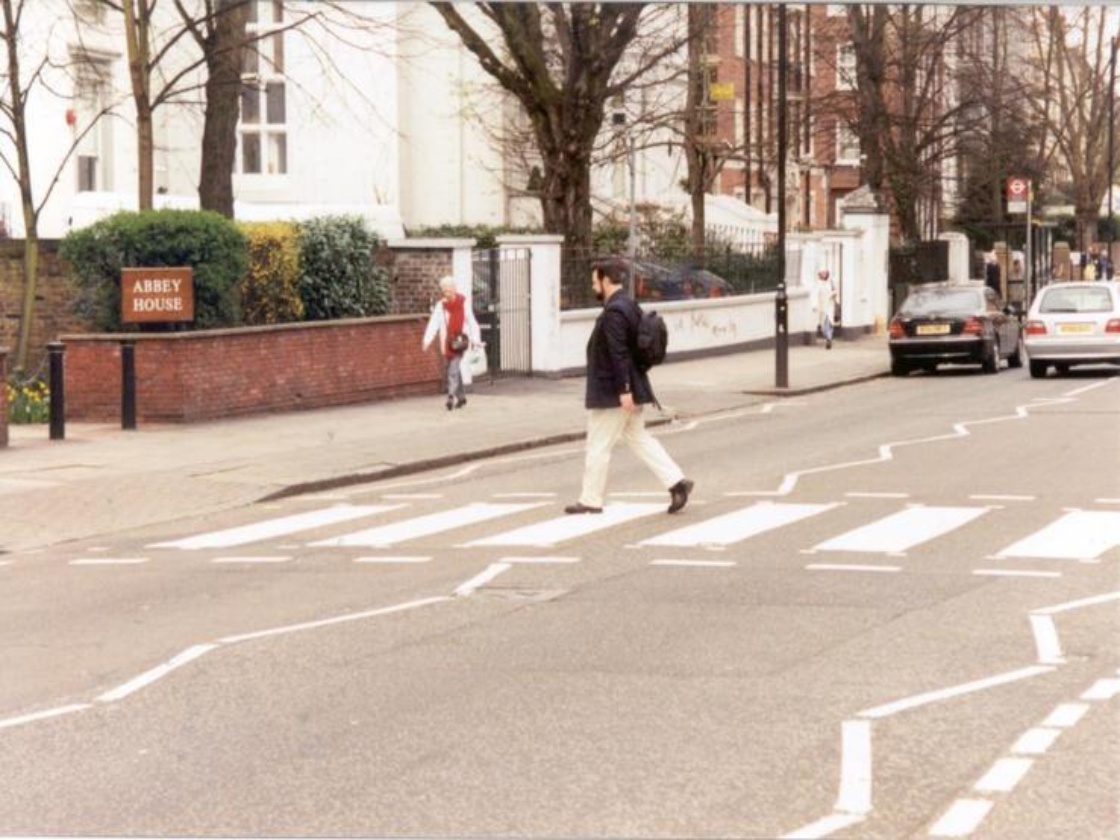} &
    \includegraphics[width=0.13\linewidth]{image/000000000785_resize.jpg} \\
    \rotatebox[origin=l]{90}{GT} &
    \includegraphics[width=0.13\linewidth]{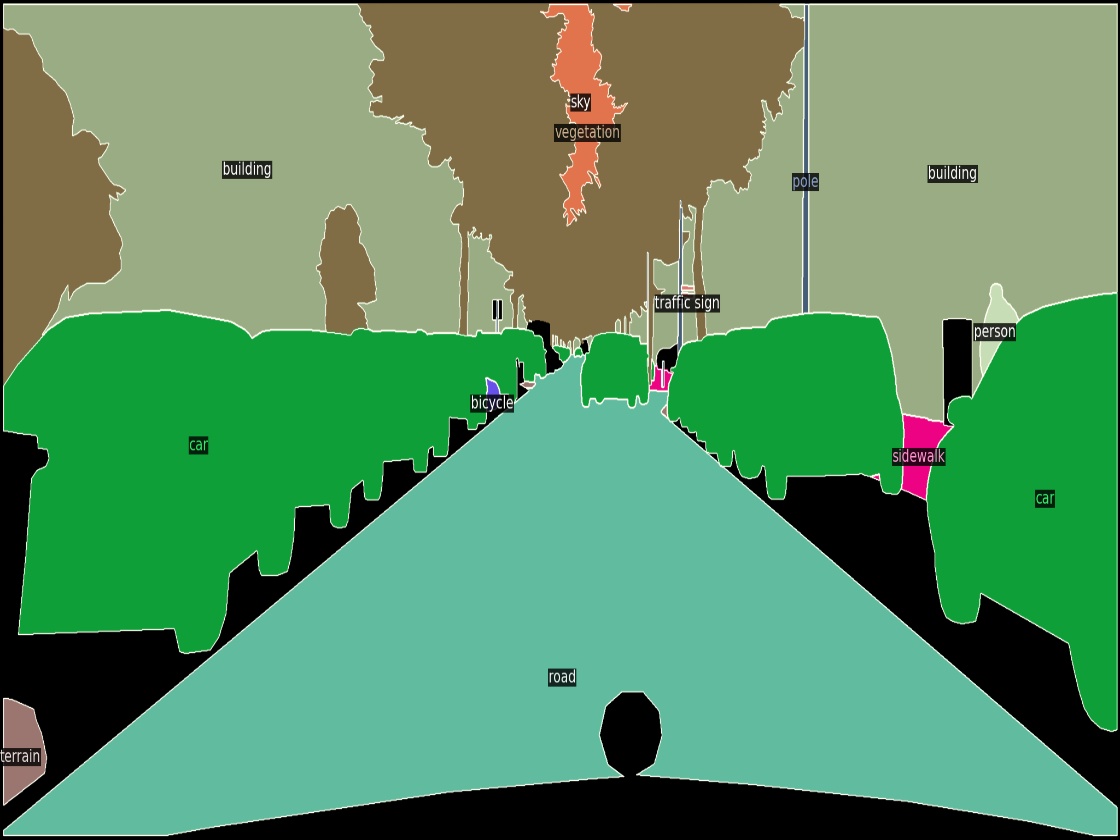} &
    \includegraphics[width=0.13\linewidth]{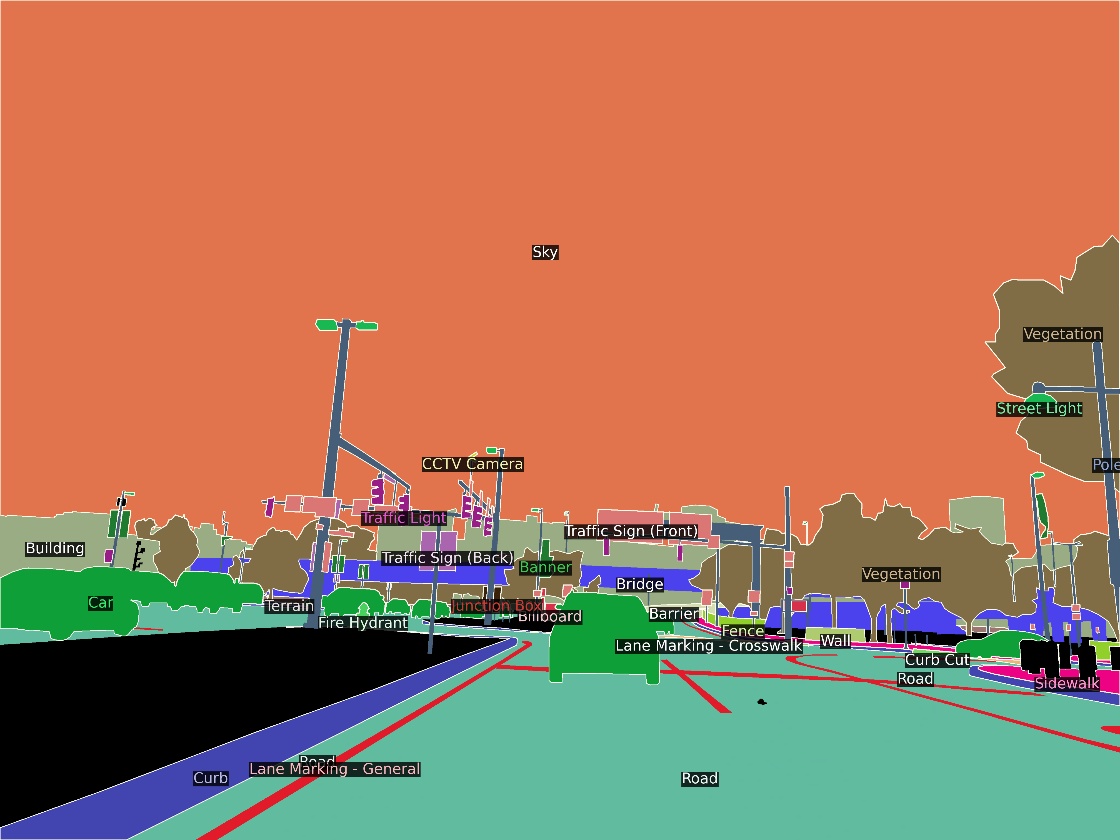} &
    \includegraphics[width=0.13\linewidth]{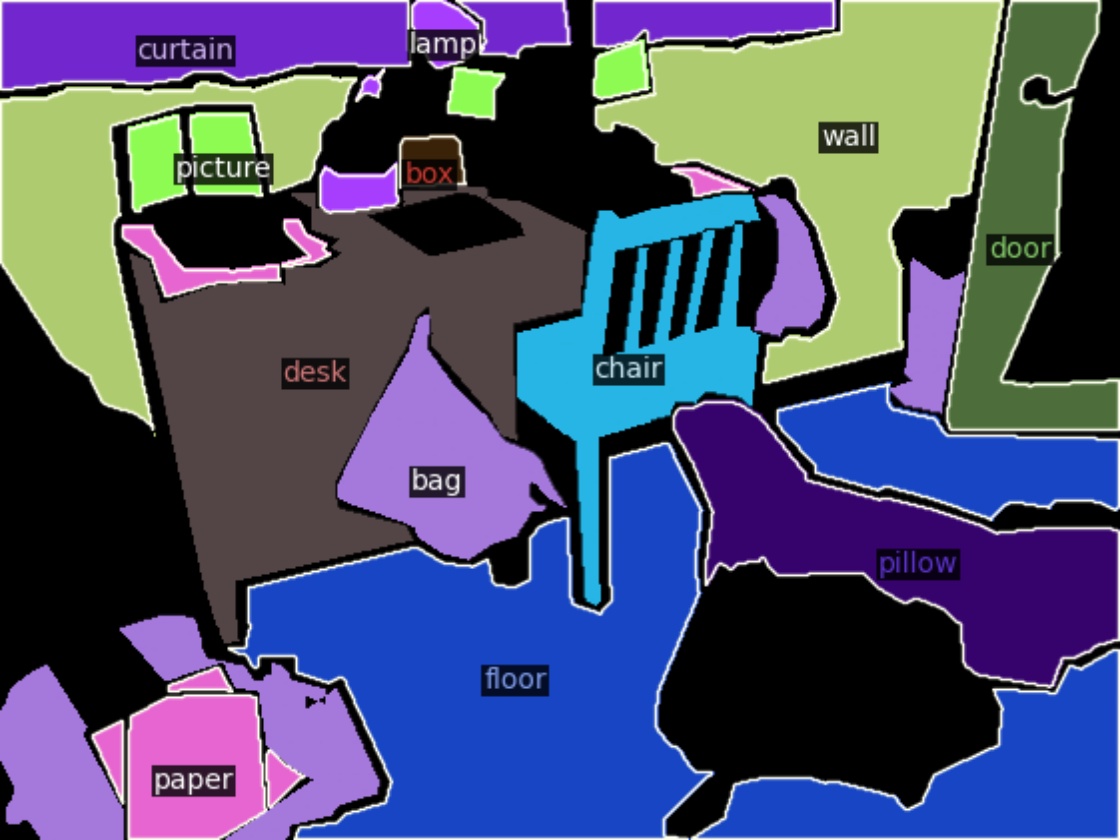} &
    \includegraphics[width=0.13\linewidth]{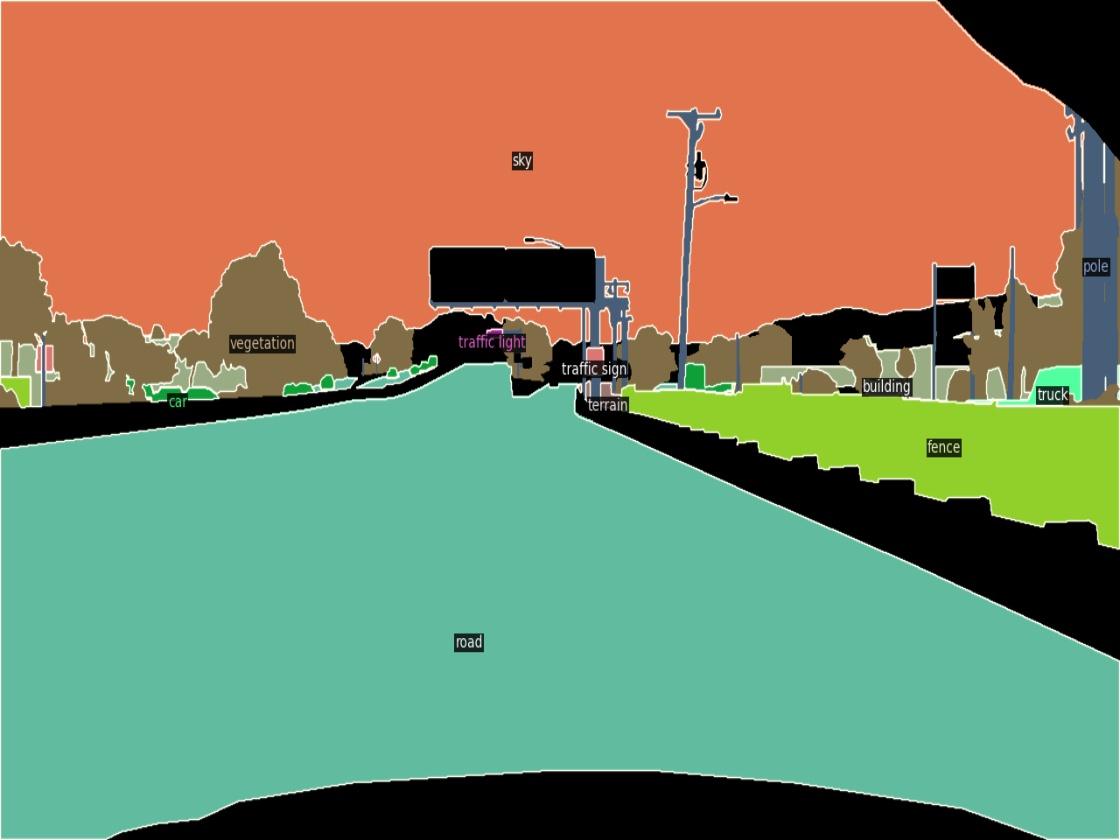} &
    \includegraphics[width=0.13\linewidth]{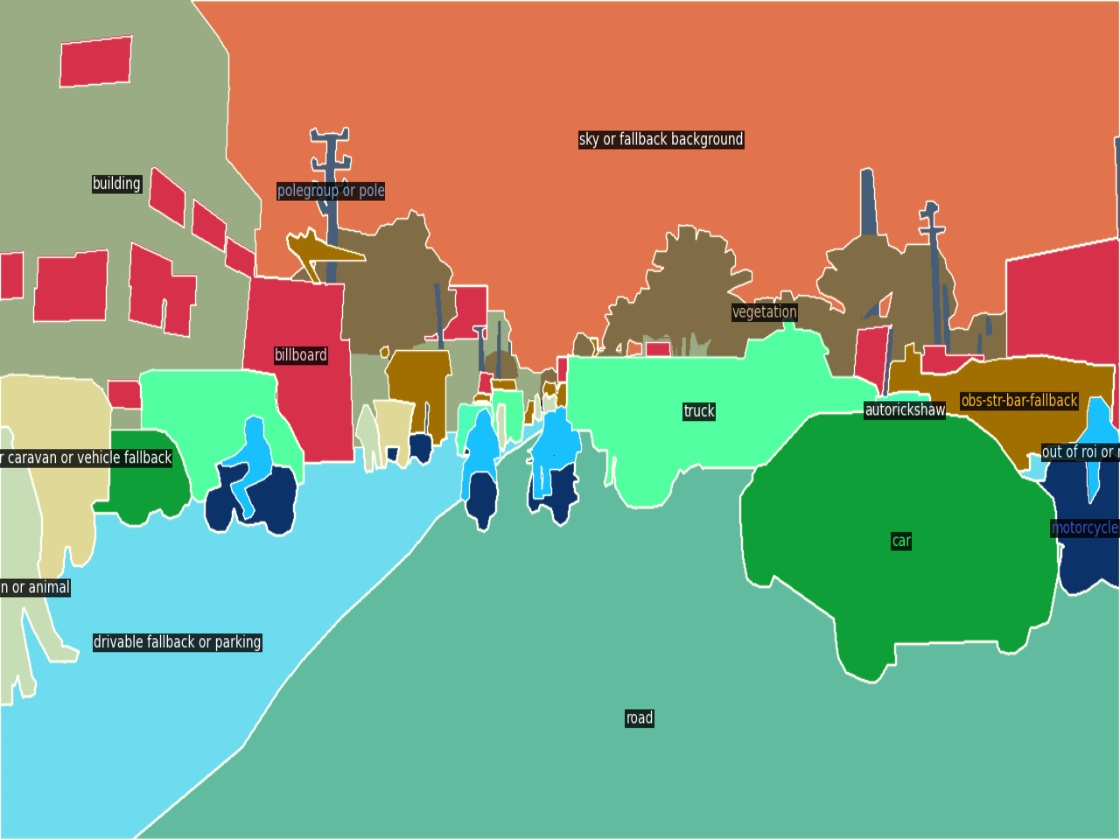} &
    \includegraphics[width=0.13\linewidth]{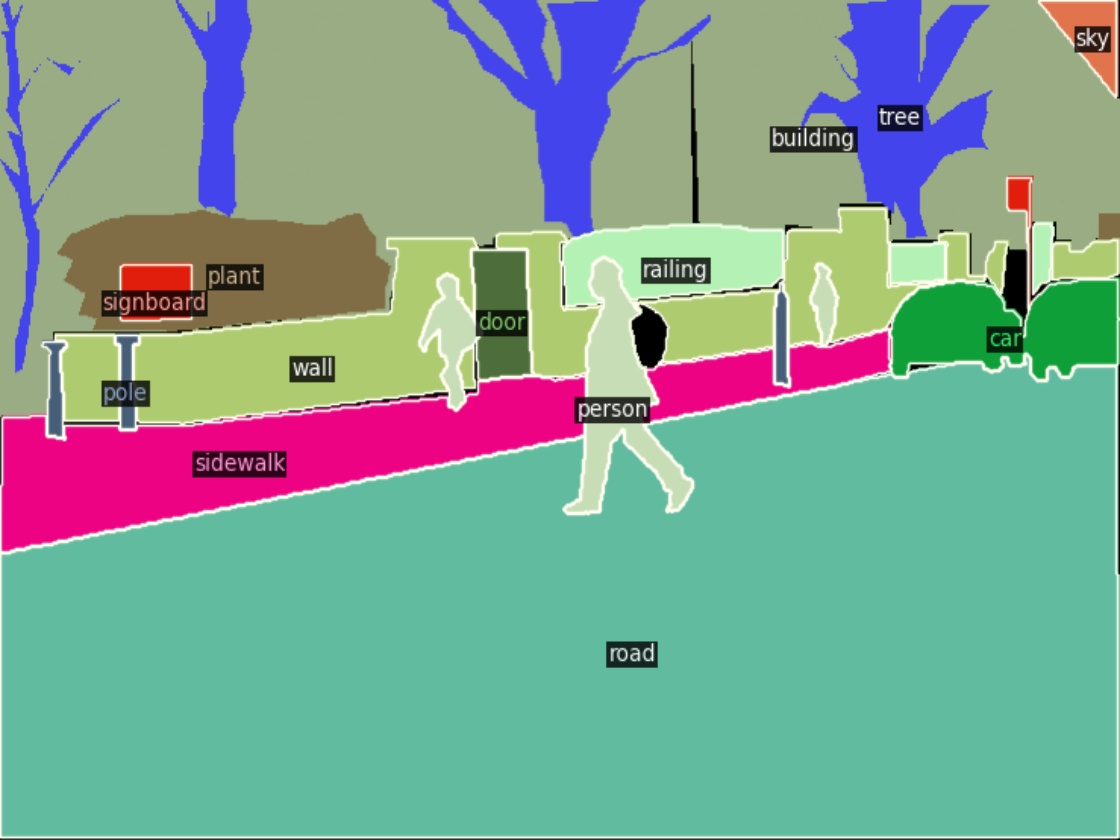} &
    \includegraphics[width=0.13\linewidth]{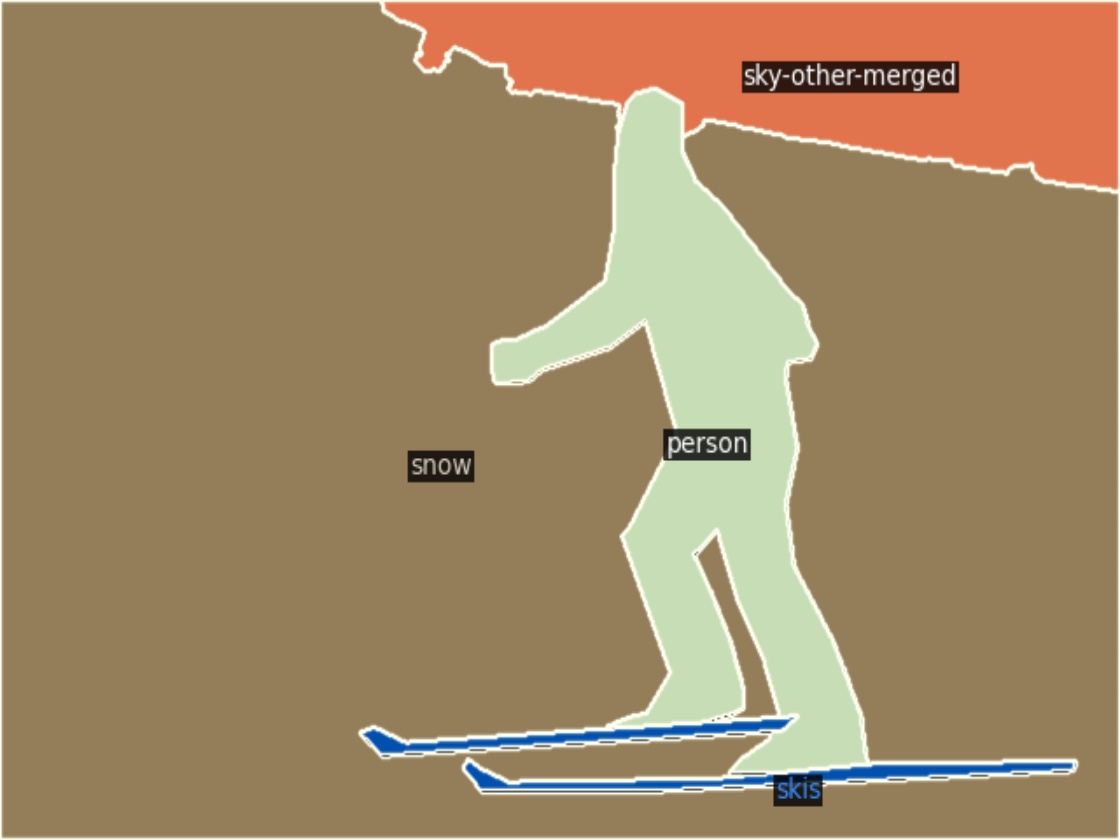} \\
    \rotatebox[origin=l]{90}{Ours} &
    \includegraphics[width=0.13\linewidth]{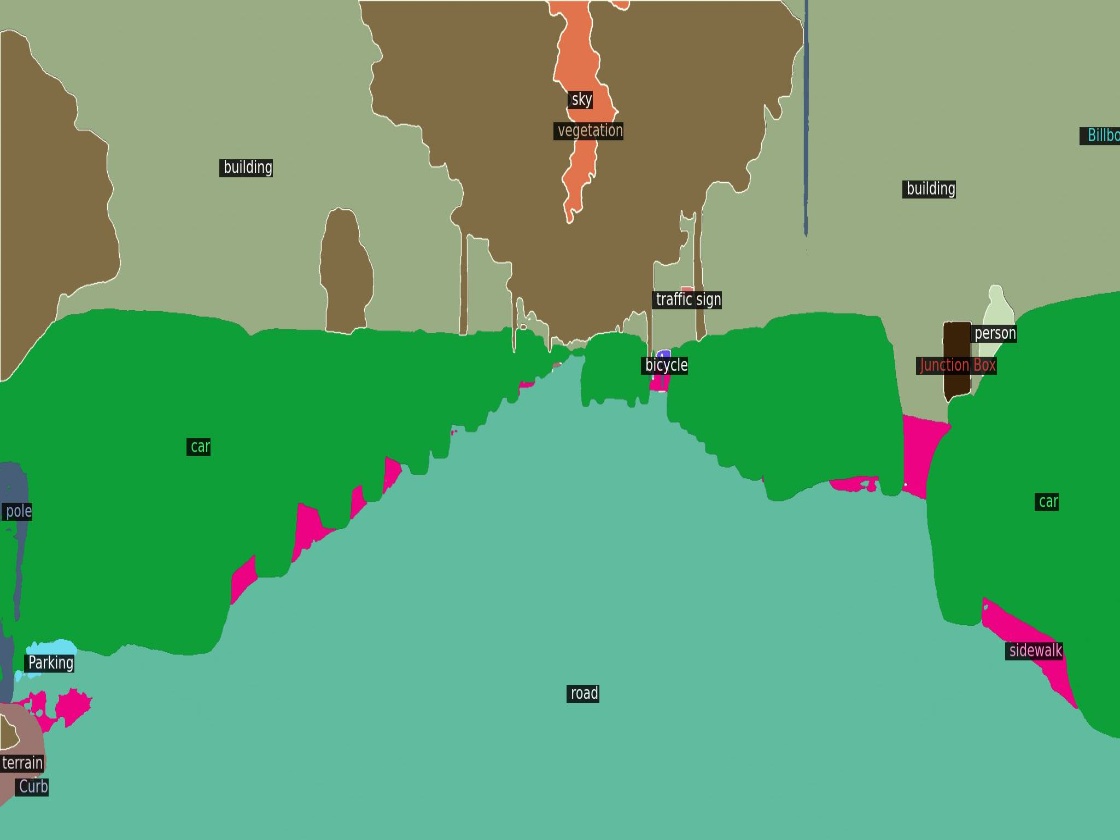} &
    \includegraphics[width=0.13\linewidth]{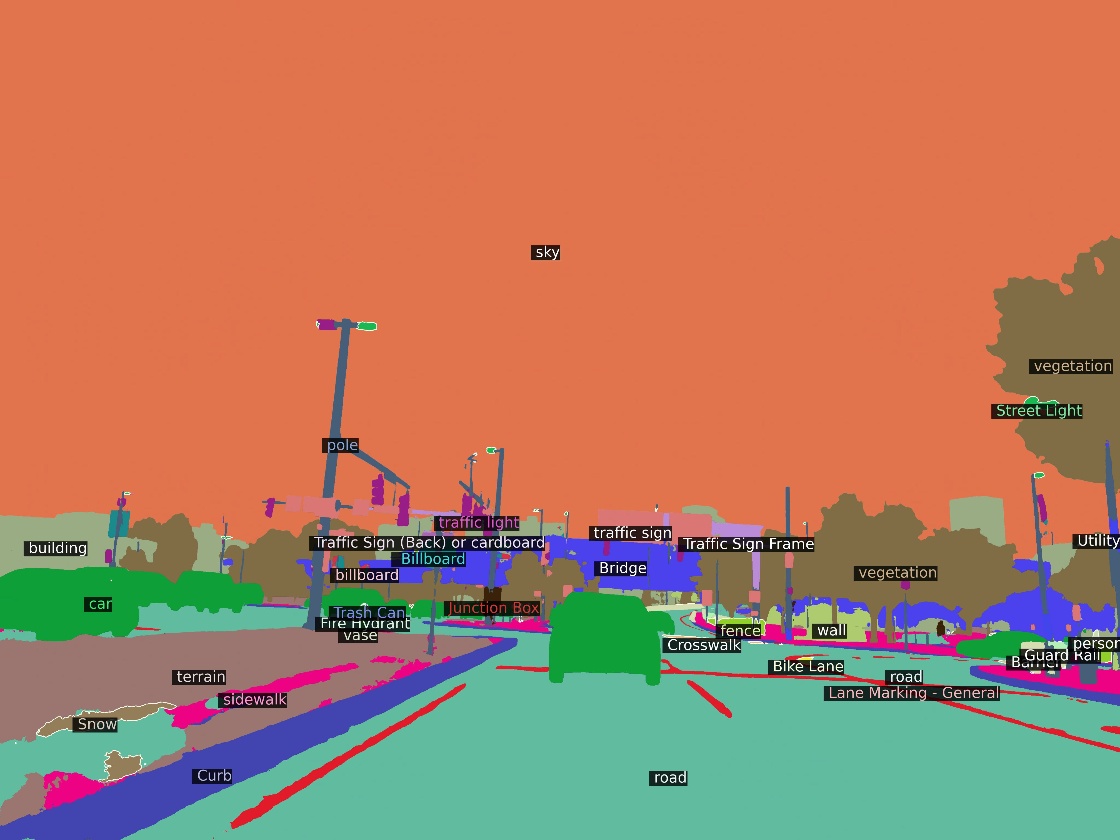} &
    \includegraphics[width=0.13\linewidth]{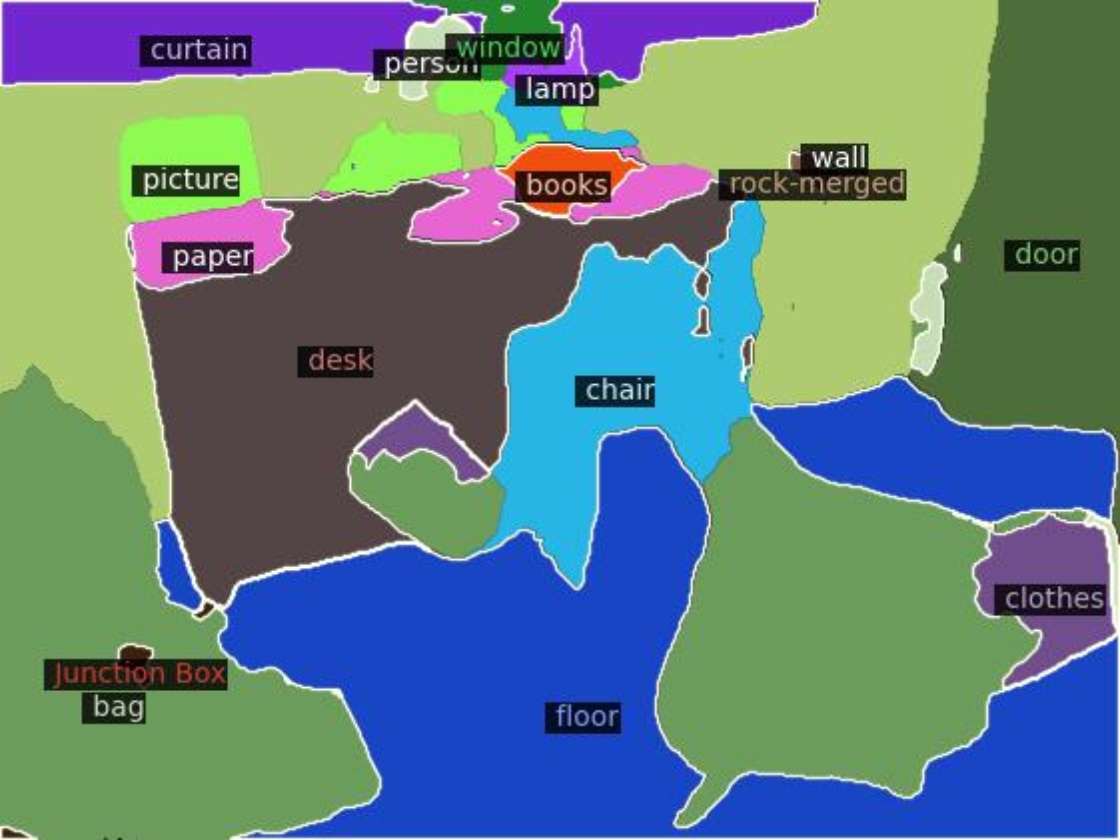} &
    \includegraphics[width=0.13\linewidth]{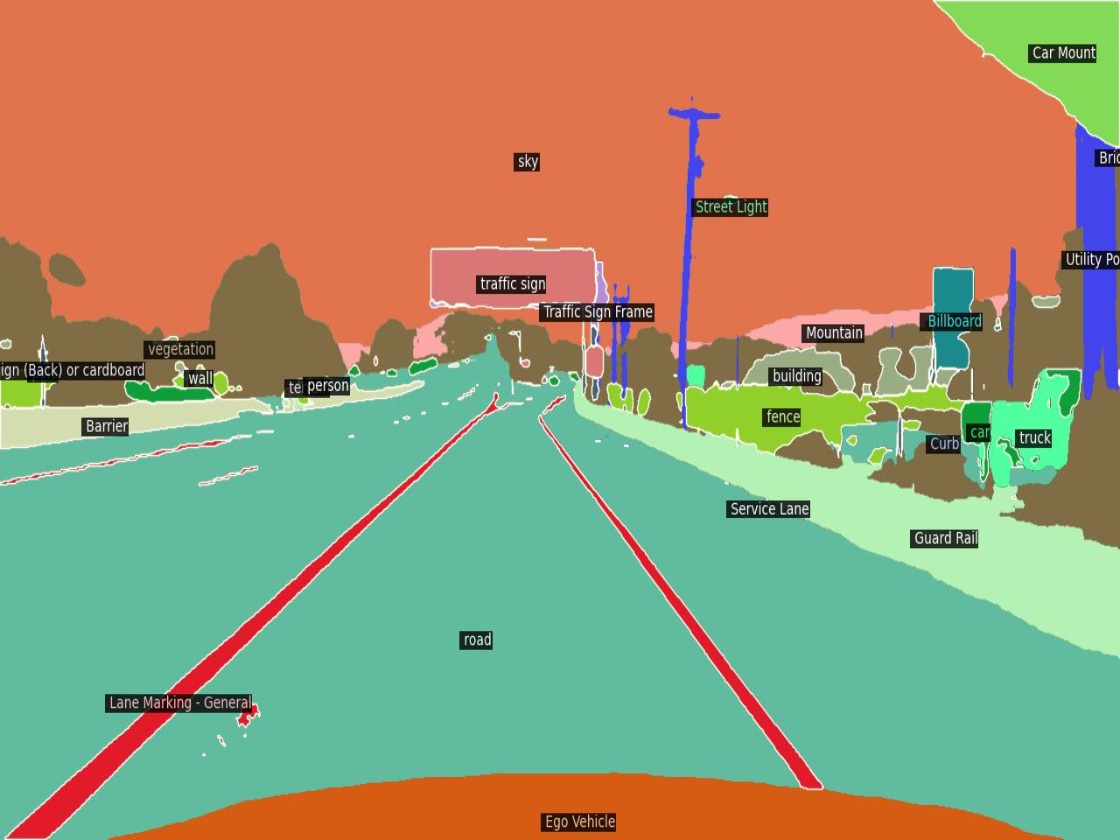} &
    \includegraphics[width=0.13\linewidth]{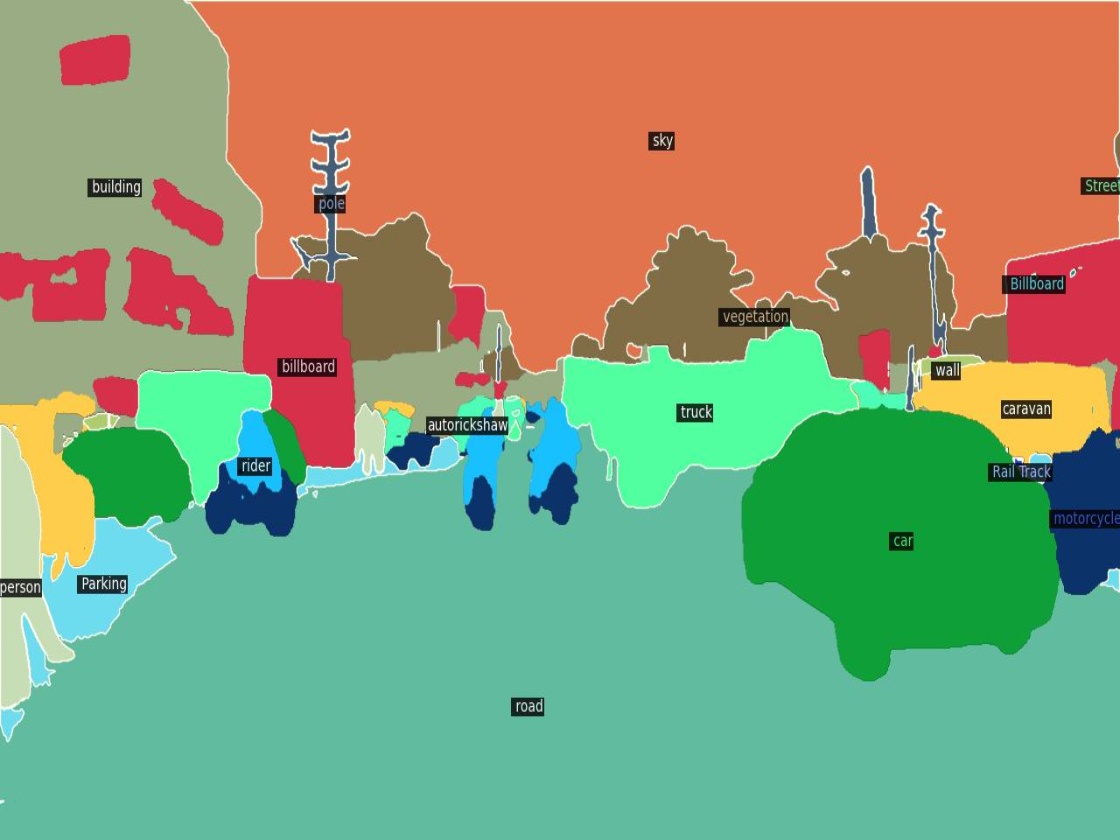} &
    \includegraphics[width=0.13\linewidth]{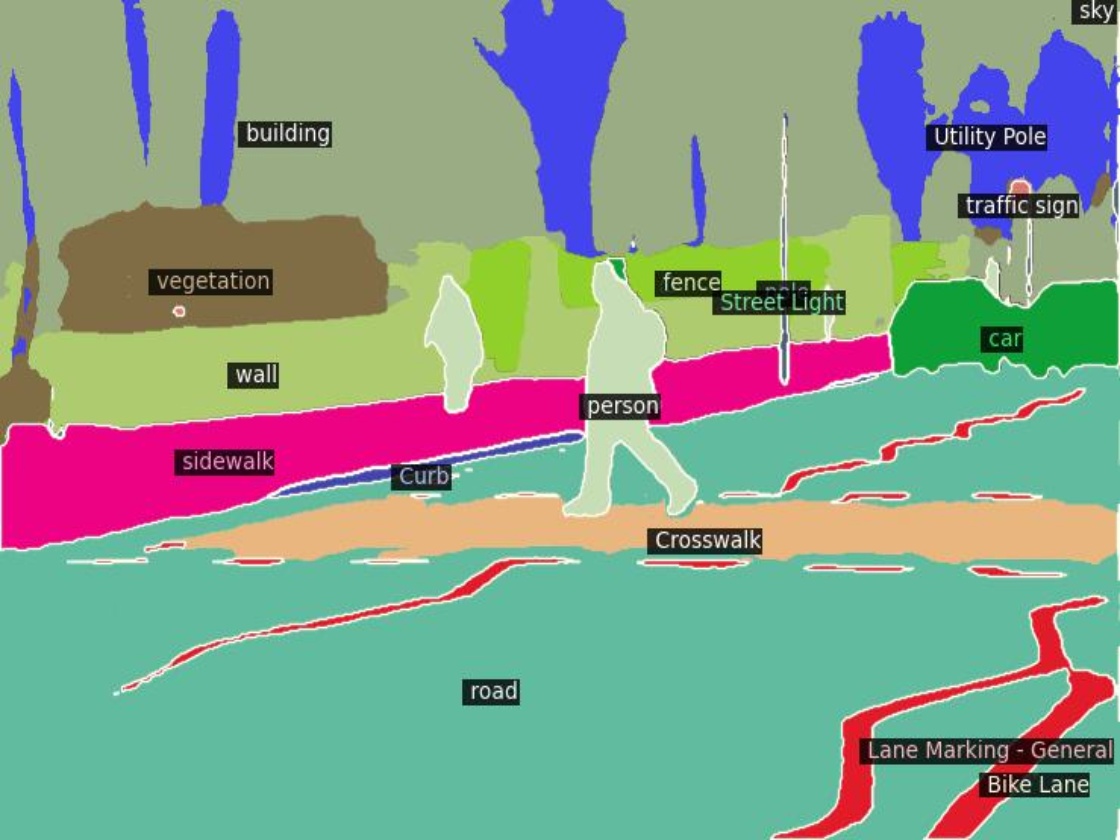} &
    \includegraphics[width=0.13\linewidth]{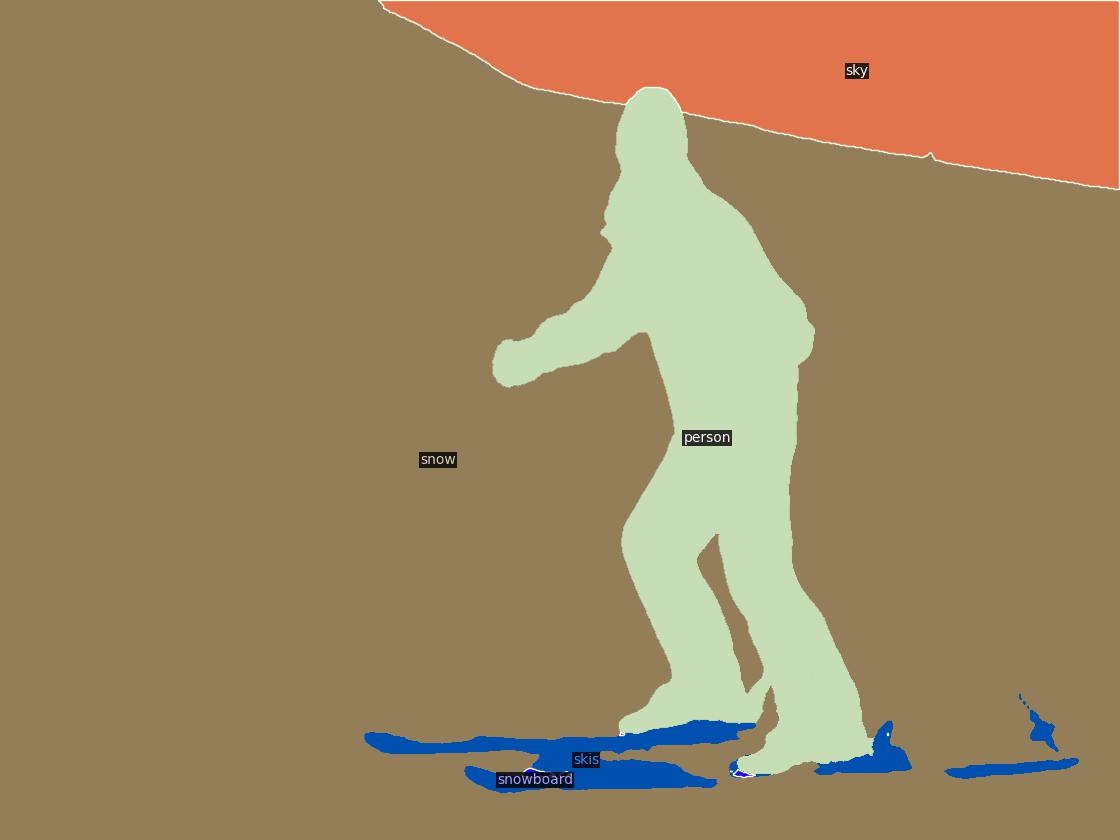} \\
    \rotatebox[origin=l]{90}{CS} &
    \includegraphics[width=0.13\linewidth]{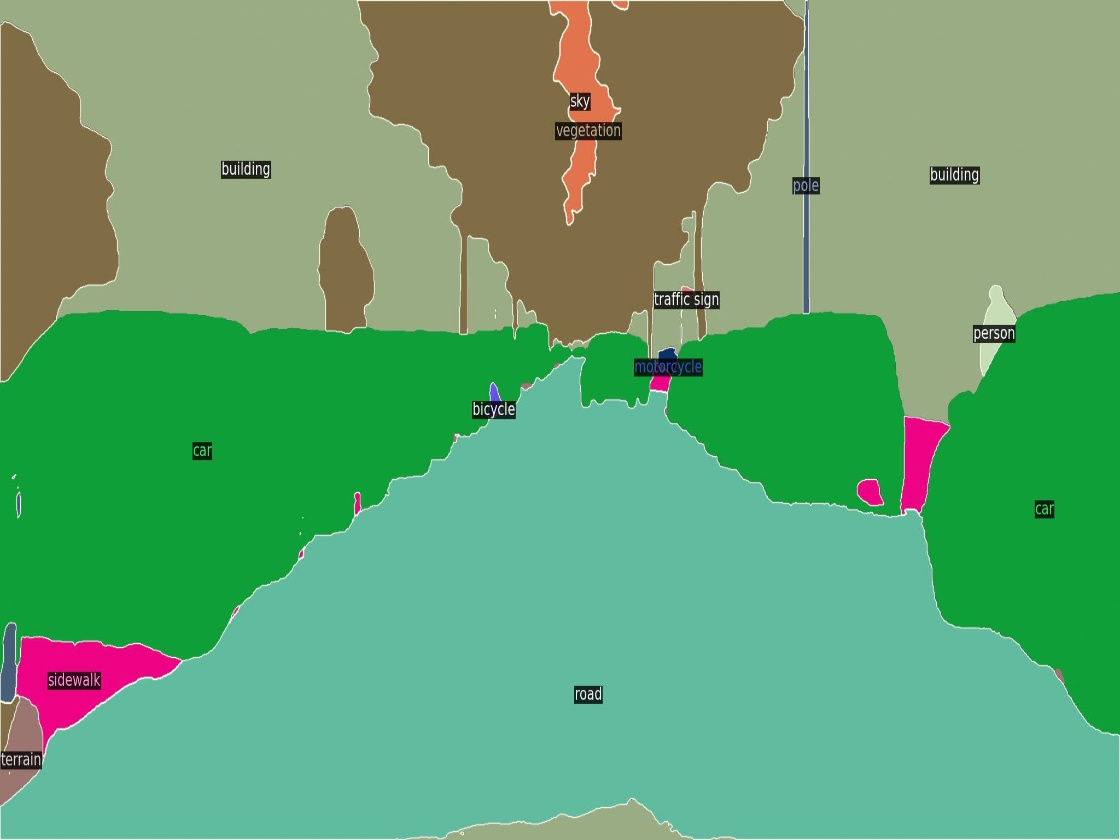} &
    \includegraphics[width=0.13\linewidth]{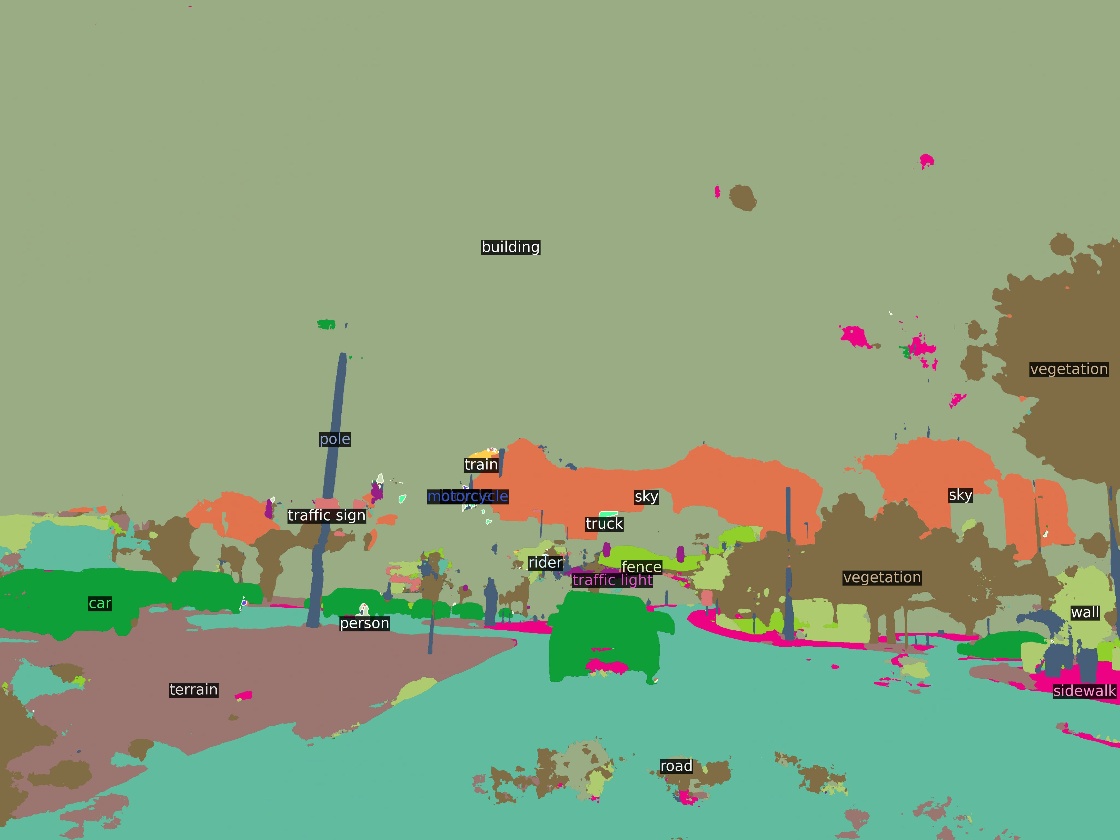} &
    \includegraphics[width=0.13\linewidth]{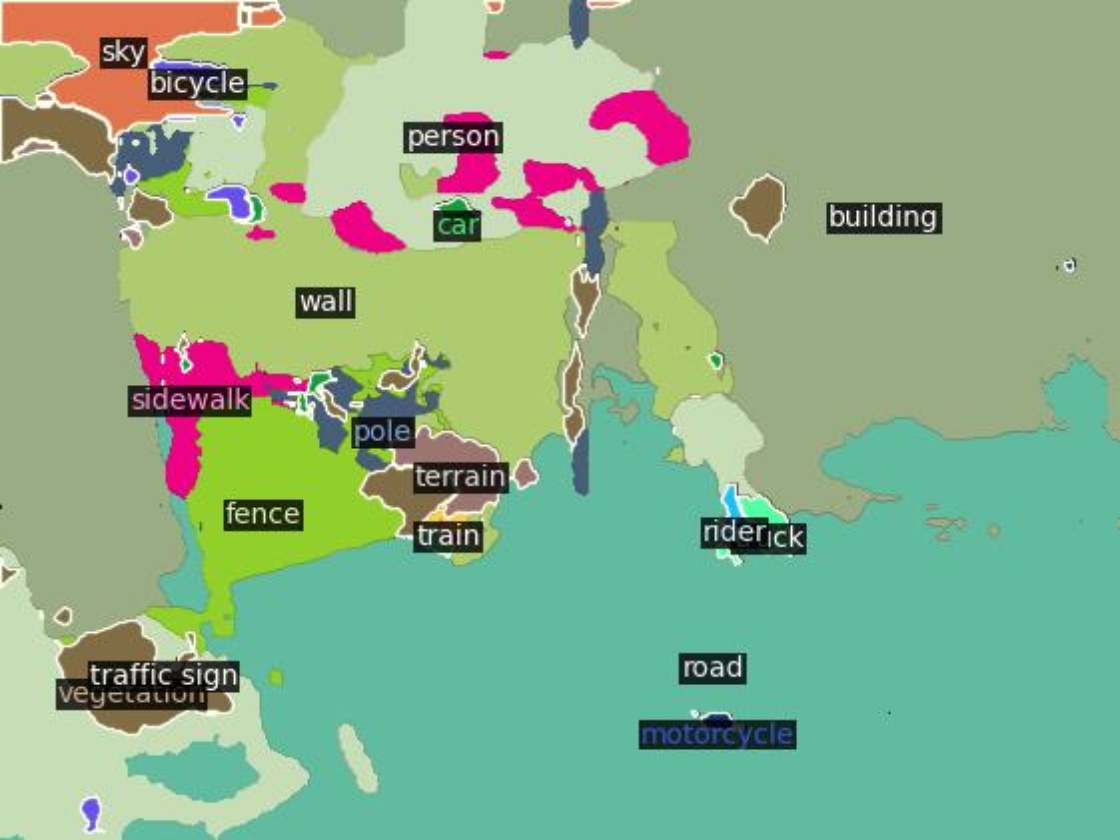} &
    \includegraphics[width=0.13\linewidth]{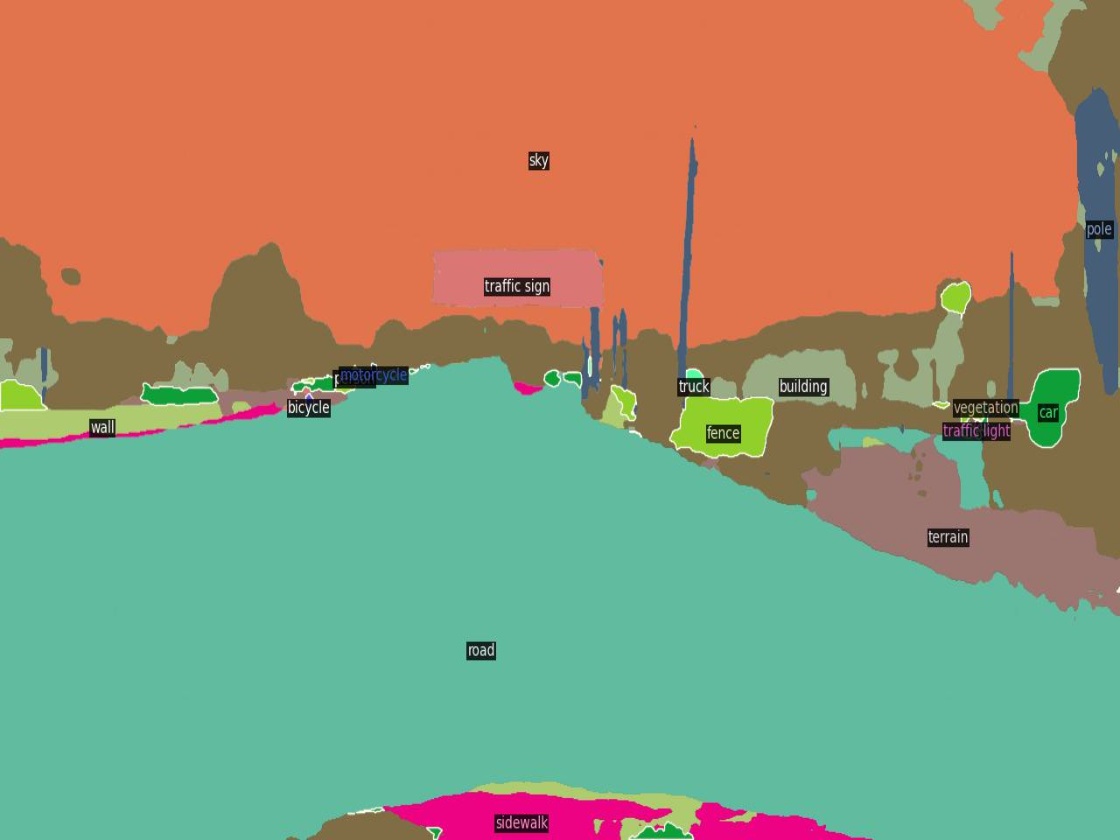} &
    \includegraphics[width=0.13\linewidth]{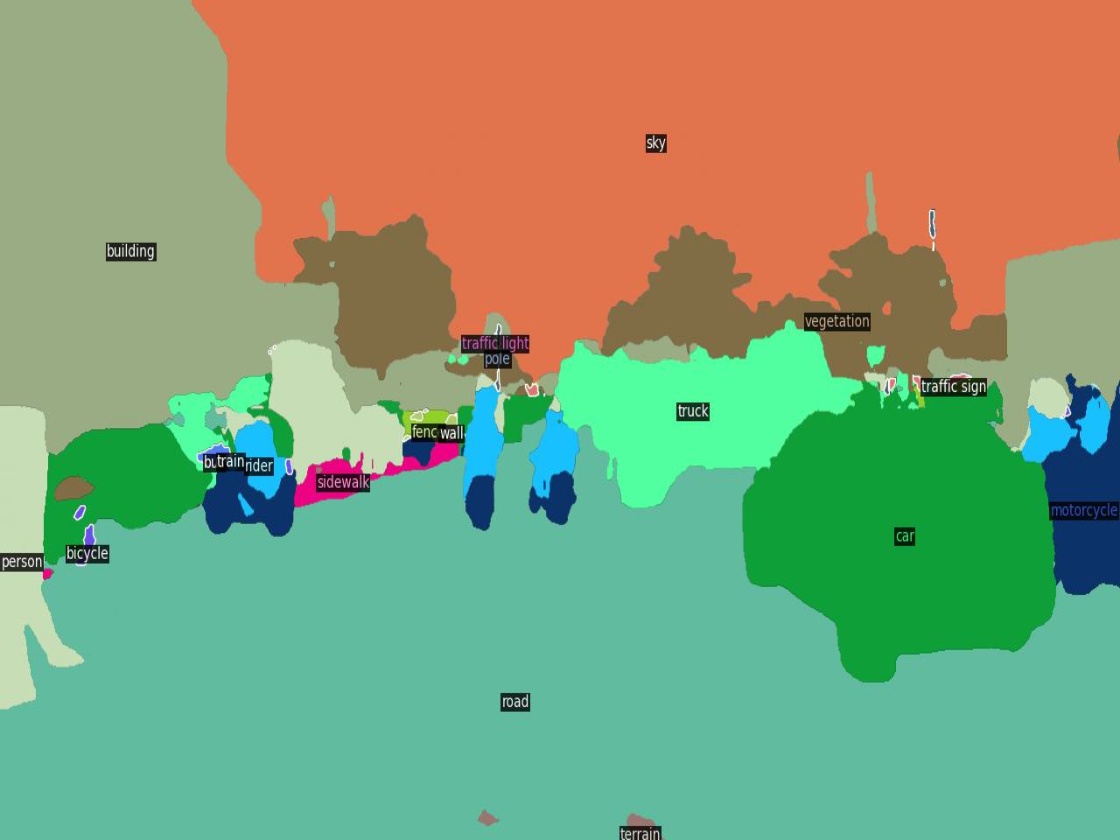} &
    \includegraphics[width=0.13\linewidth]{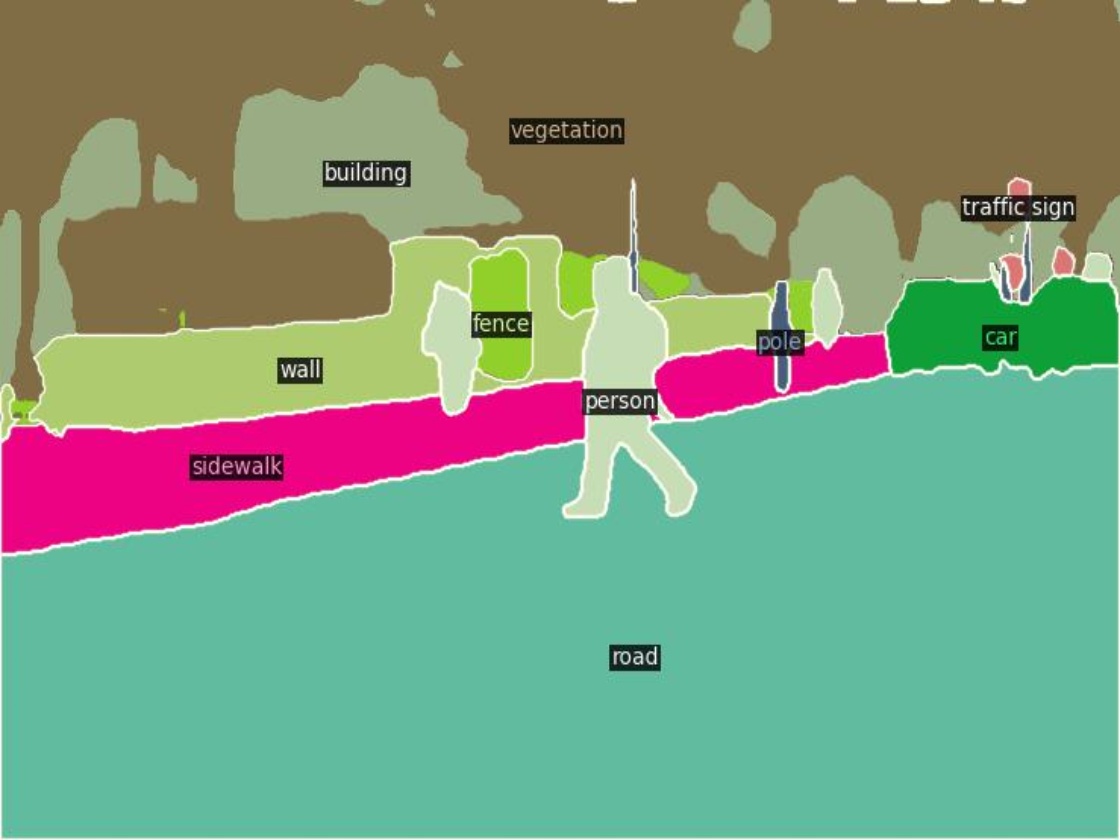} &
    \includegraphics[width=0.13\linewidth]{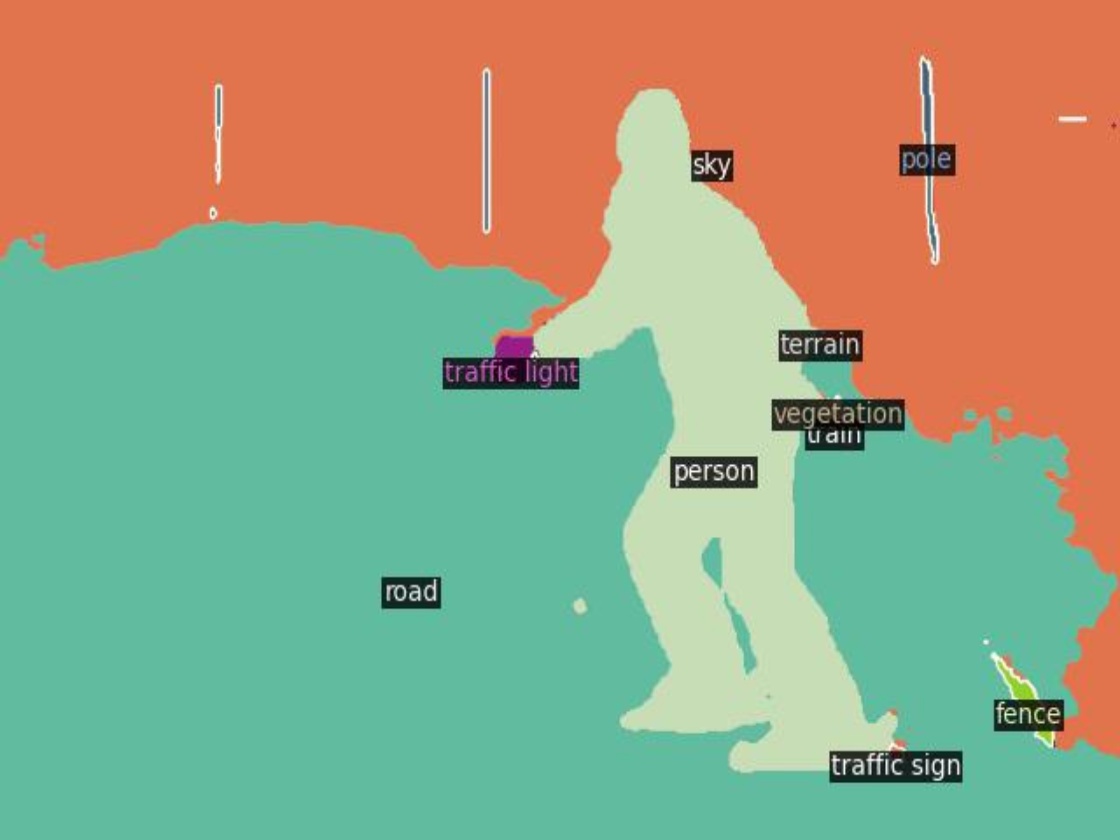} \\
    \rotatebox[origin=l]{90}{MPL} &
    \includegraphics[width=0.13\linewidth]{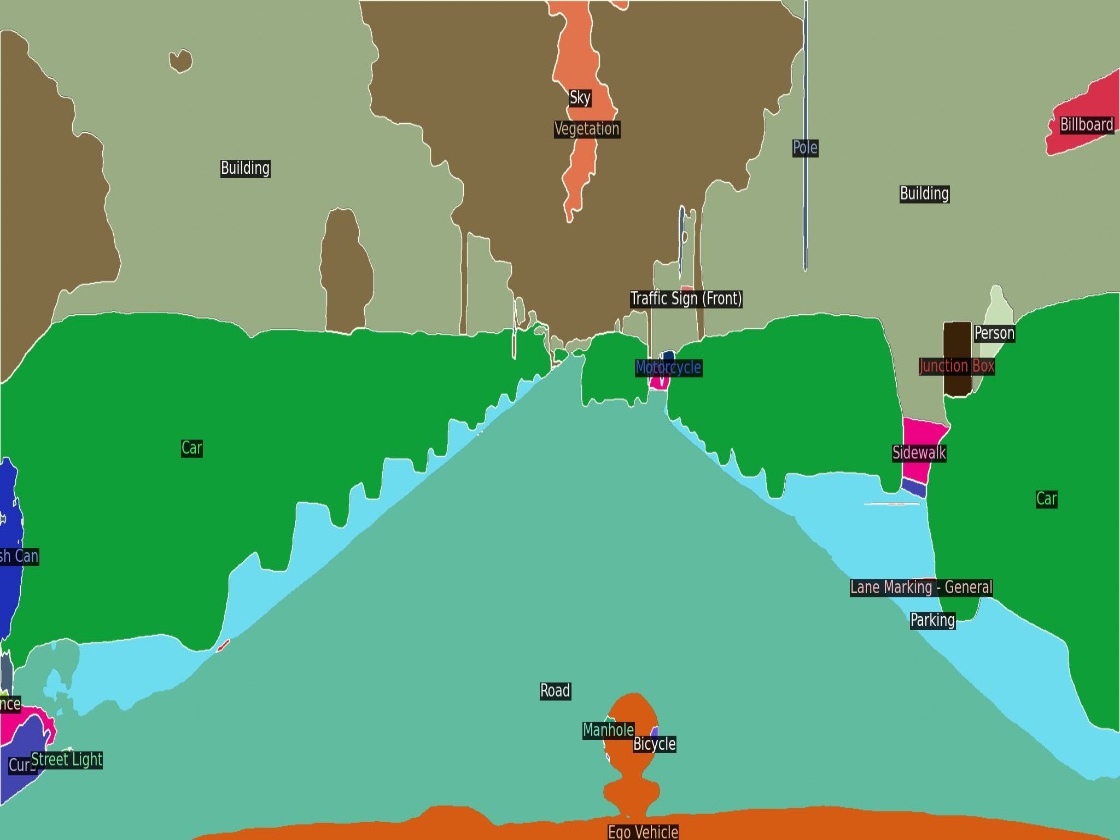} &
    \includegraphics[width=0.13\linewidth]{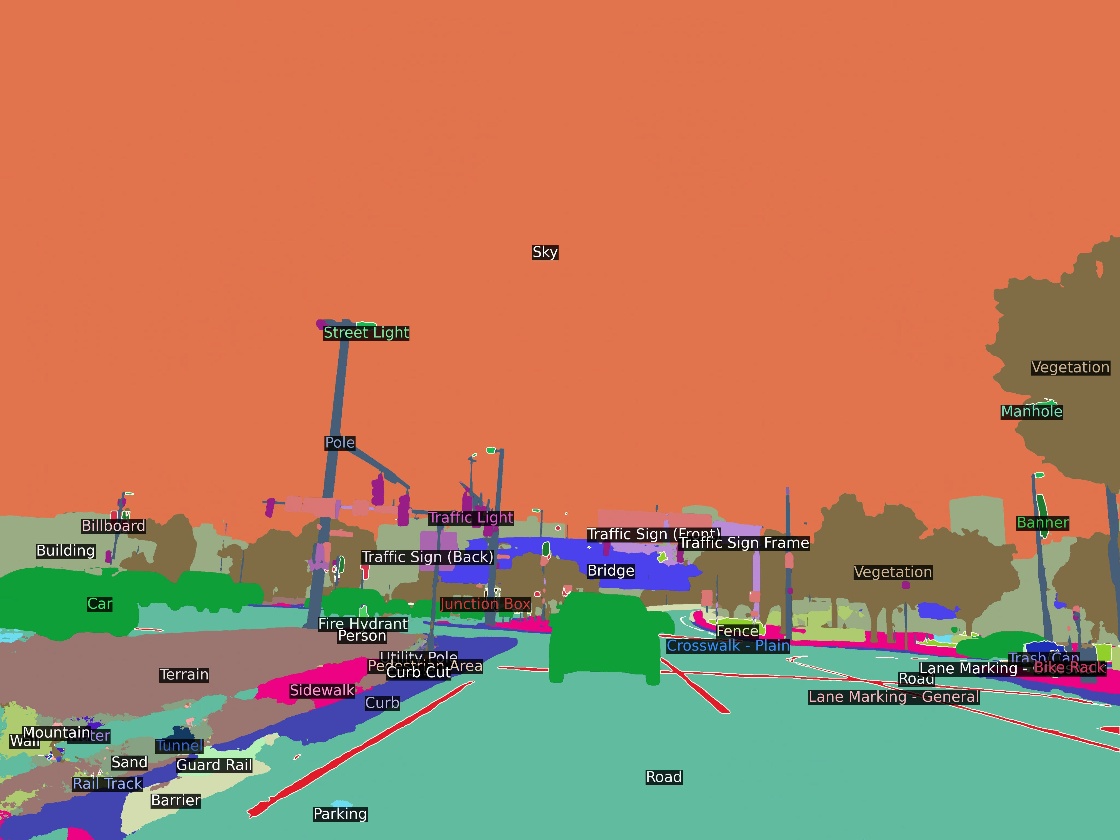} &
    \includegraphics[width=0.13\linewidth]{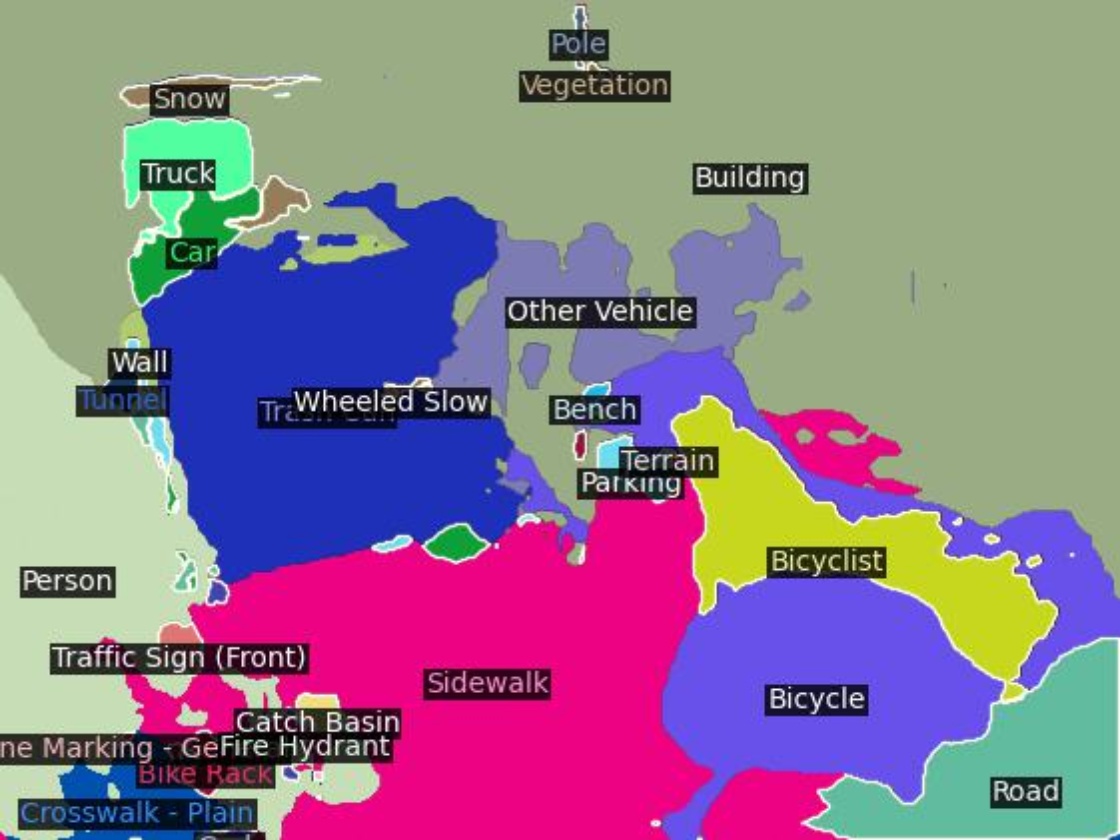} &
    \includegraphics[width=0.13\linewidth]{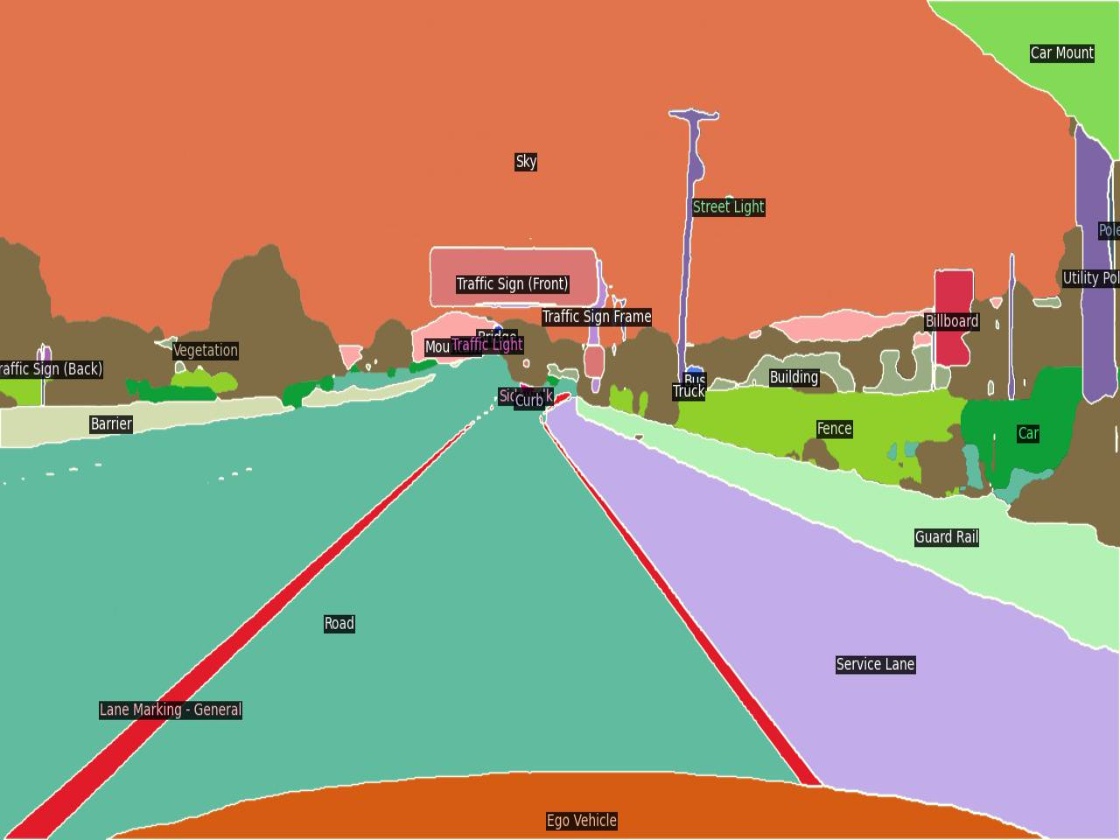} &
    \includegraphics[width=0.13\linewidth]{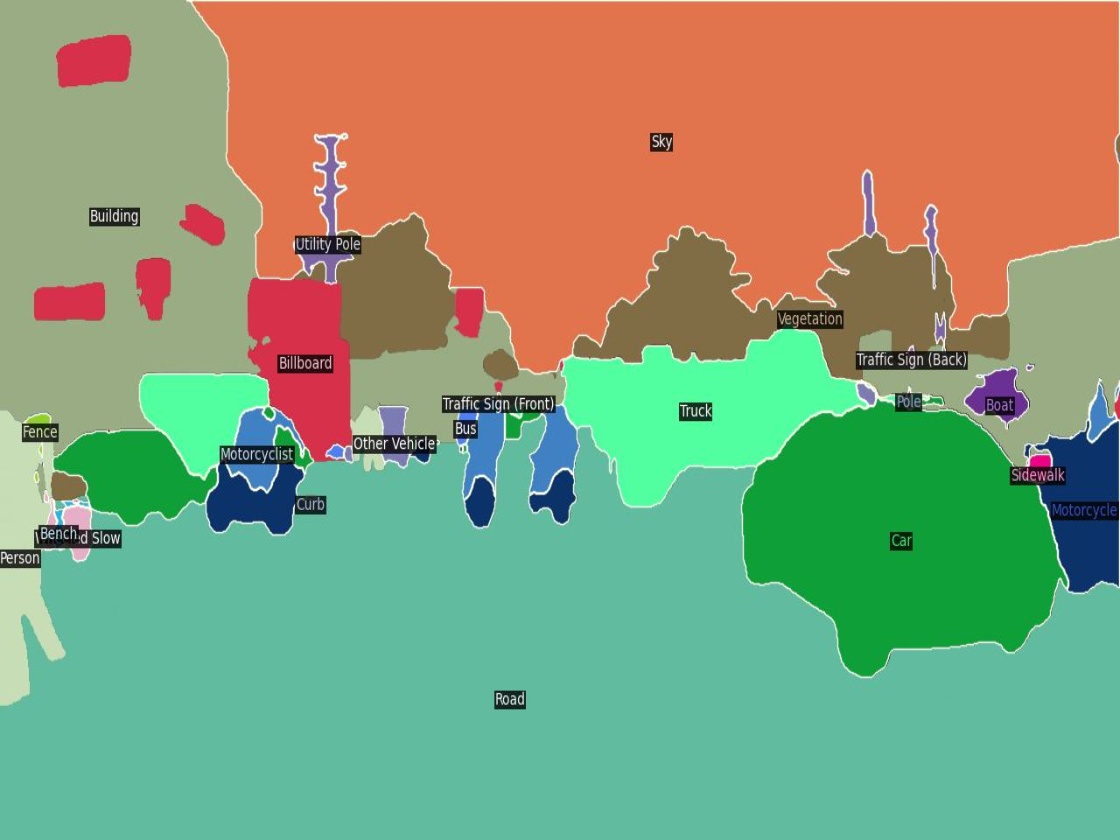} &
    \includegraphics[width=0.13\linewidth]{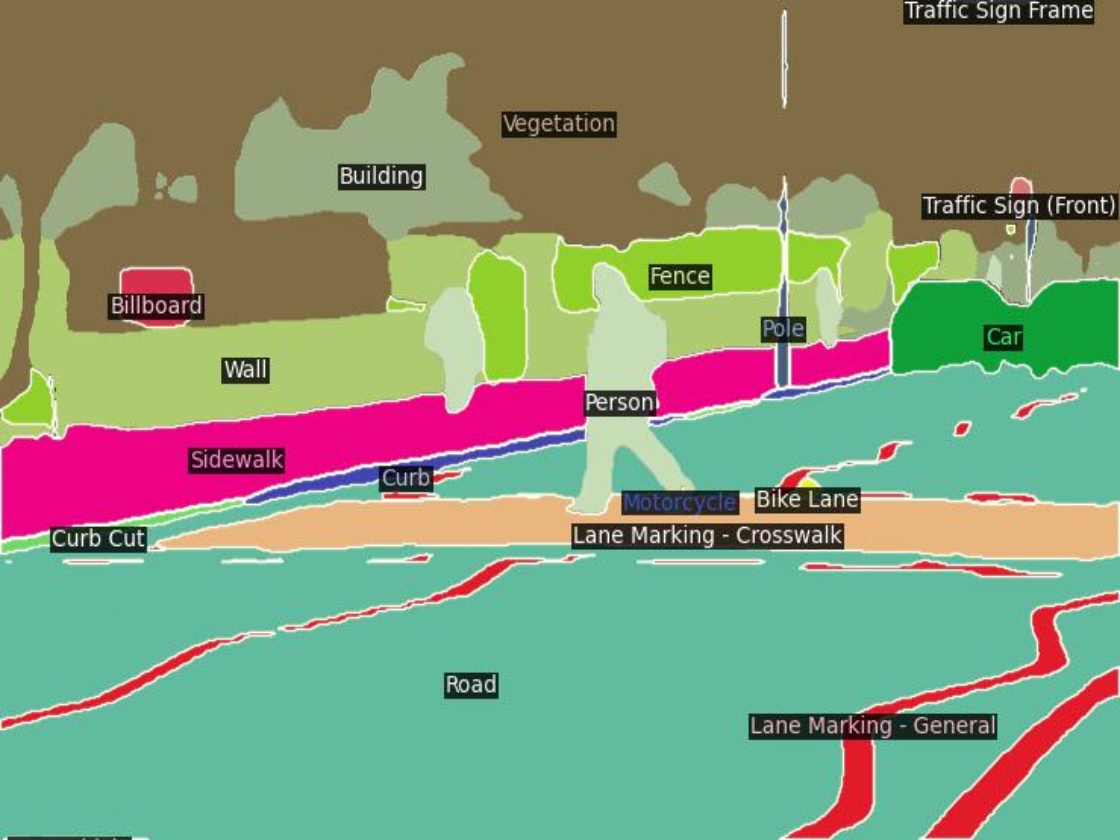} &
    \includegraphics[width=0.13\linewidth]{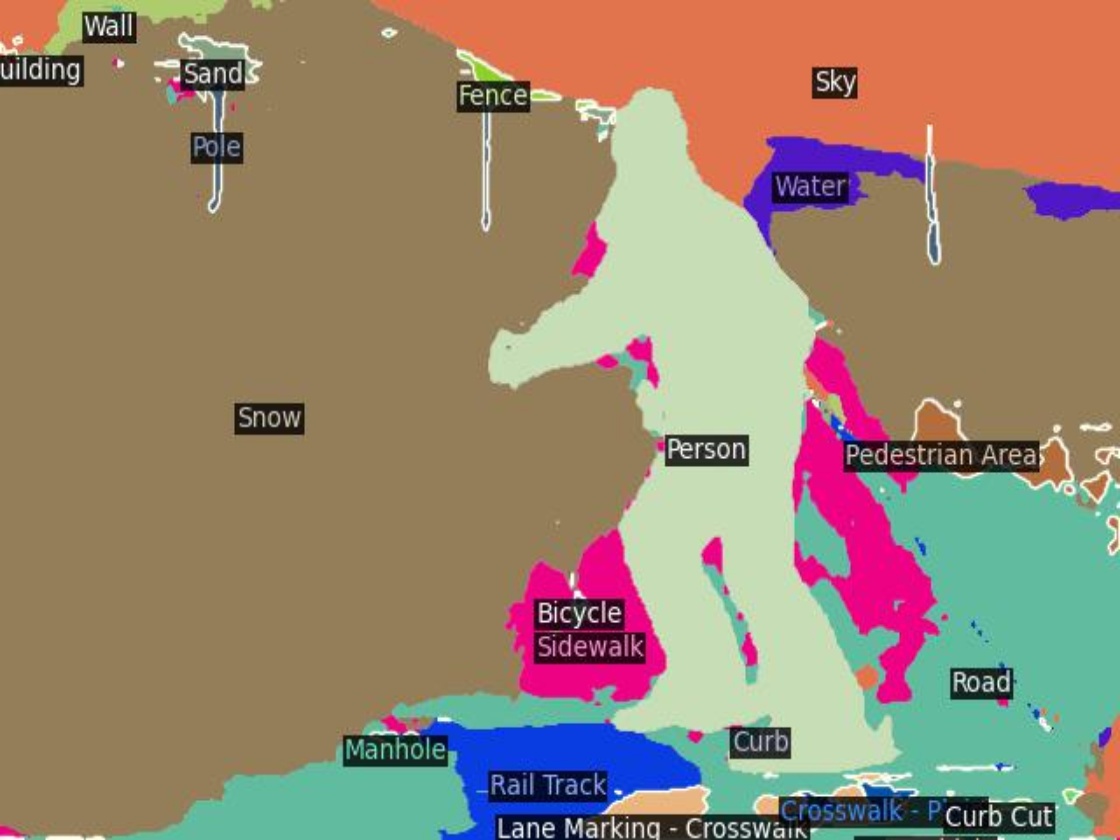} \\
    \rotatebox[origin=l]{90}{SUN} &
    \includegraphics[width=0.13\linewidth]{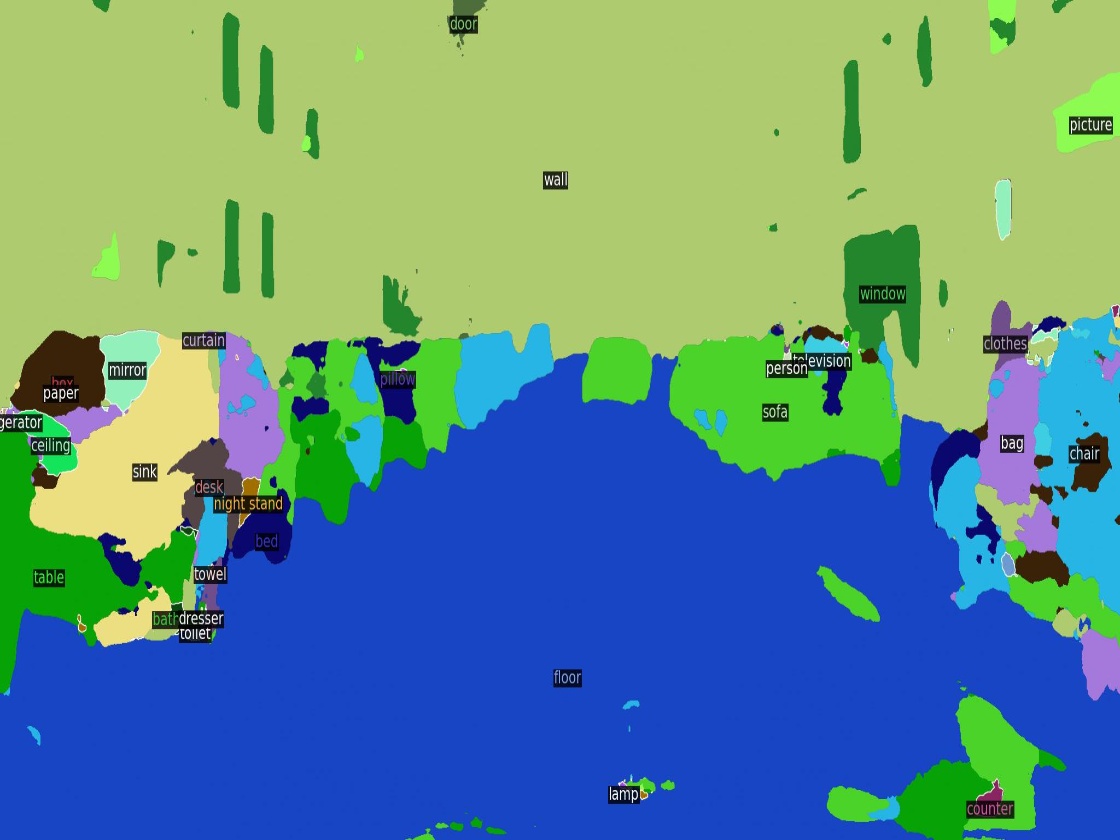} &
    \includegraphics[width=0.13\linewidth]{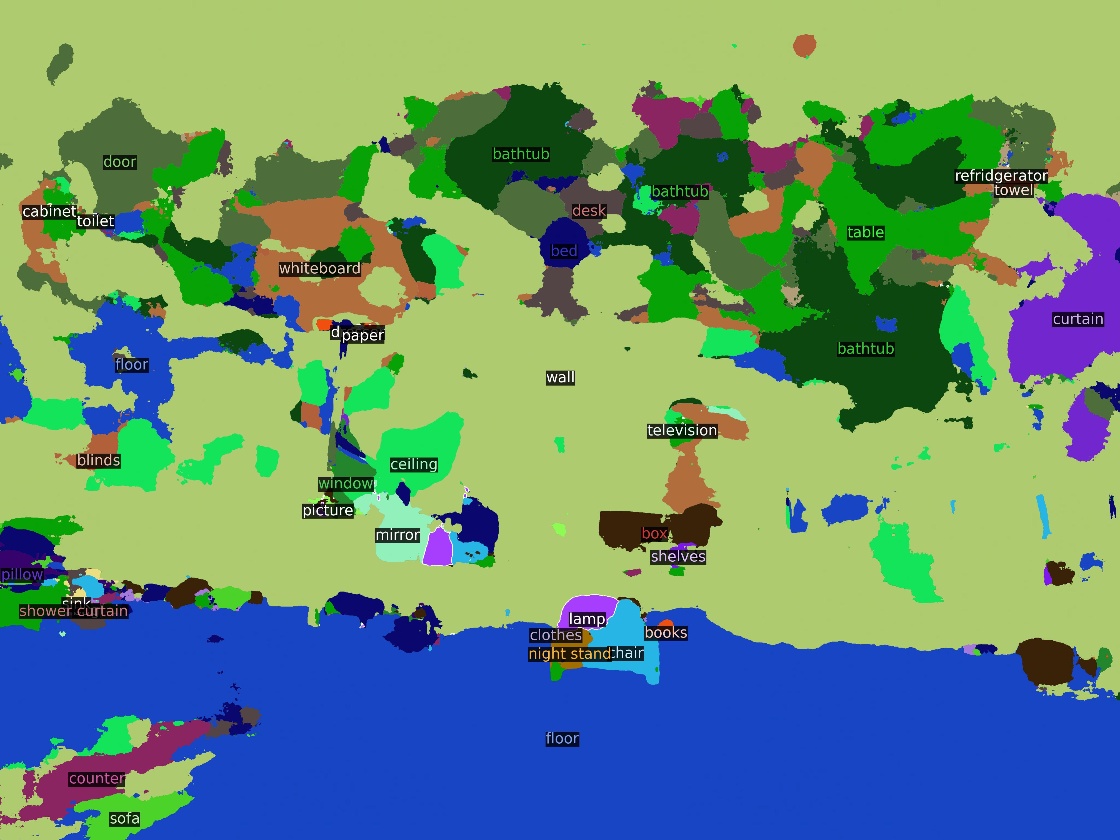} &
    \includegraphics[width=0.13\linewidth]{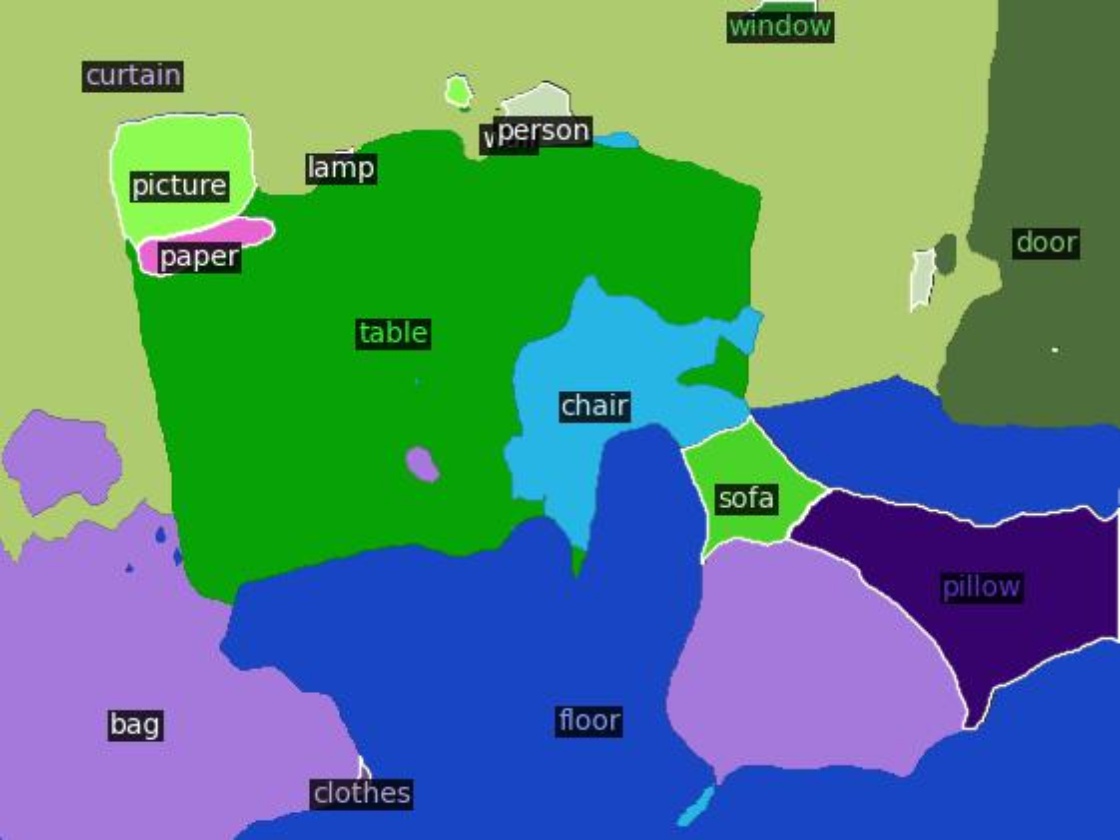} &
    \includegraphics[width=0.13\linewidth]{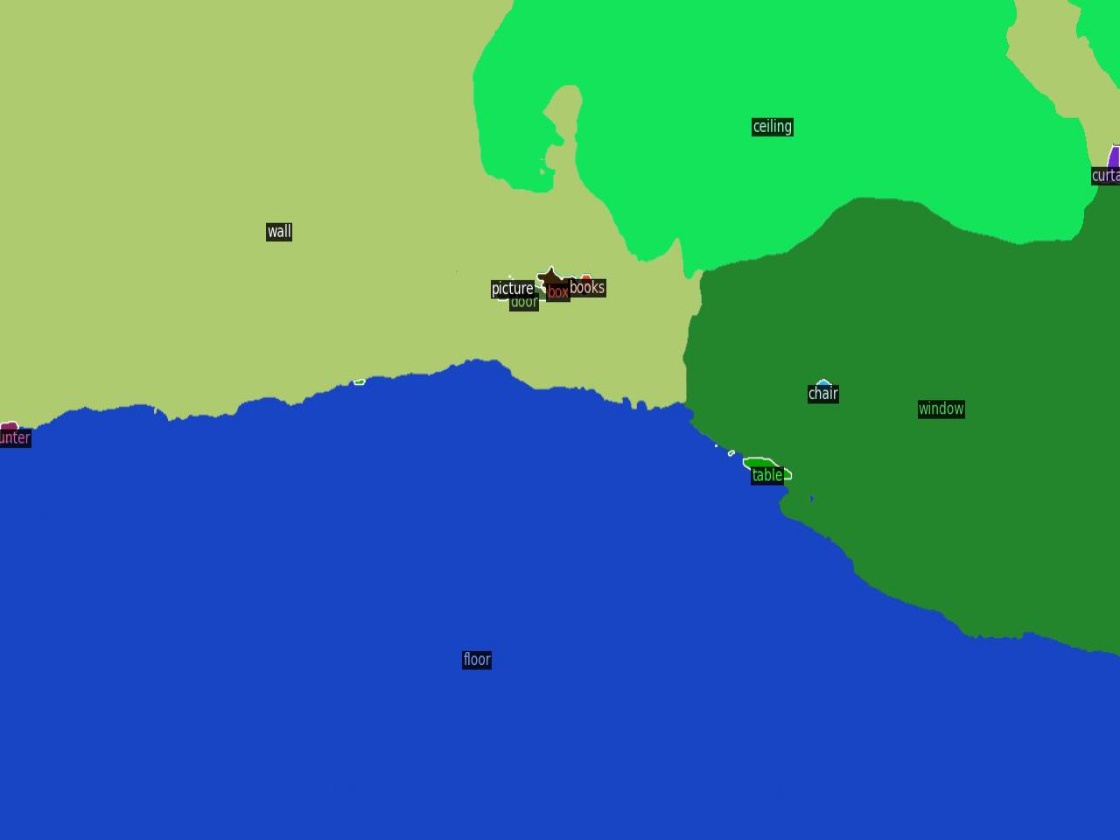} &
    \includegraphics[width=0.13\linewidth]{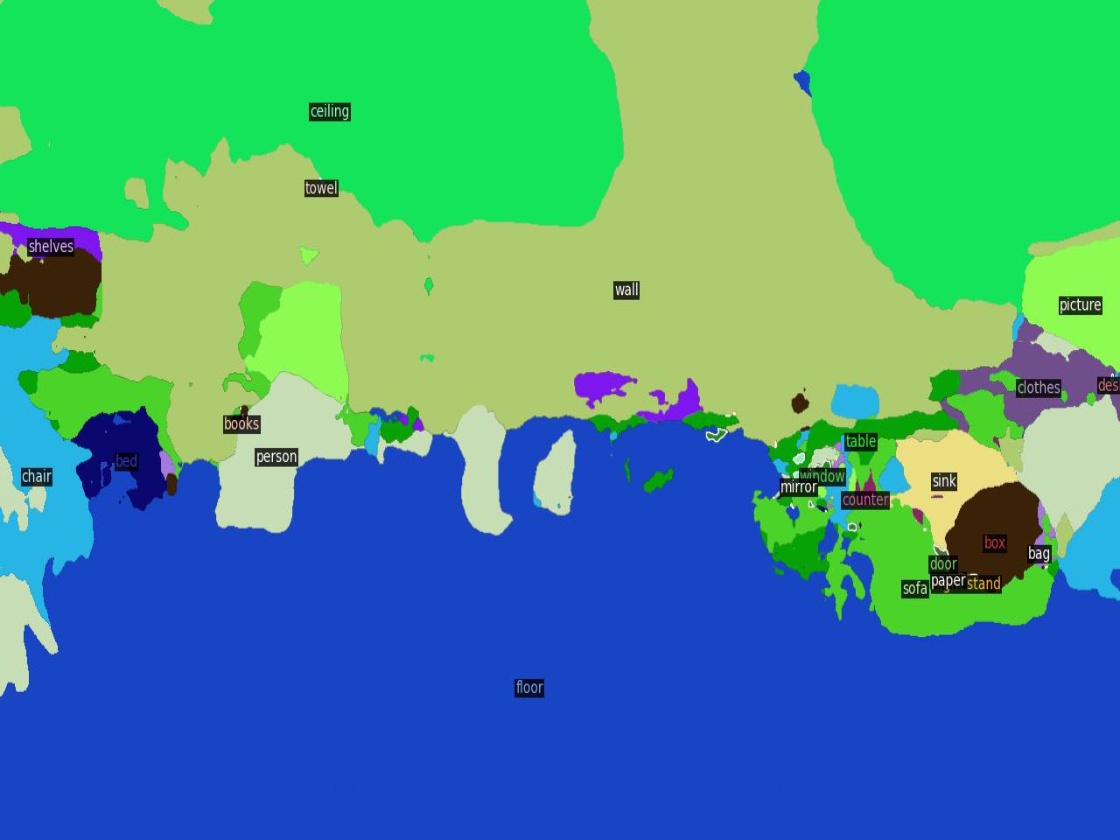} &
    \includegraphics[width=0.13\linewidth]{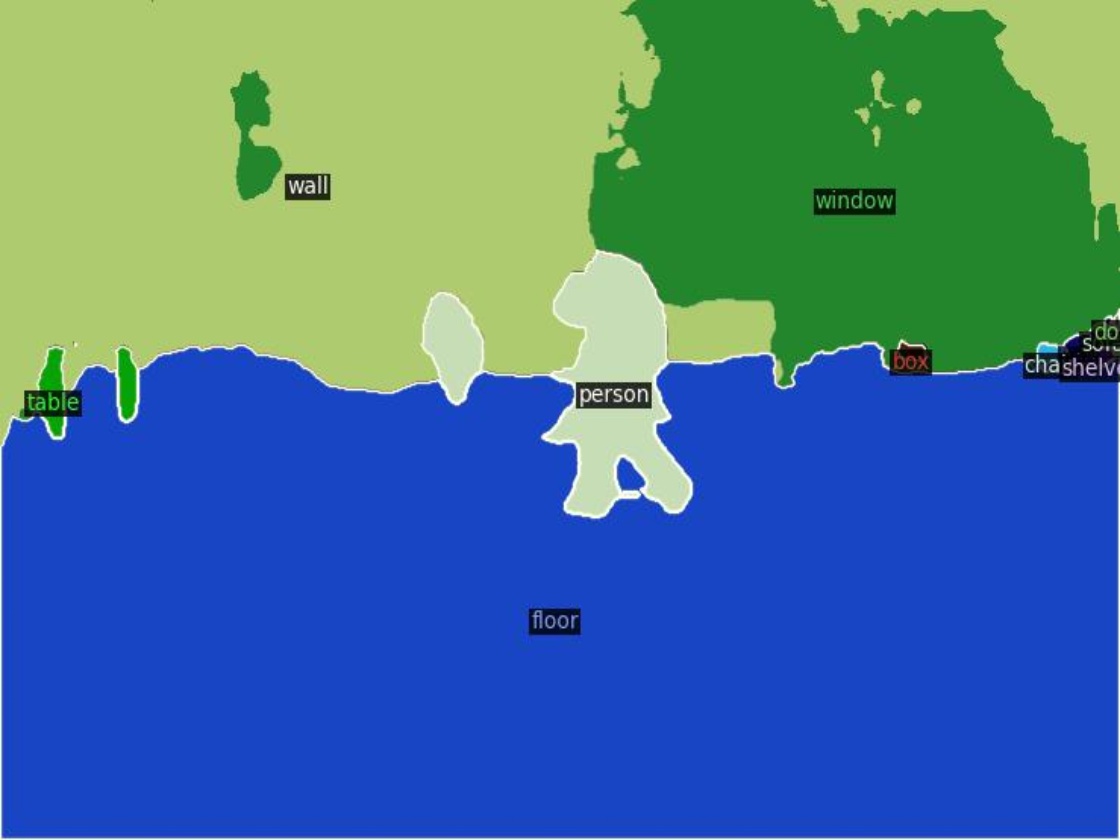} &
    \includegraphics[width=0.13\linewidth]{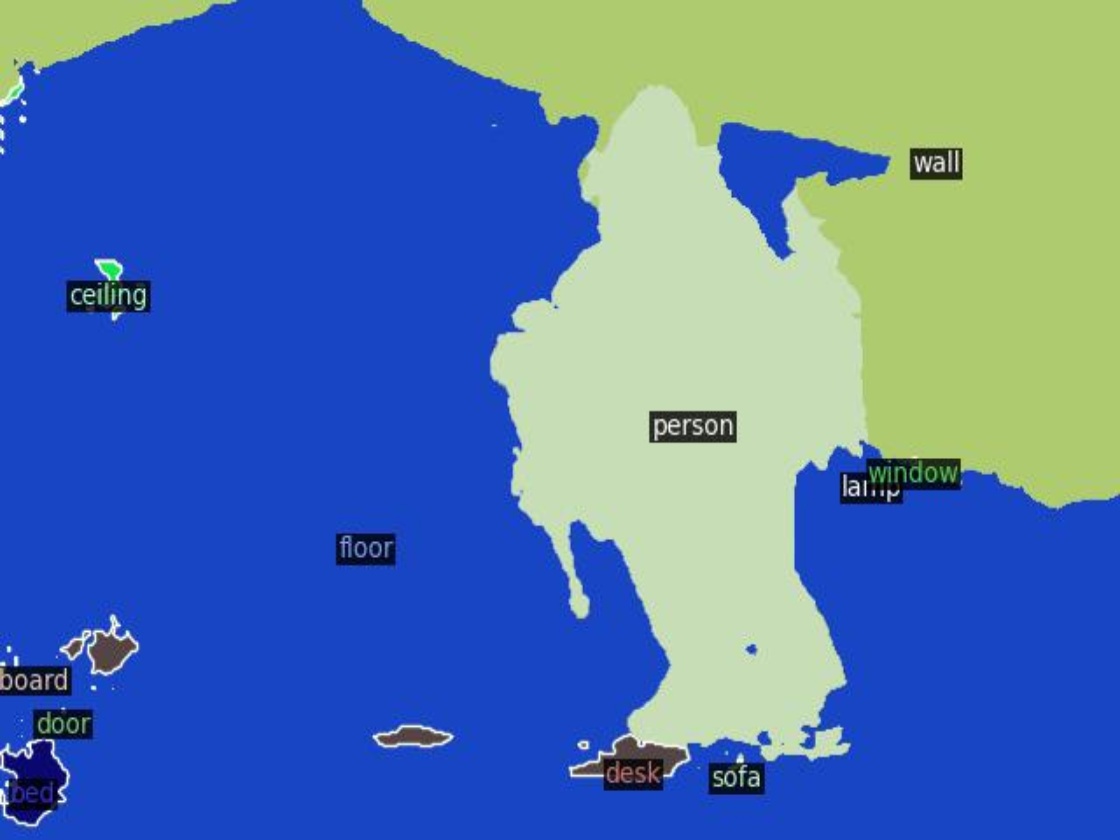} \\
        \rotatebox[origin=l]{90}{BDD} &
        \includegraphics[width=0.13\linewidth]{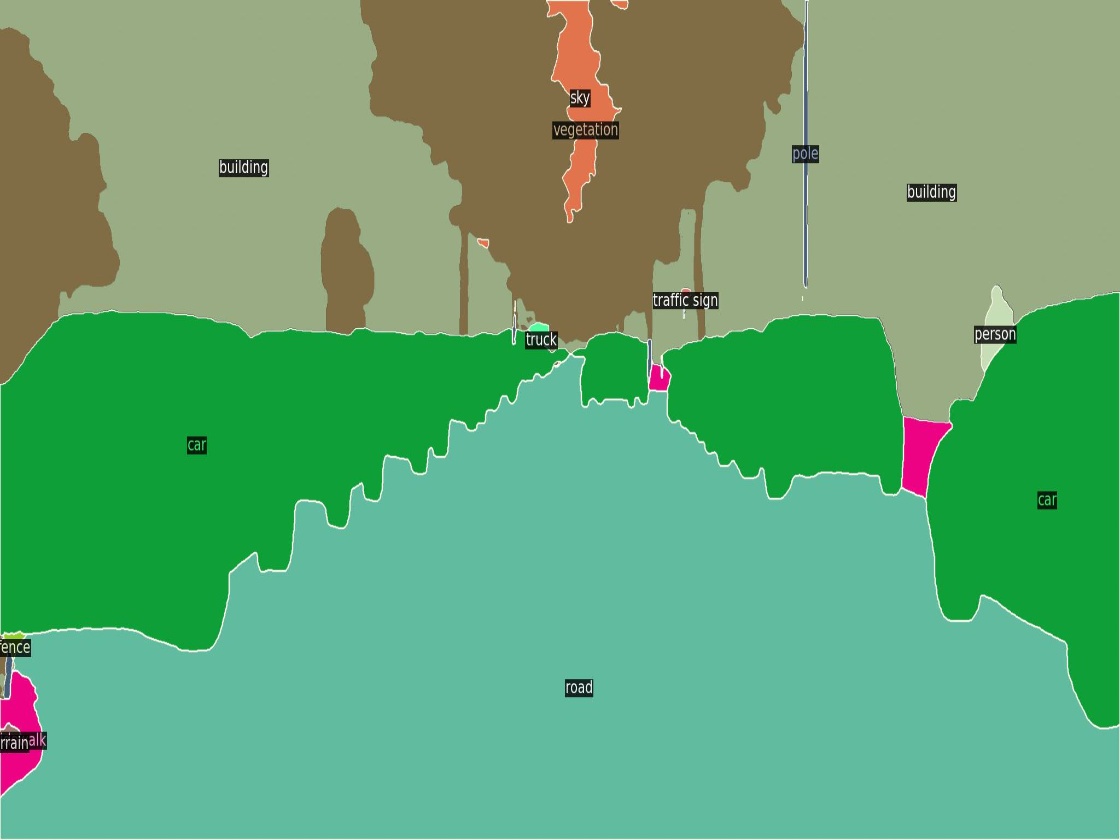} &
    \includegraphics[width=0.13\linewidth]{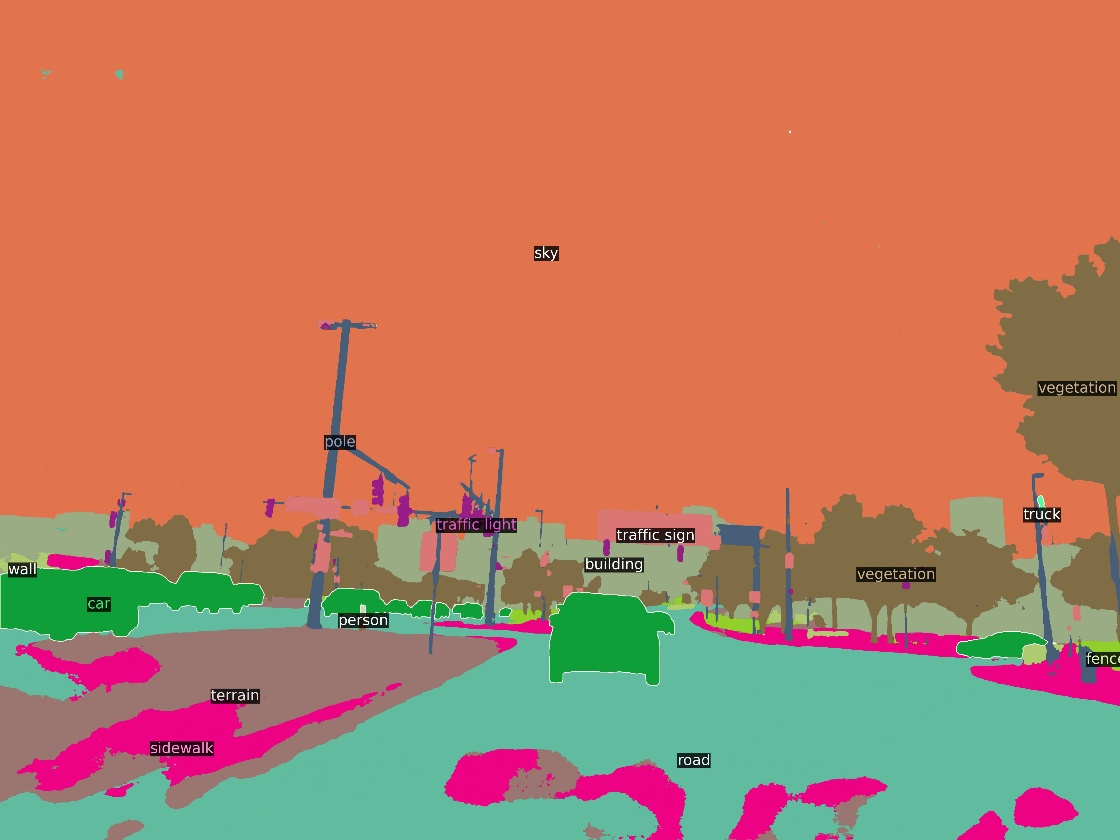} &
    \includegraphics[width=0.13\linewidth]{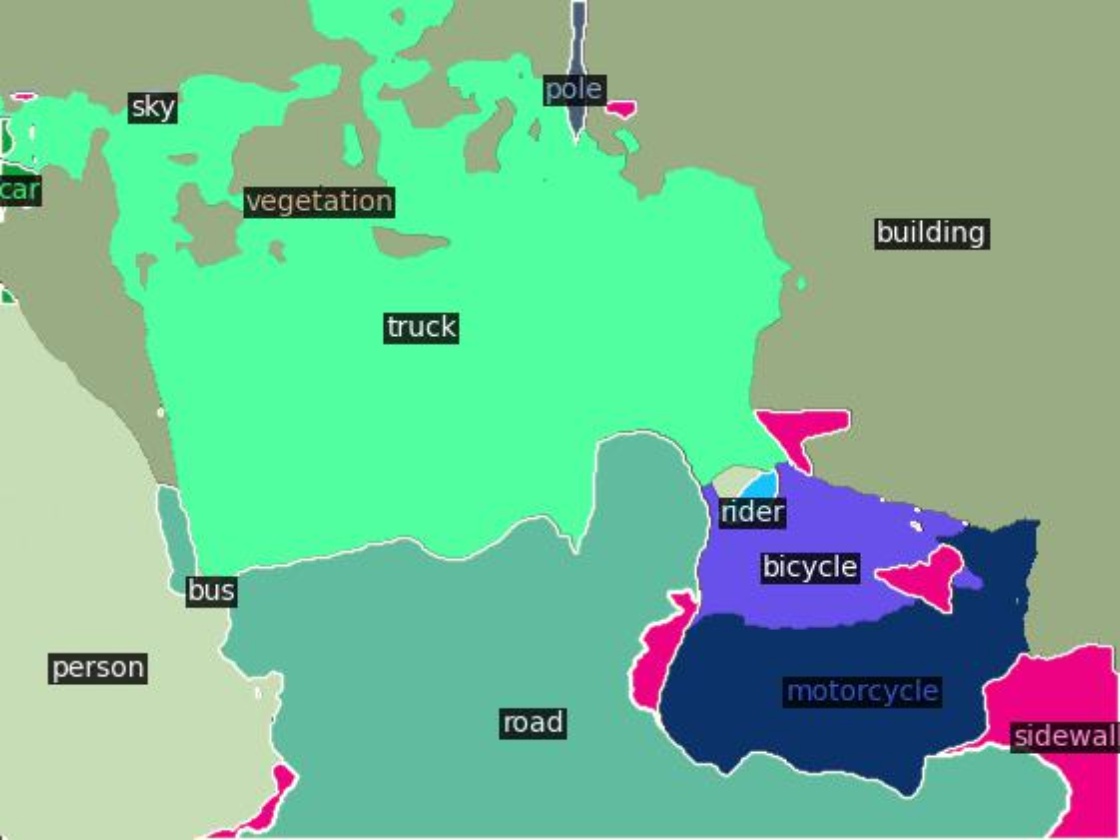} &
    \includegraphics[width=0.13\linewidth]{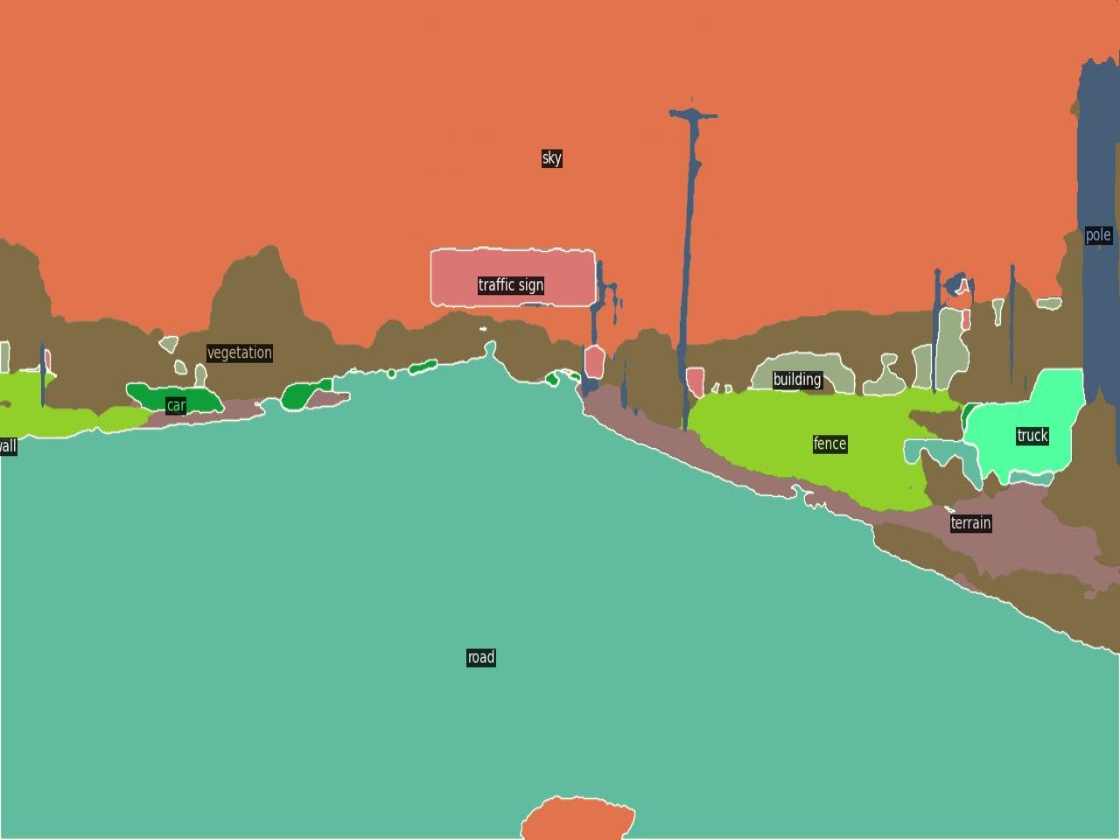} &
    \includegraphics[width=0.13\linewidth]{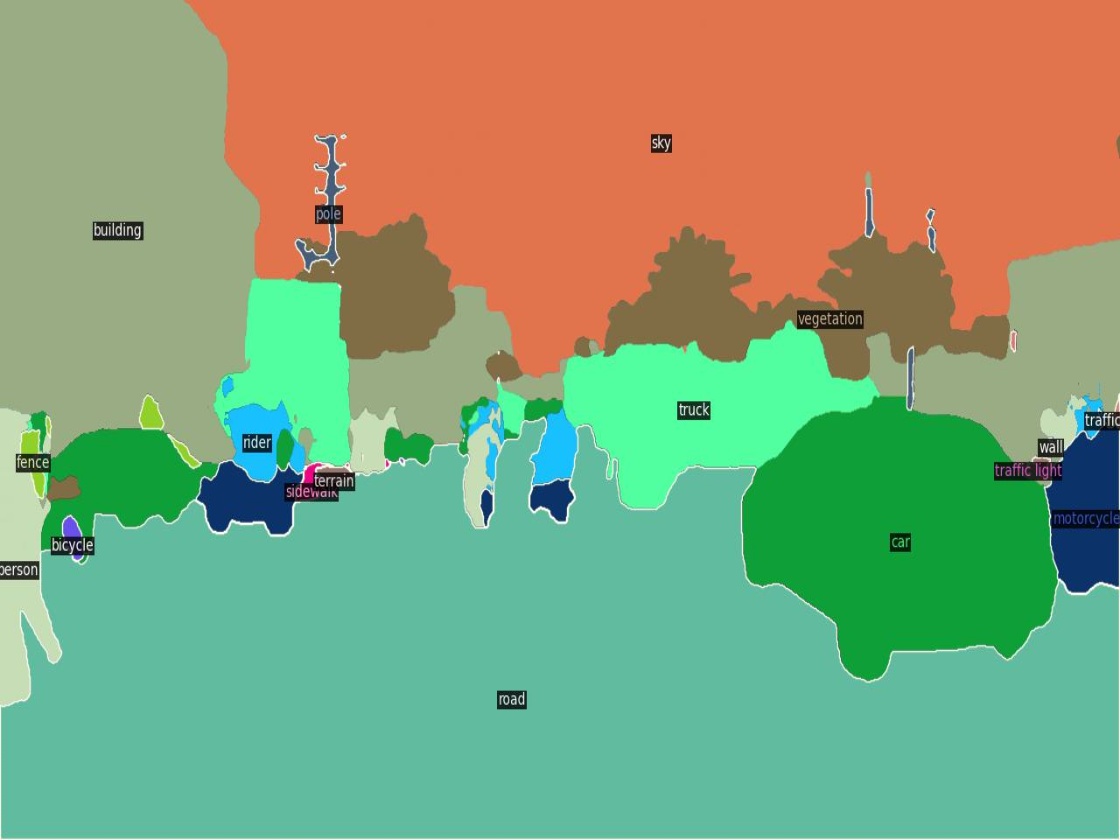} &
    \includegraphics[width=0.13\linewidth]{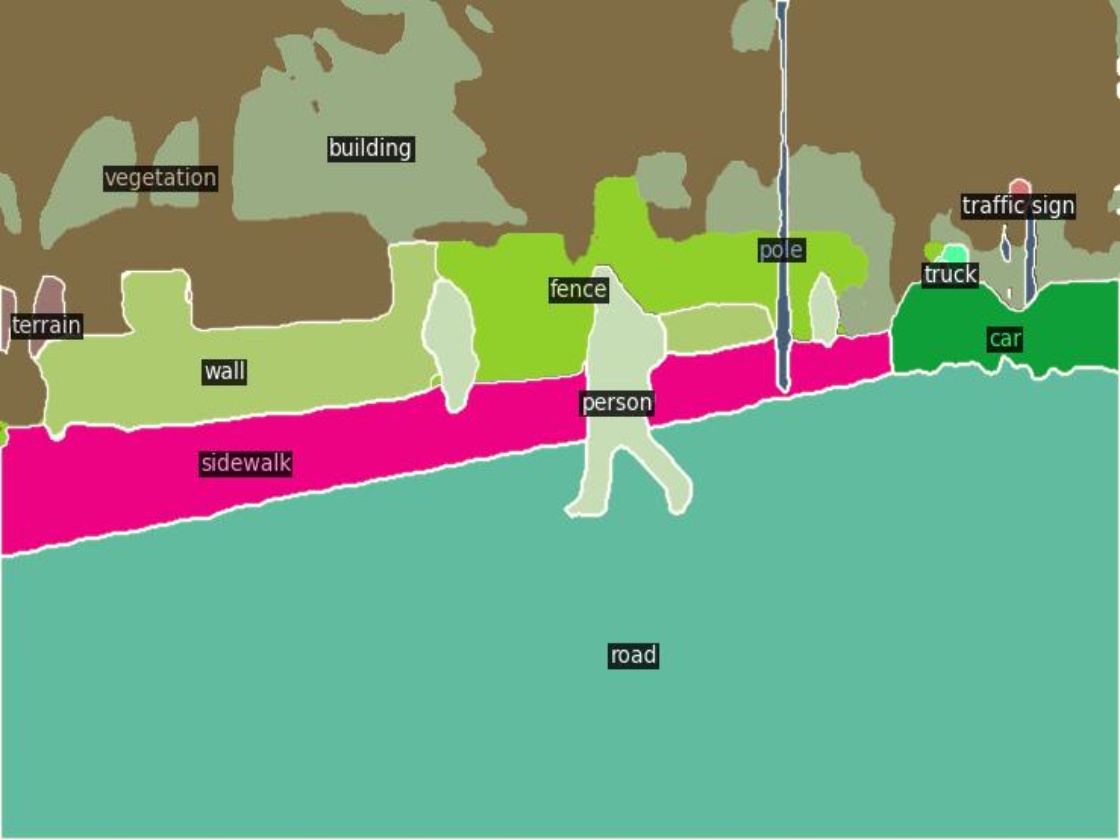} &
    \includegraphics[width=0.13\linewidth]{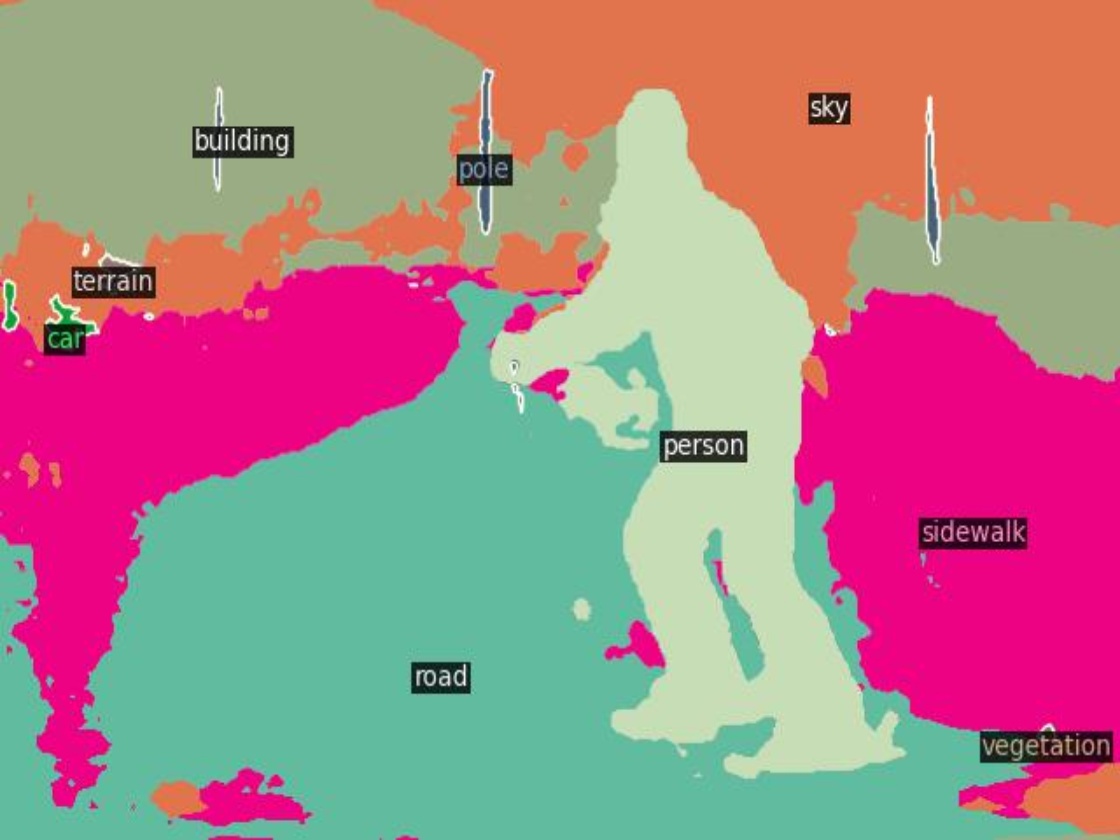} \\
        \rotatebox[origin=l]{90}{IDD} &
    \includegraphics[width=0.13\linewidth]{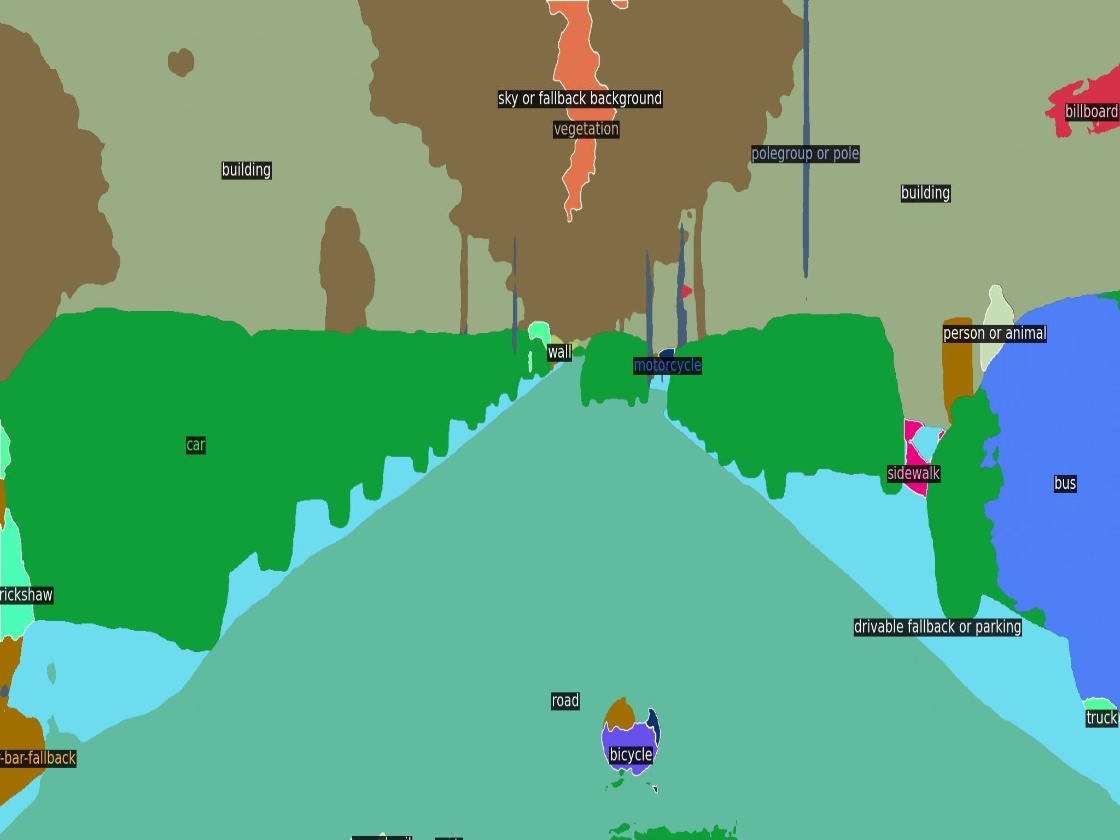} &
    \includegraphics[width=0.13\linewidth]{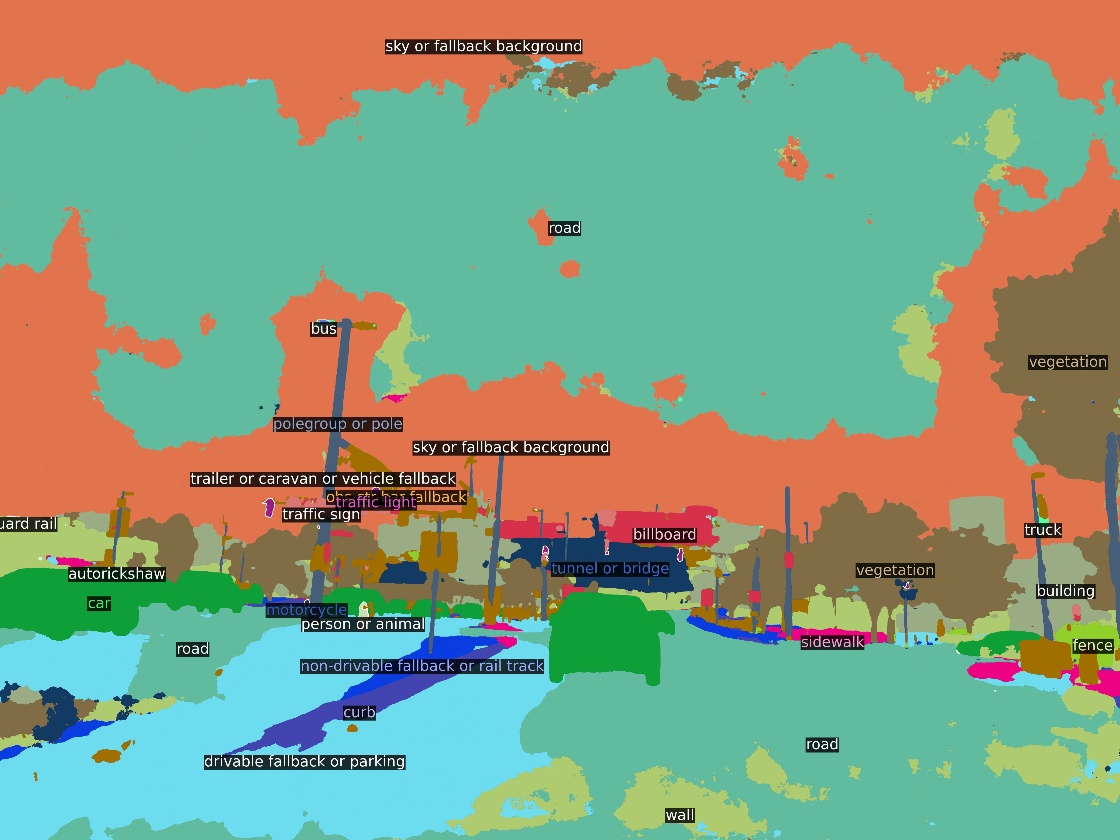} &
    \includegraphics[width=0.13\linewidth]{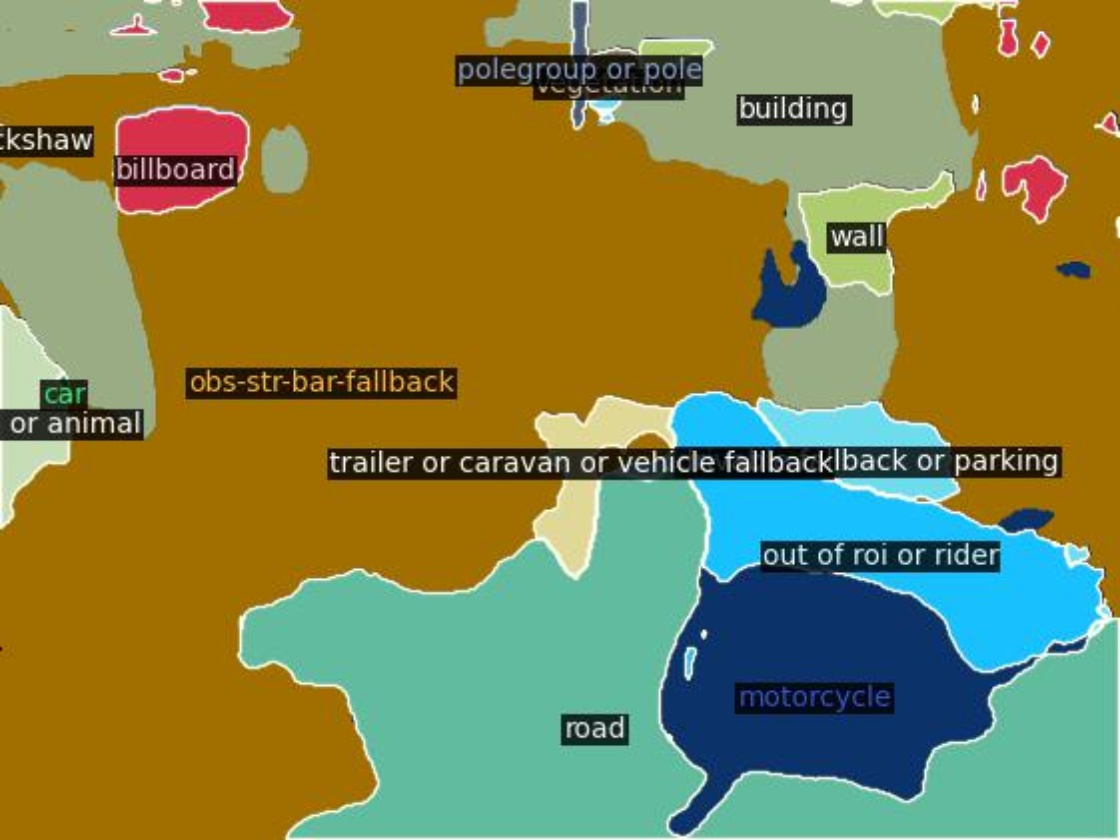} &
    \includegraphics[width=0.13\linewidth]{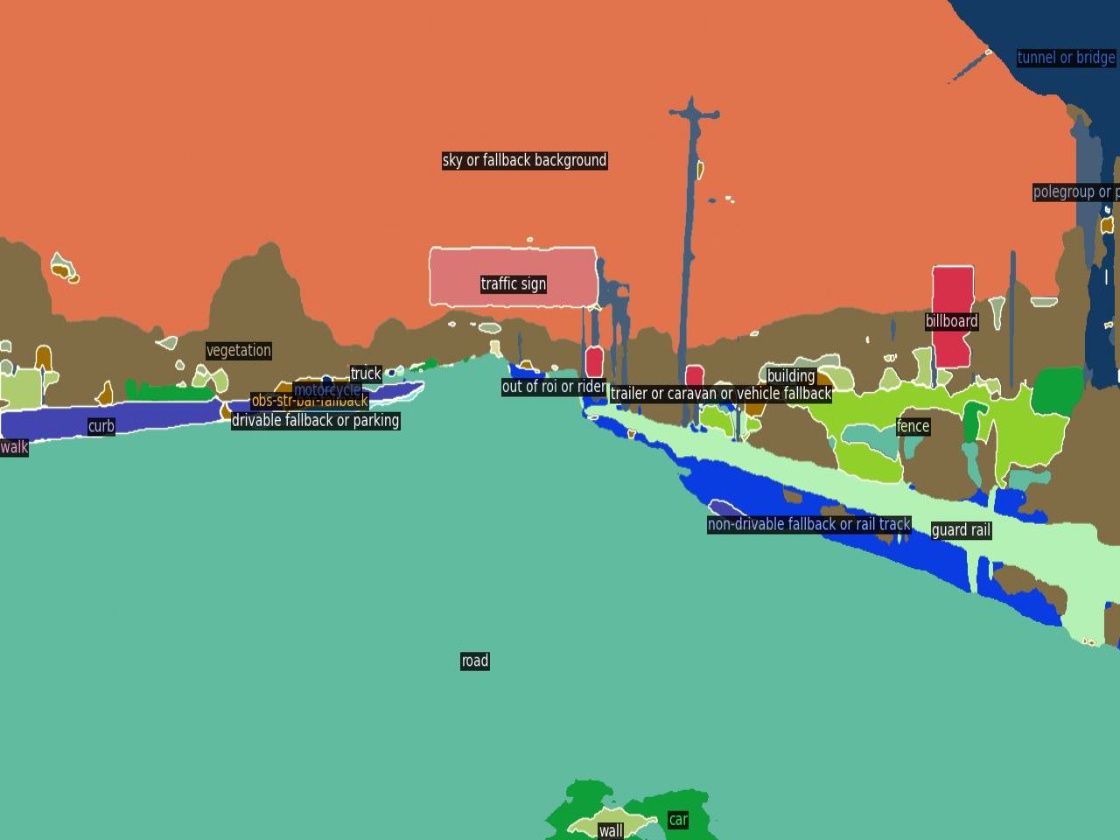} &
    \includegraphics[width=0.13\linewidth]{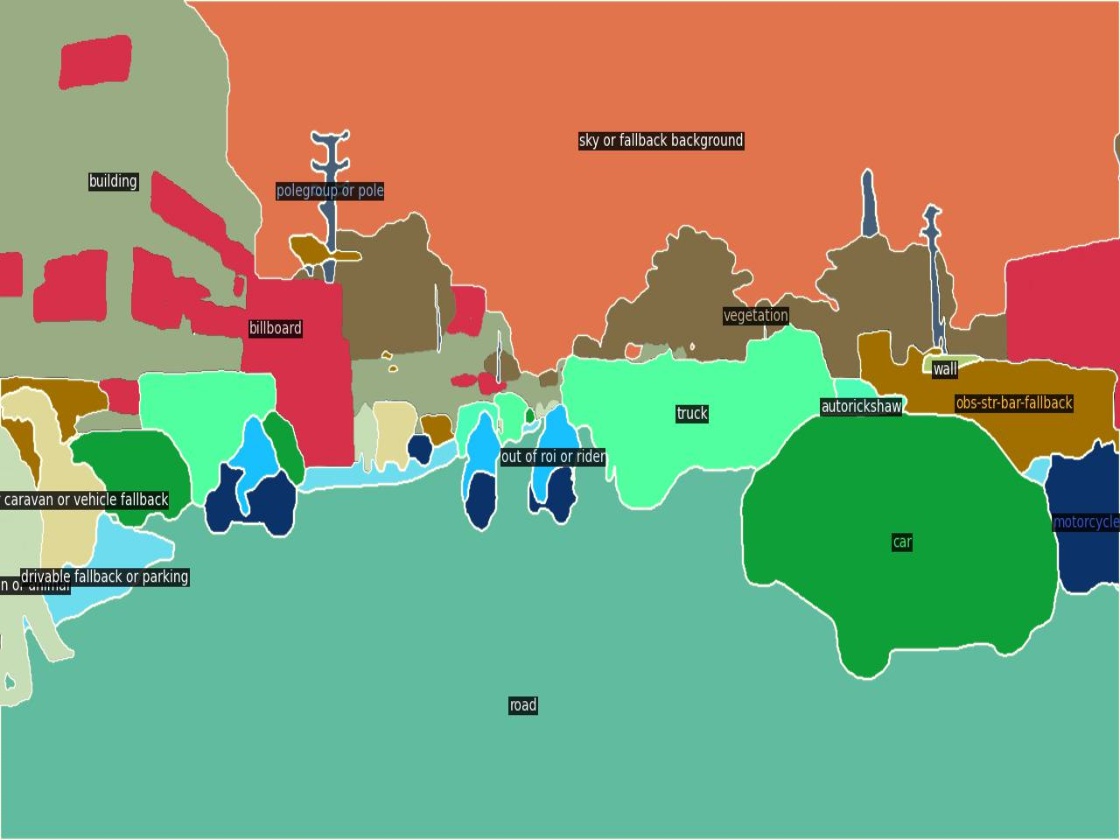} &
    \includegraphics[width=0.13\linewidth]{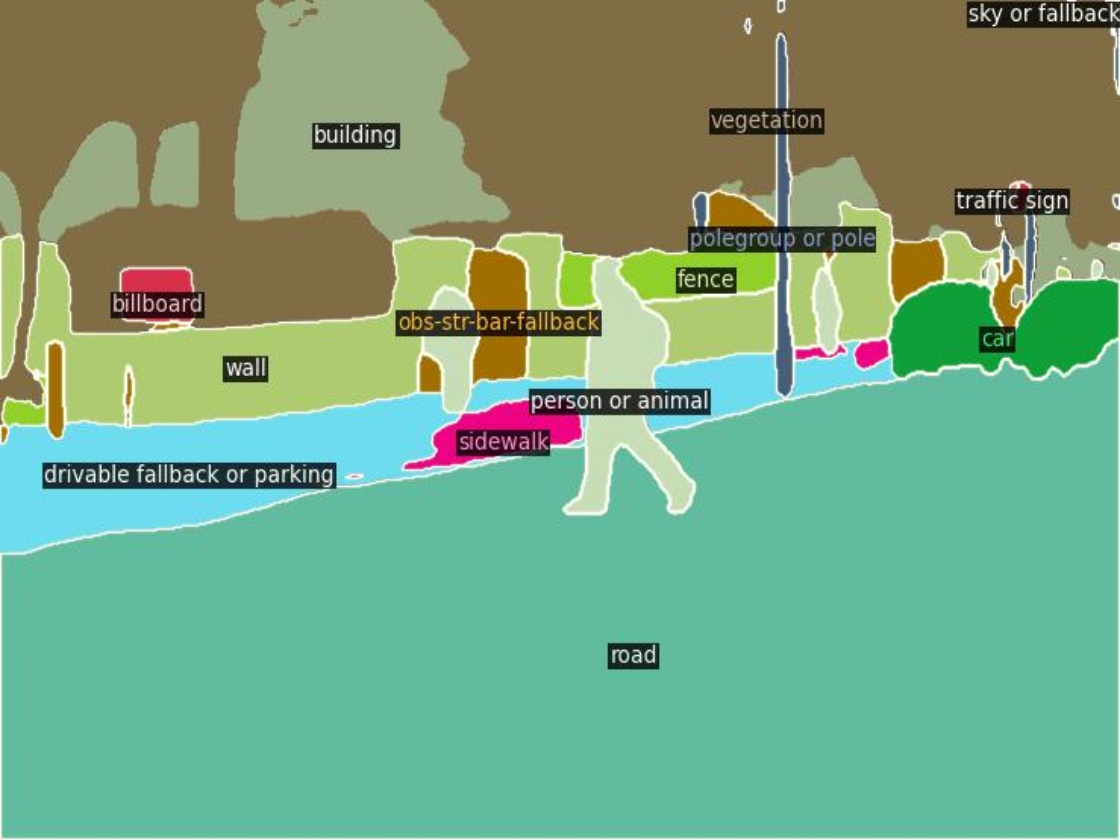} &
    \includegraphics[width=0.13\linewidth]{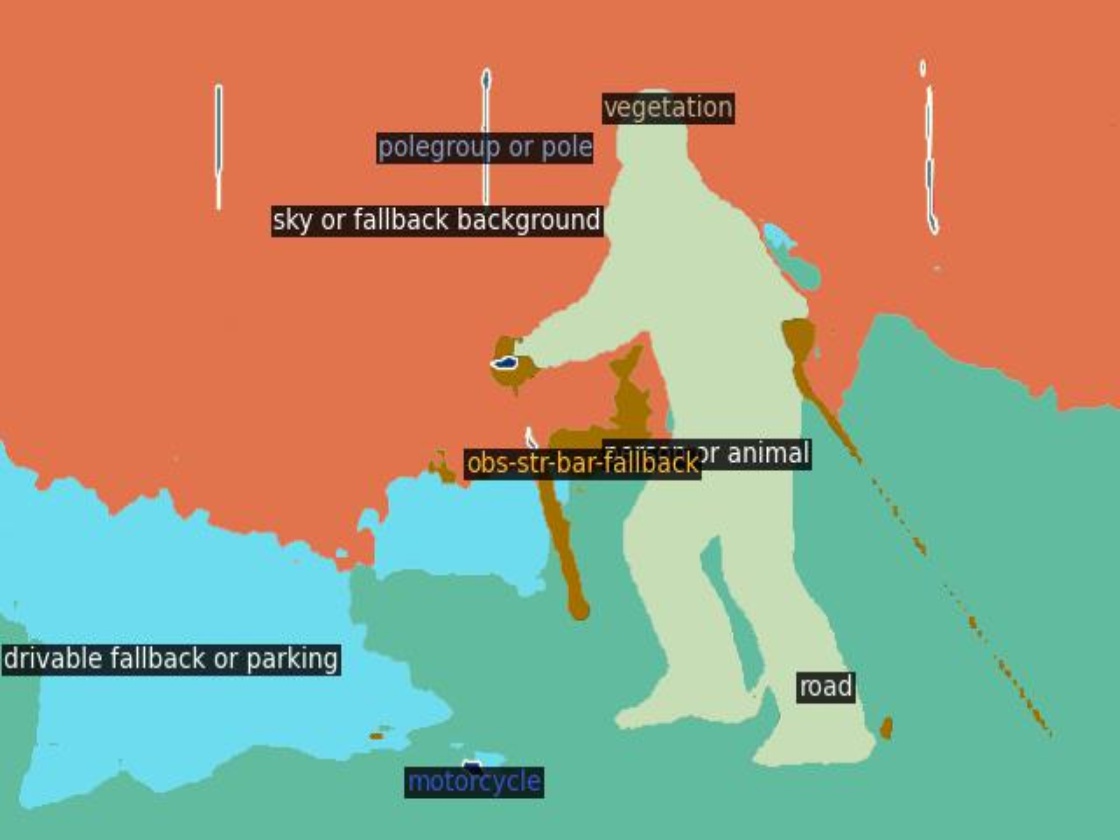} \\
        \rotatebox[origin=l]{90}{ADE} &
    \includegraphics[width=0.13\linewidth]{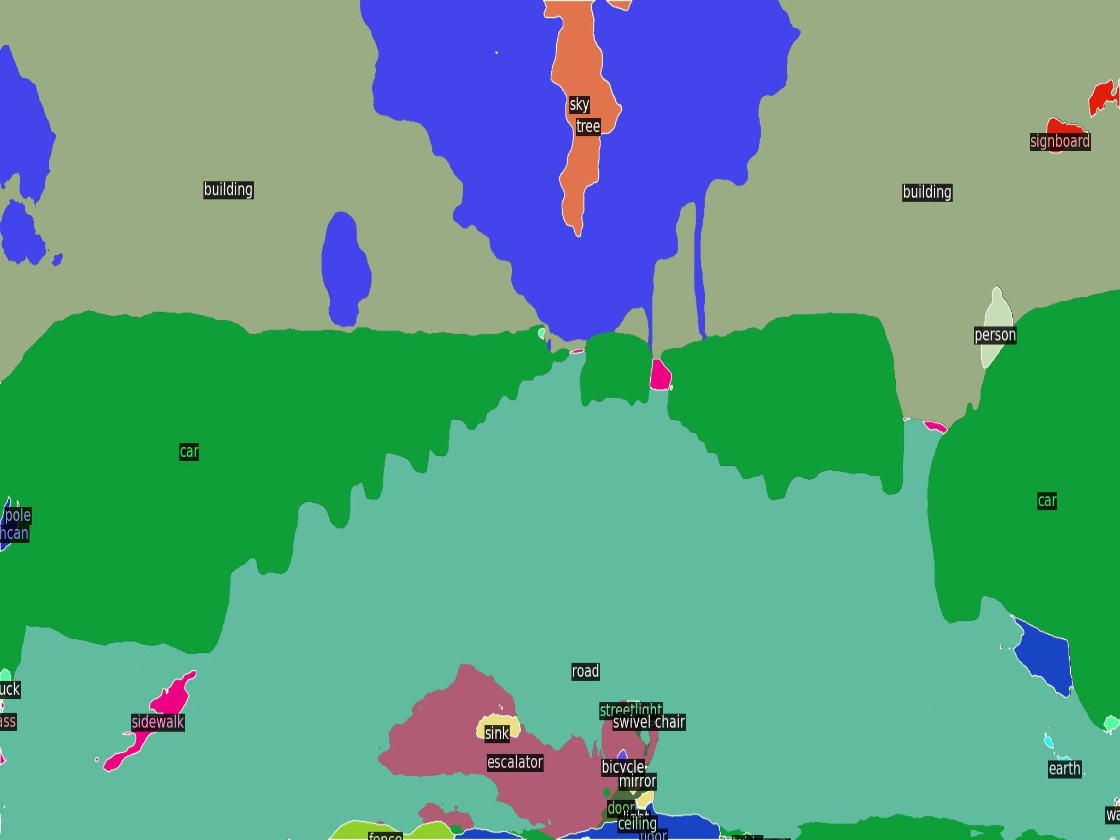} &
    \includegraphics[width=0.13\linewidth]{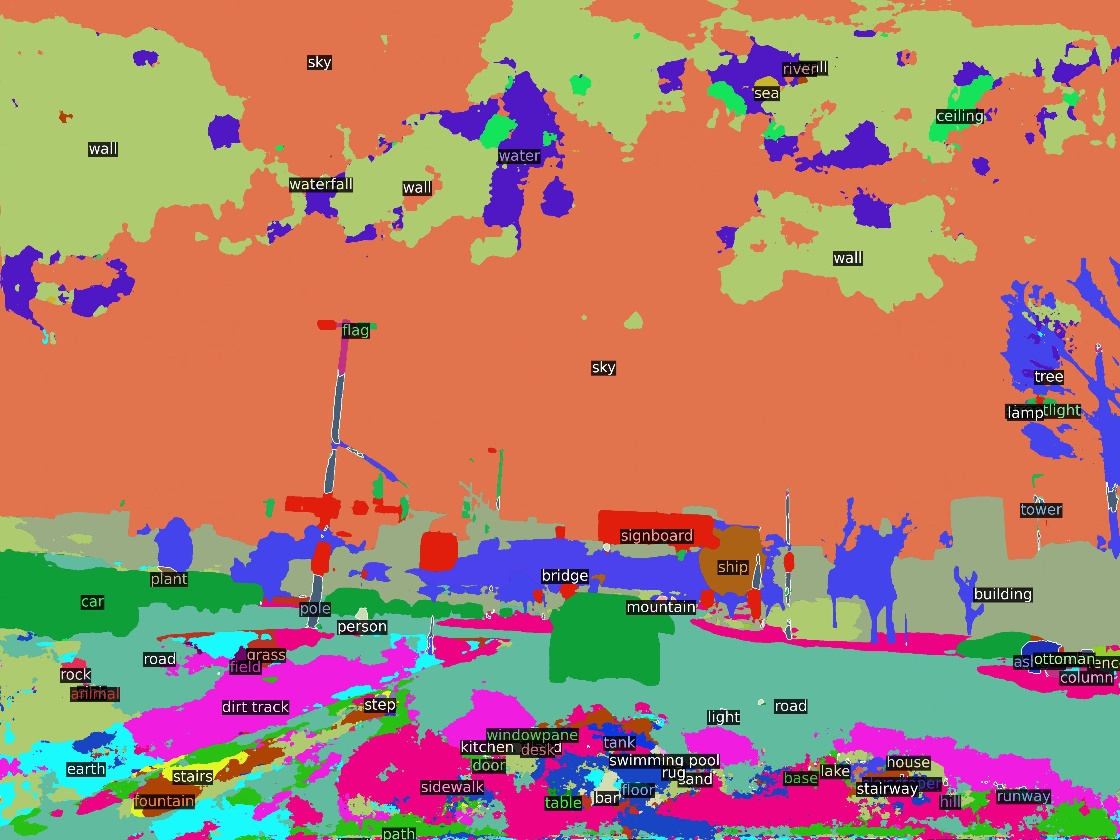} &
    \includegraphics[width=0.13\linewidth]{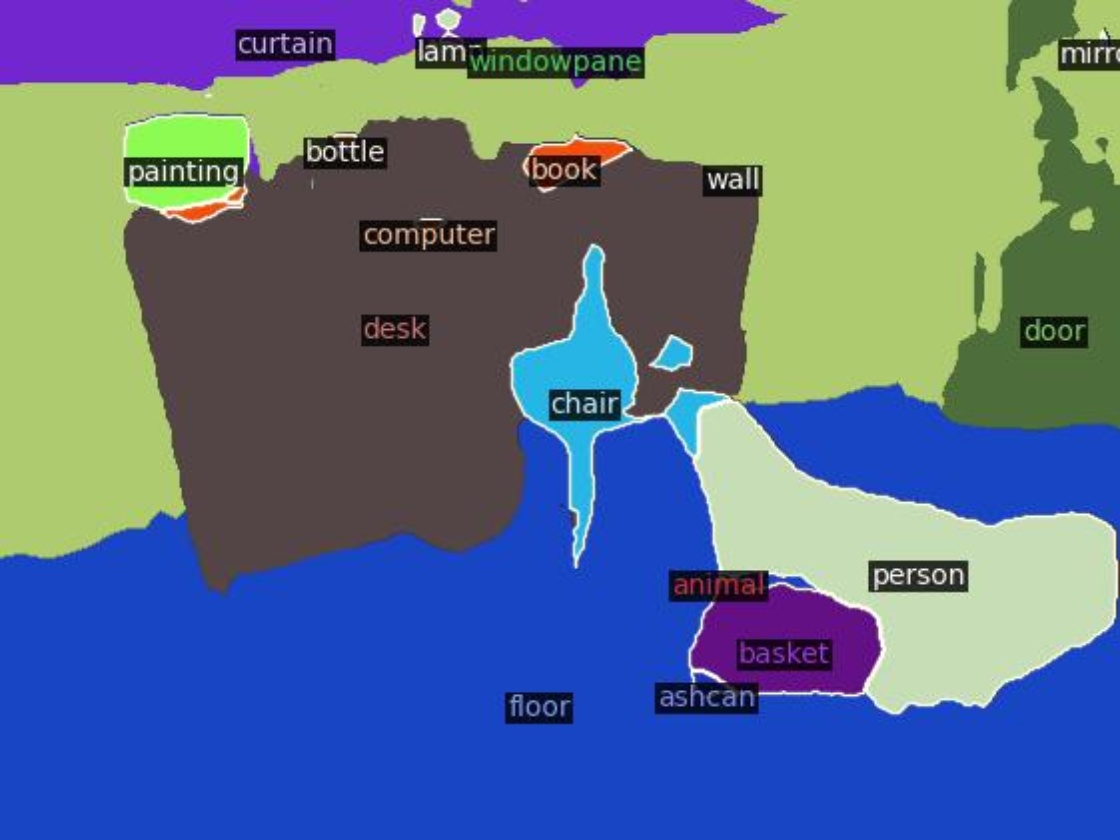} &
    \includegraphics[width=0.13\linewidth]{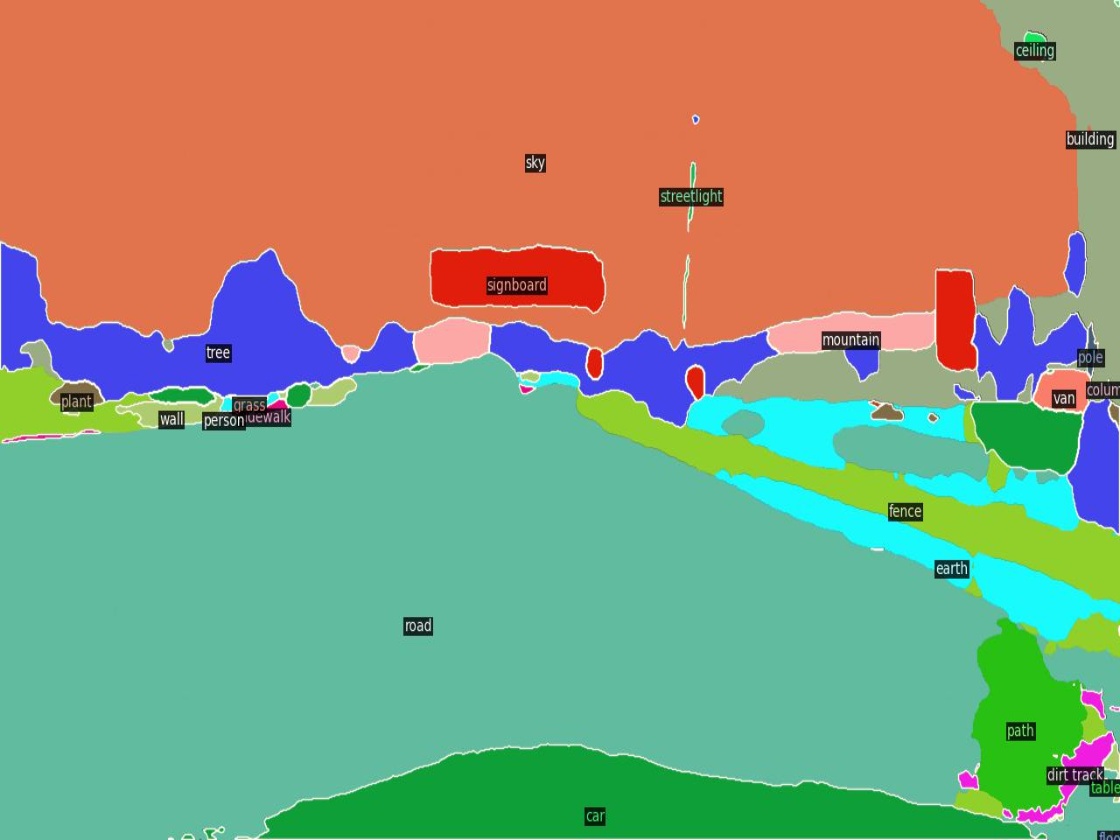} &
    \includegraphics[width=0.13\linewidth]{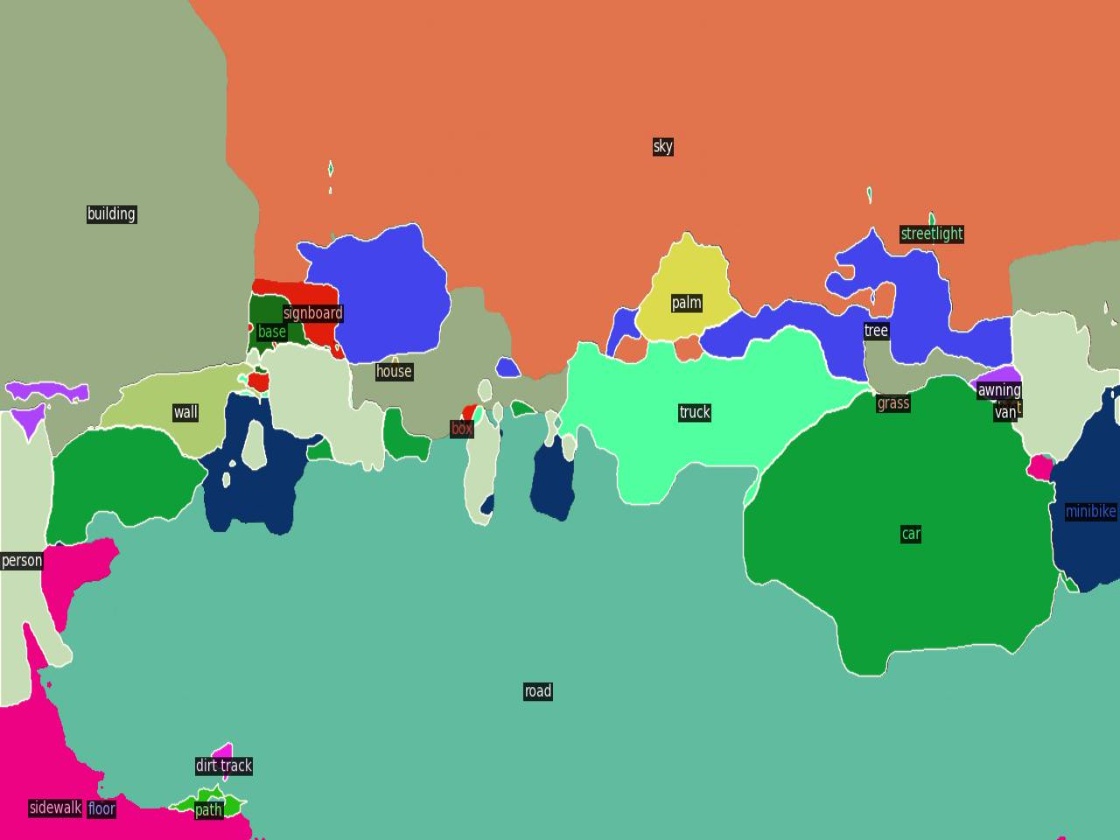} &
    \includegraphics[width=0.13\linewidth]{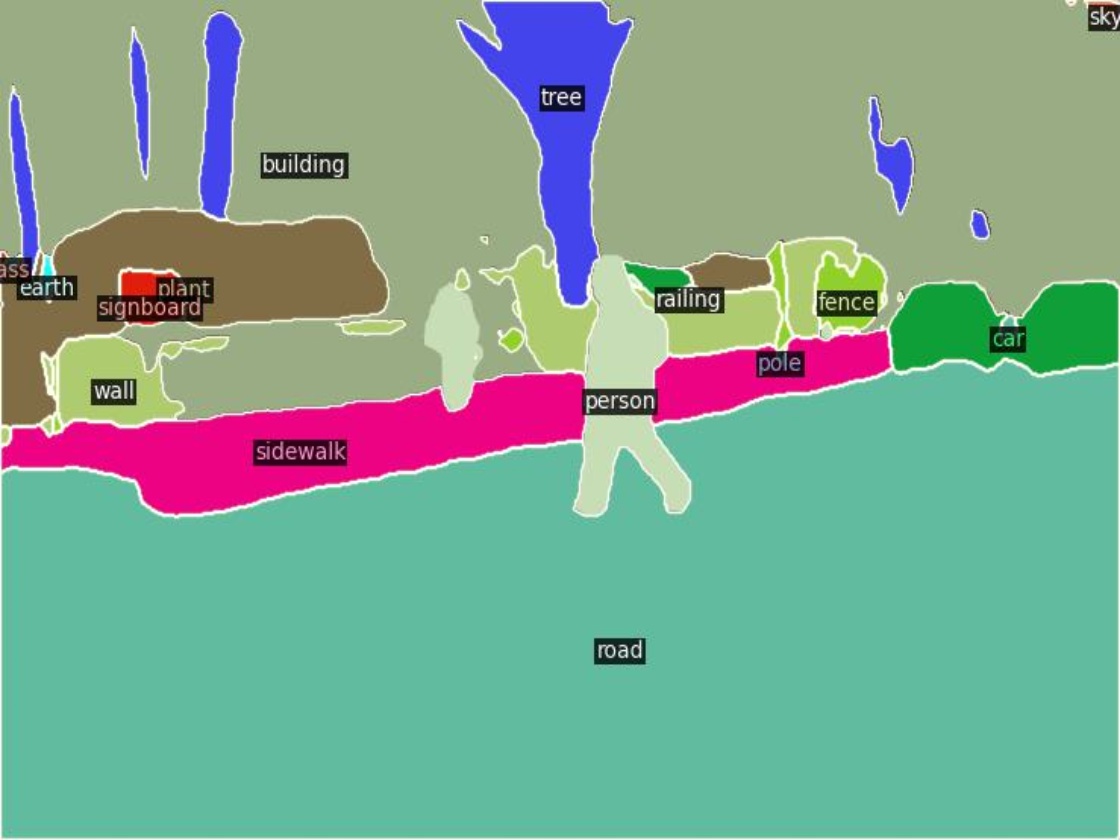} &
    \includegraphics[width=0.13\linewidth]{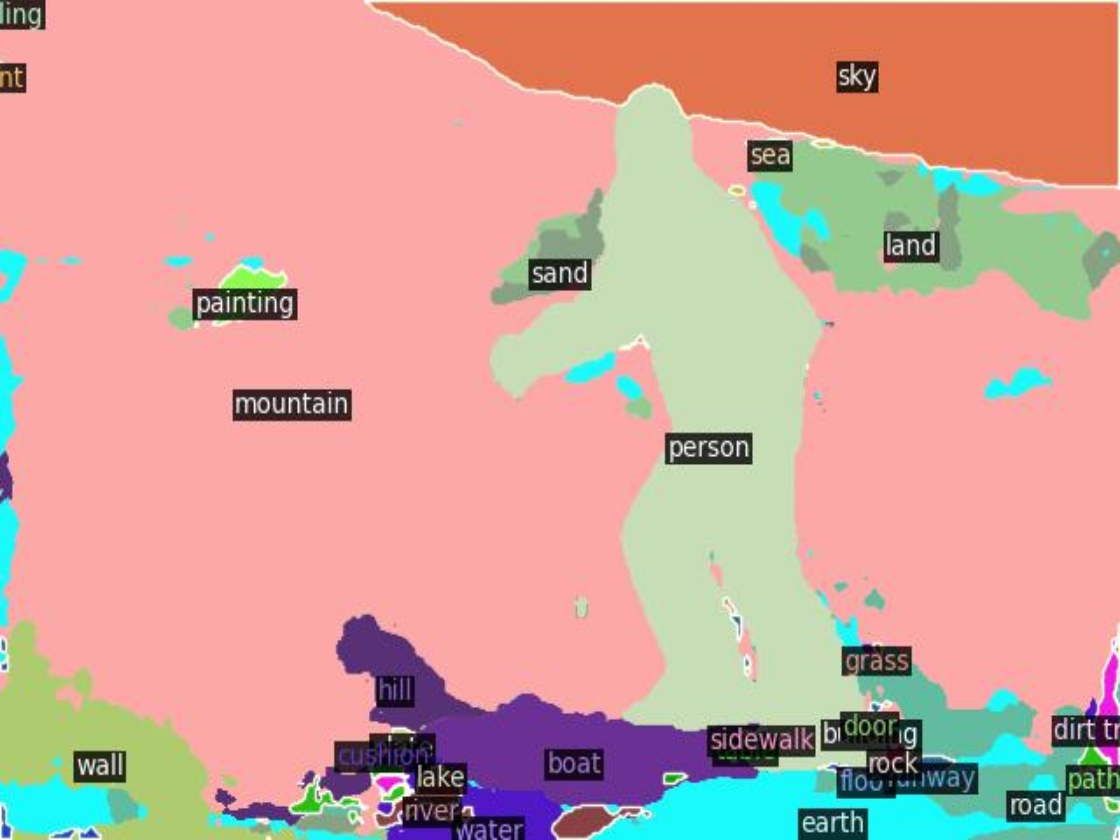} \\
        \rotatebox[origin=l]{90}{COCO} &
    \includegraphics[width=0.13\linewidth]{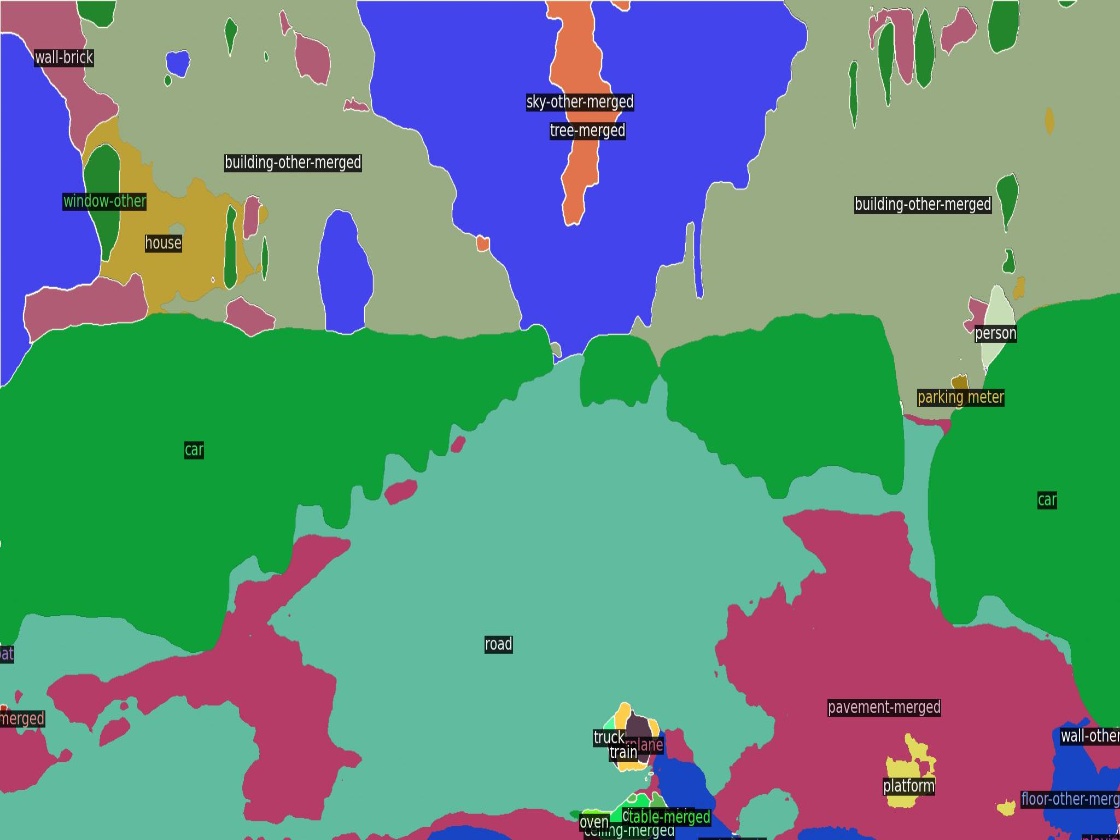} &
    \includegraphics[width=0.13\linewidth]{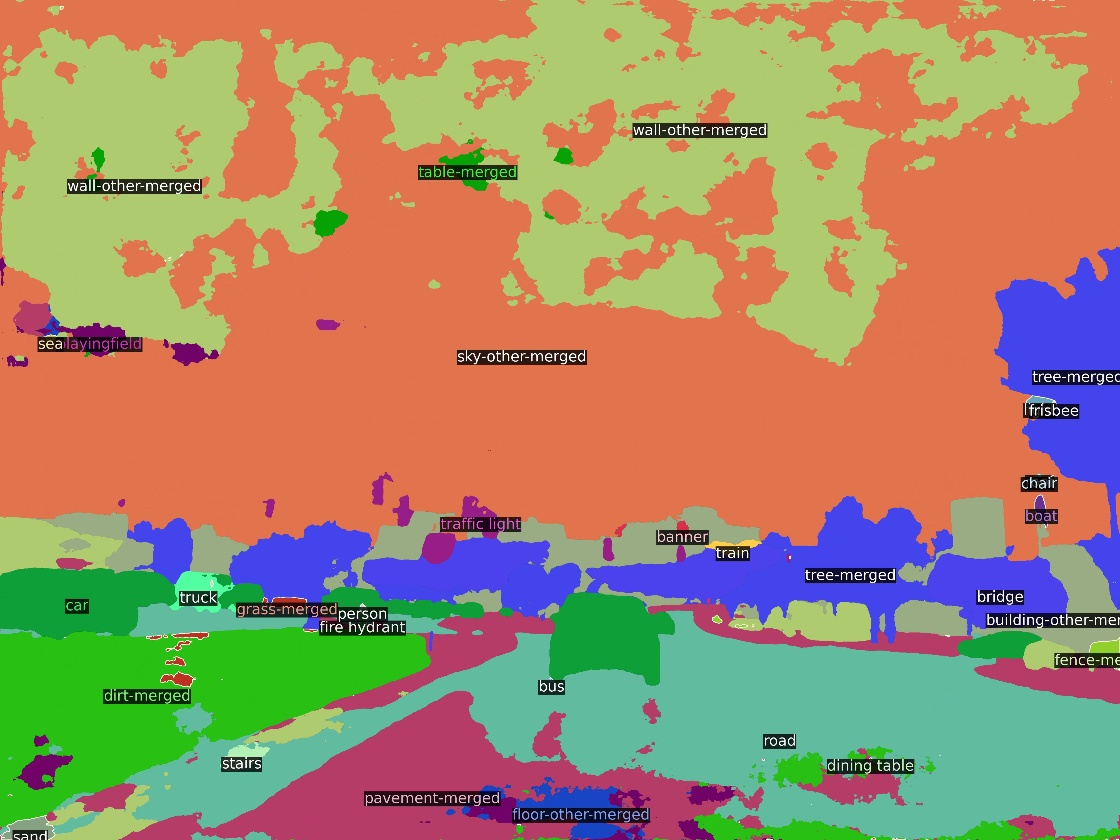} &
    \includegraphics[width=0.13\linewidth]{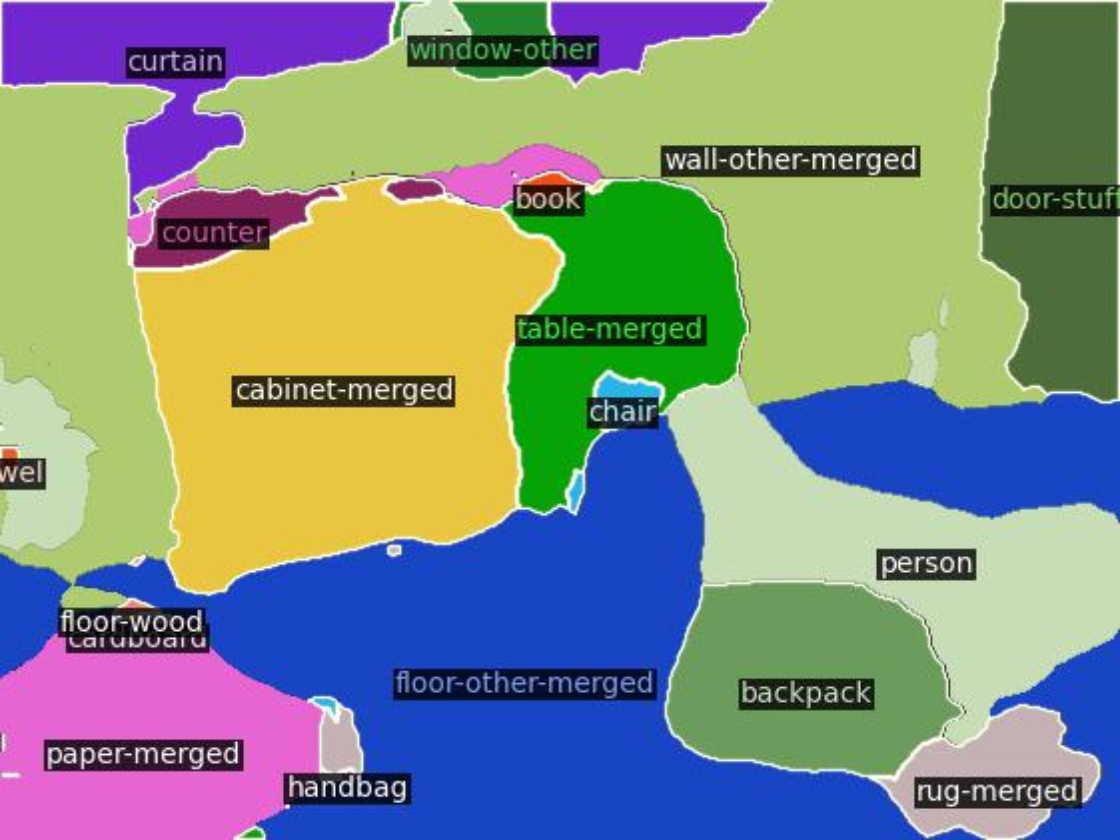} &
    \includegraphics[width=0.13\linewidth]{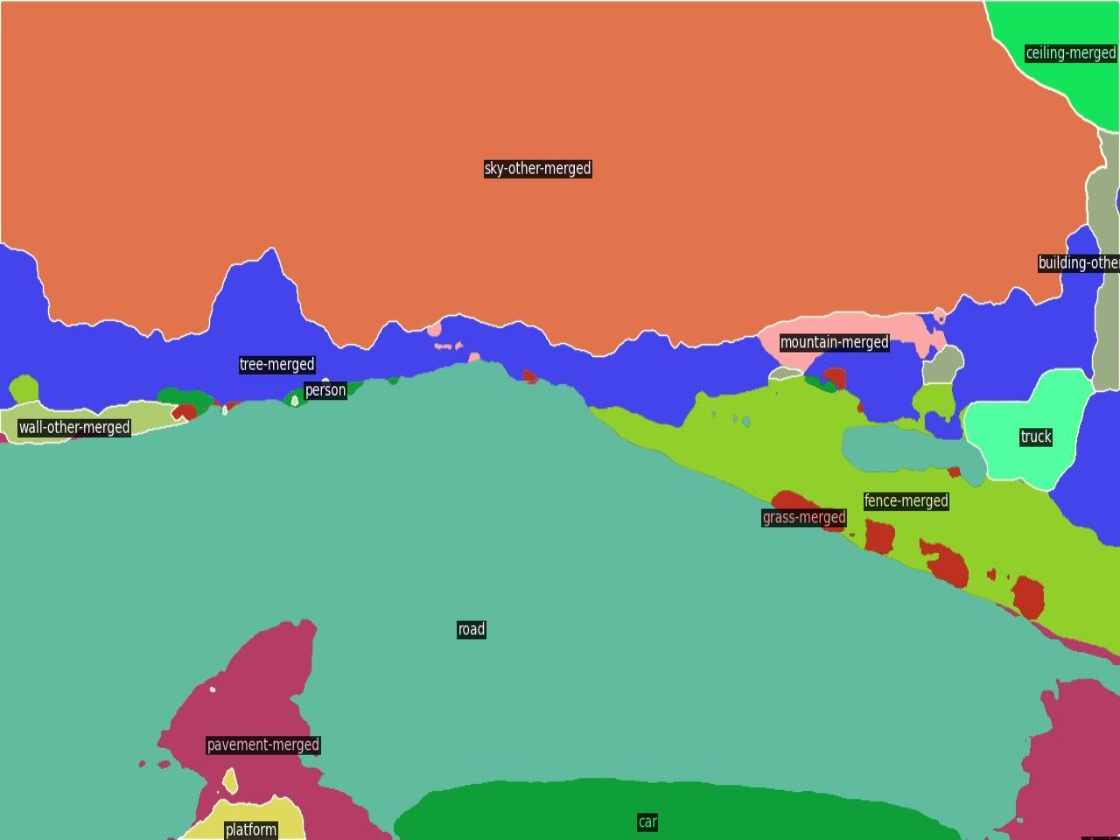} &
    \includegraphics[width=0.13\linewidth]{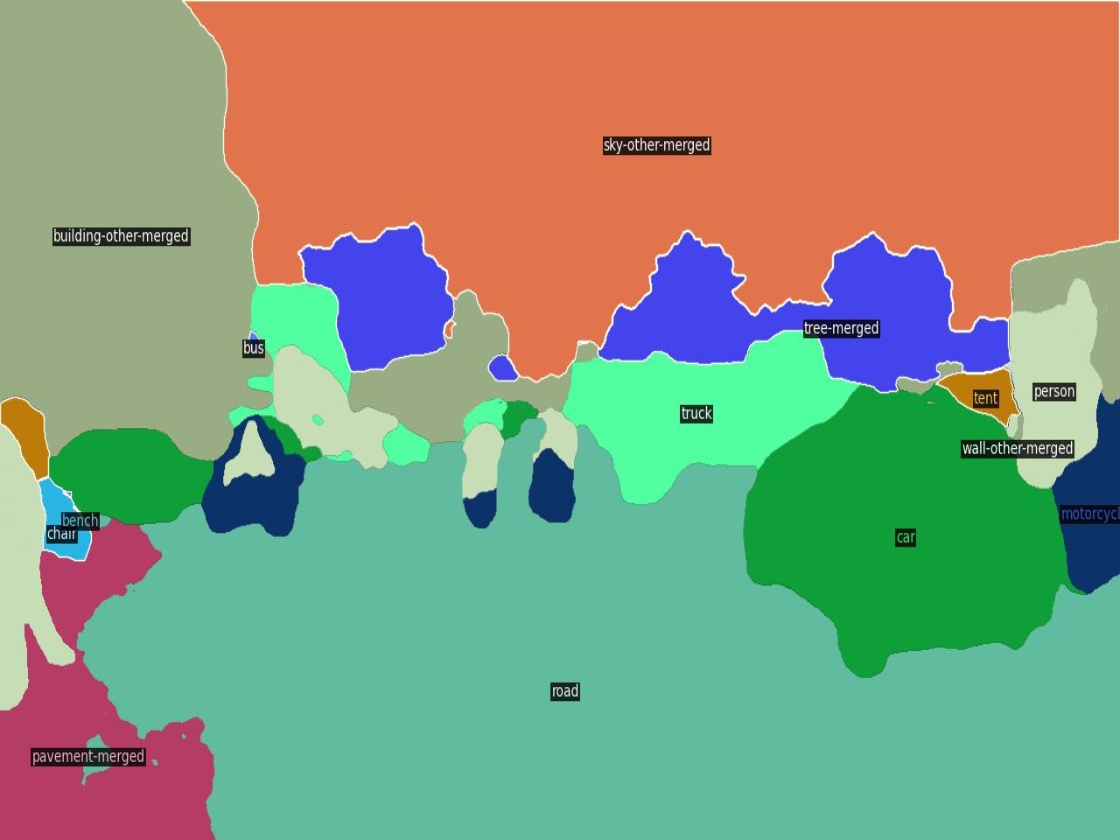} &
    \includegraphics[width=0.13\linewidth]{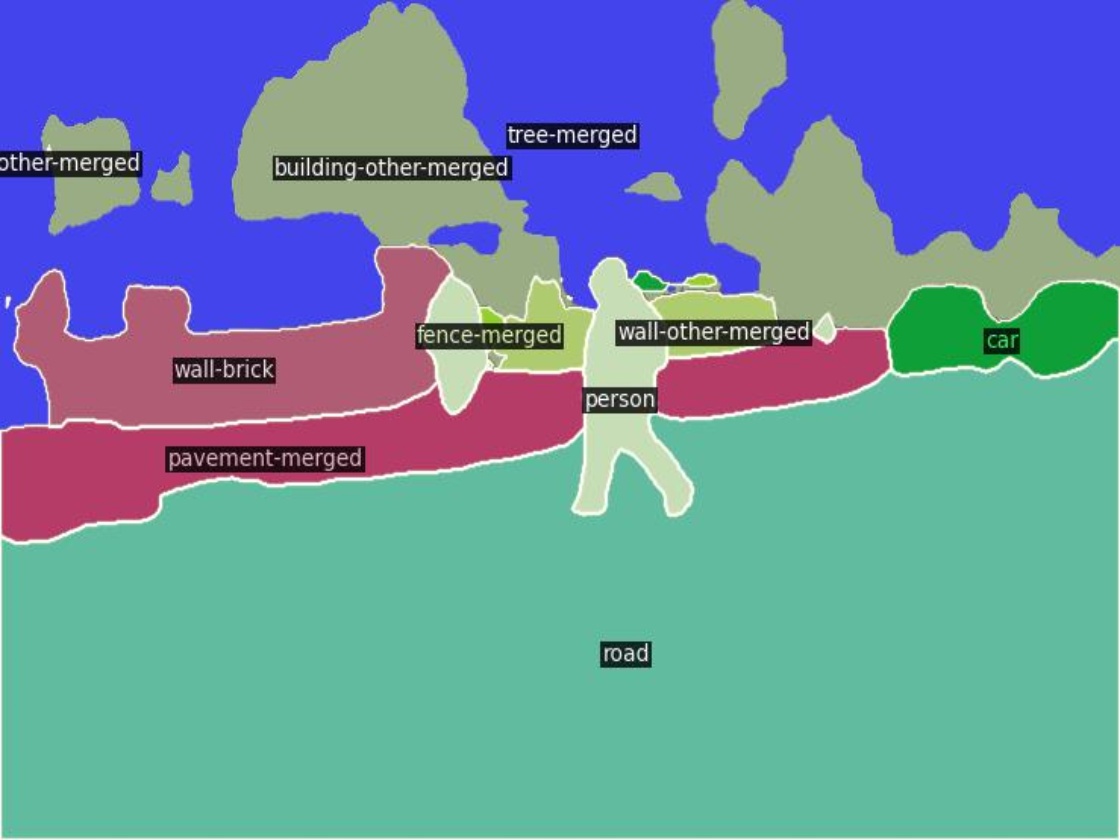} &
    \includegraphics[width=0.13\linewidth]{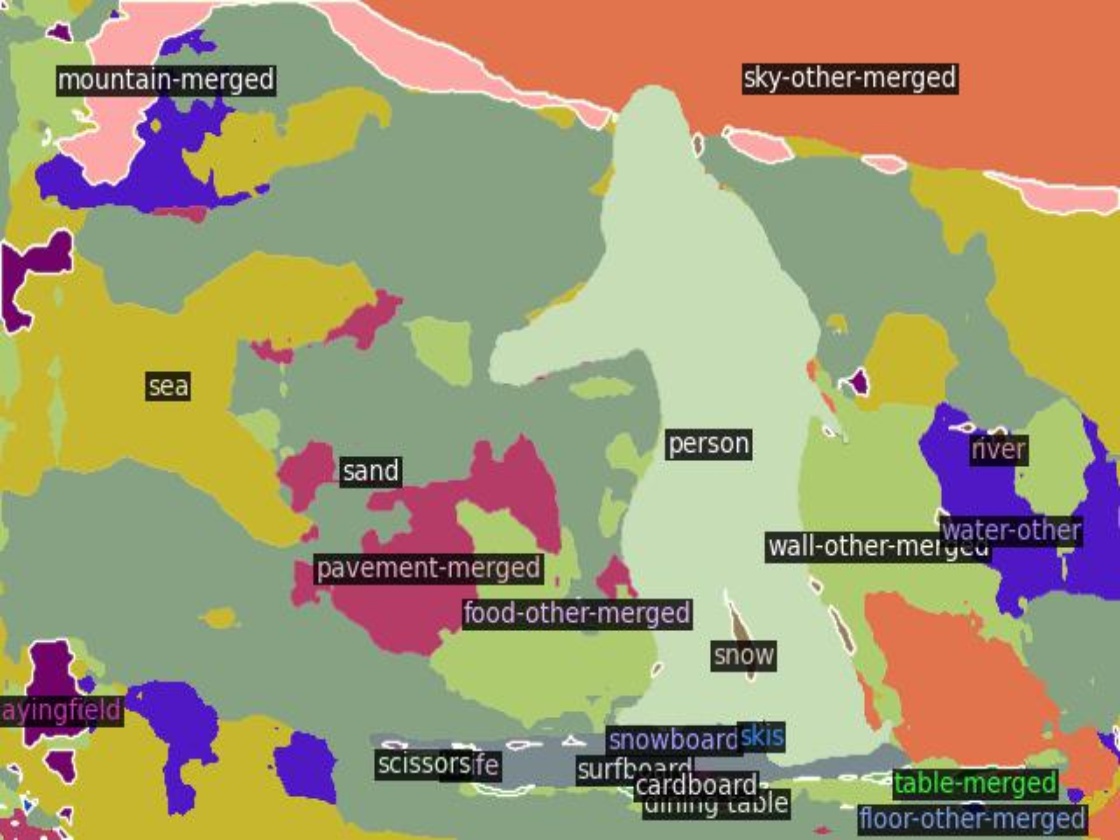} \\
  \end{tabular}
  \caption{Visual comparisons on training datasets.}
  \label{fig:train_im}
\end{figure}

\begin{figure}[h!]
  \centering
  \footnotesize
  \setlength{\tabcolsep}{0.02cm} % 设置列之间的间距
  \begin{tabular}{lcccccc}
      & KT & WD2 & SN & CV & VOC & CT  \\
    \rotatebox[origin=l]{90}{\quad Image} &
    \includegraphics[width=0.15\linewidth]{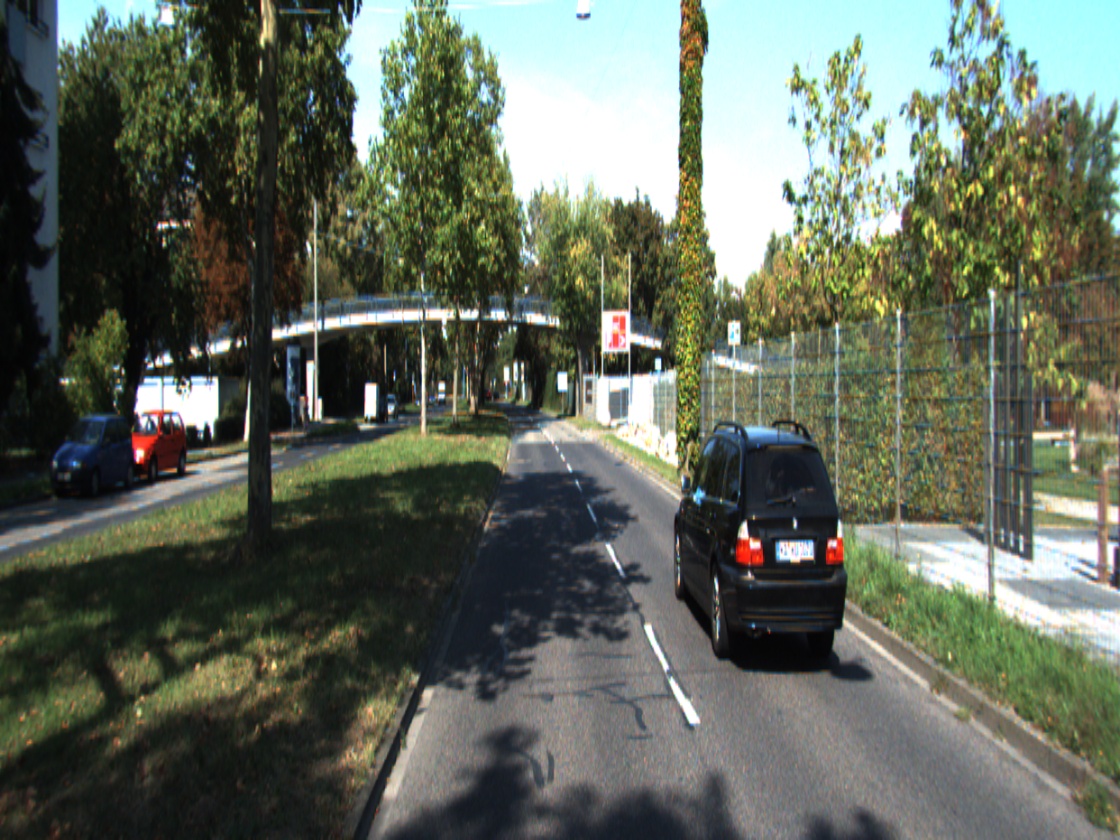} &
    \includegraphics[width=0.15\linewidth]{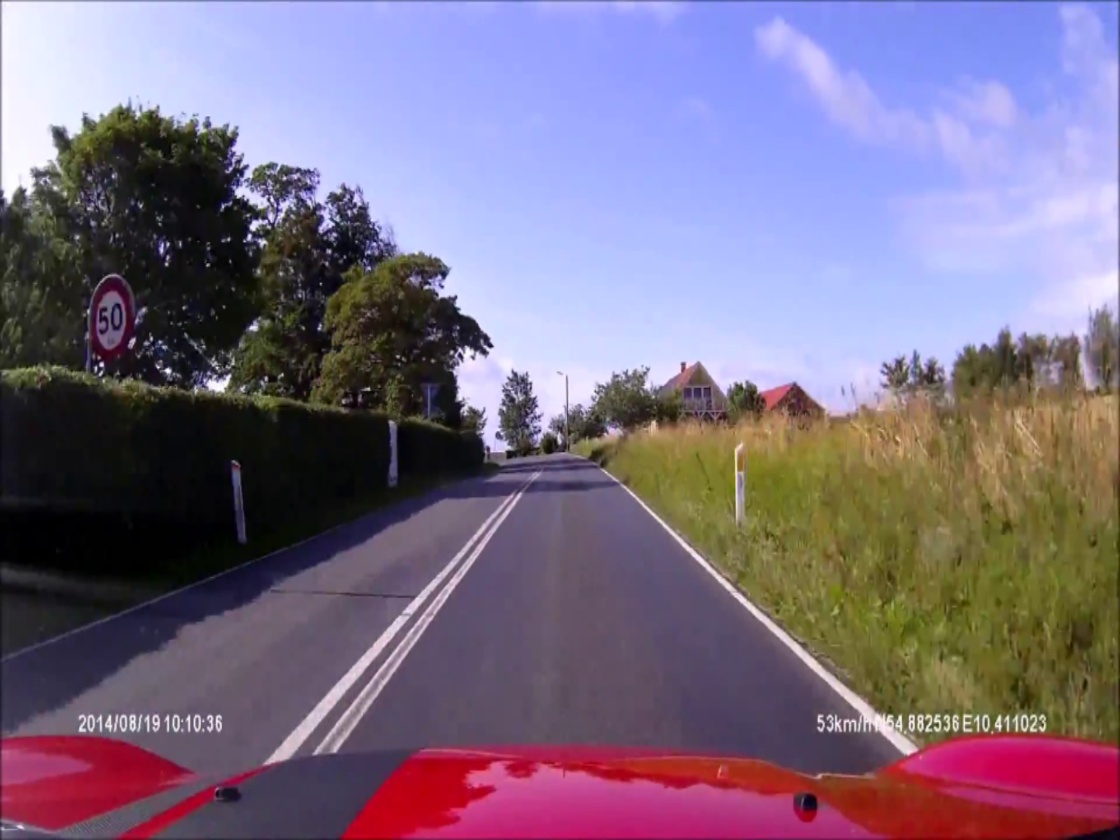} &
    \includegraphics[width=0.15\linewidth]{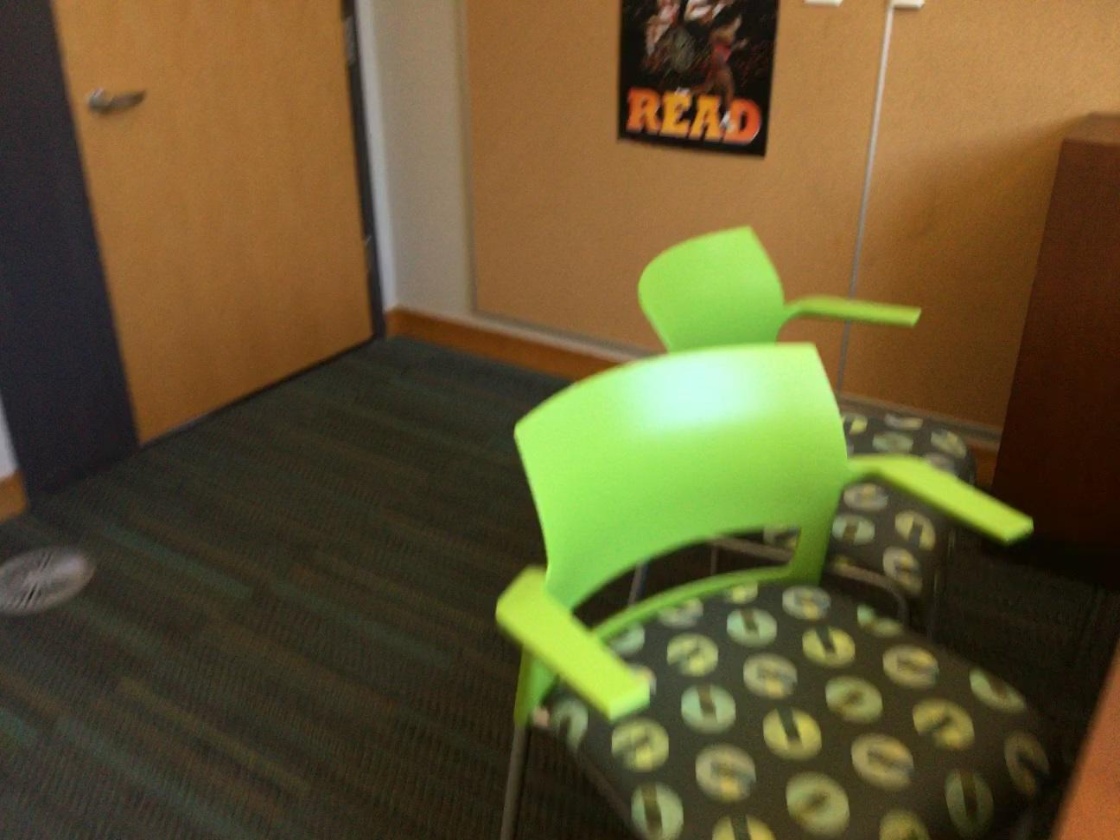} &
    \includegraphics[width=0.15\linewidth]{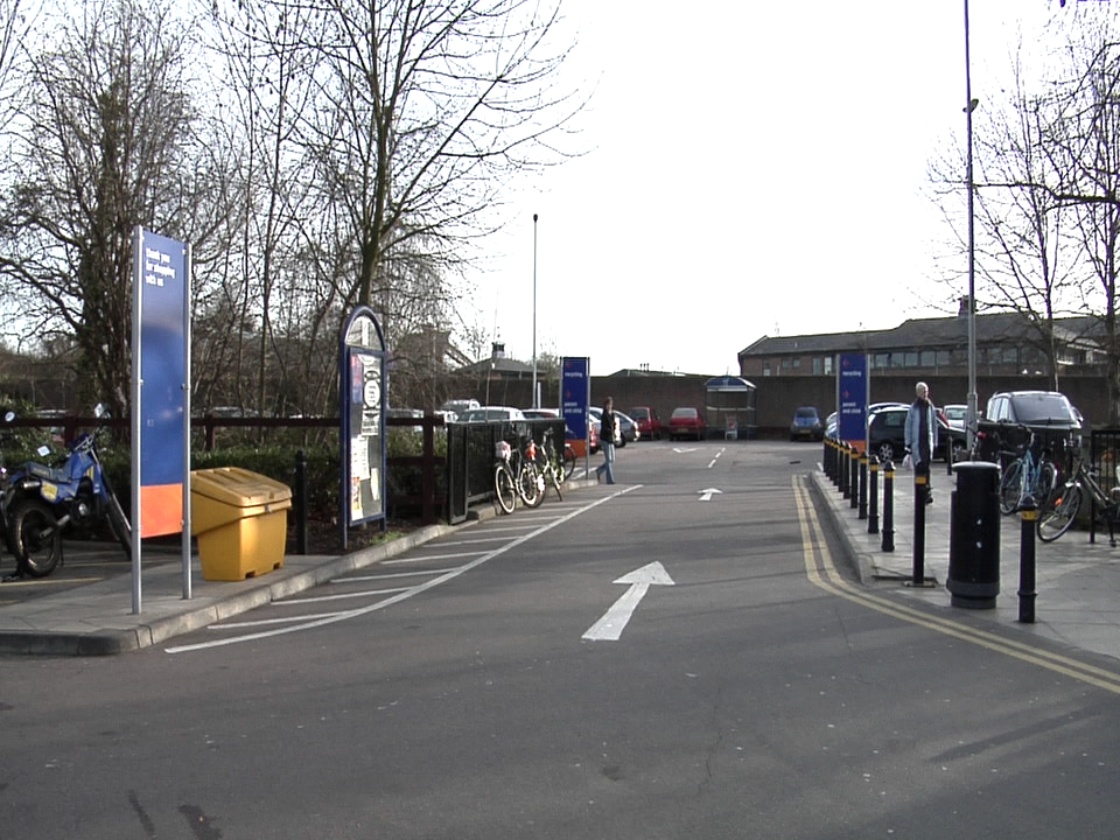} &
    \includegraphics[width=0.15\linewidth]{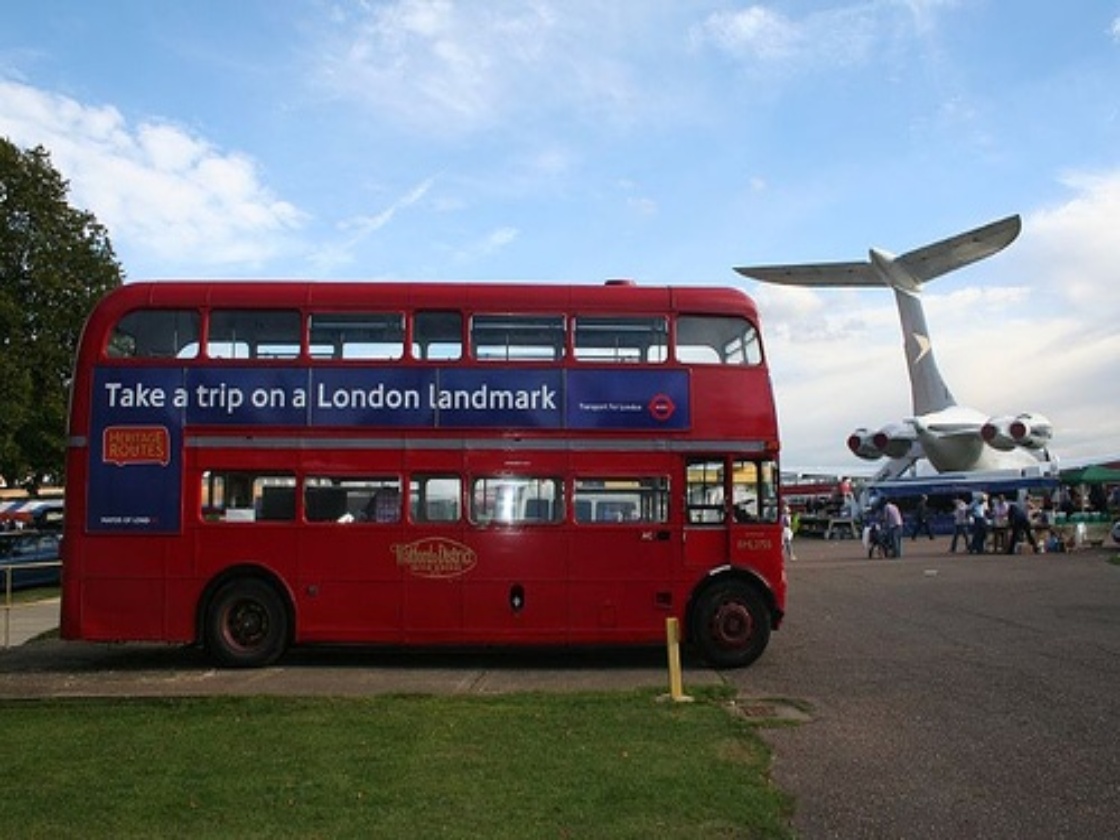} &
    \includegraphics[width=0.15\linewidth]{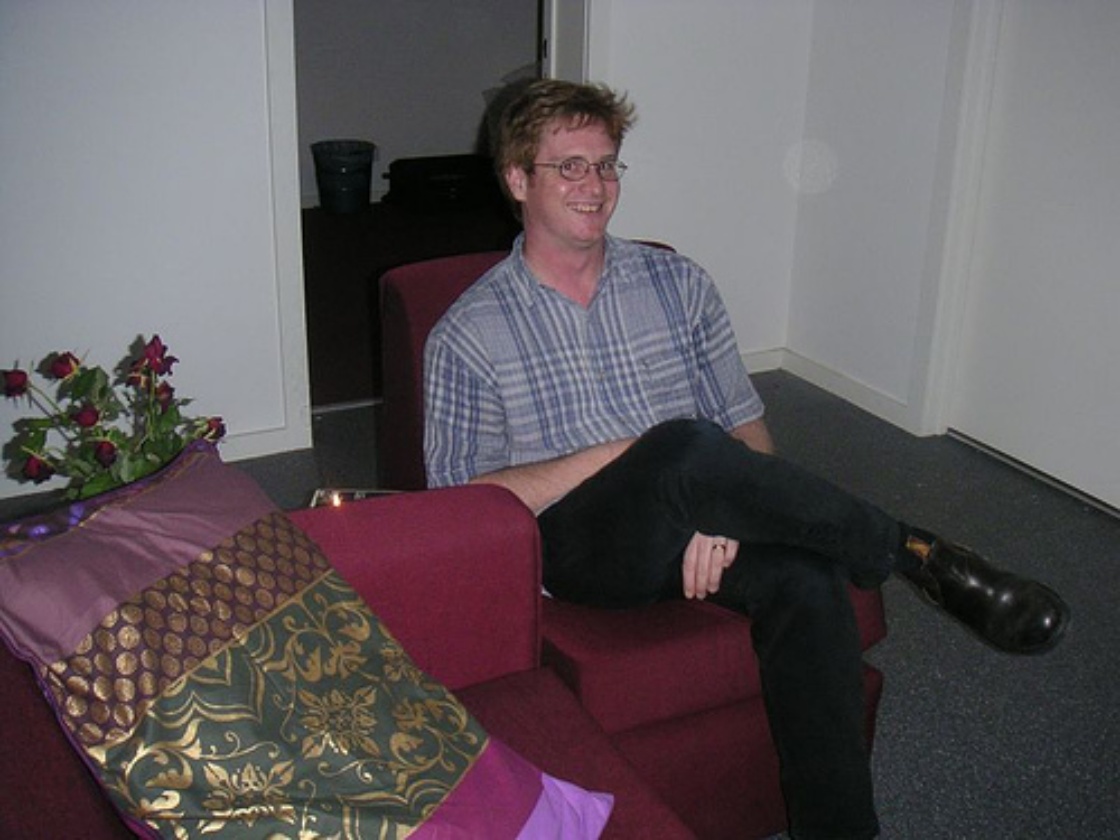} \\
    \rotatebox[origin=l]{90}{\quad GT} &
    \includegraphics[width=0.15\linewidth]{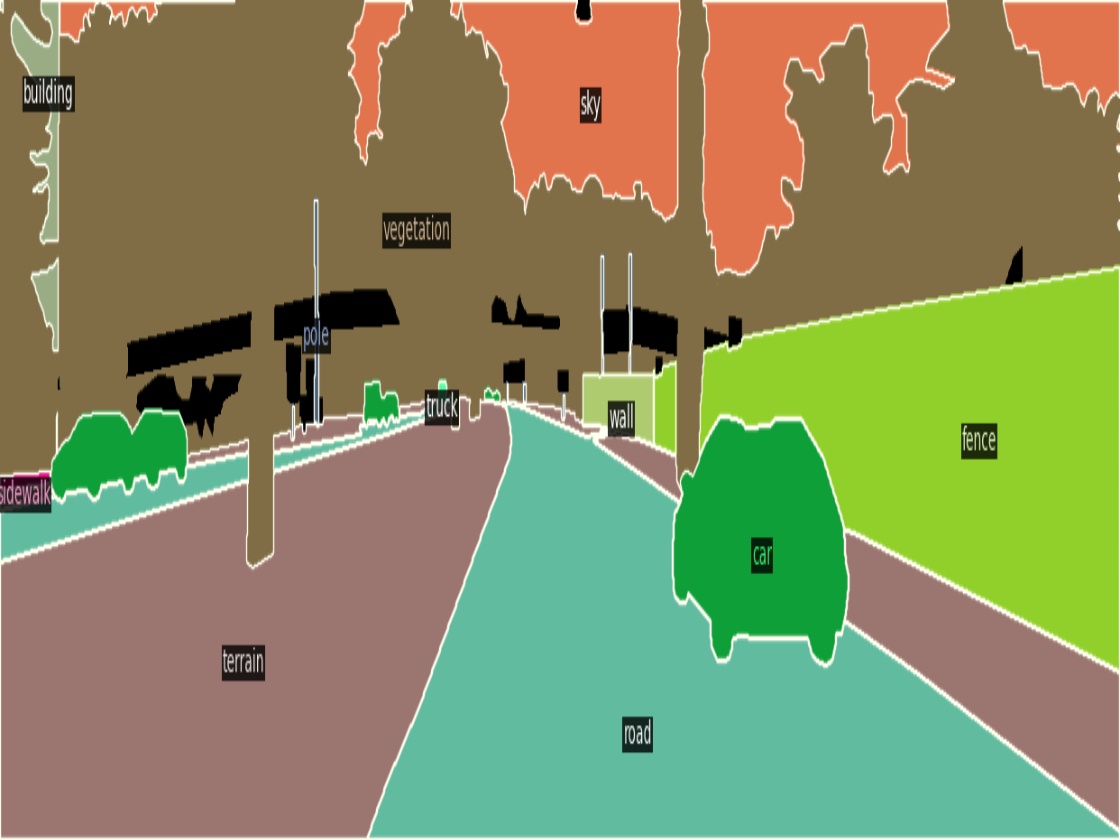} &
    \includegraphics[width=0.15\linewidth]{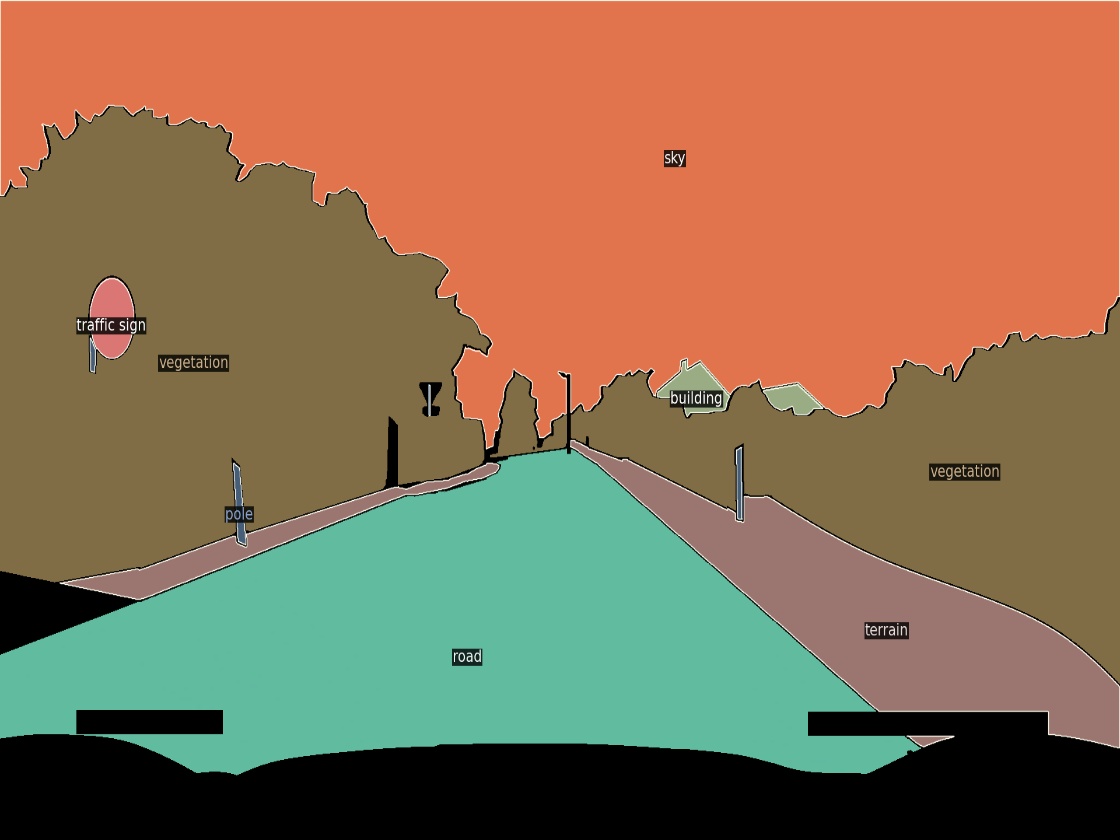} &
    \includegraphics[width=0.15\linewidth]{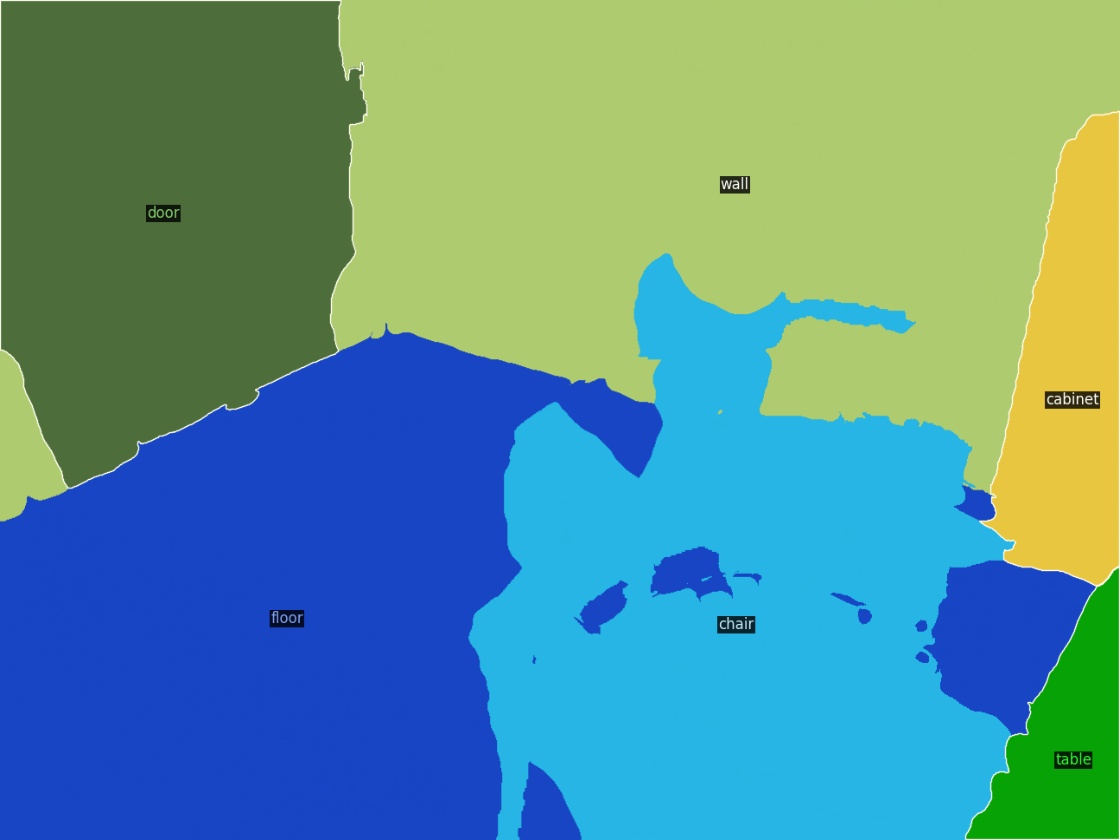} &
    \includegraphics[width=0.15\linewidth]{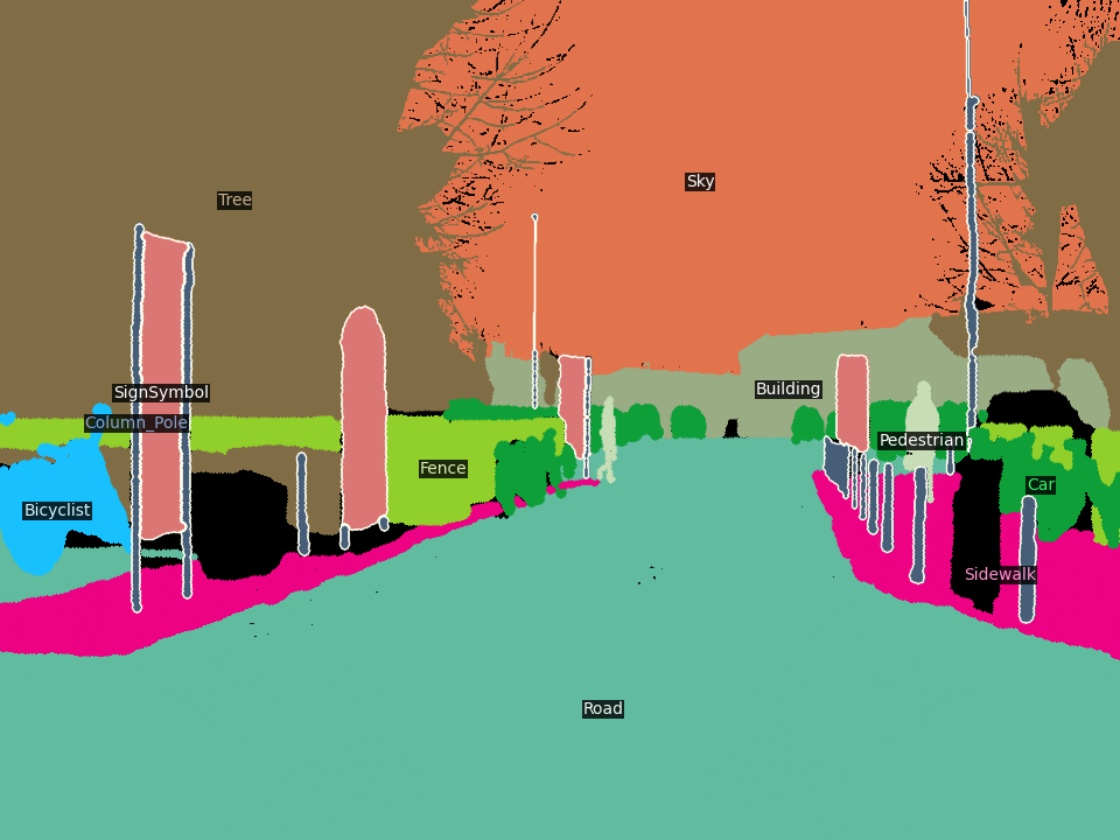} & 
    \includegraphics[width=0.15\linewidth]{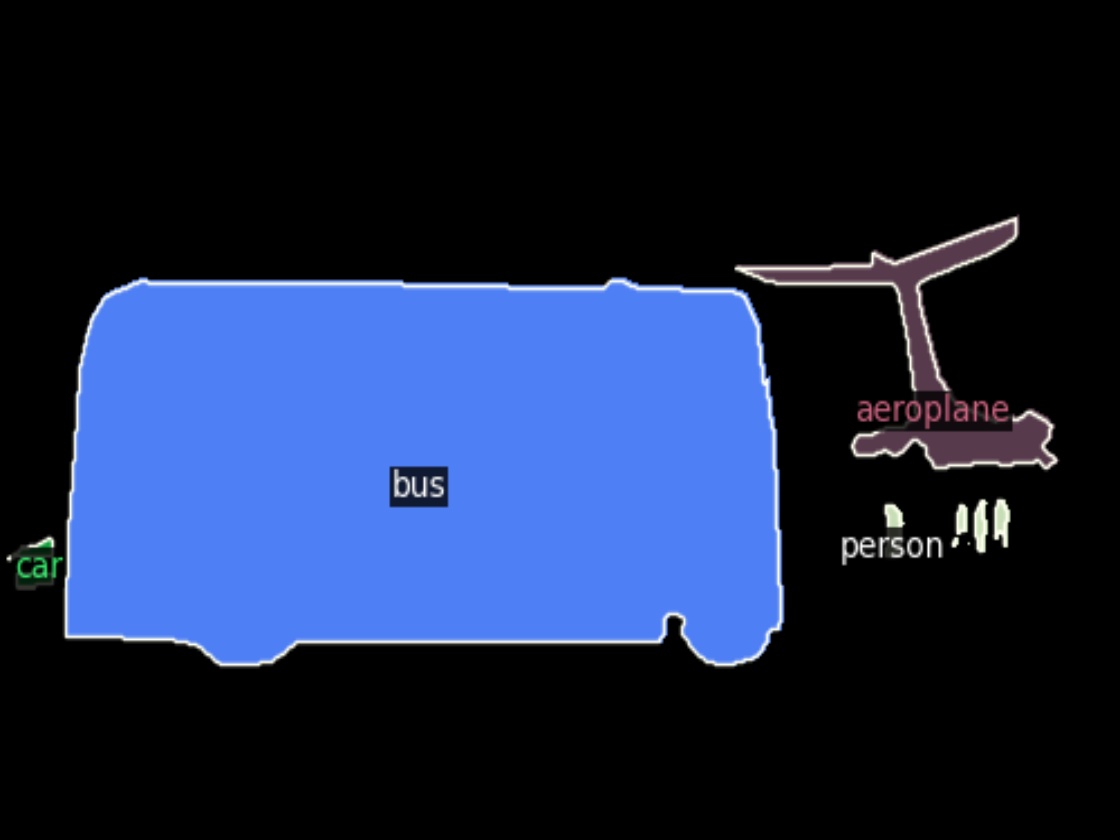} &
    \includegraphics[width=0.15\linewidth]{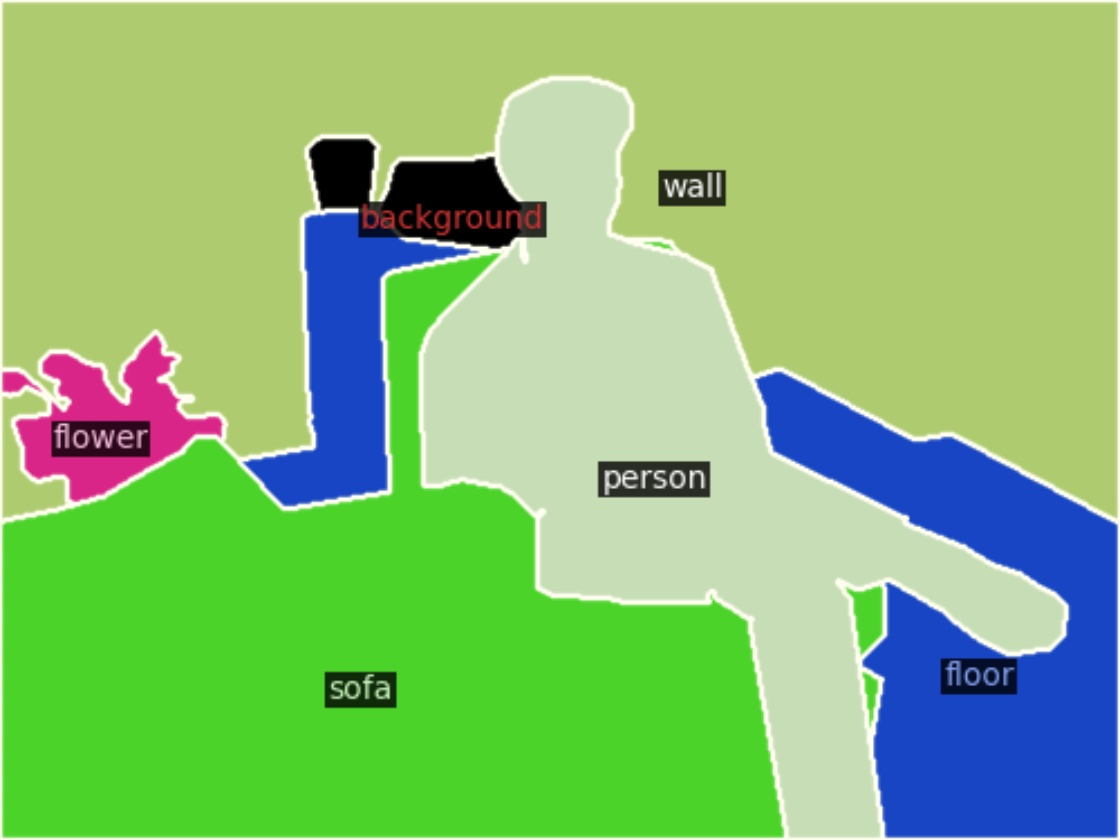} \\
    \rotatebox[origin=l]{90}{\quad Ours} &
    \includegraphics[width=0.15\linewidth]{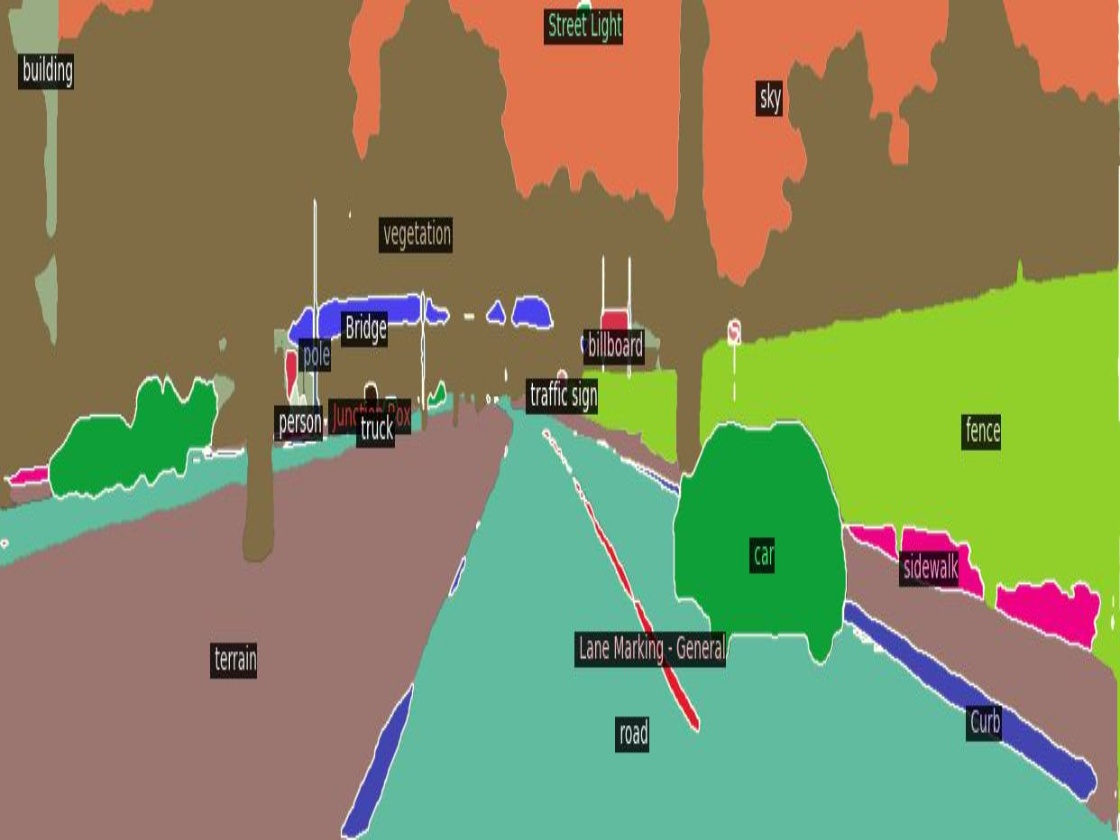} &
    \includegraphics[width=0.15\linewidth]{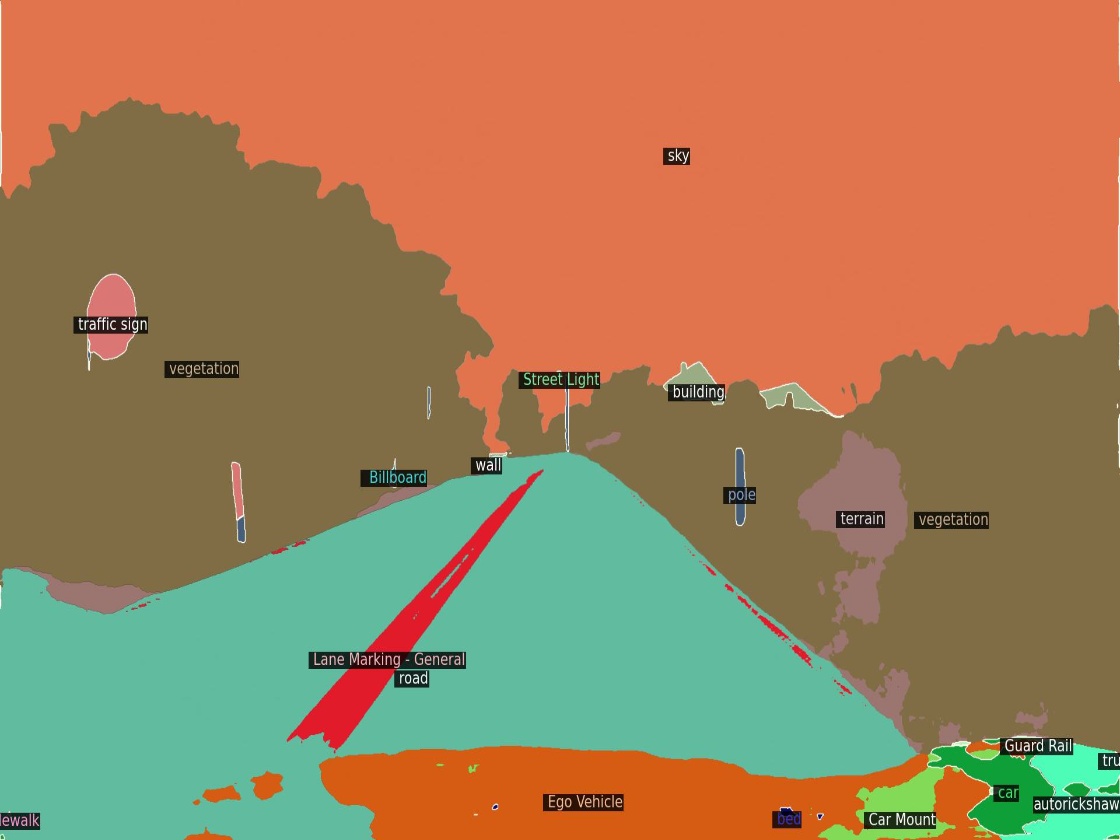} &
    \includegraphics[width=0.15\linewidth]{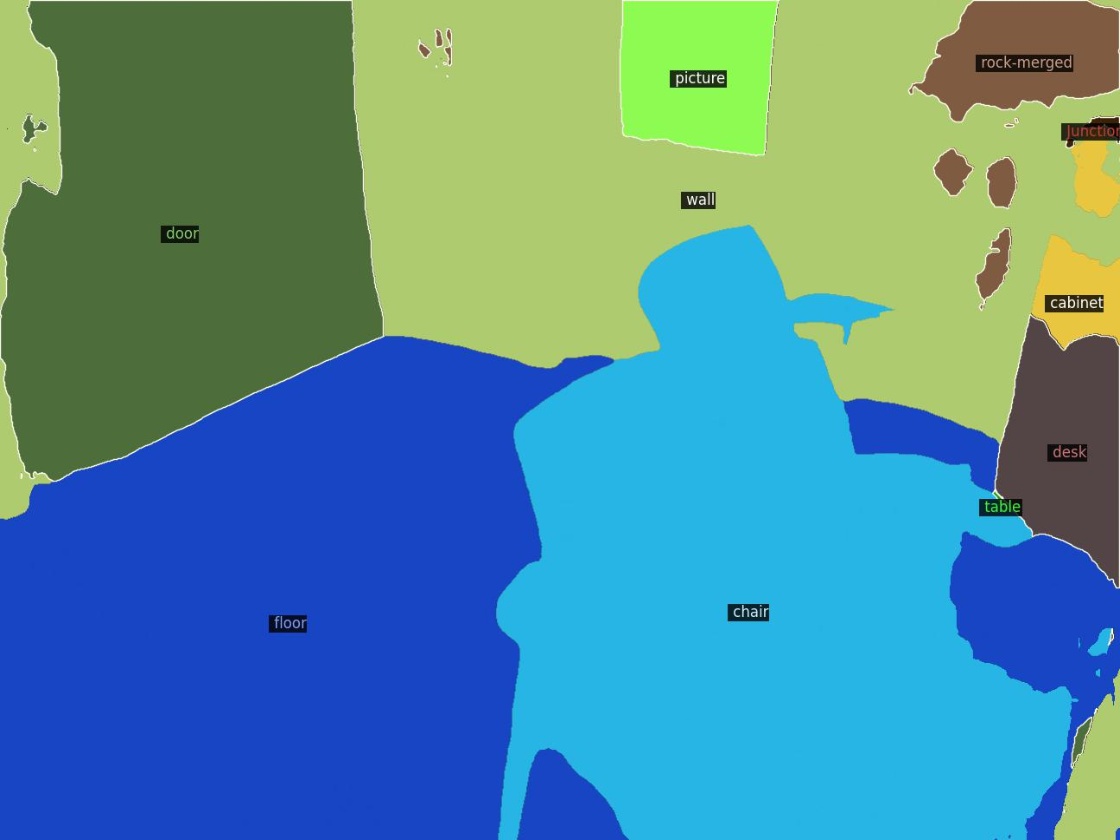} &
    \includegraphics[width=0.15\linewidth]{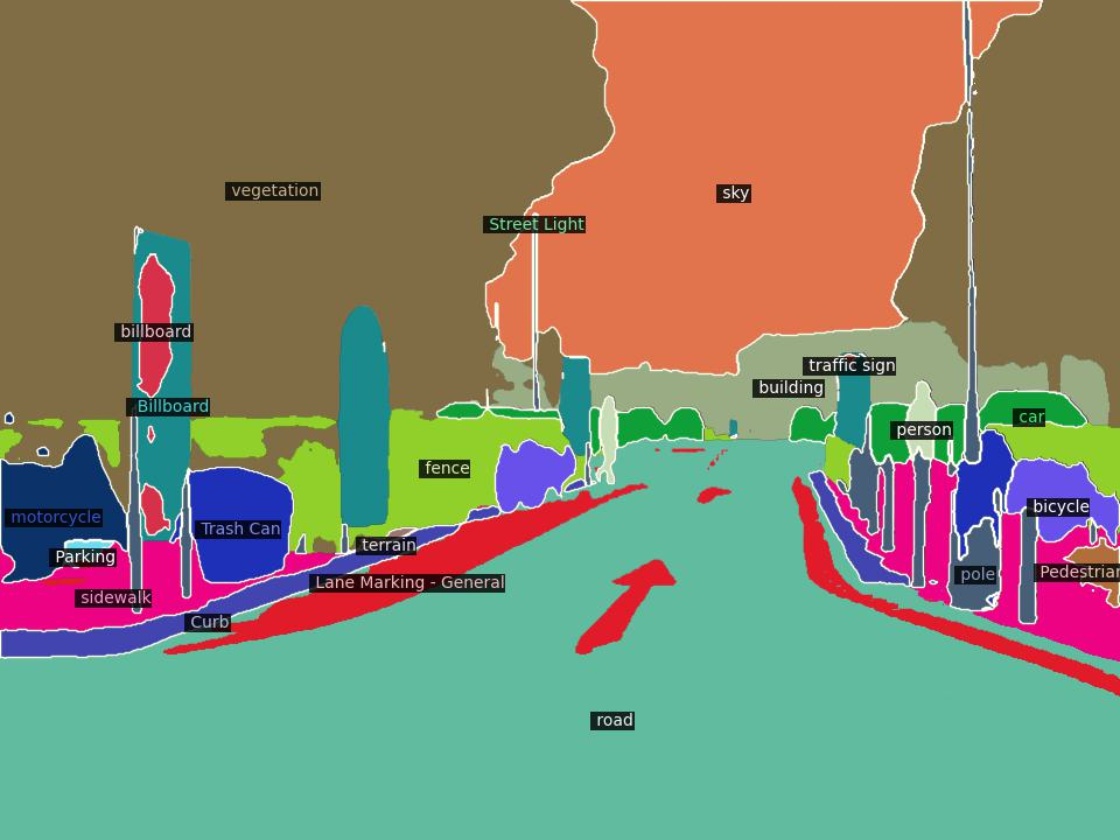} &
    \includegraphics[width=0.15\linewidth]{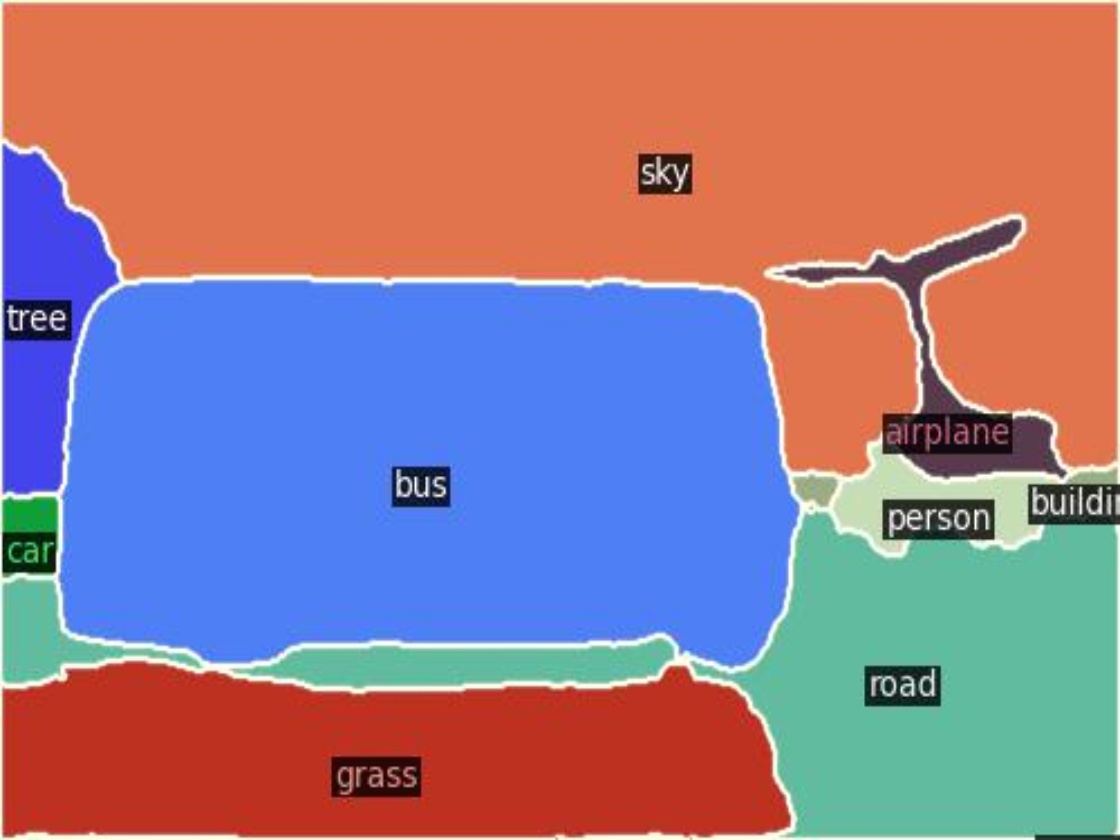} &
    \includegraphics[width=0.15\linewidth]{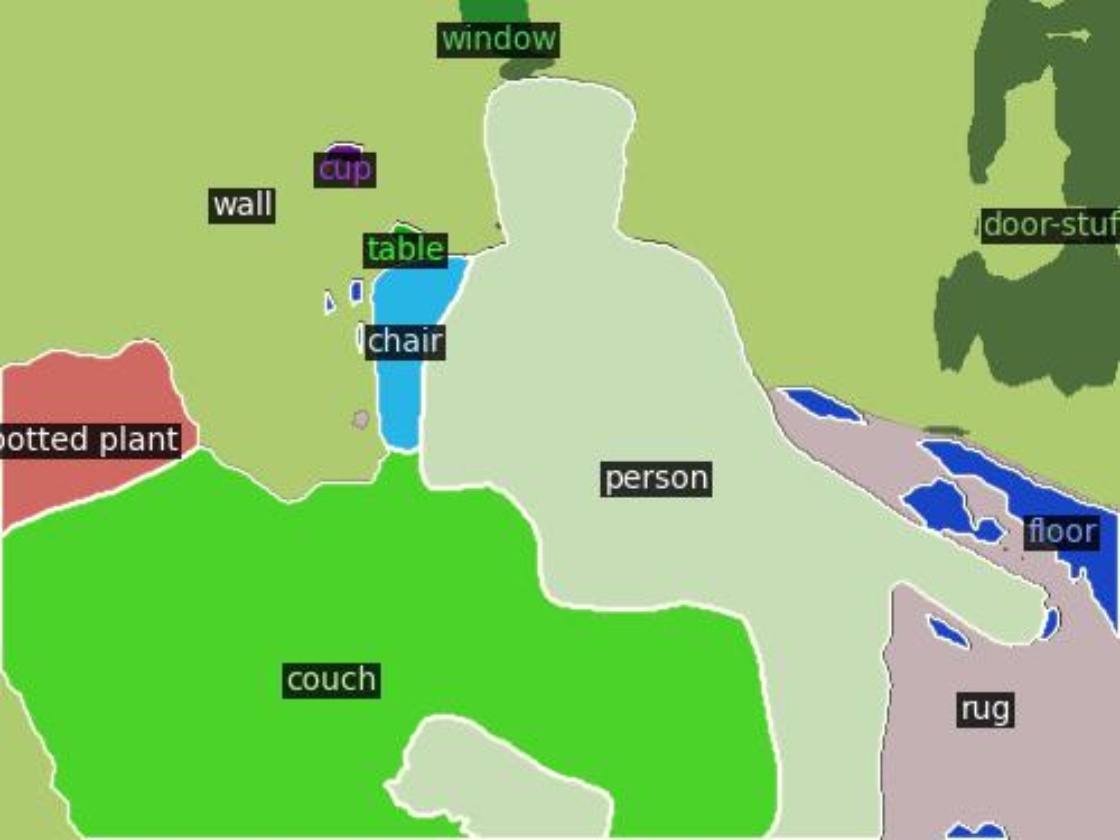} \\

  \end{tabular}
  \caption{Visual comparisons on unseen test datasets.}
  \label{fig:test_im}
\end{figure}

\begin{figure}[h]
    \centering
    \includegraphics[width=1.0\linewidth]{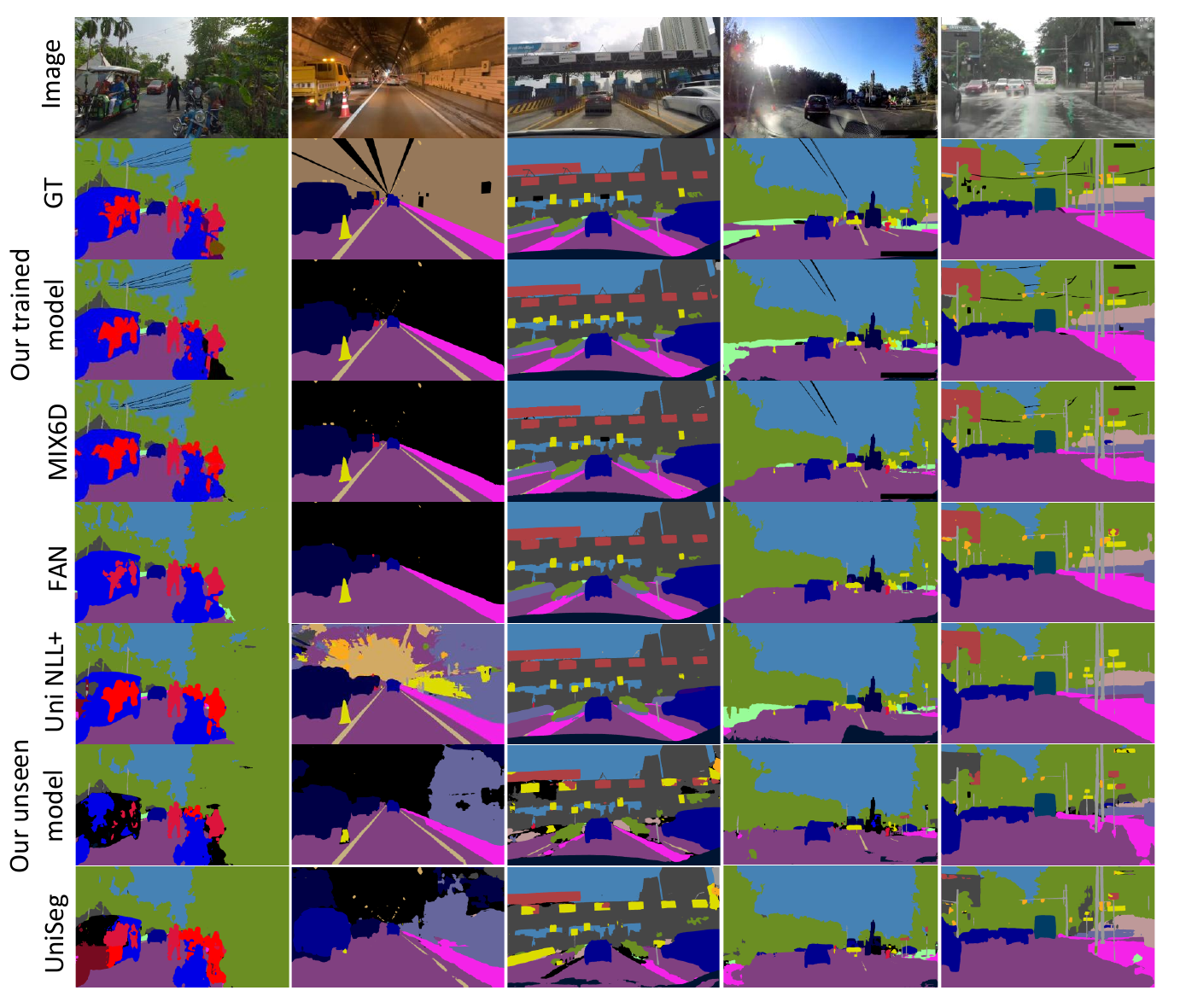}
    \caption{Visual comparisons on WildDash 2 benchmark.}
    \label{fig:wd_im}
\end{figure}

  \autoref{fig:train_im} presents the visual comparisons of models trained on single dataset and our model predicted in universal label space. From the figure, it is evident that our model achieves consistently strong performance across all training datasets while integrating label spaces from different datasets. For example, it predicts class \textit{lane marking} and \textit{crosswalk} for the ADE and BDD datasets, and predicts class \textit{books} for the SUN dataset. \autoref{fig:test_im} shows our universal predictions on test datasets. The results demonstrate that our method generalizes well across multiple unseen datasets from different domains. \autoref{fig:wd_im} shows the visual comparisons of different models on the WildDash 2 benchmark.

% %%%%%%%%%%%%%%%%%%%%%%%%%%%%%%%%%%%%%%%%%%%%%%%%%%%%%%%%%%%%
\clearpage
\newpage
\section*{NeurIPS Paper Checklist}

\begin{enumerate}

\item {\bf Claims}
    \item[] Question: Do the main claims made in the abstract and introduction accurately reflect the paper's contributions and scope?
    \item[] Answer: \answerYes{} % Replace by \answerYes{}, \answerNo{}, or \answerNA{}.
    \item[] Justification: The final paragraph of our introduction (\autoref{sec:intro}) provides a comprehensive summary of the paper's contributions, which align with the claims made in the abstract.
    % \item[] Guidelines:
    % \begin{itemize}
    %     \item The answer NA means that the abstract and introduction do not include the claims made in the paper.
    %     \item The abstract and/or introduction should clearly state the claims made, including the contributions made in the paper and important assumptions and limitations. A No or NA answer to this question will not be perceived well by the reviewers. 
    %     \item The claims made should match theoretical and experimental results, and reflect how much the results can be expected to generalize to other settings. 
    %     \item It is fine to include aspirational goals as motivation as long as it is clear that these goals are not attained by the paper. 
    % \end{itemize}

\item {\bf Limitations}
    \item[] Question: Does the paper discuss the limitations of the work performed by the authors?
    \item[] Answer: \answerYes{} % Replace by \answerYes{}, \answerNo{}, or \answerNA{}.
    \item[] Justification: In the third paragraph of our conclusion (\autoref{sec:dis}), we discuss the limitations of our work, including factors influencing performance, and computational efficiency considerations.

\item {\bf Theory Assumptions and Proofs}
    \item[] Question: For each theoretical result, does the paper provide the full set of assumptions and a complete (and correct) proof?
    \item[] Answer: \answerNA{} % Replace by \answerYes{}, \answerNo{}, or \answerNA{}.
    \item[] Justification: Our paper does not include theoretical results.
    % \item[] Guidelines:
    % \begin{itemize}
    %     \item The answer NA means that the paper does not include theoretical results. 
    %     \item All the theorems, formulas, and proofs in the paper should be numbered and cross-referenced.
    %     \item All assumptions should be clearly stated or referenced in the statement of any theorems.
    %     \item The proofs can either appear in the main paper or the supplemental material, but if they appear in the supplemental material, the authors are encouraged to provide a short proof sketch to provide intuition. 
    %     \item Inversely, any informal proof provided in the core of the paper should be complemented by formal proofs provided in appendix or supplemental material.
    %     \item Theorems and Lemmas that the proof relies upon should be properly referenced. 
    % \end{itemize}

    \item {\bf Experimental Result Reproducibility}
    \item[] Question: Does the paper fully disclose all the information needed to reproduce the main experimental results of the paper to the extent that it affects the main claims and/or conclusions of the paper (regardless of whether the code and data are provided or not)?
    \item[] Answer: \answerYes{} % Replace by \answerYes{}, \answerNo{}, or \answerNA{}.
    \item[] Justification: In \autoref{sec:method}, \autoref{apdx:train}, we provide detailed implementation steps, ensuring that our experiments can be replicated. Additionally, in \autoref{sec:exp}, we thoroughly describe the settings of our experiments, further facilitating reproducibility.

\item {\bf Open access to data and code}
    \item[] Question: Does the paper provide open access to the data and code, with sufficient instructions to faithfully reproduce the main experimental results, as described in supplemental material?
    \item[] Answer: \answerYes{} % Replace by \answerYes{}, \answerNo{}, or \answerNA{}.
    \item[] Justification: We provide reproducible code along with detailed instructions, ensuring that the main experimental results can be faithfully reproduced.
    % \item[] Guidelines:
    % \begin{itemize}
    %     \item The answer NA means that paper does not include experiments requiring code.
    %     \item Please see the NeurIPS code and data submission guidelines (\url{https://nips.cc/public/guides/CodeSubmissionPolicy}) for more details.
    %     \item While we encourage the release of code and data, we understand that this might not be possible, so “No” is an acceptable answer. Papers cannot be rejected simply for not including code, unless this is central to the contribution (e.g., for a new open-source benchmark).
    %     \item The instructions should contain the exact command and environment needed to run to reproduce the results. See the NeurIPS code and data submission guidelines (\url{https://nips.cc/public/guides/CodeSubmissionPolicy}) for more details.
    %     \item The authors should provide instructions on data access and preparation, including how to access the raw data, preprocessed data, intermediate data, and generated data, etc.
    %     \item The authors should provide scripts to reproduce all experimental results for the new proposed method and baselines. If only a subset of experiments are reproducible, they should state which ones are omitted from the script and why.
    %     \item At submission time, to preserve anonymity, the authors should release anonymized versions (if applicable).
    %     \item Providing as much information as possible in supplemental material (appended to the paper) is recommended, but including URLs to data and code is permitted.
    % \end{itemize}

\item {\bf Experimental Setting/Details}
    \item[] Question: Does the paper specify all the training and test details (e.g., data splits, hyperparameters, how they were chosen, type of optimizer, etc.) necessary to understand the results?
    \item[] Answer: \answerYes{} % Replace by \answerYes{}, \answerNo{}, or \answerNA{}.
    \item[] Justification: In \autoref{sec:exp}, we provide comprehensive details of our experimental setting.
    % \item[] Guidelines:
    % \begin{itemize}
    %     \item The answer NA means that the paper does not include experiments.
    %     \item The experimental setting should be presented in the core of the paper to a level of detail that is necessary to appreciate the results and make sense of them.
    %     \item The full details can be provided either with the code, in appendix, or as supplemental material.
    % \end{itemize}

\item {\bf Experiment Statistical Significance}
    \item[] Question: Does the paper report error bars suitably and correctly defined or other appropriate information about the statistical significance of the experiments?
    \item[] Answer: \answerNo{} % Replace by \answerYes{}, \answerNo{}, or \answerNA{}.
    \item[] Justification: Training on numerous datasets is time-consuming, with each experiment taking over a week. During our experiments, we observed consistent and stable performance in terms of mIoU for our model.
    % \item[] Guidelines:
    % \begin{itemize}
    %     \item The answer NA means that the paper does not include experiments.
    %     \item The authors should answer "Yes" if the results are accompanied by error bars, confidence intervals, or statistical significance tests, at least for the experiments that support the main claims of the paper.
    %     \item The factors of variability that the error bars are capturing should be clearly stated (for example, train/test split, initialization, random drawing of some parameter, or overall run with given experimental conditions).
    %     \item The method for calculating the error bars should be explained (closed form formula, call to a library function, bootstrap, etc.)
    %     \item The assumptions made should be given (e.g., Normally distributed errors).
    %     \item It should be clear whether the error bar is the standard deviation or the standard error of the mean.
    %     \item It is OK to report 1-sigma error bars, but one should state it. The authors should preferably report a 2-sigma error bar than state that they have a 96\% CI, if the hypothesis of Normality of errors is not verified.
    %     \item For asymmetric distributions, the authors should be careful not to show in tables or figures symmetric error bars that would yield results that are out of range (e.g. negative error rates).
    %     \item If error bars are reported in tables or plots, The authors should explain in the text how they were calculated and reference the corresponding figures or tables in the text.
    % \end{itemize}

\item {\bf Experiments Compute Resources}
    \item[] Question: For each experiment, does the paper provide sufficient information on the computer resources (type of compute workers, memory, time of execution) needed to reproduce the experiments?
    \item[] Answer: \answerYes{} % Replace by \answerYes{}, \answerNo{}, or \answerNA{}.
    \item[] Justification: In \autoref{sec:exp}, we specify that our experiments were conducted on four 80G A100 GPUs. 
    % \item[] Guidelines:
    % \begin{itemize}
    %     \item The answer NA means that the paper does not include experiments.
    %     \item The paper should indicate the type of compute workers CPU or GPU, internal cluster, or cloud provider, including relevant memory and storage.
    %     \item The paper should provide the amount of compute required for each of the individual experimental runs as well as estimate the total compute. 
    %     \item The paper should disclose whether the full research project required more compute than the experiments reported in the paper (e.g., preliminary or failed experiments that didn't make it into the paper). 
    % \end{itemize}
    
\item {\bf Code Of Ethics}
    \item[] Question: Does the research conducted in the paper conform, in every respect, with the NeurIPS Code of Ethics \url{https://neurips.cc/public/EthicsGuidelines}?
    \item[] Answer: \answerYes{} % Replace by \answerYes{}, \answerNo{}, or \answerNA{}.
    \item[] Justification: We have read and adhered to the NeurIPS Code of Ethics in conducting our research.
    % \item[] Guidelines:
    % \begin{itemize}
    %     \item The answer NA means that the authors have not reviewed the NeurIPS Code of Ethics.
    %     \item If the authors answer No, they should explain the special circumstances that require a deviation from the Code of Ethics.
    %     \item The authors should make sure to preserve anonymity (e.g., if there is a special consideration due to laws or regulations in their jurisdiction).
    % \end{itemize}

\item {\bf Broader Impacts}
    \item[] Question: Does the paper discuss both potential positive societal impacts and negative societal impacts of the work performed?
    \item[] Answer: \answerYes{} % Replace by \answerYes{}, \answerNo{}, or \answerNA{}.
    \item[] Justification: In the first paragraph of our conclusion (\autoref{sec:dis}), we thoroughly discuss the broader impacts of our work.

\item {\bf Safeguards}
    \item[] Question: Does the paper describe safeguards that have been put in place for responsible release of data or models that have a high risk for misuse (e.g., pretrained language models, image generators, or scraped datasets)?
    \item[] Answer: \answerNA{} % Replace by \answerYes{}, \answerNo{}, or \answerNA{}.
    \item[] Justification: Our research utilizes openly available datasets for training, and our models are primarily used for semantic segmentation tasks. Therefore, there are no such risks associated with our work.
    % \item[] Guidelines:
    % \begin{itemize}
    %     \item The answer NA means that the paper poses no such risks.
    %     \item Released models that have a high risk for misuse or dual-use should be released with necessary safeguards to allow for controlled use of the model, for example by requiring that users adhere to usage guidelines or restrictions to access the model or implementing safety filters. 
    %     \item Datasets that have been scraped from the Internet could pose safety risks. The authors should describe how they avoided releasing unsafe images.
    %     \item We recognize that providing effective safeguards is challenging, and many papers do not require this, but we encourage authors to take this into account and make a best faith effort.
    % \end{itemize}

\item {\bf Licenses for existing assets}
    \item[] Question: Are the creators or original owners of assets (e.g., code, data, models), used in the paper, properly credited and are the license and terms of use explicitly mentioned and properly respected?
    \item[] Answer: \answerYes{} % Replace by \answerYes{}, \answerNo{}, or \answerNA{}.
    \item[] Justification: In \autoref{sec:exp}, we clearly specify the assets used in our work, including datasets and models. We provide proper citations to the original papers and ensure that the licenses and terms of use are explicitly mentioned and respected.
    % \item[] Guidelines:
    % \begin{itemize}
    %     \item The answer NA means that the paper does not use existing assets.
    %     \item The authors should cite the original paper that produced the code package or dataset.
    %     \item The authors should state which version of the asset is used and, if possible, include a URL.
    %     \item The name of the license (e.g., CC-BY 4.0) should be included for each asset.
    %     \item For scraped data from a particular source (e.g., website), the copyright and terms of service of that source should be provided.
    %     \item If assets are released, the license, copyright information, and terms of use in the package should be provided. For popular datasets, \url{paperswithcode.com/datasets} has curated licenses for some datasets. Their licensing guide can help determine the license of a dataset.
    %     \item For existing datasets that are re-packaged, both the original license and the license of the derived asset (if it has changed) should be provided.
    %     \item If this information is not available online, the authors are encouraged to reach out to the asset's creators.
    % \end{itemize}

\item {\bf New Assets}
    \item[] Question: Are new assets introduced in the paper well documented and is the documentation provided alongside the assets?
    \item[] Answer: \answerYes{} % Replace by \answerYes{}, \answerNo{}, or \answerNA{}.
    \item[] Justification: We provide our code as an asset and include comprehensive documentation alongside it, covering details such as training procedures, licensing, limitations, and any necessary consent obtained from individuals whose assets are used.
    % \item[] Guidelines:
    % \begin{itemize}
    %     \item The answer NA means that the paper does not release new assets.
    %     \item Researchers should communicate the details of the dataset/code/model as part of their submissions via structured templates. This includes details about training, license, limitations, etc. 
    %     \item The paper should discuss whether and how consent was obtained from people whose asset is used.
    %     \item At submission time, remember to anonymize your assets (if applicable). You can either create an anonymized URL or include an anonymized zip file.
    % \end{itemize}

\item {\bf Crowdsourcing and Research with Human Subjects}
    \item[] Question: For crowdsourcing experiments and research with human subjects, does the paper include the full text of instructions given to participants and screenshots, if applicable, as well as details about compensation (if any)? 
    \item[] Answer: \answerNA{} % Replace by \answerYes{}, \answerNo{}, or \answerNA{}.
    \item[] Justification: Our paper does not involve crowdsourcing experiments or research with human subjects.
    % \item[] Guidelines:
    % \begin{itemize}
    %     \item The answer NA means that the paper does not involve crowdsourcing nor research with human subjects.
    %     \item Including this information in the supplemental material is fine, but if the main contribution of the paper involves human subjects, then as much detail as possible should be included in the main paper. 
    %     \item According to the NeurIPS Code of Ethics, workers involved in data collection, curation, or other labor should be paid at least the minimum wage in the country of the data collector. 
    % \end{itemize}

\item {\bf Institutional Review Board (IRB) Approvals or Equivalent for Research with Human Subjects}
    \item[] Question: Does the paper describe potential risks incurred by study participants, whether such risks were disclosed to the subjects, and whether Institutional Review Board (IRB) approvals (or an equivalent approval/review based on the requirements of your country or institution) were obtained?
    \item[] Answer: \answerNA{} % Replace by \answerYes{}, \answerNo{}, or \answerNA{}.
    \item[] Justification: Our work does not involve human subjects.
    % \item[] Guidelines:
    % \begin{itemize}
    %     \item The answer NA means that the paper does not involve crowdsourcing nor research with human subjects.
    %     \item Depending on the country in which research is conducted, IRB approval (or equivalent) may be required for any human subjects research. If you obtained IRB approval, you should clearly state this in the paper. 
    %     \item We recognize that the procedures for this may vary significantly between institutions and locations, and we expect authors to adhere to the NeurIPS Code of Ethics and the guidelines for their institution. 
    %     \item For initial submissions, do not include any information that would break anonymity (if applicable), such as the institution conducting the review.
    % \end{itemize}

\end{enumerate}

\end{document}